%% file: main.tex
\documentclass{article} 
\usepackage{iclr2022_conference,times}

\input{math_commands.tex}

\usepackage{hyperref}
\usepackage{url}

\title{Plant 'n' Seek:\\ Can You Find the Winning Ticket?}

\author{Jonas Fischer\\
Max Planck Institute for Informatics\\
\texttt{fischer@mpi-inf.mpg.de}\\
\And
Rebekka Burkholz\\
CISPA Helmholtz Center for Information Security\\
\texttt{burkholz@cispa.de}
}

\iclrfinalcopy 
\usepackage{dsfont} 
\usepackage{amsthm}

\newtheorem{thm}{Theorem}

\newtheorem{lemma}[thm]{Lemma}
\newtheorem*{theorem*}{Statement}
\usepackage{amsmath}

\usepackage{graphicx}
\usepackage{wrapfig}
\usepackage{subcaption}
\usepackage{fancyref}
\usepackage{bm}

\newcommand{\norm}[1]{\left\lVert#1\right\rVert}
\usepackage[ruled,vlined]{algorithm2e}
\SetKwInput{KwInput}{Input}                
\SetKwInput{KwOutput}{Output}

\SetCommentSty{mycommfont}

\newcommand{\grasp}{\textsc{GraSP}\xspace}
\newcommand{\snip}{\textsc{SNIP}\xspace}
\newcommand{\magnitude}{\textsc{Magnitude}\xspace}
\newcommand{\synflow}{\textsc{Synflow}\xspace}
\newcommand{\edgepopup}{\textsc{edge-popup}\xspace}
\newcommand{\edgepopupscaled}{\textsc{edge-pop-anneal}\xspace}
\newcommand{\rand}{\textsc{Random}\xspace}

\newcommand{\oururl}{\url{https://github.com/RelationalML/PlantNSeek/releases/tag/v1.0-beta}}

\usepackage{pgfplots}
\usepackage{subcaption}
\usepackage[notheorems]{eda} 
\usetikzlibrary{patterns,positioning}
\pgfkeys{/pgfplots/tuftelike/.style={
  semithick,
  tick style={major tick length=4pt,semithick,black},
  separate axis lines,
  axis x line*=bottom,
  axis x line shift=10pt,
  xlabel shift=5pt,
  axis y line*=left,
  axis y line shift=10pt,
  ylabel shift=5pt}}
  
\begin{document}

\maketitle

\input{abstract}
\input{intro}
\input{theory}

\input{experiments}

\input{discussion}
\bibliography{abbreviations, bib}
\bibliographystyle{iclr2022_conference}

\appendix
\input{appendixtheory}
\input{appendixexperiments}

\end{document}

%% file: math_commands.tex
\usepackage{amsmath,amsfonts,bm}

\def\eqref#1{equation~\ref{#1}}

\def\1{\bm{1}}

\DeclareMathAlphabet{\mathsfit}{\encodingdefault}{\sfdefault}{m}{sl}
\SetMathAlphabet{\mathsfit}{bold}{\encodingdefault}{\sfdefault}{bx}{n}

\DeclareMathOperator*{\argmax}{arg\,max}
\DeclareMathOperator*{\argmin}{arg\,min}

%% file: abstract.tex
\begin{abstract}

The lottery ticket hypothesis has sparked the rapid development of pruning algorithms that aim to reduce the computational costs associated with deep learning during training and model deployment. 
Currently, such algorithms are primarily evaluated on imaging data, for which we lack ground truth information and thus the understanding of how sparse lottery tickets could be. 
To fill this gap, we develop a framework that allows us to plant and hide winning tickets with desirable properties in randomly initialized neural networks.
To analyze the ability of state-of-the-art pruning to identify tickets of extreme sparsity, we design and hide such tickets solving four challenging tasks.
In extensive experiments, we observe similar trends as in imaging studies, indicating that our framework can provide transferable insights into realistic problems.
Additionally, we can now see beyond such relative trends
and highlight limitations of current pruning methods.
Based on our results, we conclude that the current limitations in ticket sparsity are likely of algorithmic rather than fundamental nature.
We anticipate that comparisons to planted tickets will facilitate future developments of efficient pruning algorithms.

\end{abstract}

%% file: intro.tex
\section{Introduction}\label{sec:introduction}
Deep learning has achieved breakthroughs in multiple challenging areas pertaining to machine learning but is computationally highly demanding. 
The benefits of overparameterization for training with SGD \citep{DD} seem to call for ever wider and deeper neural network (NN) architectures.
Training smaller, adequately regularized NNs from scratch to save resources could be a remedy but it seems to commonly fail, as \cite{frankle2019lottery} noted in their seminal paper.
However, they propose the lottery ticket hypothesis, which states that a small, well trainable subnetwork can be identified by pruning a large, randomly initialized NN, opening the field to discover such subnetworks or 'winning tickets'.
\cite{edgepopup} went even further and conjectured the existence of strong lottery tickets, i.e., subnetworks of randomly initialized NNs that do not require any further training.
This hypothesis holds the promise that training NNs could potentially be replaced by efficient pruning.  
The existence of strong lottery tickets has been proven for networks without biases \citep{malach2020proving,pensia2020optimal,orseau2020logarithmic} and with biases \citep{nonzerobiases}.

While these types of proofs show existence in realistic settings, the sparsity of the constructed tickets is likely not optimal, as they rely on pruning neurons to degree $1$.
The construction and proof of the generalized strong tickets raises two questions -- is the suboptimal sparsity merely an artifact of existence proofs or a general limitation of pruning and hence further improvements are needed to achieve effective network compression? 
And, if they exist, are current algorithms able to find very sparse tickets?
We could answer the first question by assessing the sparsity of tickets that have been found by network pruning, which however is limited to sparsities that current state-of-the-art pruning algorithms are able to achieve. 
Hence, both raised questions require to properly evaluate the state-of-the-art algorithms for pruning networks before training.
In a recent study, Frankle et al. evaluated existing pruning algorithms and concluded that there is no single winning method across considered settings and sparsity levels \citep{frankle2021review}. 
Furthermore, they raised the issue of missing baselines in the field, which renders the evaluation of current and future work difficult.

To fill this gap and generate baselines with known ground truth, we here propose an algorithm to plant and hide arbitrary winning tickets in randomly initialized NNs and construct sparse tickets that reflect three common challenges in machine learning.
We use this experimental set-up to compare state-of-the-art pruning algorithms designed to search for lottery tickets.

Our results indicate that state-of-the-art methods achieve sub-optimal sparsity levels, and are not able to recover good tickets \textit{before} training. 
The qualitative trends are consistent with previous results on image classification tasks \citep{synflow,frankle2021review} indicating that our experimental set-up exposes pruning algorithms to realistic challenges. 
In addition, we improve a state-of-the-art pruning algorithms towards finding strong lottery tickets of better sparsity.
Our proposed planting framework will enable the evaluation of future progress into this direction.

\textbf{Contributions}
1) We prove the existence of strong lottery tickets with sparse representations. 2) Inspired by the proof, we derive a framework that allows us to plant and hide strong tickets in neural networks and thus create benchmark data with known ground truth. 3) We construct sparse representations of three types of tickets that reflect typical machine learning problems. 4) We systematically evaluate state-of-the-art pruning methods that aim to discover tickets on these three problems against the ground truth tickets and highlight key challenges.


\subsection{Related work}

Pruning approaches for neural networks can be broadly categorized into three groups, pruning before, during, or after training.
While methods that sparsify the network during \citep{frankle2019lottery,srinivas2016gendropout,orthoRepair}, or after training \citep{sigmoidl0, lecun1990optimal, hassibi1992second, dong2017surgeon, li2017pruneconv, molchanov2017pruneinf} help in reducing resources required for inference, they are, however, less helpful in reducing resources at training time but can make a difference if they prune early aggressively \citep{earlybird}.\\
The lottery ticket hypothesis \citep{frankle2019lottery} has also promoted the development of neural network pruning algorithms that prune \textit{before} training \citep{grasp,snip,snipit,synflow,edgepopup}. 
Usually, these methods try to find lottery tickets in a 'weak' (but powerful) sense, that is to identify a sparse neural network architecture that is well trainable starting from its initial parameters.
These methods usually score edges in terms of network flow, which can be quantified by gradients at different stages of pruning or by edge weights, and prune edges with the lowest scores until the desired sparsity is achieved.\\
Strong lottery tickets are sparse sub-networks that perform well with the initial parameters, hence do not need further training \citep{edgepopup}.
Their existence has been proven by providing lower bounds on the width of the large, randomly initialized neural network that contains them \citep{malach2020proving,pensia2020optimal,orseau2020logarithmic,nonzerobiases,universal}.
In addition, it was shown that multiple candidate tickets exist that are also robust to parameter quantization \citep{multiprize}.



\subsection{Notation}
Let $f(x)$ denote a bounded function, without loss of generality $f: {[-1,1]}^{n_0} \rightarrow {[-1,1]}^{n_L}$, that is parameterized as a deep neural network with architecture  $\bar{n} = [n_0, n_1, ..., n_L]$, i.e., depth $L$ and widths $n_l$ for layers $l = 0, ..., L$ with ReLU activation function $\phi(x):= \max(x,0)$.
It maps an input vector $\bm{x}^{(0)}$ to neurons $x^{(l)}_i$ as $\bm{x}^{(l)} = \phi\left(\bm{W}^{(l)} \bm{x}^{(l-1)} + \bm{b}^{(l)} \right)$, where
$\bm{W}^{(l)} \in \mathbb{R}^{n_{l-1} \times n_l}$ is the weight matrix, and $\bm{b}^{(l)} \in \mathbb{R}^{n_l}$ is the bias vector of Layer $l$. We will establish approximation results with respect to the supremum norm $\norm{g}_{\infty} := \sup_{x \in {[-1,1]}^{n_0}} \norm{g}_2$ defined for any function $g$ on the domain ${[-1,1]}^{n_0}$. 
%
Assume furthermore that a ticket $f_{\epsilon}$ can be obtained by pruning a large mother network $f$, which we indicate by writing $f_{\epsilon} \subset f_{0}$. 
The sparsity level $\rho$ of $f_{\epsilon}$ is then defined as the fraction of non-zero weights that remain after pruning, i.e., $\rho = \left(\sum_l \norm{\bm{W}^{(l)}_{\epsilon}}_0 \right)/\left(\sum_l  \norm{\bm{W}^{(l)}_{0}}_0 \right)$, where $\norm{\cdot}_{0}$ denotes the $l_0$-norm, which counts the number of non-zero elements in a vector or matrix.
Another important quantity that influences the existence probability of lottery tickets is the in-degree of a node $i$ in layer $l$ of the target $f$, which we define as the number of non-zero connections of a neuron to the previous layer plus $1$ if the bias is non-zero, i.e., $k^{(l)}_i := \norm{\bm{W}^{(l)}_{i,:}}_{0} + \norm{b^{(l)}_{i}}_{0}$, where $\bm{W}^{(l)}_{i,:}$ is the $i$-th row of $\bm{W}^{(l)}$.
The maximum degree of all neurons in layer $l$ is denoted as $k_{l,\text{max}}$.

%% file: theory.tex
\section{Existence of lottery tickets}
Pruning algorithms that search for strong lottery tickets achieve sparsity levels of at best $0.5$ when the resulting models should be able to compete with the accuracy of the entire, trained mother network \citep{edgepopup}.
Proofs of the existence of strong lottery tickets give no clear indication whether this is an algorithmic shortcoming, which could be overcome, or a fundamental limitation of pruning randomly initialized networks alone.
The reason is that existing proofs \citep{malach2020proving,pensia2020optimal,orseau2020logarithmic,nonzerobiases} guarantee high existence probabilities of subnetworks that have double the depth and $2-30$ times the width of the target network and thus non-optimal sparsity. 
Based on their $2L$ construction, \cite{malach2020proving} even went so far to conclude that training by pruning might be computationally at least as hard as training shallower neural networks.
However, it is well known that specific function classes can be approximated in significantly more parameter efficient ways by deeper neural networks rather than shallower ones \citep{deepOverShallow,approx} and also be learned more efficiently \citep{hieberCompose}.
Thus, by leveraging its full depth, the randomly initialized $2L$ deep neural network might contain a much sparser lottery ticket than any of the ones whose existence has been proven.

As a first step towards making claims about the existence of very sparse representations, we therefore prove next a lower bound on the probability that a target network of general architecture is contained in a larger, randomly initialized neural network with the same depth as the target network.
%
%
%
As many relevant targets have known representations of lower sparsity than what is covered by this bound, we will afterwards propose a planting algorithm to design experiments that can distinguish between algorithmic and fundamental limitations of pruning for strong lottery tickets.




\subsection{Lower bound on existence probability}\label{sec:fulldepth}
 Pruning a randomly initialized neural network usually recovers a network that has parameters close to, but not exactly like a target network.
First, we need to understand how these errors in the parameters affect the final network output and what error ranges are acceptable. 
For completeness, we restate Lem.~1 of \citep{nonzerobiases} that guarantees an $\epsilon$ approximation of the entire network. 
%
\begin{lemma}[Error propagation]\label{thm:approx}
Assume $\epsilon > 0$ and let the target network $f$ and its approximation $f_{\epsilon}$ have the same architecture.
If every parameter $\theta$ of $f$ and corresponding $\theta_{\epsilon}$ of $f_{\epsilon}$ in layer $l$ fulfils $|\theta_{\epsilon} - \theta| \leq \epsilon_{l}$ for 
\begin{align}\label{eq:epsTheta}
\epsilon_l := \epsilon \left(L \sqrt{n_l k_{l,\text{max}}} \left(1+  \sup_{x \in {[-1,1]}^{n_0}} \norm{\bm{x}^{(l)}}_{1}\right) \prod^L_{k=l+1} \left(\norm{\bm{W}^{(l)}}_{\infty} + \epsilon/L \right) \right)^{-1},
\end{align}
then it follows that $\norm{f-f_{\epsilon}}_{\infty} \leq \epsilon$. 
\end{lemma}
Respecting the allowed errors $\epsilon_l$, we can next establish a lower bound on the existence probability of a specific target network assuming standard initialization schemes with necessary non-zero bias initialization \citep{nonzerobiases}. 
The main argument is a union bound over matching each target neuron $i$ (with $k_i$ parameters) with neurons of the mother network in the corresponding layer. 

\begin{thm}[Lower bound on existence probability]\label{thm:LTexistFullDepth}
Assume that ${\epsilon \in {(0,1)}}$ and a target network $f$ with depth $L$ and architecture $\bar{n}$ are given. Each parameter of the larger deep neural network $f_{0}$ with depth $L$ and architecture $\bar{n}_0$ is initialized independently, uniformly at random with $w^{(l)}_{ij} \sim U{\left(\left[-\sigma^{(l)}_w, \sigma^{(l)}_w\right]\right)}$ and $b^{(l)}_{i} \sim U{\left(\left[- \prod^l_{k=1}\sigma^{(k)}_w, \prod^l_{k=1} \sigma^{(k)}_w\right]\right)}$. Then, $f_{0}$ contains a rescaled approximation $f_{\epsilon}$ of $f$ with probability at least
\begin{align}
  \mathbb{P}\left(\exists f_{\epsilon} \subset f_{0}: \ \norm{f- \lambda f_{\epsilon}}_{\infty} \leq \epsilon \right) \geq \prod^L_{l=1} \left(1 - \sum^{n_l}_{i=1}(1 - \epsilon^{k_i}_l )^{n_{l,0}} \right),
\end{align}
where $\epsilon_l$ is defined as in Eq.~(\ref{eq:epsTheta}) and the scaling factor is given by $\lambda = \prod^L_{l=1}1/\sigma^{(l)}_w$.
\end{thm}
We could obtain similar results for initially normally distributed weights and biases, we would just have to substitute $\epsilon_l$ by $\epsilon_l/2$. 
A proof is provided in Appendix~\ref{sec:existproof}.

\input{figs/data_plots.tex}

Thm.~\ref{thm:LTexistFullDepth} provides us with an intuition for what kind of lottery ticket architectures we can expect to find.
First of all, it tells us that a large number of nodes in a layer, and more importantly nodes with large in-degree $k_i$, render the existence of a specific network architecture as winning lottery ticket less likely. 
Each additional layer reduces the probability further.
Moreover, we observe that the last layer is a bottleneck, as it usually has the same width as in the large initial network.
To circumvent this problem we could assume that the last layer can be trained, which is common practice.

A higher width of the mother network is clearly advantageous. 
Note that we could turn this theorem also into a lower bound on the width $n_{l,0}$ of the larger mother network as it is common in existence proofs. 
Assuming the same width $n_{l,0} = n_0$ and $n_l = n$ across layers, we would receive roughly $n_0 \geq C \log(L n/\delta) \max_l\left(\epsilon^{-k_{\text{max}}}_l\right)$.
Even though it is polynomial in the relevant parameters, it only provides a practical existence proof for extremely sparse architectures. 
We therefore have to resort to planting to answer fundamental questions about abilities of pruning algorithms.
In fact, the proof of the above theorem inspires the planting algorithm introduced next. 
\subsection{Planting lottery tickets}

As we have discussed, the tickets that exist with high probability rarely fulfill criteria of interest, such as low sparsity, favorable generalization properties, or adversarial robustness. 
We therefore propose to plant winning tickets with such desirable properties 
within randomly initialized neural networks.
This approach offers the flexibility to design experiments of different degrees of difficulty and generate training and test data based on a baseline ticket.

A simple approach to planting a target $f$ in a network $f_0$ would be to select a random subset of neurons in each layer and set them to their target values and otherwise randomly initialize the rest.
This, however, would usually lead to a trivially detectable ticket because the target parameters are much larger than the initialized parameters of the larger mother network.
The reason is that both networks produce output that lies in a similar range (ideally the one of the training labels).
Yet, the target network has to achieve this by adding up a much smaller number of parameters.
A different perspective on the same issue is that a pruned lottery ticket needs to be scaled up to compensate for the lost parameters.
Note the scaling factor $\lambda$ in Theorem~\ref{thm:LTexistFullDepth} for that purpose.
Thus, at least, we would need to scale the target parameters appropriately during planting. 

We follow a more general approach that also applies to networks $f_0$ whose parameters have not been randomly initialized and that captures the full variability of possible target solutions by allowing for different scaling factors per neuron.
We search in each layer of $f_0$ for suitable neurons that best match a target neuron in $f$, starting from the first layer.
Given that we matched all neurons in layer $l-1$, we try to establish their connections to neurons in the current layer $l$.
A best match is decided by minimizing the l2-distance to its potential input parameters thereby adjusting for an optimal scaling factor.
For example, let neuron $i$ in Layer $l$ of the target $f$ have non-zero parameters $\bm{\theta} = (b, \bm{w})$ that point to already matched neurons in Layer $l-1$.
Each neuron $j$ in Layer $l$ of $f_0$ that we have not matched yet could be a potential match for $i$.
Let the corresponding parameters of $j$ be $\bm{m}$.
The match quality between $i$ and $j$ is assessed by $q_{\theta}(m)=\norm{\bm{\theta}- \lambda(m)\bm{m}}_2$, where $\lambda(m) =  \bm{\theta^T m}/\norm{\bm{m}}^2_2 $ is the optimal scaling factor. 
The best matching parameters $m^* = \argmin_{m} q_{\theta}(m)$ are replaced by rescaled target parameters $\theta/\lambda(m^*)$ in $f_0$. 
We provide pseudocode and details in Supp. \ref{sec:plantingalgo}.
\subsection{Construction of targets for planting}
Based on the proposed planting algorithm, we generate sparse tickets for three problems that expose general pruning algorithms to common challenges in machine learning.
On purpose, these are designed to avoid high computational burdens.

\textbf{Regression of a ReLU unit} (\texttt{ReLU}).
The ReLU unit $\phi(x)$ is an essential building block of state-of-the-art neural networks.
It is particularly interesting to study because we know the optimal solution and can guarantee that it exists with high probability. 
Assuming a mother network $f_0$ of depth $L$, a ReLU can be implemented with a single neuron per layer.
Any path through the network with positive weights $\prod^L_{l=1} \phi(w_{i_{l-1} i_l} x)$ defines a ReLU with scaling factor $\lambda = \prod^L_{l=1} w_{i_{l-1} i_l}$ for indices $i_l$ in Layer $l$ with $w_{i_{l-1} i_l} > 0$.
Note that each random path fulfills this criterion with probability $0.5^L$ so that even random pruning has a considerable chance to find an optimal ticket.
A winning path exists with probability $\prod^{L}_{l=1} (1-0.5^{n_{l,0}})$, which is almost $1$ even in relatively small networks.
As a ticket with optimal sparsity exists with high probability, we would not need to plant it.
However, to give also pruning algorithms that do not specifically prune biases a chance, we set ticket biases  to zero.
As we will see in experiments, despite this simplification, pruning algorithms are severely challenged and cannot find an optimally sparse ticket.

\begin{figure}
  \begin{subfigure}[t]{0.22\textwidth}
  \centering
  \includegraphics[width=0.85\textwidth]{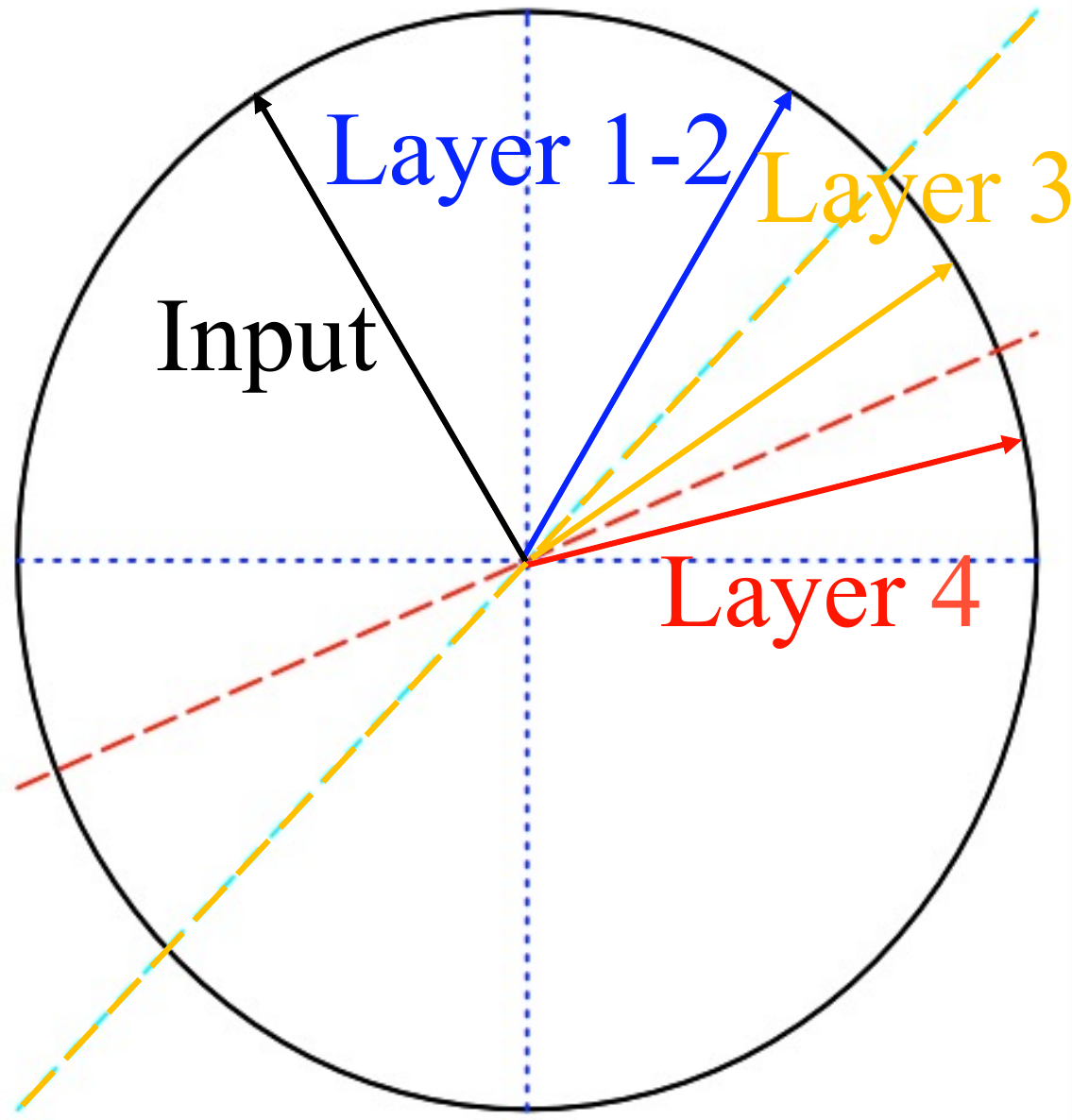}
  \caption{Mirroring along axes for \texttt{Circle}.}
  \end{subfigure}
 \hspace{0.5em}
  \begin{subfigure}[t]{0.22\textwidth}
  \centering
 \includegraphics[width=0.9\textwidth]{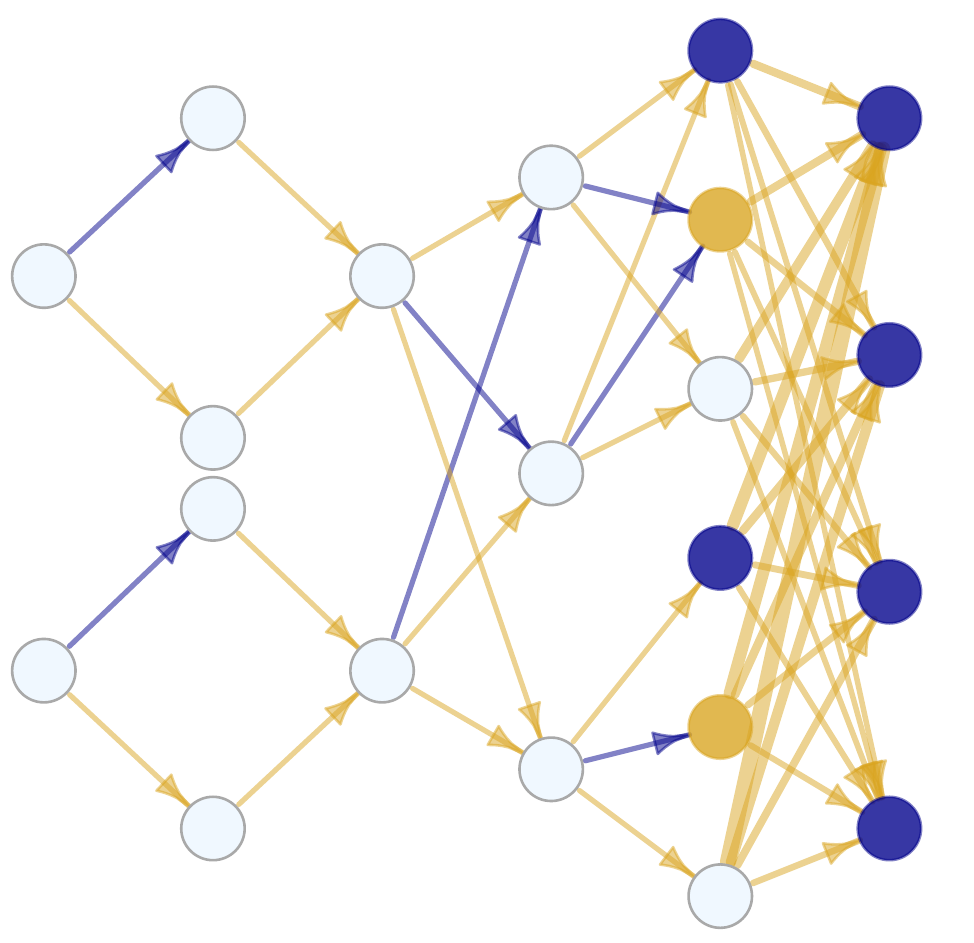}
  \caption{\texttt{Circle} architecture with depth $L=5$.}
  \end{subfigure}
  \hspace{0.5em}
    \begin{subfigure}[t]{0.22\textwidth}
  \centering
  \includegraphics[width=0.9\textwidth]{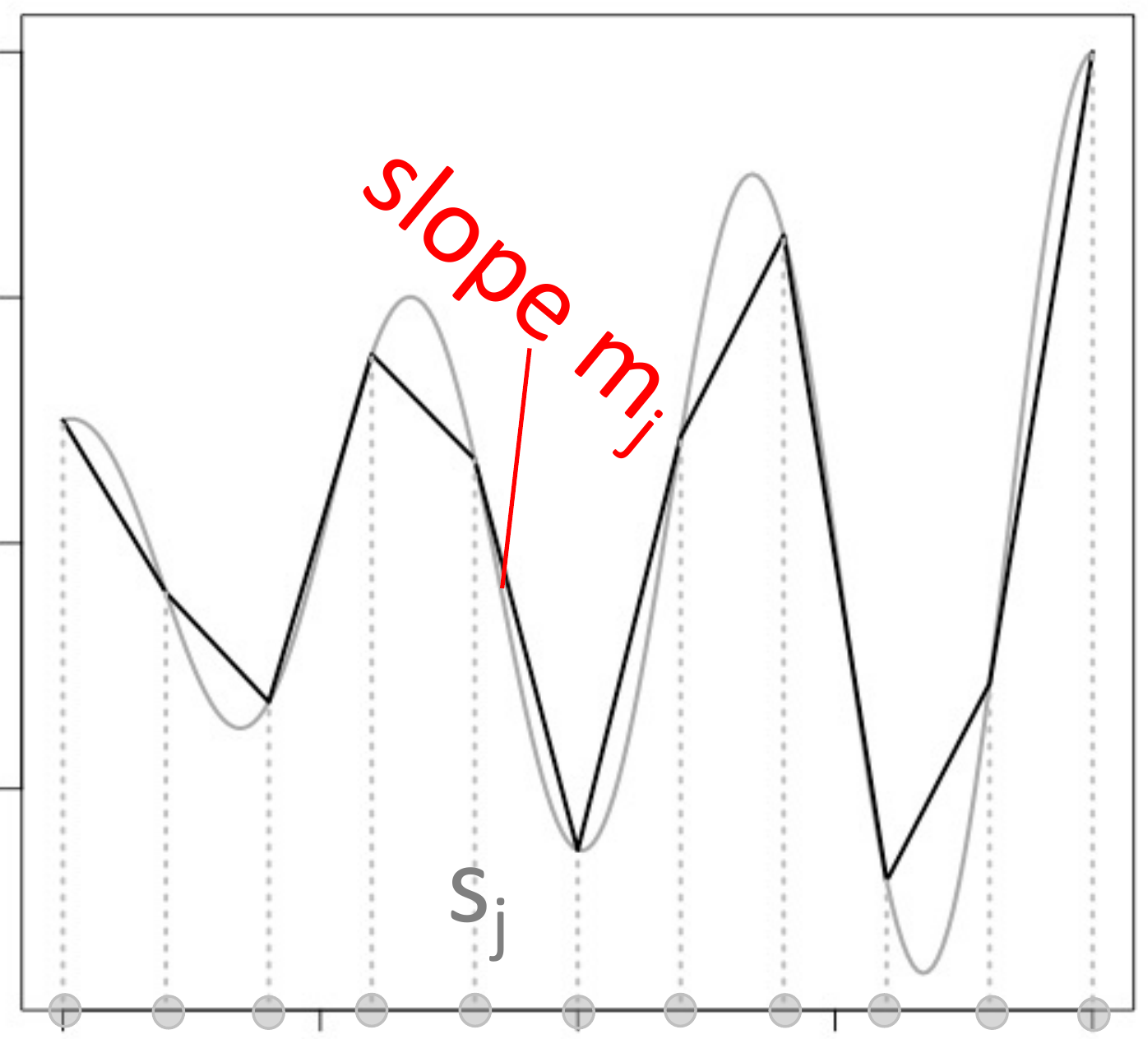}
  \caption{Univariate function approximation.}
  \end{subfigure}
  \hspace{0.5em}
  \begin{subfigure}[t]{0.22\textwidth}
  \centering
  \includegraphics[width=0.9\textwidth]{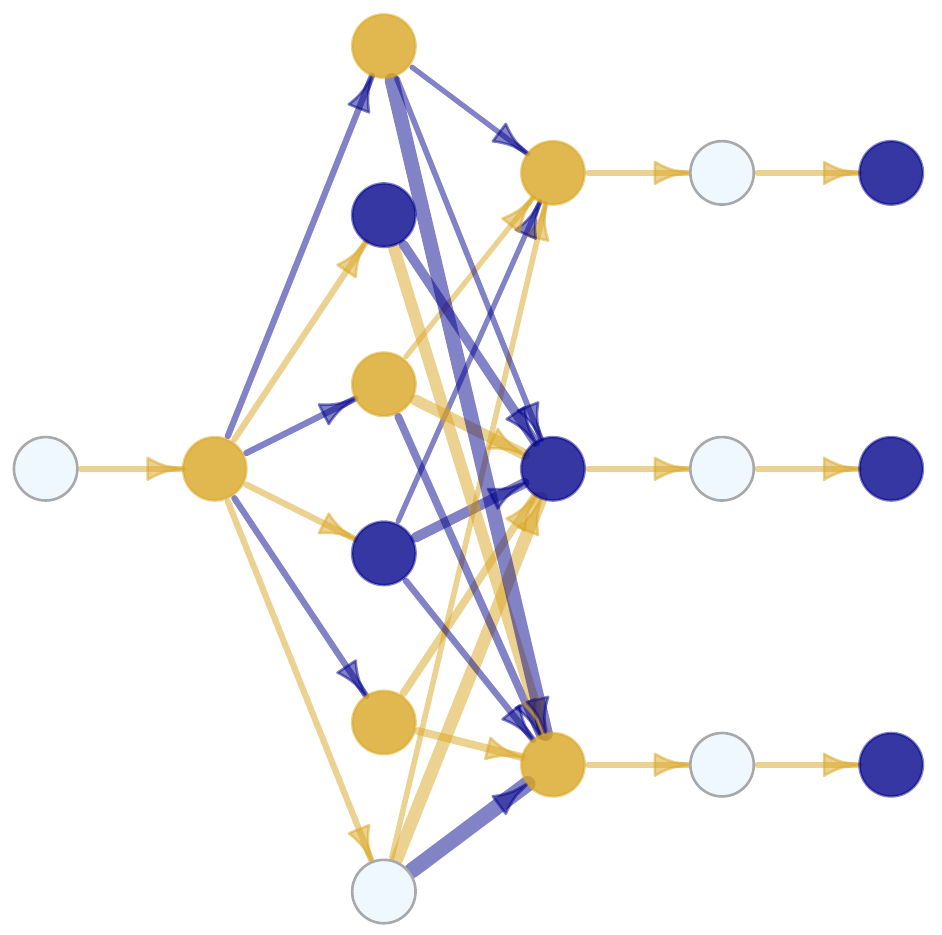}
  \caption{\texttt{Helix} architecture with depth $L=5$.}
  \end{subfigure}
  \caption{(a) Visualization of the first layers of \texttt{Circle} representing $g(x_1, x_2) = x^2_1 + x^2_2$. (c) Univariate deep neural network parametrization with outer weights $a^{(i)}_{j} = \Delta m_j = m_j-m_{j-1}$. (b+d) Ticket architectures, edge width is proportional to the absolute weight value, blue indicates a negative sign, yellow a positive sign. Neurons are colored by bias sign, gray indicates zero biases.}
  \label{fig:circleExplainMain}
\end{figure}

\textbf{Classification of rings} (\texttt{Circle}). 
Another basic building block of many functions, in particular, radial symmetric functions, is the radius or l2-norm of a vector.
It is also an important operation to represent products via the relation $xy = 0.25 ((x+y)^2 - (x-y)^2)$. 
We derived a sparse representation that leverages the full depth of a given network, as its sparsity improves with increasing depth. 
We visualize the related 4-class classification problem with 2-dimensional inputs in Fig.~\ref{fig:dataplot} (left).
The output is 4-dimensional, where each output unit $f_c(x)$ corresponds to the probability $f_c(x)$ that an input $(x_1, x_2) \in [-1,1]^2$ belongs to class $c$ that is computed with softmax activation functions. 
The decision boundaries are defined in the last layer based on inputs of the form $g(x_1, x_2) = x^2_1 + x^2_2$. 
The high symmetry of $g(x_1, x_2)$ allows us to construct a particularly sparse representation by mirroring data points along axes as visualized in Figure~\ref{fig:circleExplainMain}a.
Each consecutive layer $l$ mirrors the previous layer along the axis $\bm{a^{(l)}} = (\cos(\pi/2^{l-1}), \sin(\pi/2^{l-1}))$.
To enable higher precision for networks of smaller depth, the second to last layer approximates $h(x) = x^2$ for each component.
This is unnecessary for representations of high enough depth.
The details of our construction are explained in Supp.~\ref{sec:circleExplain}.
Fig.~\ref{fig:circleExplainMain}b shows an exemplary architecture of the planted ticket, for which we can vary the depth and width of the second to last layer. 

%
\textbf{Identification of a submanifold} (\texttt{Helix}).  
Another common problem in machine learning is to learn lower dimensional functions that are embedded in a higher dimensional space.
Fig.~\ref{fig:dataplot} (right) shows our minimal regression example in form of a helix. 
As we have observed that many pruning algorithms have the tendency to keep a higher number of neurons closer to the input (and sometimes also the output layer), we construct a ticket that has similar properties, see Fig.~\ref{fig:circleExplainMain}d.
This should ease the task for pruning algorithms to find the planted winning ticket but, as we show in our experiments, this \texttt{Helix} problem is surprisingly challenging. 
The helix has three output coordinates $f_1(x)=(5\pi + 3\pi x) * \cos(5\pi + 3\pi x)/(8\pi)$, $f_2(x)=(5\pi + 3\pi x) * \sin(5\pi + 3\pi x)/(8\pi)$, and $f_3(x)=(5\pi + 3\pi x)/(8\pi)$ for 1-dimensional input $x \in [-1,1]$.
We can approximate each of the components $f_i(x)$ by an univariate deep neural network $n_i(x) = \sum^{N}_{j = 1} a^{(i)}_{j} \phi(x - s_j) + b^{(i)}$ with depth $L=2$ (see Fig.~\ref{fig:circleExplainMain}c and Supp.~\ref{sec:helixExplain}), which is achieved by the first layers, while the remaining layers model the identity.
An interesting feature of this example is that the largest width could be assigned to almost any layer of the architecture, which makes \texttt{Helix} a good candidate to provoke layer collapse in different pruning algorithms.

\textbf{Strong tickets based on trained neural networks} 
To answer our original question whether state-of-the-art pruning algorithms can find extremely sparse strong lottery tickets, we plant a trained weak lottery ticket in a randomly initialized (VGG like) neural network.
Note that the proposed pruning algorithm can also be applied to convolutional layers in addition to fully connected ones.
%
%

%% file: figs/data_plots.tex
\begin{figure}
\begin{subfigure}[t]{0.45\textwidth}
\begin{tikzpicture}
\begin{axis}[
    jonas scatter color,
    colormap={circle}{[0.5cm]
    color(1cm)=(internationalorange)
    color(2cm)=(twilightlavender)
    color(3cm)=(uscgold)
    color(4cm)=(electricblue)
    },
	width=4.5cm,
	height=3.5cm,
    only marks,
    xtick={-1,-0.5,0,0.5,1},
    ytick={-1,-0.5,0,0.5,1},
    x label style = {at={(axis description cs:0.5,-0.1)}, anchor=north, font=\scriptsize},
    y label style = {at={(axis description cs:0,0.5)},  anchor=south, font=\scriptsize},
    xlabel=$x$,
    ylabel=$y$,
    ]
    \addplot [point meta=explicit, mark=*] table[x index = 0, y index = 1, meta index = 2] {figs/circlesmall.txt};
\end{axis}
\end{tikzpicture}
\end{subfigure}
\hfill
\begin{subfigure}[t]{0.53\textwidth}
\begin{tikzpicture}
\begin{axis}[
    jonas scatter3d color,
    colormap/viridis,
    colorbar,
	colorbar style={
	  at={(1.2,0.4)},
	  anchor=west,
	  title={\scriptsize $x$},
	  height=1.5cm,
	  width=.2cm,
	  yticklabel style={
            font=\tiny,
            text width=1.5em,
            align=right,
            /pgf/number format/.cd,
                fixed,
                fixed zerofill
        },
	},
	width=5.0cm,
	height=4.5cm,
    only marks,
    xtick={-1,-0.5,0,0.5,1},
    ytick={-1,-0.5,0,0.5,1},
    ztick={0,0.25,0.5,0.75,1},
    x label style = {at={(axis description cs:0.3,-0.1)}, anchor=north, font=\scriptsize},
    y label style = {at={(axis description cs:1.2,-0.1)}, anchor=south, font=\scriptsize},
    z label style = {at={(axis description cs:-0.3,0.4)}, anchor=south, font=\scriptsize},
    xlabel=$f_1$,
    ylabel=$f_2$,
    zlabel=$f_3$,
    ]
    \addplot3+ [point meta=explicit, mark=*, point meta min={-1}, point meta max={1}] table[meta index = 0, x index = 1, y index=2, z index = 3] {figs/helixsmall.txt};
\end{axis}
\end{tikzpicture}
\end{subfigure}
\caption{\textit{Benchmark data.} Shown are samples from the \texttt{Circle} (left) and \texttt{Helix} (right) task.}
\label{fig:dataplot}
\end{figure}

%% file: experiments.tex
\section{Experiments}\label{sec:experiments}

We evaluate existing approaches that seek winning tickets on the previously introduced problems with known ground truth. To show that our experiments reflect realistic conditions we also compare the general trends to results on image tasks.
For reproducibility, we provide details on data, networks, and experimental setup in Supp. B and code online.\!\footnote{\oururl}

For comparison, we consider \grasp, \snip, \synflow, \magnitude pruning, and  \rand pruning as baseline \citep{grasp, snip, synflow, frankle2019lottery}, which are all algorithms to discover weak tickets.
We do not consider partially trained tickets, such as by rewinding \citep{rewind}.
Additionally, we consider \edgepopup that was designed to find strong tickets \citep{edgepopup}. 
We use existing implementations of weak ticket pruners \citep{synflow} and implement \edgepopup ourselves, extending its score values to bias parameters.
Moreover, we test different pruning strategies.

\subsection{Pruning without intermediate training}
\label{sec:singleshot}

First, we consider pruning strategies without intermediate training, where tickets are discovered by pruning a network using solemnly scores of initial network parameters.

\input{figs/singleshot}

\textbf{Singleshot pruning} 
In singleshot pruning, which is originally applied in \snip and \grasp,  edges are scored in a single pass and then pruned to the desired sparsity.
Compared to multishot pruning, this saves significant amounts of resources by preventing training entirely, if a strong ticket is found, or only training a small subnetwork once, in case a weak ticket is found.
For our benchmark data, we construct networks of depth $5$ and width $100$ and test the ability of algorithms to discover both strong and weak tickets. The key results are visualized in Fig. \ref{fig:singleshot}, reporting performance of the algorithms trying to discover tickets at varying sparsity levels across 25 repetitions.
Results for varying network depths and widths can be found in Supp. B noting that they are consistent also with larger depth, but the methods fail to find any ticket in more shallow networks of depth 3.

We find that all approaches, including \rand pruning, are able to find weak tickets for moderate sparsity levels for \texttt{Circle} and \texttt{ReLU}, but fail to recover them on the manifold learning task \texttt{Helix} entirely.
Although \magnitude pruning was originally not designed for this pruning strategy, it is on par with state-of-the-art singleshot methods.
For lower sparsity levels $\leq 0.01$, in particular baseline ticket sparsity, all methods fail to recover good subnetworks.
Dissecting the results, we observe layer collapse, meaning that entire layers are masked, thus disrupting flow through the network.
Despite that \synflow was proposed as a solution to this issue, we observe that it also experiences layer collapse for extreme sparsities, even when pruning for $100$ rounds as suggested in the original paper, performing only slightly better than with one round of pruning (see Supp. B).
In summary, with only a single pruning round, most pruning algorithms discover weak tickets at moderate sparsity, however fail to recover weak tickets of low sparsity and any strong tickets.

\textbf{Robustness to noise}
To model real-world settings, all our datasets contain small amounts of noise as described in Supp. B. To rule out that noise in the data is the primary source for issues with discovering tickets, we generated \texttt{Circle} datasets with varying levels of noise.
The results indicate that on the one hand, without noise we do not see much of an improvement in terms of discovered tickets, but on the other hand observe that the algorithms are robust to even large amounts of noise, finding tickets with almost similar performance as with no noise at all (see Supp. B).


\textbf{Comparison to results on image data}
The reported results are in line with experiments on image data as reported in the literature \citep{synflow}.
In particular, for all these methods a similar drop around $0.01$ sparsity is observed for different VGG and Resnet architectures and image data sets.
Similarly, layer collapse has been reported for image data.
The main difference is that for our data, we know the obtainable sparsity as well as performance of tickets, setting these results into a context beyond trendline differences of methods for selected sparsity values.

\subsection{Pruning combined with training}

While much more resource intensive, iteratively training followed by pruning and resetting to initial weights, slowly annealing to the desired sparsity, has been proven a successful approach to discover lottery tickets.
Here we extend this pruning scheme to other approaches beyond magnitude pruning.

\input{figs/multishot}
\textbf{Multishot pruning}
To investigate the effect of multishot pruning, we run each of the previous methods iteratively for $10$ rounds on our benchmark datasets (see Fig. \ref{fig:multishot}).
Analogous to iterative magnitude pruning, for each round $r$, we iteratively reduce the sparsity to $\rho^{r / 10}$, where $\rho$ is the desired network sparsity.
Within each round, the current subnetwork is first shortly trained, then pruned to the current target sparsity, and then reset to initial parameters for the next round.
Compared to the singleshot results, we observe that for the classification task, \magnitude, \snip, and \synflow are able to retrieve weak tickets of much higher sparsity. Furthermore, these three approaches are now able to recover weak tickets of moderate sparsities also for the challenging \texttt{Helix} dataset.
Overall, \synflow consistently performs best in discovering weak tickets, even recovering the extremely sparse baseline ticket for \texttt{Circle}.
We also observe that \grasp performs poor overall, noting that it is a method designed for singleshot pruning. Examining the results, we see that \grasp experiences layer collapse already in early iterations with large target sparsities.
None of the above approaches is able to discover a strong ticket.


To discover strong tickets, \cite{edgepopup} proposed \edgepopup, which 
 falls in the same category of multishot pruning approaches.
\edgepopup assigns each model parameter a score value, which is then actively trained for several rounds while freezing all original parameters, requiring a similar computational effort as multishot pruning.
Training \edgepopup for $10$ rounds, we observe that it discovers a strong ticket of sparsity $0.5$ for \texttt{Circle} however fails to discover tickets of different sparsity, which is in line with their original results \cite{edgepopup}.
Similar to other algorithms, we observe layer collapse.
We can extend their original algorithm using annealing as in multishot pruning by slowly decreasing the  sparsity in every round, which increases performance allowing it to discover a subnetwork with reasonable accuracy at $0.1$ sparsity.
Interestingly, \edgepopup is not able to find any good subnetwork for the regression tasks.


\textbf{\grasp with local sparsity constraints}
As observed before, \grasp seems unsuitable for multishot pruning due to early layer collapse.
Several works \citep{earlybird,synflow} considered local sparsity constraints, having a target sparsity for each layer, or even channel.
These however impose unrealistic architecture constraints as layer sparsity is usually imbalanced \citep{synflow}, which also holds true for our \texttt{Circle} and \texttt{Helix} benchmarks.
With the goal to avoid layer collapse, we still equipped \grasp with local sparsity constraints per layer (see Supp. B).
Yet, the flow through the layers stays interrupted.
A possible explanation is that \grasp incorporates information about weight couplings via the hessian in its pruning strategy, which makes it more sensitive to removing individual connections from the mask as it happens during iterative pruning.

\textbf{Comparison to results on image data}
\input{figs/vgg_cifar_fig}
Our results are coherent with the reported relative trends on image tasks both for strong and weak tickets  \citep{edgepopup, synflow}.
We further reproduced results for VGG16 on CIFAR10 with non-zero bias initialization (Fig. \ref{fig:vgg_edgepopup} left).
In particular, for strong tickets we observe the same trends for \edgepopup spiking at 0.5 sparsity, performing less well at sparsity 0.1, and not recovering tickets for higher and lower sparsity. Again consistent with previously reported results, other methods are neither suited nor designed to find strong tickets.
For weak tickets, \synflow  and \magnitude are on part, which is different than originally reported in \cite{synflow}, as we allow both \snip and \magnitude to learn in a multishot fashion. This approach closely resembles the one of iterative magnitude pruning, where we do not observe layer collapse for these methods.

\textbf{VGG with strong tickets}
Finally, to put the hypothesis to a test that \edgepopup is limited to discover strong tickets of suboptimal sparsity of around $0.5$, we investigate its capabilities to recover a planted baseline ticket from VGG16. For that, we use \synflow to discover a weak ticket of sparsity $0.01$ from VGG16 with multishot pruning, trained the weak ticket, and planted it back into the network.
Running \edgepopup on this network, we observe that it indeed cannot retrieve the baseline ticket of desired sparsity in this real world setting (see Fig. \ref{fig:vgg_edgepopup} right).

%% file: figs/singleshot.tex
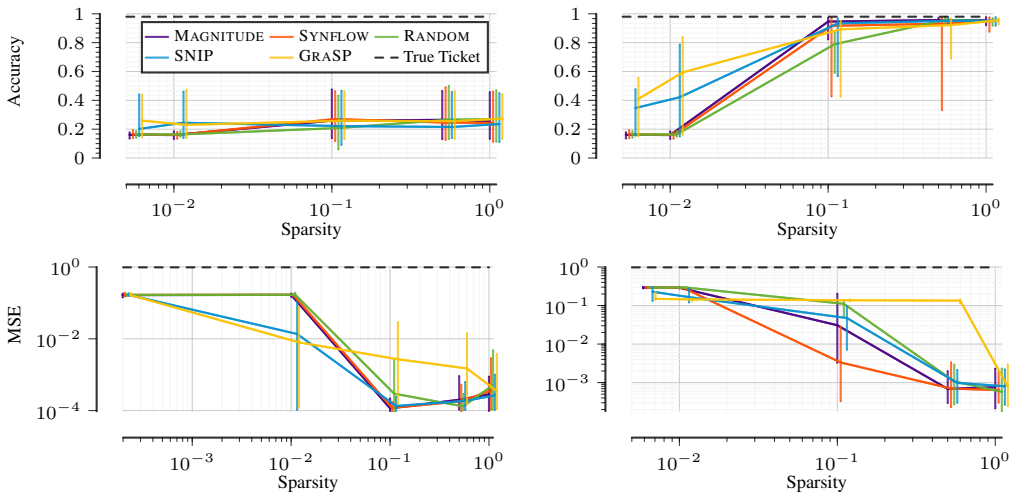
\begin{figure}
\centering
    \begin{subfigure}[t]{0.49\textwidth}
    \centering
		\ifpdf
		\begin{tikzpicture}
		\begin{axis}[
		jonas line,
		xmode=log,
		width = 6.5cm,
		height = 3.5cm,
		xlabel = {Sparsity}, 
		ylabel= {Accuracy},
		xmin=0.005, xmax=1.1,
		ymin=0, ymax=1,
		legend columns=3,
		x label style 		= {at={(axis description cs:0.5,-0.1)}, anchor=north, font=\scriptsize},
		y label style 		= {at={(axis description cs:0.0,0.6)},  anchor=south, font=\scriptsize},
		legend style={nodes={scale=0.9, transform shape}, at={(0.98,0.95)}, anchor=north east, row sep=-1.4pt, font=\tiny}
		]
		
		\addplot+[forget plot, eda errorbarcolored, y dir=plus, y explicit]
		table[x=magSparsity, y=magMeanPostprune, y error=magErrorPlusPostprune] {expres/grid/rescircleGridSingleshotDepth5.tsv};
		\addplot+[eda errorbarcolored, y dir=minus, y explicit]
		table[x=magSparsity, y=magMeanPostprune, y error=magErrorMinusPostprune] {expres/grid/rescircleGridSingleshotDepth5.tsv};
		\addlegendentry{\magnitude}
		
		\addplot+[forget plot, eda errorbarcolored, y dir=plus, y explicit]
		table[x expr=\thisrow{magSparsity}*1.05, y=synflowMeanPostprune, y error=synflowErrorPlusPostprune] {expres/grid/rescircleGridSingleshotDepth5.tsv};
		\addplot+[eda errorbarcolored, y dir=minus, y explicit]
		table[x expr=\thisrow{magSparsity}*1.05, y=synflowMeanPostprune, y error=synflowErrorMinusPostprune] {expres/grid/rescircleGridSingleshotDepth5.tsv};
		\addlegendentry{\synflow}
		
		\addplot+[forget plot, eda errorbarcolored, y dir=plus, y explicit]
		table[x expr=\thisrow{magSparsity}*1.1, y=randMeanPostprune, y error=randErrorPlusPostprune] {expres/grid/rescircleGridSingleshotDepth5.tsv};
		\addplot+[eda errorbarcolored, y dir=minus, y explicit]
		table[x expr=\thisrow{magSparsity}*1.1, y=randMeanPostprune, y error=randErrorMinusPostprune] {expres/grid/rescircleGridSingleshotDepth5.tsv};
		\addlegendentry{\rand}
		
		\addplot+[forget plot, eda errorbarcolored, y dir=plus, y explicit]
		table[x expr=\thisrow{magSparsity}*1.15, y=snipMeanPostprune, y error=snipErrorPlusPostprune] {expres/grid/rescircleGridSingleshotDepth5.tsv};
		\addplot+[eda errorbarcolored, y dir=minus, y explicit]
		table[x expr=\thisrow{magSparsity}*1.15, y=snipMeanPostprune, y error=snipErrorMinusPostprune] {expres/grid/rescircleGridSingleshotDepth5.tsv};
		\addlegendentry{\snip}
		
		\addplot+[forget plot, eda errorbarcolored, y dir=plus, y explicit]
		table[x expr=\thisrow{magSparsity}*1.2, y=graspMeanPostprune, y error=graspErrorPlusPostprune] {expres/grid/rescircleGridSingleshotDepth5.tsv};
		\addplot+[eda errorbarcolored, y dir=minus, y explicit]
		table[x expr=\thisrow{magSparsity}*1.2, y=graspMeanPostprune, y error=graspErrorMinusPostprune] {expres/grid/rescircleGridSingleshotDepth5.tsv};
		\addlegendentry{\grasp}
		
        \addplot[mark=none, dashed, black, samples=2] coordinates {(0.005,0.98) (1.0,0.98)};
        \addlegendentry{True Ticket}
		
		\end{axis}
		\end{tikzpicture}
		\fi
    \end{subfigure}
    \begin{subfigure}[t]{0.49\textwidth}
    \centering
		\ifpdf
		\begin{tikzpicture}
		\begin{axis}[
		jonas line,
		xmode=log,
		width = 6.5cm,
		height = 3.5cm,
		xlabel = {Sparsity}, 
		ylabel= {},
		xmin=0.005, xmax=1.1,
		ymin=0, ymax=1,
		legend columns=2,
		x label style 		= {at={(axis description cs:0.5,-0.1)}, anchor=north, font=\scriptsize},
		y label style 		= {at={(axis description cs:0.0,0.6)},  anchor=south, font=\scriptsize},
		legend style={nodes={scale=0.9, transform shape}, at={(0.98,0.95)}, anchor=north east, row sep=-1.4pt, font=\tiny}
		]
		
		\addplot+[forget plot, eda errorbarcolored, y dir=plus, y explicit]
		table[x=magSparsity, y=magMeanPosttrain, y error=magErrorPlusPosttrain] {expres/grid/rescircleGridSingleshotDepth5.tsv};
		\addplot+[eda errorbarcolored, y dir=minus, y explicit]
		table[x=magSparsity, y=magMeanPosttrain, y error=magErrorMinusPosttrain] {expres/grid/rescircleGridSingleshotDepth5.tsv};
		
		\addplot+[forget plot, eda errorbarcolored, y dir=plus, y explicit]
		table[x expr=\thisrow{magSparsity}*1.05, y=synflowMeanPosttrain, y error=synflowErrorPlusPosttrain] {expres/grid/rescircleGridSingleshotDepth5.tsv};
		\addplot+[eda errorbarcolored, y dir=minus, y explicit]
		table[x expr=\thisrow{magSparsity}*1.05, y=synflowMeanPosttrain, y error=synflowErrorMinusPosttrain] {expres/grid/rescircleGridSingleshotDepth5.tsv};
		
		\addplot+[forget plot, eda errorbarcolored, y dir=plus, y explicit]
		table[x expr=\thisrow{magSparsity}*1.1, y=randMeanPosttrain, y error=randErrorPlusPosttrain] {expres/grid/rescircleGridSingleshotDepth5.tsv};
		\addplot+[eda errorbarcolored, y dir=minus, y explicit]
		table[x expr=\thisrow{magSparsity}*1.1, y=randMeanPosttrain, y error=randErrorMinusPosttrain] {expres/grid/rescircleGridSingleshotDepth5.tsv};
		
		\addplot+[forget plot, eda errorbarcolored, y dir=plus, y explicit]
		table[x expr=\thisrow{magSparsity}*1.15, y=snipMeanPosttrain, y error=snipErrorPlusPosttrain] {expres/grid/rescircleGridSingleshotDepth5.tsv};
		\addplot+[eda errorbarcolored, y dir=minus, y explicit]
		table[x expr=\thisrow{magSparsity}*1.15, y=snipMeanPosttrain, y error=snipErrorMinusPosttrain] {expres/grid/rescircleGridSingleshotDepth5.tsv};
		
		\addplot+[forget plot, eda errorbarcolored, y dir=plus, y explicit]
		table[x expr=\thisrow{magSparsity}*1.2, y=graspMeanPosttrain, y error=graspErrorPlusPosttrain] {expres/grid/rescircleGridSingleshotDepth5.tsv};
		\addplot+[eda errorbarcolored, y dir=minus, y explicit]
		table[x expr=\thisrow{magSparsity}*1.2, y=graspMeanPosttrain, y error=graspErrorMinusPosttrain] {expres/grid/rescircleGridSingleshotDepth5.tsv};
		
        \addplot[mark=none, dashed, black, samples=2] coordinates {(0.005,0.98) (1.0,0.98)};
		
		\end{axis}
		\end{tikzpicture}
		\fi
    \end{subfigure}
    \begin{subfigure}[t]{0.49\textwidth}
    \centering
		\ifpdf
		\begin{tikzpicture}
		\begin{axis}[
		jonas line,
		xmode=log,
		ymode=log,
		width = 6.5cm,
		height = 3.5cm,
		xlabel = {Sparsity}, 
		ylabel= {MSE},
		xmin=0.0002, xmax=1.1,
		ymax=1,
		legend columns=2,
		x label style 		= {at={(axis description cs:0.5,-0.1)}, anchor=north, font=\scriptsize},
		y label style 		= {at={(axis description cs:0.0,0.6)},  anchor=south, font=\scriptsize},
		legend style={nodes={scale=0.9, transform shape}, at={(0.98,0.95)}, anchor=north east, row sep=-1.4pt, font=\tiny}
		]
		
		\addplot+[forget plot, eda errorbarcolored, y dir=plus, y explicit]
		table[x=magSparsity, y=magMeanPosttrain, y error=magErrorPlusPosttrain] {expres/grid/resreluGridSingleshotDepth5.tsv};
		\addplot+[eda errorbarcolored, y dir=minus, y explicit]
		table[x=magSparsity, y=magMeanPosttrain, y error=magErrorMinusPosttrain] {expres/grid/resreluGridSingleshotDepth5.tsv};
		
		\addplot+[forget plot, eda errorbarcolored, y dir=plus, y explicit]
		table[x expr=\thisrow{magSparsity}*1.05, y=synflowMeanPosttrain, y error=synflowErrorPlusPosttrain] {expres/grid/resreluGridSingleshotDepth5.tsv};
		\addplot+[eda errorbarcolored, y dir=minus, y explicit]
		table[x expr=\thisrow{magSparsity}*1.05, y=synflowMeanPosttrain, y error=synflowErrorMinusPosttrain] {expres/grid/resreluGridSingleshotDepth5.tsv};
		
		\addplot+[forget plot, eda errorbarcolored, y dir=plus, y explicit]
		table[x expr=\thisrow{magSparsity}*1.1, y=randMeanPosttrain, y error=randErrorPlusPosttrain] {expres/grid/resreluGridSingleshotDepth5.tsv};
		\addplot+[eda errorbarcolored, y dir=minus, y explicit]
		table[x expr=\thisrow{magSparsity}*1.1, y=randMeanPosttrain, y error=randErrorMinusPosttrain] {expres/grid/resreluGridSingleshotDepth5.tsv};
		
		\addplot+[forget plot, eda errorbarcolored, y dir=plus, y explicit]
		table[x expr=\thisrow{magSparsity}*1.15, y=snipMeanPosttrain, y error=snipErrorPlusPosttrain] {expres/grid/resreluGridSingleshotDepth5.tsv};
		\addplot+[eda errorbarcolored, y dir=minus, y explicit]
		table[x expr=\thisrow{magSparsity}*1.15, y=snipMeanPosttrain, y error=snipErrorMinusPosttrain] {expres/grid/resreluGridSingleshotDepth5.tsv};
		
		\addplot+[forget plot, eda errorbarcolored, y dir=plus, y explicit]
		table[x expr=\thisrow{magSparsity}*1.2, y=graspMeanPosttrain, y error=graspErrorPlusPosttrain] {expres/grid/resreluGridSingleshotDepth5.tsv};
		\addplot+[eda errorbarcolored, y dir=minus, y explicit]
		table[x expr=\thisrow{magSparsity}*1.2, y=graspMeanPosttrain, y error=graspErrorMinusPosttrain] {expres/grid/resreluGridSingleshotDepth5.tsv};
		
        \addplot[mark=none, dashed, black, samples=2] coordinates {(0.0002,0.98) (1.0,0.98)};
		
		\end{axis}
		\end{tikzpicture}
		\fi
    \end{subfigure}
    \begin{subfigure}[t]{0.49\textwidth}
    \centering
		\ifpdf
		\begin{tikzpicture}
		\begin{axis}[
		jonas line,
		xmode=log,
		ymode=log,
		minor x tick num = 5,
		width = 6.5cm,
		height = 3.5cm,
		xlabel = {Sparsity}, 
		ylabel= {},
		xmin=0.005, xmax=1.1,
		ymax=1,
		legend columns=2,
		x label style 		= {at={(axis description cs:0.5,-0.1)}, anchor=north, font=\scriptsize},
		y label style 		= {at={(axis description cs:0.0,0.6)},  anchor=south, font=\scriptsize},
		legend style={nodes={scale=0.9, transform shape}, at={(0.98,0.95)}, anchor=north east, row sep=-1.4pt, font=\tiny}
		]
		
		\addplot+[forget plot, eda errorbarcolored, y dir=plus, y explicit]
		table[x=magSparsity, y=magMeanPosttrain, y error=magErrorPlusPosttrain] {expres/grid/reshelixGridSingleshotDepth5.tsv};
		\addplot+[eda errorbarcolored, y dir=minus, y explicit]
		table[x=magSparsity, y=magMeanPosttrain, y error=magErrorMinusPosttrain] {expres/grid/reshelixGridSingleshotDepth5.tsv};
		
		\addplot+[forget plot, eda errorbarcolored, y dir=plus, y explicit]
		table[x expr=\thisrow{magSparsity}*1.05, y=synflowMeanPosttrain, y error=synflowErrorPlusPosttrain] {expres/grid/reshelixGridSingleshotDepth5.tsv};
		\addplot+[eda errorbarcolored, y dir=minus, y explicit]
		table[x expr=\thisrow{magSparsity}*1.05, y=synflowMeanPosttrain, y error=synflowErrorMinusPosttrain] {expres/grid/reshelixGridSingleshotDepth5.tsv};
		
		\addplot+[forget plot, eda errorbarcolored, y dir=plus, y explicit]
		table[x expr=\thisrow{magSparsity}*1.1, y=randMeanPosttrain, y error=randErrorPlusPosttrain] {expres/grid/reshelixGridSingleshotDepth5.tsv};
		\addplot+[eda errorbarcolored, y dir=minus, y explicit]
		table[x expr=\thisrow{magSparsity}*1.1, y=randMeanPosttrain, y error=randErrorMinusPosttrain] {expres/grid/reshelixGridSingleshotDepth5.tsv};
		
		\addplot+[forget plot, eda errorbarcolored, y dir=plus, y explicit]
		table[x expr=\thisrow{magSparsity}*1.15, y=snipMeanPosttrain, y error=snipErrorPlusPosttrain] {expres/grid/reshelixGridSingleshotDepth5.tsv};
		\addplot+[eda errorbarcolored, y dir=minus, y explicit]
		table[x expr=\thisrow{magSparsity}*1.15, y=snipMeanPosttrain, y error=snipErrorMinusPosttrain] {expres/grid/reshelixGridSingleshotDepth5.tsv};
		
		\addplot+[forget plot, eda errorbarcolored, y dir=plus, y explicit]
		table[x expr=\thisrow{magSparsity}*1.2, y=graspMeanPosttrain, y error=graspErrorPlusPosttrain] {expres/grid/reshelixGridSingleshotDepth5.tsv};
		\addplot+[eda errorbarcolored, y dir=minus, y explicit]
		table[x expr=\thisrow{magSparsity}*1.2, y=graspMeanPosttrain, y error=graspErrorMinusPosttrain] {expres/grid/reshelixGridSingleshotDepth5.tsv};
		
        \addplot[mark=none, dashed, black, samples=2] coordinates {(0.005,0.98) (1.0,0.98)};
		
		\end{axis}
		\end{tikzpicture}
		\fi
    \end{subfigure}
    \caption{\textit{Singleshot results.} Performance of discovered tickets against target sparsities as mean and value ranges (minimum and maximum) across $25$ runs. In order of appearance: \texttt{Circle} after pruning (strong ticket), \texttt{Circle}, \texttt{ReLU}, and \texttt{Helix} after training (weak ticket). Results after pruning look similar across tasks. Baseline ticket (leftmost sparsity) is given by black dashed line.} \label{fig:singleshot}
\end{figure}

%% file: figs/results_singleshot_5_circle_postprune.tikz
\begin{tikzpicture}[x=1pt,y=1pt,baseline={(0,0.4)}]
\definecolor{fillColor}{RGB}{255,255,255}
\begin{scope}
\definecolor{drawColor}{RGB}{218,165,32}

\path[draw=drawColor,line width= 0.6pt,line join=round] ( 51.33, 38.77) --
	( 51.41, 38.77);

\path[draw=drawColor,line width= 0.6pt,line join=round] ( 51.37, 38.77) --
	( 51.37, 38.40);

\path[draw=drawColor,line width= 0.6pt,line join=round] ( 51.33, 38.40) --
	( 51.41, 38.40);

\path[draw=drawColor,line width= 0.6pt,line join=round] ( 89.94, 44.28) --
	( 90.02, 44.28);

\path[draw=drawColor,line width= 0.6pt,line join=round] ( 89.98, 44.28) --
	( 89.98, 40.81);

\path[draw=drawColor,line width= 0.6pt,line join=round] ( 89.94, 40.81) --
	( 90.02, 40.81);

\path[draw=drawColor,line width= 0.6pt,line join=round] (116.93, 44.42) --
	(117.01, 44.42);

\path[draw=drawColor,line width= 0.6pt,line join=round] (116.97, 44.42) --
	(116.97, 41.13);

\path[draw=drawColor,line width= 0.6pt,line join=round] (116.93, 41.13) --
	(117.01, 41.13);

\path[draw=drawColor,line width= 0.6pt,line join=round] (128.55, 44.14) --
	(128.63, 44.14);

\path[draw=drawColor,line width= 0.6pt,line join=round] (128.59, 44.14) --
	(128.59, 40.64);

\path[draw=drawColor,line width= 0.6pt,line join=round] (128.55, 40.64) --
	(128.63, 40.64);

\path[draw=drawColor,line width= 0.6pt,line join=round] ( 40.50, 38.84) --
	( 40.58, 38.84);

\path[draw=drawColor,line width= 0.6pt,line join=round] ( 40.54, 38.84) --
	( 40.54, 38.46);

\path[draw=drawColor,line width= 0.6pt,line join=round] ( 40.50, 38.46) --
	( 40.58, 38.46);
\definecolor{drawColor}{RGB}{0,0,139}

\path[draw=drawColor,line width= 0.6pt,line join=round] ( 51.41, 38.91) --
	( 51.48, 38.91);

\path[draw=drawColor,line width= 0.6pt,line join=round] ( 51.44, 38.91) --
	( 51.44, 38.58);

\path[draw=drawColor,line width= 0.6pt,line join=round] ( 51.41, 38.58) --
	( 51.48, 38.58);

\path[draw=drawColor,line width= 0.6pt,line join=round] ( 90.02, 42.05) --
	( 90.09, 42.05);

\path[draw=drawColor,line width= 0.6pt,line join=round] ( 90.06, 42.05) --
	( 90.06, 38.98);

\path[draw=drawColor,line width= 0.6pt,line join=round] ( 90.02, 38.98) --
	( 90.09, 38.98);

\path[draw=drawColor,line width= 0.6pt,line join=round] (117.01, 44.64) --
	(117.08, 44.64);

\path[draw=drawColor,line width= 0.6pt,line join=round] (117.04, 44.64) --
	(117.04, 40.98);

\path[draw=drawColor,line width= 0.6pt,line join=round] (117.01, 40.98) --
	(117.08, 40.98);

\path[draw=drawColor,line width= 0.6pt,line join=round] (128.63, 44.82) --
	(128.71, 44.82);

\path[draw=drawColor,line width= 0.6pt,line join=round] (128.67, 44.82) --
	(128.67, 41.17);

\path[draw=drawColor,line width= 0.6pt,line join=round] (128.63, 41.17) --
	(128.71, 41.17);

\path[draw=drawColor,line width= 0.6pt,line join=round] ( 40.58, 38.91) --
	( 40.66, 38.91);

\path[draw=drawColor,line width= 0.6pt,line join=round] ( 40.62, 38.91) --
	( 40.62, 38.53);

\path[draw=drawColor,line width= 0.6pt,line join=round] ( 40.58, 38.53) --
	( 40.66, 38.53);
\definecolor{drawColor}{RGB}{255,0,255}

\path[draw=drawColor,line width= 0.6pt,line join=round] ( 51.48, 38.86) --
	( 51.56, 38.86);

\path[draw=drawColor,line width= 0.6pt,line join=round] ( 51.52, 38.86) --
	( 51.52, 38.43);

\path[draw=drawColor,line width= 0.6pt,line join=round] ( 51.48, 38.43) --
	( 51.56, 38.43);

\path[draw=drawColor,line width= 0.6pt,line join=round] ( 90.09, 44.62) --
	( 90.17, 44.62);

\path[draw=drawColor,line width= 0.6pt,line join=round] ( 90.13, 44.62) --
	( 90.13, 41.38);

\path[draw=drawColor,line width= 0.6pt,line join=round] ( 90.09, 41.38) --
	( 90.17, 41.38);

\path[draw=drawColor,line width= 0.6pt,line join=round] (117.08, 43.69) --
	(117.16, 43.69);

\path[draw=drawColor,line width= 0.6pt,line join=round] (117.12, 43.69) --
	(117.12, 40.33);

\path[draw=drawColor,line width= 0.6pt,line join=round] (117.08, 40.33) --
	(117.16, 40.33);

\path[draw=drawColor,line width= 0.6pt,line join=round] (128.71, 43.65) --
	(128.78, 43.65);

\path[draw=drawColor,line width= 0.6pt,line join=round] (128.74, 43.65) --
	(128.74, 40.19);

\path[draw=drawColor,line width= 0.6pt,line join=round] (128.71, 40.19) --
	(128.78, 40.19);

\path[draw=drawColor,line width= 0.6pt,line join=round] ( 40.66, 39.06) --
	( 40.73, 39.06);

\path[draw=drawColor,line width= 0.6pt,line join=round] ( 40.69, 39.06) --
	( 40.69, 38.57);

\path[draw=drawColor,line width= 0.6pt,line join=round] ( 40.66, 38.57) --
	( 40.73, 38.57);
\definecolor{drawColor}{RGB}{178,34,34}

\path[draw=drawColor,line width= 0.6pt,line join=round] ( 51.56, 43.22) --
	( 51.64, 43.22);

\path[draw=drawColor,line width= 0.6pt,line join=round] ( 51.60, 43.22) --
	( 51.60, 39.58);

\path[draw=drawColor,line width= 0.6pt,line join=round] ( 51.56, 39.58) --
	( 51.64, 39.58);

\path[draw=drawColor,line width= 0.6pt,line join=round] ( 90.17, 44.51) --
	( 90.25, 44.51);

\path[draw=drawColor,line width= 0.6pt,line join=round] ( 90.21, 44.51) --
	( 90.21, 40.57);

\path[draw=drawColor,line width= 0.6pt,line join=round] ( 90.17, 40.57) --
	( 90.25, 40.57);

\path[draw=drawColor,line width= 0.6pt,line join=round] (117.16, 44.25) --
	(117.24, 44.25);

\path[draw=drawColor,line width= 0.6pt,line join=round] (117.20, 44.25) --
	(117.20, 40.55);

\path[draw=drawColor,line width= 0.6pt,line join=round] (117.16, 40.55) --
	(117.24, 40.55);

\path[draw=drawColor,line width= 0.6pt,line join=round] (128.78, 44.66) --
	(128.86, 44.66);

\path[draw=drawColor,line width= 0.6pt,line join=round] (128.82, 44.66) --
	(128.82, 41.59);

\path[draw=drawColor,line width= 0.6pt,line join=round] (128.78, 41.59) --
	(128.86, 41.59);

\path[draw=drawColor,line width= 0.6pt,line join=round] ( 40.73, 44.25) --
	( 40.81, 44.25);

\path[draw=drawColor,line width= 0.6pt,line join=round] ( 40.77, 44.25) --
	( 40.77, 40.85);

\path[draw=drawColor,line width= 0.6pt,line join=round] ( 40.73, 40.85) --
	( 40.81, 40.85);
\definecolor{drawColor}{RGB}{67,205,128}

\path[draw=drawColor,line width= 0.6pt,line join=round] ( 51.64, 43.54) --
	( 51.72, 43.54);

\path[draw=drawColor,line width= 0.6pt,line join=round] ( 51.68, 43.54) --
	( 51.68, 40.40);

\path[draw=drawColor,line width= 0.6pt,line join=round] ( 51.64, 40.40) --
	( 51.72, 40.40);

\path[draw=drawColor,line width= 0.6pt,line join=round] ( 90.25, 42.77) --
	( 90.33, 42.77);

\path[draw=drawColor,line width= 0.6pt,line join=round] ( 90.29, 42.77) --
	( 90.29, 39.31);

\path[draw=drawColor,line width= 0.6pt,line join=round] ( 90.25, 39.31) --
	( 90.33, 39.31);

\path[draw=drawColor,line width= 0.6pt,line join=round] (117.24, 42.08) --
	(117.31, 42.08);

\path[draw=drawColor,line width= 0.6pt,line join=round] (117.28, 42.08) --
	(117.28, 39.53);

\path[draw=drawColor,line width= 0.6pt,line join=round] (117.24, 39.53) --
	(117.31, 39.53);

\path[draw=drawColor,line width= 0.6pt,line join=round] (128.86, 42.98) --
	(128.94, 42.98);

\path[draw=drawColor,line width= 0.6pt,line join=round] (128.90, 42.98) --
	(128.90, 40.26);

\path[draw=drawColor,line width= 0.6pt,line join=round] (128.86, 40.26) --
	(128.94, 40.26);

\path[draw=drawColor,line width= 0.6pt,line join=round] ( 40.81, 41.52) --
	( 40.89, 41.52);

\path[draw=drawColor,line width= 0.6pt,line join=round] ( 40.85, 41.52) --
	( 40.85, 39.02);

\path[draw=drawColor,line width= 0.6pt,line join=round] ( 40.81, 39.02) --
	( 40.89, 39.02);
\definecolor{drawColor}{RGB}{218,165,32}
\definecolor{fillColor}{RGB}{218,165,32}

\path[draw=drawColor,draw opacity=0.80,line width= 0.4pt,line join=round,line cap=round,fill=fillColor,fill opacity=0.80] ( 51.52, 41.64) --
	( 54.16, 37.06) --
	( 48.88, 37.06) --
	cycle;

\path[draw=drawColor,draw opacity=0.80,line width= 0.4pt,line join=round,line cap=round,fill=fillColor,fill opacity=0.80] ( 90.13, 45.60) --
	( 92.78, 41.02) --
	( 87.49, 41.02) --
	cycle;

\path[draw=drawColor,draw opacity=0.80,line width= 0.4pt,line join=round,line cap=round,fill=fillColor,fill opacity=0.80] (117.12, 45.82) --
	(119.76, 41.25) --
	(114.48, 41.25) --
	cycle;

\path[draw=drawColor,draw opacity=0.80,line width= 0.4pt,line join=round,line cap=round,fill=fillColor,fill opacity=0.80] (128.74, 45.44) --
	(131.39, 40.86) --
	(126.10, 40.86) --
	cycle;

\path[draw=drawColor,draw opacity=0.80,line width= 0.4pt,line join=round,line cap=round,fill=fillColor,fill opacity=0.80] ( 40.69, 41.70) --
	( 43.34, 37.13) --
	( 38.05, 37.13) --
	cycle;
\definecolor{drawColor}{RGB}{0,0,139}
\definecolor{fillColor}{RGB}{0,0,139}

\path[draw=drawColor,draw opacity=0.80,line width= 0.4pt,line join=round,line cap=round,fill=fillColor,fill opacity=0.80] ( 51.52, 36.29) --
	( 53.98, 38.74) --
	( 51.52, 41.20) --
	( 49.06, 38.74) --
	cycle;

\path[draw=drawColor,draw opacity=0.80,line width= 0.4pt,line join=round,line cap=round,fill=fillColor,fill opacity=0.80] ( 90.13, 38.06) --
	( 92.59, 40.52) --
	( 90.13, 42.97) --
	( 87.67, 40.52) --
	cycle;

\path[draw=drawColor,draw opacity=0.80,line width= 0.4pt,line join=round,line cap=round,fill=fillColor,fill opacity=0.80] (117.12, 40.35) --
	(119.58, 42.81) --
	(117.12, 45.27) --
	(114.66, 42.81) --
	cycle;

\path[draw=drawColor,draw opacity=0.80,line width= 0.4pt,line join=round,line cap=round,fill=fillColor,fill opacity=0.80] (128.74, 40.53) --
	(131.20, 42.99) --
	(128.74, 45.45) --
	(126.29, 42.99) --
	cycle;

\path[draw=drawColor,draw opacity=0.80,line width= 0.4pt,line join=round,line cap=round,fill=fillColor,fill opacity=0.80] ( 40.69, 36.26) --
	( 43.15, 38.72) --
	( 40.69, 41.18) --
	( 38.23, 38.72) --
	cycle;
\definecolor{fillColor}{RGB}{255,0,255}

\path[fill=fillColor,fill opacity=0.80] ( 49.56, 36.69) --
	( 53.48, 36.69) --
	( 53.48, 40.61) --
	( 49.56, 40.61) --
	cycle;

\path[fill=fillColor,fill opacity=0.80] ( 88.17, 41.04) --
	( 92.10, 41.04) --
	( 92.10, 44.96) --
	( 88.17, 44.96) --
	cycle;

\path[fill=fillColor,fill opacity=0.80] (115.16, 40.05) --
	(119.08, 40.05) --
	(119.08, 43.97) --
	(115.16, 43.97) --
	cycle;

\path[fill=fillColor,fill opacity=0.80] (126.78, 39.96) --
	(130.71, 39.96) --
	(130.71, 43.88) --
	(126.78, 43.88) --
	cycle;

\path[fill=fillColor,fill opacity=0.80] ( 38.73, 36.85) --
	( 42.66, 36.85) --
	( 42.66, 40.78) --
	( 38.73, 40.78) --
	cycle;
\definecolor{drawColor}{RGB}{178,34,34}
\definecolor{fillColor}{RGB}{178,34,34}

\path[draw=drawColor,draw opacity=0.80,line width= 0.4pt,line join=round,line cap=round,fill=fillColor,fill opacity=0.80] ( 51.52, 38.35) --
	( 54.16, 42.92) --
	( 48.88, 42.92) --
	cycle;

\path[draw=drawColor,draw opacity=0.80,line width= 0.4pt,line join=round,line cap=round,fill=fillColor,fill opacity=0.80] ( 90.13, 39.49) --
	( 92.78, 44.07) --
	( 87.49, 44.07) --
	cycle;

\path[draw=drawColor,draw opacity=0.80,line width= 0.4pt,line join=round,line cap=round,fill=fillColor,fill opacity=0.80] (117.12, 39.35) --
	(119.76, 43.93) --
	(114.48, 43.93) --
	cycle;

\path[draw=drawColor,draw opacity=0.80,line width= 0.4pt,line join=round,line cap=round,fill=fillColor,fill opacity=0.80] (128.74, 40.07) --
	(131.39, 44.65) --
	(126.10, 44.65) --
	cycle;

\path[draw=drawColor,draw opacity=0.80,line width= 0.4pt,line join=round,line cap=round,fill=fillColor,fill opacity=0.80] ( 40.69, 39.50) --
	( 43.34, 44.08) --
	( 38.05, 44.08) --
	cycle;
\definecolor{drawColor}{RGB}{67,205,128}
\definecolor{fillColor}{RGB}{67,205,128}

\path[draw=drawColor,draw opacity=0.80,line width= 0.4pt,line join=round,line cap=round,fill=fillColor,fill opacity=0.80] ( 51.52, 39.51) --
	( 53.98, 41.97) --
	( 51.52, 44.43) --
	( 49.06, 41.97) --
	cycle;

\path[draw=drawColor,draw opacity=0.80,line width= 0.4pt,line join=round,line cap=round,fill=fillColor,fill opacity=0.80] ( 90.13, 38.58) --
	( 92.59, 41.04) --
	( 90.13, 43.50) --
	( 87.67, 41.04) --
	cycle;

\path[draw=drawColor,draw opacity=0.80,line width= 0.4pt,line join=round,line cap=round,fill=fillColor,fill opacity=0.80] (117.12, 38.35) --
	(119.58, 40.81) --
	(117.12, 43.27) --
	(114.66, 40.81) --
	cycle;

\path[draw=drawColor,draw opacity=0.80,line width= 0.4pt,line join=round,line cap=round,fill=fillColor,fill opacity=0.80] (128.74, 39.16) --
	(131.20, 41.62) --
	(128.74, 44.07) --
	(126.29, 41.62) --
	cycle;

\path[draw=drawColor,draw opacity=0.80,line width= 0.4pt,line join=round,line cap=round,fill=fillColor,fill opacity=0.80] ( 40.69, 37.81) --
	( 43.15, 40.27) --
	( 40.69, 42.73) --
	( 38.23, 40.27) --
	cycle;
\definecolor{drawColor}{RGB}{218,165,32}

\path[draw=drawColor,draw opacity=0.80,line width= 0.5pt,line join=round] ( 40.69, 38.65) --
	( 51.52, 38.59) --
	( 90.13, 42.55) --
	(117.12, 42.77) --
	(128.74, 42.39);
\definecolor{drawColor}{RGB}{0,0,139}

\path[draw=drawColor,draw opacity=0.80,line width= 0.5pt,line join=round] ( 40.69, 38.72) --
	( 51.52, 38.74) --
	( 90.13, 40.52) --
	(117.12, 42.81) --
	(128.74, 42.99);
\definecolor{drawColor}{RGB}{255,0,255}

\path[draw=drawColor,draw opacity=0.80,line width= 0.5pt,line join=round] ( 40.69, 38.81) --
	( 51.52, 38.65) --
	( 90.13, 43.00) --
	(117.12, 42.01) --
	(128.74, 41.92);
\definecolor{drawColor}{RGB}{178,34,34}

\path[draw=drawColor,draw opacity=0.80,line width= 0.5pt,line join=round] ( 40.69, 42.55) --
	( 51.52, 41.40) --
	( 90.13, 42.54) --
	(117.12, 42.40) --
	(128.74, 43.12);
\definecolor{drawColor}{RGB}{67,205,128}

\path[draw=drawColor,draw opacity=0.80,line width= 0.5pt,line join=round] ( 40.69, 40.27) --
	( 51.52, 41.97) --
	( 90.13, 41.04) --
	(117.12, 40.81) --
	(128.74, 41.62);
\definecolor{drawColor}{RGB}{0,0,0}

\path[draw=drawColor,line width= 0.5pt,line join=round] ( 40.69, 71.19) --
	( 51.52, 71.19) --
	( 90.13, 71.21) --
	(117.12, 71.18) --
	(128.74, 71.19);

\path[draw=drawColor,line width= 0.5pt,line join=round] ( 40.69, 71.19) --
	( 51.52, 71.19) --
	( 90.13, 71.19) --
	(117.12, 71.16) --
	(128.74, 71.24);

\path[draw=drawColor,line width= 0.5pt,line join=round] ( 40.69, 71.17) --
	( 51.52, 71.17) --
	( 90.13, 71.22) --
	(117.12, 71.20) --
	(128.74, 71.26);

\path[draw=drawColor,line width= 0.5pt,line join=round] ( 40.69, 71.22) --
	( 51.52, 71.18) --
	( 90.13, 71.17) --
	(117.12, 71.17) --
	(128.74, 71.20);

\path[draw=drawColor,line width= 0.5pt,line join=round] ( 40.69, 71.24) --
	( 51.52, 71.15) --
	( 90.13, 71.17) --
	(117.12, 71.16) --
	(128.74, 71.20);
\definecolor{drawColor}{gray}{0.60}

\path[draw=drawColor,line width= 0.5pt,line join=round,line cap=round] ( 36.16, 30.28) -- ( 36.16, 27.43);

\path[draw=drawColor,line width= 0.5pt,line join=round,line cap=round] ( 39.90, 30.28) -- ( 39.90, 27.43);

\path[draw=drawColor,line width= 0.5pt,line join=round,line cap=round] ( 42.96, 30.28) -- ( 42.96, 27.43);

\path[draw=drawColor,line width= 0.5pt,line join=round,line cap=round] ( 45.54, 30.28) -- ( 45.54, 27.43);

\path[draw=drawColor,line width= 0.5pt,line join=round,line cap=round] ( 47.78, 30.28) -- ( 47.78, 27.43);

\path[draw=drawColor,line width= 0.5pt,line join=round,line cap=round] ( 49.76, 30.28) -- ( 49.76, 27.43);

\path[draw=drawColor,line width= 0.5pt,line join=round,line cap=round] ( 51.52, 30.28) -- ( 51.52, 24.59);

\path[draw=drawColor,line width= 0.5pt,line join=round,line cap=round] ( 63.15, 30.28) -- ( 63.15, 27.43);

\path[draw=drawColor,line width= 0.5pt,line join=round,line cap=round] ( 69.94, 30.28) -- ( 69.94, 27.43);

\path[draw=drawColor,line width= 0.5pt,line join=round,line cap=round] ( 74.77, 30.28) -- ( 74.77, 27.43);

\path[draw=drawColor,line width= 0.5pt,line join=round,line cap=round] ( 78.51, 30.28) -- ( 78.51, 27.43);

\path[draw=drawColor,line width= 0.5pt,line join=round,line cap=round] ( 81.57, 30.28) -- ( 81.57, 27.43);

\path[draw=drawColor,line width= 0.5pt,line join=round,line cap=round] ( 84.15, 30.28) -- ( 84.15, 27.43);

\path[draw=drawColor,line width= 0.5pt,line join=round,line cap=round] ( 86.39, 30.28) -- ( 86.39, 27.43);

\path[draw=drawColor,line width= 0.5pt,line join=round,line cap=round] ( 88.37, 30.28) -- ( 88.37, 27.43);

\path[draw=drawColor,line width= 0.5pt,line join=round,line cap=round] ( 90.13, 30.28) -- ( 90.13, 24.59);

\path[draw=drawColor,line width= 0.5pt,line join=round,line cap=round] (101.76, 30.28) -- (101.76, 27.43);

\path[draw=drawColor,line width= 0.5pt,line join=round,line cap=round] (108.56, 30.28) -- (108.56, 27.43);

\path[draw=drawColor,line width= 0.5pt,line join=round,line cap=round] (113.38, 30.28) -- (113.38, 27.43);

\path[draw=drawColor,line width= 0.5pt,line join=round,line cap=round] (117.12, 30.28) -- (117.12, 27.43);

\path[draw=drawColor,line width= 0.5pt,line join=round,line cap=round] (120.18, 30.28) -- (120.18, 27.43);

\path[draw=drawColor,line width= 0.5pt,line join=round,line cap=round] (122.76, 30.28) -- (122.76, 27.43);

\path[draw=drawColor,line width= 0.5pt,line join=round,line cap=round] (125.00, 30.28) -- (125.00, 27.43);

\path[draw=drawColor,line width= 0.5pt,line join=round,line cap=round] (126.98, 30.28) -- (126.98, 27.43);

\path[draw=drawColor,line width= 0.5pt,line join=round,line cap=round] (128.74, 30.28) -- (128.74, 24.59);
\end{scope}
\begin{scope}
\definecolor{drawColor}{gray}{0.30}

\node[text=drawColor,anchor=base east,inner sep=0pt, outer sep=0pt, scale=  0.80] at ( 31.13, 29.51) {0.00};

\node[text=drawColor,anchor=base east,inner sep=0pt, outer sep=0pt, scale=  0.80] at ( 31.13, 39.43) {0.25};

\node[text=drawColor,anchor=base east,inner sep=0pt, outer sep=0pt, scale=  0.80] at ( 31.13, 49.36) {0.50};

\node[text=drawColor,anchor=base east,inner sep=0pt, outer sep=0pt, scale=  0.80] at ( 31.13, 59.28) {0.75};

\node[text=drawColor,anchor=base east,inner sep=0pt, outer sep=0pt, scale=  0.80] at ( 31.13, 69.21) {1.00};
\end{scope}
\begin{scope}
\definecolor{drawColor}{gray}{0.60}

\path[draw=drawColor,line width= 0.6pt,line join=round] ( 33.33, 32.26) --
	( 36.08, 32.26);

\path[draw=drawColor,line width= 0.6pt,line join=round] ( 33.33, 42.19) --
	( 36.08, 42.19);

\path[draw=drawColor,line width= 0.6pt,line join=round] ( 33.33, 52.11) --
	( 36.08, 52.11);

\path[draw=drawColor,line width= 0.6pt,line join=round] ( 33.33, 62.04) --
	( 36.08, 62.04);

\path[draw=drawColor,line width= 0.6pt,line join=round] ( 33.33, 71.96) --
	( 36.08, 71.96);
\end{scope}
\begin{scope}
\definecolor{drawColor}{gray}{0.60}

\path[draw=drawColor,line width= 0.6pt,line join=round] ( 51.52, 27.53) --
	( 51.52, 30.28);

\path[draw=drawColor,line width= 0.6pt,line join=round] ( 90.13, 27.53) --
	( 90.13, 30.28);

\path[draw=drawColor,line width= 0.6pt,line join=round] (128.74, 27.53) --
	(128.74, 30.28);
\end{scope}
\begin{scope}
\definecolor{drawColor}{gray}{0.30}

\node[text=drawColor,anchor=base west,inner sep=0pt, outer sep=0pt, scale=  0.80] at ( 45.19, 16.13) {10};

\node[text=drawColor,anchor=base west,inner sep=0pt, outer sep=0pt, scale=  0.56] at ( 53.19, 19.40) {-2};

\node[text=drawColor,anchor=base west,inner sep=0pt, outer sep=0pt, scale=  0.80] at ( 83.80, 16.13) {10};

\node[text=drawColor,anchor=base west,inner sep=0pt, outer sep=0pt, scale=  0.56] at ( 91.80, 19.40) {-1};

\node[text=drawColor,anchor=base west,inner sep=0pt, outer sep=0pt, scale=  0.80] at (123.35, 16.13) {10};

\node[text=drawColor,anchor=base west,inner sep=0pt, outer sep=0pt, scale=  0.56] at (131.34, 19.40) {0};
\end{scope}
\begin{scope}
\definecolor{drawColor}{RGB}{0,0,0}

\node[text=drawColor,anchor=base,inner sep=0pt, outer sep=0pt, scale=  0.80] at ( 84.72,  4.40) {Sparsity};
\end{scope}
\begin{scope}
\definecolor{drawColor}{RGB}{0,0,0}

\node[text=drawColor,rotate= 90.00,anchor=base,inner sep=0pt, outer sep=0pt, scale=  0.80] at (  8.36, 52.11) {Accuracy};
\end{scope}
\begin{scope}
\definecolor{drawColor}{RGB}{0,0,0}

\node[text=drawColor,anchor=base west,inner sep=0pt, outer sep=0pt, scale=  0.80] at (149.86, 66.81) {Method};
\end{scope}
\begin{scope}
\definecolor{drawColor}{RGB}{218,165,32}

\path[draw=drawColor,line width= 0.6pt,line join=round] (150.46, 58.94) -- (155.28, 58.94);
\end{scope}
\begin{scope}
\definecolor{drawColor}{RGB}{218,165,32}
\definecolor{fillColor}{RGB}{218,165,32}

\path[draw=drawColor,draw opacity=0.80,line width= 0.4pt,line join=round,line cap=round,fill=fillColor,fill opacity=0.80] (152.87, 62.00) --
	(155.51, 57.42) --
	(150.23, 57.42) --
	cycle;
\end{scope}
\begin{scope}
\definecolor{drawColor}{RGB}{218,165,32}

\path[draw=drawColor,draw opacity=0.80,line width= 0.5pt,line join=round] (150.46, 58.94) -- (155.28, 58.94);
\end{scope}
\begin{scope}
\definecolor{drawColor}{RGB}{0,0,139}

\path[draw=drawColor,line width= 0.6pt,line join=round] (150.46, 52.76) -- (155.28, 52.76);
\end{scope}
\begin{scope}
\definecolor{drawColor}{RGB}{0,0,139}
\definecolor{fillColor}{RGB}{0,0,139}

\path[draw=drawColor,draw opacity=0.80,line width= 0.4pt,line join=round,line cap=round,fill=fillColor,fill opacity=0.80] (152.87, 50.30) --
	(155.33, 52.76) --
	(152.87, 55.22) --
	(150.41, 52.76) --
	cycle;
\end{scope}
\begin{scope}
\definecolor{drawColor}{RGB}{0,0,139}

\path[draw=drawColor,draw opacity=0.80,line width= 0.5pt,line join=round] (150.46, 52.76) -- (155.28, 52.76);
\end{scope}
\begin{scope}
\definecolor{drawColor}{RGB}{255,0,255}

\path[draw=drawColor,line width= 0.6pt,line join=round] (150.46, 46.58) -- (155.28, 46.58);
\end{scope}
\begin{scope}
\definecolor{fillColor}{RGB}{255,0,255}

\path[fill=fillColor,fill opacity=0.80] (150.91, 44.62) --
	(154.83, 44.62) --
	(154.83, 48.54) --
	(150.91, 48.54) --
	cycle;
\end{scope}
\begin{scope}
\definecolor{drawColor}{RGB}{255,0,255}

\path[draw=drawColor,draw opacity=0.80,line width= 0.5pt,line join=round] (150.46, 46.58) -- (155.28, 46.58);
\end{scope}
\begin{scope}
\definecolor{drawColor}{RGB}{178,34,34}

\path[draw=drawColor,line width= 0.6pt,line join=round] (150.46, 40.40) -- (155.28, 40.40);
\end{scope}
\begin{scope}
\definecolor{drawColor}{RGB}{178,34,34}
\definecolor{fillColor}{RGB}{178,34,34}

\path[draw=drawColor,draw opacity=0.80,line width= 0.4pt,line join=round,line cap=round,fill=fillColor,fill opacity=0.80] (152.87, 37.35) --
	(155.51, 41.93) --
	(150.23, 41.93) --
	cycle;
\end{scope}
\begin{scope}
\definecolor{drawColor}{RGB}{178,34,34}

\path[draw=drawColor,draw opacity=0.80,line width= 0.5pt,line join=round] (150.46, 40.40) -- (155.28, 40.40);
\end{scope}
\begin{scope}
\definecolor{drawColor}{RGB}{67,205,128}

\path[draw=drawColor,line width= 0.6pt,line join=round] (150.46, 34.22) -- (155.28, 34.22);
\end{scope}
\begin{scope}
\definecolor{drawColor}{RGB}{67,205,128}
\definecolor{fillColor}{RGB}{67,205,128}

\path[draw=drawColor,draw opacity=0.80,line width= 0.4pt,line join=round,line cap=round,fill=fillColor,fill opacity=0.80] (152.87, 31.76) --
	(155.33, 34.22) --
	(152.87, 36.68) --
	(150.41, 34.22) --
	cycle;
\end{scope}
\begin{scope}
\definecolor{drawColor}{RGB}{67,205,128}

\path[draw=drawColor,draw opacity=0.80,line width= 0.5pt,line join=round] (150.46, 34.22) -- (155.28, 34.22);
\end{scope}
\begin{scope}
\definecolor{drawColor}{RGB}{0,0,0}

\node[text=drawColor,anchor=base west,inner sep=0pt, outer sep=0pt, scale=  0.70] at (159.88, 56.53) {{ \footnotesize \textsc{Magnitude}}};
\end{scope}
\begin{scope}
\definecolor{drawColor}{RGB}{0,0,0}

\node[text=drawColor,anchor=base west,inner sep=0pt, outer sep=0pt, scale=  0.70] at (159.88, 50.35) {{ \footnotesize \textsc{Random}}};
\end{scope}
\begin{scope}
\definecolor{drawColor}{RGB}{0,0,0}

\node[text=drawColor,anchor=base west,inner sep=0pt, outer sep=0pt, scale=  0.70] at (159.88, 44.17) {{ \footnotesize \textsc{Synflow}}};
\end{scope}
\begin{scope}
\definecolor{drawColor}{RGB}{0,0,0}

\node[text=drawColor,anchor=base west,inner sep=0pt, outer sep=0pt, scale=  0.70] at (159.88, 37.99) {{ \footnotesize \textsc{GraSP}}};
\end{scope}
\begin{scope}
\definecolor{drawColor}{RGB}{0,0,0}

\node[text=drawColor,anchor=base west,inner sep=0pt, outer sep=0pt, scale=  0.70] at (159.88, 31.81) {{ \footnotesize \textsc{SNIP}}};
\end{scope}
\end{tikzpicture}

%% file: figs/results_singleshot_5_circle_posttrain.tikz
\begin{tikzpicture}[x=1pt,y=1pt,baseline={(0,0.4)}]
\definecolor{fillColor}{RGB}{255,255,255}
\begin{scope}
\definecolor{drawColor}{RGB}{218,165,32}

\path[draw=drawColor,line width= 0.6pt,line join=round] ( 61.70, 38.77) --
	( 61.83, 38.77);

\path[draw=drawColor,line width= 0.6pt,line join=round] ( 61.76, 38.77) --
	( 61.76, 38.40);

\path[draw=drawColor,line width= 0.6pt,line join=round] ( 61.70, 38.40) --
	( 61.83, 38.40);

\path[draw=drawColor,line width= 0.6pt,line join=round] (126.57, 70.28) --
	(126.70, 70.28);

\path[draw=drawColor,line width= 0.6pt,line join=round] (126.63, 70.28) --
	(126.63, 69.36);

\path[draw=drawColor,line width= 0.6pt,line join=round] (126.57, 69.36) --
	(126.70, 69.36);

\path[draw=drawColor,line width= 0.6pt,line join=round] (171.91, 70.42) --
	(172.04, 70.42);

\path[draw=drawColor,line width= 0.6pt,line join=round] (171.97, 70.42) --
	(171.97, 70.09);

\path[draw=drawColor,line width= 0.6pt,line join=round] (171.91, 70.09) --
	(172.04, 70.09);

\path[draw=drawColor,line width= 0.6pt,line join=round] (191.43, 70.40) --
	(191.56, 70.40);

\path[draw=drawColor,line width= 0.6pt,line join=round] (191.50, 70.40) --
	(191.50, 69.95);

\path[draw=drawColor,line width= 0.6pt,line join=round] (191.43, 69.95) --
	(191.56, 69.95);

\path[draw=drawColor,line width= 0.6pt,line join=round] ( 43.51, 38.84) --
	( 43.64, 38.84);

\path[draw=drawColor,line width= 0.6pt,line join=round] ( 43.57, 38.84) --
	( 43.57, 38.46);

\path[draw=drawColor,line width= 0.6pt,line join=round] ( 43.51, 38.46) --
	( 43.64, 38.46);
\definecolor{drawColor}{RGB}{0,0,139}

\path[draw=drawColor,line width= 0.6pt,line join=round] ( 61.83, 38.91) --
	( 61.96, 38.91);

\path[draw=drawColor,line width= 0.6pt,line join=round] ( 61.89, 38.91) --
	( 61.89, 38.58);

\path[draw=drawColor,line width= 0.6pt,line join=round] ( 61.83, 38.58) --
	( 61.96, 38.58);

\path[draw=drawColor,line width= 0.6pt,line join=round] (126.70, 64.66) --
	(126.83, 64.66);

\path[draw=drawColor,line width= 0.6pt,line join=round] (126.76, 64.66) --
	(126.76, 62.54);

\path[draw=drawColor,line width= 0.6pt,line join=round] (126.70, 62.54) --
	(126.83, 62.54);

\path[draw=drawColor,line width= 0.6pt,line join=round] (172.04, 70.46) --
	(172.17, 70.46);

\path[draw=drawColor,line width= 0.6pt,line join=round] (172.10, 70.46) --
	(172.10, 70.12);

\path[draw=drawColor,line width= 0.6pt,line join=round] (172.04, 70.12) --
	(172.17, 70.12);

\path[draw=drawColor,line width= 0.6pt,line join=round] (191.56, 70.16) --
	(191.69, 70.16);

\path[draw=drawColor,line width= 0.6pt,line join=round] (191.63, 70.16) --
	(191.63, 69.68);

\path[draw=drawColor,line width= 0.6pt,line join=round] (191.56, 69.68) --
	(191.69, 69.68);

\path[draw=drawColor,line width= 0.6pt,line join=round] ( 43.64, 38.91) --
	( 43.77, 38.91);

\path[draw=drawColor,line width= 0.6pt,line join=round] ( 43.70, 38.91) --
	( 43.70, 38.53);

\path[draw=drawColor,line width= 0.6pt,line join=round] ( 43.64, 38.53) --
	( 43.77, 38.53);
\definecolor{drawColor}{RGB}{255,0,255}

\path[draw=drawColor,line width= 0.6pt,line join=round] ( 61.96, 38.86) --
	( 62.09, 38.86);

\path[draw=drawColor,line width= 0.6pt,line join=round] ( 62.02, 38.86) --
	( 62.02, 38.43);

\path[draw=drawColor,line width= 0.6pt,line join=round] ( 61.96, 38.43) --
	( 62.09, 38.43);

\path[draw=drawColor,line width= 0.6pt,line join=round] (126.83, 71.05) --
	(126.96, 71.05);

\path[draw=drawColor,line width= 0.6pt,line join=round] (126.89, 71.05) --
	(126.89, 66.27);

\path[draw=drawColor,line width= 0.6pt,line join=round] (126.83, 66.27) --
	(126.96, 66.27);

\path[draw=drawColor,line width= 0.6pt,line join=round] (172.17, 71.28) --
	(172.30, 71.28);

\path[draw=drawColor,line width= 0.6pt,line join=round] (172.23, 71.28) --
	(172.23, 67.18);

\path[draw=drawColor,line width= 0.6pt,line join=round] (172.17, 67.18) --
	(172.30, 67.18);

\path[draw=drawColor,line width= 0.6pt,line join=round] (191.69, 70.43) --
	(191.82, 70.43);

\path[draw=drawColor,line width= 0.6pt,line join=round] (191.76, 70.43) --
	(191.76, 69.63);

\path[draw=drawColor,line width= 0.6pt,line join=round] (191.69, 69.63) --
	(191.82, 69.63);

\path[draw=drawColor,line width= 0.6pt,line join=round] ( 43.77, 39.06) --
	( 43.90, 39.06);

\path[draw=drawColor,line width= 0.6pt,line join=round] ( 43.83, 39.06) --
	( 43.83, 38.57);

\path[draw=drawColor,line width= 0.6pt,line join=round] ( 43.77, 38.57) --
	( 43.90, 38.57);
\definecolor{drawColor}{RGB}{178,34,34}

\path[draw=drawColor,line width= 0.6pt,line join=round] ( 62.09, 58.71) --
	( 62.22, 58.71);

\path[draw=drawColor,line width= 0.6pt,line join=round] ( 62.15, 58.71) --
	( 62.15, 52.97);

\path[draw=drawColor,line width= 0.6pt,line join=round] ( 62.09, 52.97) --
	( 62.22, 52.97);

\path[draw=drawColor,line width= 0.6pt,line join=round] (126.96, 69.44) --
	(127.09, 69.44);

\path[draw=drawColor,line width= 0.6pt,line join=round] (127.02, 69.44) --
	(127.02, 65.96);

\path[draw=drawColor,line width= 0.6pt,line join=round] (126.96, 65.96) --
	(127.09, 65.96);

\path[draw=drawColor,line width= 0.6pt,line join=round] (172.30, 69.75) --
	(172.43, 69.75);

\path[draw=drawColor,line width= 0.6pt,line join=round] (172.36, 69.75) --
	(172.36, 67.97);

\path[draw=drawColor,line width= 0.6pt,line join=round] (172.30, 67.97) --
	(172.43, 67.97);

\path[draw=drawColor,line width= 0.6pt,line join=round] (191.82, 70.37) --
	(191.95, 70.37);

\path[draw=drawColor,line width= 0.6pt,line join=round] (191.89, 70.37) --
	(191.89, 70.01);

\path[draw=drawColor,line width= 0.6pt,line join=round] (191.82, 70.01) --
	(191.95, 70.01);

\path[draw=drawColor,line width= 0.6pt,line join=round] ( 43.90, 50.26) --
	( 44.03, 50.26);

\path[draw=drawColor,line width= 0.6pt,line join=round] ( 43.96, 50.26) --
	( 43.96, 46.83);

\path[draw=drawColor,line width= 0.6pt,line join=round] ( 43.90, 46.83) --
	( 44.03, 46.83);
\definecolor{drawColor}{RGB}{67,205,128}

\path[draw=drawColor,line width= 0.6pt,line join=round] ( 62.22, 51.16) --
	( 62.35, 51.16);

\path[draw=drawColor,line width= 0.6pt,line join=round] ( 62.28, 51.16) --
	( 62.28, 46.84);

\path[draw=drawColor,line width= 0.6pt,line join=round] ( 62.22, 46.84) --
	( 62.35, 46.84);

\path[draw=drawColor,line width= 0.6pt,line join=round] (127.09, 70.60) --
	(127.21, 70.60);

\path[draw=drawColor,line width= 0.6pt,line join=round] (127.15, 70.60) --
	(127.15, 68.06);

\path[draw=drawColor,line width= 0.6pt,line join=round] (127.09, 68.06) --
	(127.21, 68.06);

\path[draw=drawColor,line width= 0.6pt,line join=round] (172.43, 70.18) --
	(172.56, 70.18);

\path[draw=drawColor,line width= 0.6pt,line join=round] (172.49, 70.18) --
	(172.49, 69.89);

\path[draw=drawColor,line width= 0.6pt,line join=round] (172.43, 69.89) --
	(172.56, 69.89);

\path[draw=drawColor,line width= 0.6pt,line join=round] (191.95, 70.35) --
	(192.08, 70.35);

\path[draw=drawColor,line width= 0.6pt,line join=round] (192.02, 70.35) --
	(192.02, 69.76);

\path[draw=drawColor,line width= 0.6pt,line join=round] (191.95, 69.76) --
	(192.08, 69.76);

\path[draw=drawColor,line width= 0.6pt,line join=round] ( 44.03, 48.10) --
	( 44.16, 48.10);

\path[draw=drawColor,line width= 0.6pt,line join=round] ( 44.09, 48.10) --
	( 44.09, 43.93);

\path[draw=drawColor,line width= 0.6pt,line join=round] ( 44.03, 43.93) --
	( 44.16, 43.93);
\definecolor{drawColor}{RGB}{218,165,32}
\definecolor{fillColor}{RGB}{218,165,32}

\path[draw=drawColor,draw opacity=0.80,line width= 0.4pt,line join=round,line cap=round,fill=fillColor,fill opacity=0.80] ( 62.02, 41.64) --
	( 64.67, 37.06) --
	( 59.38, 37.06) --
	cycle;

\path[draw=drawColor,draw opacity=0.80,line width= 0.4pt,line join=round,line cap=round,fill=fillColor,fill opacity=0.80] (126.89, 72.87) --
	(129.53, 68.29) --
	(124.25, 68.29) --
	cycle;

\path[draw=drawColor,draw opacity=0.80,line width= 0.4pt,line join=round,line cap=round,fill=fillColor,fill opacity=0.80] (172.23, 73.30) --
	(174.87, 68.73) --
	(169.59, 68.73) --
	cycle;

\path[draw=drawColor,draw opacity=0.80,line width= 0.4pt,line join=round,line cap=round,fill=fillColor,fill opacity=0.80] (191.76, 73.23) --
	(194.40, 68.65) --
	(189.12, 68.65) --
	cycle;

\path[draw=drawColor,draw opacity=0.80,line width= 0.4pt,line join=round,line cap=round,fill=fillColor,fill opacity=0.80] ( 43.83, 41.70) --
	( 46.47, 37.13) --
	( 41.19, 37.13) --
	cycle;
\definecolor{drawColor}{RGB}{0,0,139}
\definecolor{fillColor}{RGB}{0,0,139}

\path[draw=drawColor,draw opacity=0.80,line width= 0.4pt,line join=round,line cap=round,fill=fillColor,fill opacity=0.80] ( 62.02, 36.29) --
	( 64.48, 38.74) --
	( 62.02, 41.20) --
	( 59.56, 38.74) --
	cycle;

\path[draw=drawColor,draw opacity=0.80,line width= 0.4pt,line join=round,line cap=round,fill=fillColor,fill opacity=0.80] (126.89, 61.14) --
	(129.35, 63.60) --
	(126.89, 66.06) --
	(124.43, 63.60) --
	cycle;

\path[draw=drawColor,draw opacity=0.80,line width= 0.4pt,line join=round,line cap=round,fill=fillColor,fill opacity=0.80] (172.23, 67.83) --
	(174.69, 70.29) --
	(172.23, 72.75) --
	(169.77, 70.29) --
	cycle;

\path[draw=drawColor,draw opacity=0.80,line width= 0.4pt,line join=round,line cap=round,fill=fillColor,fill opacity=0.80] (191.76, 67.46) --
	(194.22, 69.92) --
	(191.76, 72.38) --
	(189.30, 69.92) --
	cycle;

\path[draw=drawColor,draw opacity=0.80,line width= 0.4pt,line join=round,line cap=round,fill=fillColor,fill opacity=0.80] ( 43.83, 36.26) --
	( 46.29, 38.72) --
	( 43.83, 41.18) --
	( 41.37, 38.72) --
	cycle;
\definecolor{fillColor}{RGB}{255,0,255}

\path[fill=fillColor,fill opacity=0.80] ( 60.06, 36.69) --
	( 63.99, 36.69) --
	( 63.99, 40.61) --
	( 60.06, 40.61) --
	cycle;

\path[fill=fillColor,fill opacity=0.80] (124.93, 66.70) --
	(128.85, 66.70) --
	(128.85, 70.62) --
	(124.93, 70.62) --
	cycle;

\path[fill=fillColor,fill opacity=0.80] (170.27, 67.27) --
	(174.19, 67.27) --
	(174.19, 71.19) --
	(170.27, 71.19) --
	cycle;

\path[fill=fillColor,fill opacity=0.80] (189.80, 68.07) --
	(193.72, 68.07) --
	(193.72, 71.99) --
	(189.80, 71.99) --
	cycle;

\path[fill=fillColor,fill opacity=0.80] ( 41.87, 36.85) --
	( 45.79, 36.85) --
	( 45.79, 40.78) --
	( 41.87, 40.78) --
	cycle;
\definecolor{drawColor}{RGB}{178,34,34}
\definecolor{fillColor}{RGB}{178,34,34}

\path[draw=drawColor,draw opacity=0.80,line width= 0.4pt,line join=round,line cap=round,fill=fillColor,fill opacity=0.80] ( 62.02, 52.79) --
	( 64.67, 57.36) --
	( 59.38, 57.36) --
	cycle;

\path[draw=drawColor,draw opacity=0.80,line width= 0.4pt,line join=round,line cap=round,fill=fillColor,fill opacity=0.80] (126.89, 64.65) --
	(129.53, 69.23) --
	(124.25, 69.23) --
	cycle;

\path[draw=drawColor,draw opacity=0.80,line width= 0.4pt,line join=round,line cap=round,fill=fillColor,fill opacity=0.80] (172.23, 65.81) --
	(174.87, 70.39) --
	(169.59, 70.39) --
	cycle;

\path[draw=drawColor,draw opacity=0.80,line width= 0.4pt,line join=round,line cap=round,fill=fillColor,fill opacity=0.80] (191.76, 67.14) --
	(194.40, 71.72) --
	(189.12, 71.72) --
	cycle;

\path[draw=drawColor,draw opacity=0.80,line width= 0.4pt,line join=round,line cap=round,fill=fillColor,fill opacity=0.80] ( 43.83, 45.49) --
	( 46.47, 50.07) --
	( 41.19, 50.07) --
	cycle;
\definecolor{drawColor}{RGB}{67,205,128}
\definecolor{fillColor}{RGB}{67,205,128}

\path[draw=drawColor,draw opacity=0.80,line width= 0.4pt,line join=round,line cap=round,fill=fillColor,fill opacity=0.80] ( 62.02, 46.54) --
	( 64.48, 49.00) --
	( 62.02, 51.46) --
	( 59.56, 49.00) --
	cycle;

\path[draw=drawColor,draw opacity=0.80,line width= 0.4pt,line join=round,line cap=round,fill=fillColor,fill opacity=0.80] (126.89, 66.87) --
	(129.35, 69.33) --
	(126.89, 71.79) --
	(124.43, 69.33) --
	cycle;

\path[draw=drawColor,draw opacity=0.80,line width= 0.4pt,line join=round,line cap=round,fill=fillColor,fill opacity=0.80] (172.23, 67.58) --
	(174.69, 70.03) --
	(172.23, 72.49) --
	(169.77, 70.03) --
	cycle;

\path[draw=drawColor,draw opacity=0.80,line width= 0.4pt,line join=round,line cap=round,fill=fillColor,fill opacity=0.80] (191.76, 67.60) --
	(194.22, 70.06) --
	(191.76, 72.52) --
	(189.30, 70.06) --
	cycle;

\path[draw=drawColor,draw opacity=0.80,line width= 0.4pt,line join=round,line cap=round,fill=fillColor,fill opacity=0.80] ( 43.83, 43.56) --
	( 46.29, 46.02) --
	( 43.83, 48.48) --
	( 41.37, 46.02) --
	cycle;
\definecolor{drawColor}{RGB}{218,165,32}

\path[draw=drawColor,draw opacity=0.80,line width= 0.5pt,line join=round] ( 43.83, 38.65) --
	( 62.02, 38.59) --
	(126.89, 69.82) --
	(172.23, 70.25) --
	(191.76, 70.18);
\definecolor{drawColor}{RGB}{0,0,139}

\path[draw=drawColor,draw opacity=0.80,line width= 0.5pt,line join=round] ( 43.83, 38.72) --
	( 62.02, 38.74) --
	(126.89, 63.60) --
	(172.23, 70.29) --
	(191.76, 69.92);
\definecolor{drawColor}{RGB}{255,0,255}

\path[draw=drawColor,draw opacity=0.80,line width= 0.5pt,line join=round] ( 43.83, 38.81) --
	( 62.02, 38.65) --
	(126.89, 68.66) --
	(172.23, 69.23) --
	(191.76, 70.03);
\definecolor{drawColor}{RGB}{178,34,34}

\path[draw=drawColor,draw opacity=0.80,line width= 0.5pt,line join=round] ( 43.83, 48.55) --
	( 62.02, 55.84) --
	(126.89, 67.70) --
	(172.23, 68.86) --
	(191.76, 70.19);
\definecolor{drawColor}{RGB}{67,205,128}

\path[draw=drawColor,draw opacity=0.80,line width= 0.5pt,line join=round] ( 43.83, 46.02) --
	( 62.02, 49.00) --
	(126.89, 69.33) --
	(172.23, 70.03) --
	(191.76, 70.06);
\definecolor{drawColor}{RGB}{0,0,0}

\path[draw=drawColor,line width= 0.5pt,line join=round] ( 43.83, 71.19) --
	( 62.02, 71.19) --
	(126.89, 71.21) --
	(172.23, 71.18) --
	(191.76, 71.19);

\path[draw=drawColor,line width= 0.5pt,line join=round] ( 43.83, 71.19) --
	( 62.02, 71.19) --
	(126.89, 71.19) --
	(172.23, 71.16) --
	(191.76, 71.24);

\path[draw=drawColor,line width= 0.5pt,line join=round] ( 43.83, 71.17) --
	( 62.02, 71.17) --
	(126.89, 71.22) --
	(172.23, 71.20) --
	(191.76, 71.26);

\path[draw=drawColor,line width= 0.5pt,line join=round] ( 43.83, 71.22) --
	( 62.02, 71.18) --
	(126.89, 71.17) --
	(172.23, 71.17) --
	(191.76, 71.20);

\path[draw=drawColor,line width= 0.5pt,line join=round] ( 43.83, 71.24) --
	( 62.02, 71.15) --
	(126.89, 71.17) --
	(172.23, 71.16) --
	(191.76, 71.20);
\definecolor{drawColor}{gray}{0.60}

\path[draw=drawColor,line width= 0.5pt,line join=round,line cap=round] ( 36.21, 30.28) -- ( 36.21, 27.43);

\path[draw=drawColor,line width= 0.5pt,line join=round,line cap=round] ( 42.50, 30.28) -- ( 42.50, 27.43);

\path[draw=drawColor,line width= 0.5pt,line join=round,line cap=round] ( 47.63, 30.28) -- ( 47.63, 27.43);

\path[draw=drawColor,line width= 0.5pt,line join=round,line cap=round] ( 51.98, 30.28) -- ( 51.98, 27.43);

\path[draw=drawColor,line width= 0.5pt,line join=round,line cap=round] ( 55.74, 30.28) -- ( 55.74, 27.43);

\path[draw=drawColor,line width= 0.5pt,line join=round,line cap=round] ( 59.06, 30.28) -- ( 59.06, 27.43);

\path[draw=drawColor,line width= 0.5pt,line join=round,line cap=round] ( 62.02, 30.28) -- ( 62.02, 24.59);

\path[draw=drawColor,line width= 0.5pt,line join=round,line cap=round] ( 81.55, 30.28) -- ( 81.55, 27.43);

\path[draw=drawColor,line width= 0.5pt,line join=round,line cap=round] ( 92.97, 30.28) -- ( 92.97, 27.43);

\path[draw=drawColor,line width= 0.5pt,line join=round,line cap=round] (101.08, 30.28) -- (101.08, 27.43);

\path[draw=drawColor,line width= 0.5pt,line join=round,line cap=round] (107.36, 30.28) -- (107.36, 27.43);

\path[draw=drawColor,line width= 0.5pt,line join=round,line cap=round] (112.50, 30.28) -- (112.50, 27.43);

\path[draw=drawColor,line width= 0.5pt,line join=round,line cap=round] (116.84, 30.28) -- (116.84, 27.43);

\path[draw=drawColor,line width= 0.5pt,line join=round,line cap=round] (120.60, 30.28) -- (120.60, 27.43);

\path[draw=drawColor,line width= 0.5pt,line join=round,line cap=round] (123.92, 30.28) -- (123.92, 27.43);

\path[draw=drawColor,line width= 0.5pt,line join=round,line cap=round] (126.89, 30.28) -- (126.89, 24.59);

\path[draw=drawColor,line width= 0.5pt,line join=round,line cap=round] (146.42, 30.28) -- (146.42, 27.43);

\path[draw=drawColor,line width= 0.5pt,line join=round,line cap=round] (157.84, 30.28) -- (157.84, 27.43);

\path[draw=drawColor,line width= 0.5pt,line join=round,line cap=round] (165.94, 30.28) -- (165.94, 27.43);

\path[draw=drawColor,line width= 0.5pt,line join=round,line cap=round] (172.23, 30.28) -- (172.23, 27.43);

\path[draw=drawColor,line width= 0.5pt,line join=round,line cap=round] (177.37, 30.28) -- (177.37, 27.43);

\path[draw=drawColor,line width= 0.5pt,line join=round,line cap=round] (181.71, 30.28) -- (181.71, 27.43);

\path[draw=drawColor,line width= 0.5pt,line join=round,line cap=round] (185.47, 30.28) -- (185.47, 27.43);

\path[draw=drawColor,line width= 0.5pt,line join=round,line cap=round] (188.79, 30.28) -- (188.79, 27.43);

\path[draw=drawColor,line width= 0.5pt,line join=round,line cap=round] (191.76, 30.28) -- (191.76, 24.59);
\end{scope}
\begin{scope}
\definecolor{drawColor}{gray}{0.30}

\node[text=drawColor,anchor=base east,inner sep=0pt, outer sep=0pt, scale=  0.80] at ( 31.13, 29.51) {0.00};

\node[text=drawColor,anchor=base east,inner sep=0pt, outer sep=0pt, scale=  0.80] at ( 31.13, 39.43) {0.25};

\node[text=drawColor,anchor=base east,inner sep=0pt, outer sep=0pt, scale=  0.80] at ( 31.13, 49.36) {0.50};

\node[text=drawColor,anchor=base east,inner sep=0pt, outer sep=0pt, scale=  0.80] at ( 31.13, 59.28) {0.75};

\node[text=drawColor,anchor=base east,inner sep=0pt, outer sep=0pt, scale=  0.80] at ( 31.13, 69.21) {1.00};
\end{scope}
\begin{scope}
\definecolor{drawColor}{gray}{0.60}

\path[draw=drawColor,line width= 0.6pt,line join=round] ( 33.33, 32.26) --
	( 36.08, 32.26);

\path[draw=drawColor,line width= 0.6pt,line join=round] ( 33.33, 42.19) --
	( 36.08, 42.19);

\path[draw=drawColor,line width= 0.6pt,line join=round] ( 33.33, 52.11) --
	( 36.08, 52.11);

\path[draw=drawColor,line width= 0.6pt,line join=round] ( 33.33, 62.04) --
	( 36.08, 62.04);

\path[draw=drawColor,line width= 0.6pt,line join=round] ( 33.33, 71.96) --
	( 36.08, 71.96);
\end{scope}
\begin{scope}
\definecolor{drawColor}{gray}{0.60}

\path[draw=drawColor,line width= 0.6pt,line join=round] ( 62.02, 27.53) --
	( 62.02, 30.28);

\path[draw=drawColor,line width= 0.6pt,line join=round] (126.89, 27.53) --
	(126.89, 30.28);

\path[draw=drawColor,line width= 0.6pt,line join=round] (191.76, 27.53) --
	(191.76, 30.28);
\end{scope}
\begin{scope}
\definecolor{drawColor}{gray}{0.30}

\node[text=drawColor,anchor=base west,inner sep=0pt, outer sep=0pt, scale=  0.80] at ( 55.69, 16.13) {10};

\node[text=drawColor,anchor=base west,inner sep=0pt, outer sep=0pt, scale=  0.56] at ( 63.69, 19.40) {-2};

\node[text=drawColor,anchor=base west,inner sep=0pt, outer sep=0pt, scale=  0.80] at (120.56, 16.13) {10};

\node[text=drawColor,anchor=base west,inner sep=0pt, outer sep=0pt, scale=  0.56] at (128.56, 19.40) {-1};

\node[text=drawColor,anchor=base west,inner sep=0pt, outer sep=0pt, scale=  0.80] at (186.36, 16.13) {10};

\node[text=drawColor,anchor=base west,inner sep=0pt, outer sep=0pt, scale=  0.56] at (194.36, 19.40) {0};
\end{scope}
\begin{scope}
\definecolor{drawColor}{RGB}{0,0,0}

\node[text=drawColor,anchor=base,inner sep=0pt, outer sep=0pt, scale=  0.80] at (117.79,  4.40) {Sparsity};
\end{scope}
\begin{scope}
\definecolor{drawColor}{RGB}{0,0,0}

\node[text=drawColor,rotate= 90.00,anchor=base,inner sep=0pt, outer sep=0pt, scale=  0.80] at (  8.36, 52.11) {Accuracy};
\end{scope}
\end{tikzpicture}

%% file: figs/results_singleshot_5_relu_posttrain.tikz
\begin{tikzpicture}[x=1pt,y=1pt,baseline={(0,0.4)}]
\definecolor{fillColor}{RGB}{255,255,255}
\begin{scope}
\definecolor{drawColor}{RGB}{218,165,32}

\path[draw=drawColor,line width= 0.6pt,line join=round] (110.82, 71.90) --
	(110.90, 71.90);

\path[draw=drawColor,line width= 0.6pt,line join=round] (110.86, 71.90) --
	(110.86, 71.68);

\path[draw=drawColor,line width= 0.6pt,line join=round] (110.82, 71.68) --
	(110.90, 71.68);

\path[draw=drawColor,line width= 0.6pt,line join=round] (151.22, 37.81) --
	(151.30, 37.81);

\path[draw=drawColor,line width= 0.6pt,line join=round] (151.26, 37.81) --
	(151.26, 37.02);

\path[draw=drawColor,line width= 0.6pt,line join=round] (151.22, 37.02) --
	(151.30, 37.02);

\path[draw=drawColor,line width= 0.6pt,line join=round] (179.45, 41.54) --
	(179.53, 41.54);

\path[draw=drawColor,line width= 0.6pt,line join=round] (179.49, 41.54) --
	(179.49, 37.78);

\path[draw=drawColor,line width= 0.6pt,line join=round] (179.45, 37.78) --
	(179.53, 37.78);

\path[draw=drawColor,line width= 0.6pt,line join=round] (191.61, 43.08) --
	(191.69, 43.08);

\path[draw=drawColor,line width= 0.6pt,line join=round] (191.65, 43.08) --
	(191.65, 39.67);

\path[draw=drawColor,line width= 0.6pt,line join=round] (191.61, 39.67) --
	(191.69, 39.67);

\path[draw=drawColor,line width= 0.6pt,line join=round] ( 42.02, 71.85) --
	( 42.10, 71.85);

\path[draw=drawColor,line width= 0.6pt,line join=round] ( 42.06, 71.85) --
	( 42.06, 71.60);

\path[draw=drawColor,line width= 0.6pt,line join=round] ( 42.02, 71.60) --
	( 42.10, 71.60);
\definecolor{drawColor}{RGB}{0,0,139}

\path[draw=drawColor,line width= 0.6pt,line join=round] (110.90, 71.96) --
	(110.98, 71.96);

\path[draw=drawColor,line width= 0.6pt,line join=round] (110.94, 71.96) --
	(110.94, 71.69);

\path[draw=drawColor,line width= 0.6pt,line join=round] (110.90, 71.69) --
	(110.98, 71.69);

\path[draw=drawColor,line width= 0.6pt,line join=round] (151.30, 44.83) --
	(151.38, 44.83);

\path[draw=drawColor,line width= 0.6pt,line join=round] (151.34, 44.83) --
	(151.34, 32.26);

\path[draw=drawColor,line width= 0.6pt,line join=round] (151.30, 32.26) --
	(151.38, 32.26);

\path[draw=drawColor,line width= 0.6pt,line join=round] (179.53, 38.72) --
	(179.61, 38.72);

\path[draw=drawColor,line width= 0.6pt,line join=round] (179.57, 38.72) --
	(179.57, 37.18);

\path[draw=drawColor,line width= 0.6pt,line join=round] (179.53, 37.18) --
	(179.61, 37.18);

\path[draw=drawColor,line width= 0.6pt,line join=round] (191.69, 47.42) --
	(191.77, 47.42);

\path[draw=drawColor,line width= 0.6pt,line join=round] (191.73, 47.42) --
	(191.73, 35.05);

\path[draw=drawColor,line width= 0.6pt,line join=round] (191.69, 35.05) --
	(191.77, 35.05);

\path[draw=drawColor,line width= 0.6pt,line join=round] ( 42.10, 71.86) --
	( 42.18, 71.86);

\path[draw=drawColor,line width= 0.6pt,line join=round] ( 42.14, 71.86) --
	( 42.14, 71.70);

\path[draw=drawColor,line width= 0.6pt,line join=round] ( 42.10, 71.70) --
	( 42.18, 71.70);
\definecolor{drawColor}{RGB}{255,0,255}

\path[draw=drawColor,line width= 0.6pt,line join=round] (110.98, 71.87) --
	(111.07, 71.87);

\path[draw=drawColor,line width= 0.6pt,line join=round] (111.02, 71.87) --
	(111.02, 71.73);

\path[draw=drawColor,line width= 0.6pt,line join=round] (110.98, 71.73) --
	(111.07, 71.73);

\path[draw=drawColor,line width= 0.6pt,line join=round] (151.38, 37.86) --
	(151.46, 37.86);

\path[draw=drawColor,line width= 0.6pt,line join=round] (151.42, 37.86) --
	(151.42, 37.21);

\path[draw=drawColor,line width= 0.6pt,line join=round] (151.38, 37.21) --
	(151.46, 37.21);

\path[draw=drawColor,line width= 0.6pt,line join=round] (179.61, 40.67) --
	(179.69, 40.67);

\path[draw=drawColor,line width= 0.6pt,line join=round] (179.65, 40.67) --
	(179.65, 38.10);

\path[draw=drawColor,line width= 0.6pt,line join=round] (179.61, 38.10) --
	(179.69, 38.10);

\path[draw=drawColor,line width= 0.6pt,line join=round] (191.77, 45.43) --
	(191.85, 45.43);

\path[draw=drawColor,line width= 0.6pt,line join=round] (191.81, 45.43) --
	(191.81, 36.74);

\path[draw=drawColor,line width= 0.6pt,line join=round] (191.77, 36.74) --
	(191.85, 36.74);

\path[draw=drawColor,line width= 0.6pt,line join=round] ( 42.18, 71.90) --
	( 42.27, 71.90);

\path[draw=drawColor,line width= 0.6pt,line join=round] ( 42.23, 71.90) --
	( 42.23, 71.69);

\path[draw=drawColor,line width= 0.6pt,line join=round] ( 42.18, 71.69) --
	( 42.27, 71.69);
\definecolor{drawColor}{RGB}{178,34,34}

\path[draw=drawColor,line width= 0.6pt,line join=round] (111.07, 60.27) --
	(111.15, 60.27);

\path[draw=drawColor,line width= 0.6pt,line join=round] (111.11, 60.27) --
	(111.11, 49.05);

\path[draw=drawColor,line width= 0.6pt,line join=round] (111.07, 49.05) --
	(111.15, 49.05);

\path[draw=drawColor,line width= 0.6pt,line join=round] (151.46, 55.44) --
	(151.54, 55.44);

\path[draw=drawColor,line width= 0.6pt,line join=round] (151.50, 55.44) --
	(151.50, 38.58);

\path[draw=drawColor,line width= 0.6pt,line join=round] (151.46, 38.58) --
	(151.54, 38.58);

\path[draw=drawColor,line width= 0.6pt,line join=round] (179.69, 52.21) --
	(179.77, 52.21);

\path[draw=drawColor,line width= 0.6pt,line join=round] (179.73, 52.21) --
	(179.73, 42.51);

\path[draw=drawColor,line width= 0.6pt,line join=round] (179.69, 42.51) --
	(179.77, 42.51);

\path[draw=drawColor,line width= 0.6pt,line join=round] (191.85, 45.56) --
	(191.93, 45.56);

\path[draw=drawColor,line width= 0.6pt,line join=round] (191.89, 45.56) --
	(191.89, 33.90);

\path[draw=drawColor,line width= 0.6pt,line join=round] (191.85, 33.90) --
	(191.93, 33.90);

\path[draw=drawColor,line width= 0.6pt,line join=round] ( 42.27, 71.85) --
	( 42.35, 71.85);

\path[draw=drawColor,line width= 0.6pt,line join=round] ( 42.31, 71.85) --
	( 42.31, 71.63);

\path[draw=drawColor,line width= 0.6pt,line join=round] ( 42.27, 71.63) --
	( 42.35, 71.63);
\definecolor{drawColor}{RGB}{67,205,128}

\path[draw=drawColor,line width= 0.6pt,line join=round] (111.15, 63.16) --
	(111.23, 63.16);

\path[draw=drawColor,line width= 0.6pt,line join=round] (111.19, 63.16) --
	(111.19, 43.40);

\path[draw=drawColor,line width= 0.6pt,line join=round] (111.15, 43.40) --
	(111.23, 43.40);

\path[draw=drawColor,line width= 0.6pt,line join=round] (151.54, 38.64) --
	(151.62, 38.64);

\path[draw=drawColor,line width= 0.6pt,line join=round] (151.58, 38.64) --
	(151.58, 37.60);

\path[draw=drawColor,line width= 0.6pt,line join=round] (151.54, 37.60) --
	(151.62, 37.60);

\path[draw=drawColor,line width= 0.6pt,line join=round] (179.77, 40.74) --
	(179.85, 40.74);

\path[draw=drawColor,line width= 0.6pt,line join=round] (179.81, 40.74) --
	(179.81, 38.30);

\path[draw=drawColor,line width= 0.6pt,line join=round] (179.77, 38.30) --
	(179.85, 38.30);

\path[draw=drawColor,line width= 0.6pt,line join=round] (191.93, 42.87) --
	(192.01, 42.87);

\path[draw=drawColor,line width= 0.6pt,line join=round] (191.97, 42.87) --
	(191.97, 39.10);

\path[draw=drawColor,line width= 0.6pt,line join=round] (191.93, 39.10) --
	(192.01, 39.10);

\path[draw=drawColor,line width= 0.6pt,line join=round] ( 42.35, 71.87) --
	( 42.43, 71.87);

\path[draw=drawColor,line width= 0.6pt,line join=round] ( 42.39, 71.87) --
	( 42.39, 71.70);

\path[draw=drawColor,line width= 0.6pt,line join=round] ( 42.35, 71.70) --
	( 42.43, 71.70);
\definecolor{drawColor}{RGB}{218,165,32}
\definecolor{fillColor}{RGB}{218,165,32}

\path[draw=drawColor,draw opacity=0.80,line width= 0.4pt,line join=round,line cap=round,fill=fillColor,fill opacity=0.80] (111.02, 74.84) --
	(113.67, 70.27) --
	(108.38, 70.27) --
	cycle;

\path[draw=drawColor,draw opacity=0.80,line width= 0.4pt,line join=round,line cap=round,fill=fillColor,fill opacity=0.80] (151.42, 40.49) --
	(154.06, 35.91) --
	(148.77, 35.91) --
	cycle;

\path[draw=drawColor,draw opacity=0.80,line width= 0.4pt,line join=round,line cap=round,fill=fillColor,fill opacity=0.80] (179.65, 43.08) --
	(182.29, 38.50) --
	(177.01, 38.50) --
	cycle;

\path[draw=drawColor,draw opacity=0.80,line width= 0.4pt,line join=round,line cap=round,fill=fillColor,fill opacity=0.80] (191.81, 44.73) --
	(194.45, 40.15) --
	(189.17, 40.15) --
	cycle;

\path[draw=drawColor,draw opacity=0.80,line width= 0.4pt,line join=round,line cap=round,fill=fillColor,fill opacity=0.80] ( 42.23, 74.78) --
	( 44.87, 70.20) --
	( 39.58, 70.20) --
	cycle;
\definecolor{drawColor}{RGB}{0,0,139}
\definecolor{fillColor}{RGB}{0,0,139}

\path[draw=drawColor,draw opacity=0.80,line width= 0.4pt,line join=round,line cap=round,fill=fillColor,fill opacity=0.80] (111.02, 69.37) --
	(113.48, 71.83) --
	(111.02, 74.29) --
	(108.57, 71.83) --
	cycle;

\path[draw=drawColor,draw opacity=0.80,line width= 0.4pt,line join=round,line cap=round,fill=fillColor,fill opacity=0.80] (151.42, 39.41) --
	(153.88, 41.87) --
	(151.42, 44.33) --
	(148.96, 41.87) --
	cycle;

\path[draw=drawColor,draw opacity=0.80,line width= 0.4pt,line join=round,line cap=round,fill=fillColor,fill opacity=0.80] (179.65, 35.55) --
	(182.11, 38.01) --
	(179.65, 40.47) --
	(177.19, 38.01) --
	cycle;

\path[draw=drawColor,draw opacity=0.80,line width= 0.4pt,line join=round,line cap=round,fill=fillColor,fill opacity=0.80] (191.81, 42.02) --
	(194.27, 44.48) --
	(191.81, 46.94) --
	(189.35, 44.48) --
	cycle;

\path[draw=drawColor,draw opacity=0.80,line width= 0.4pt,line join=round,line cap=round,fill=fillColor,fill opacity=0.80] ( 42.23, 69.32) --
	( 44.68, 71.78) --
	( 42.23, 74.24) --
	( 39.77, 71.78) --
	cycle;
\definecolor{fillColor}{RGB}{255,0,255}

\path[fill=fillColor,fill opacity=0.80] (109.06, 69.84) --
	(112.99, 69.84) --
	(112.99, 73.76) --
	(109.06, 73.76) --
	cycle;

\path[fill=fillColor,fill opacity=0.80] (149.45, 35.58) --
	(153.38, 35.58) --
	(153.38, 39.51) --
	(149.45, 39.51) --
	cycle;

\path[fill=fillColor,fill opacity=0.80] (177.69, 37.60) --
	(181.61, 37.60) --
	(181.61, 41.52) --
	(177.69, 41.52) --
	cycle;

\path[fill=fillColor,fill opacity=0.80] (189.85, 40.89) --
	(193.77, 40.89) --
	(193.77, 44.82) --
	(189.85, 44.82) --
	cycle;

\path[fill=fillColor,fill opacity=0.80] ( 40.26, 69.83) --
	( 44.19, 69.83) --
	( 44.19, 73.76) --
	( 40.26, 73.76) --
	cycle;
\definecolor{drawColor}{RGB}{178,34,34}
\definecolor{fillColor}{RGB}{178,34,34}

\path[draw=drawColor,draw opacity=0.80,line width= 0.4pt,line join=round,line cap=round,fill=fillColor,fill opacity=0.80] (111.02, 54.37) --
	(113.67, 58.94) --
	(108.38, 58.94) --
	cycle;

\path[draw=drawColor,draw opacity=0.80,line width= 0.4pt,line join=round,line cap=round,fill=fillColor,fill opacity=0.80] (151.42, 49.25) --
	(154.06, 53.82) --
	(148.77, 53.82) --
	cycle;

\path[draw=drawColor,draw opacity=0.80,line width= 0.4pt,line join=round,line cap=round,fill=fillColor,fill opacity=0.80] (179.65, 46.46) --
	(182.29, 51.03) --
	(177.01, 51.03) --
	cycle;

\path[draw=drawColor,draw opacity=0.80,line width= 0.4pt,line join=round,line cap=round,fill=fillColor,fill opacity=0.80] (191.81, 39.62) --
	(194.45, 44.20) --
	(189.17, 44.20) --
	cycle;

\path[draw=drawColor,draw opacity=0.80,line width= 0.4pt,line join=round,line cap=round,fill=fillColor,fill opacity=0.80] ( 42.23, 68.69) --
	( 44.87, 73.27) --
	( 39.58, 73.27) --
	cycle;
\definecolor{drawColor}{RGB}{67,205,128}
\definecolor{fillColor}{RGB}{67,205,128}

\path[draw=drawColor,draw opacity=0.80,line width= 0.4pt,line join=round,line cap=round,fill=fillColor,fill opacity=0.80] (111.02, 57.49) --
	(113.48, 59.95) --
	(111.02, 62.41) --
	(108.57, 59.95) --
	cycle;

\path[draw=drawColor,draw opacity=0.80,line width= 0.4pt,line join=round,line cap=round,fill=fillColor,fill opacity=0.80] (151.42, 35.69) --
	(153.88, 38.15) --
	(151.42, 40.61) --
	(148.96, 38.15) --
	cycle;

\path[draw=drawColor,draw opacity=0.80,line width= 0.4pt,line join=round,line cap=round,fill=fillColor,fill opacity=0.80] (179.65, 37.21) --
	(182.11, 39.67) --
	(179.65, 42.13) --
	(177.19, 39.67) --
	cycle;

\path[draw=drawColor,draw opacity=0.80,line width= 0.4pt,line join=round,line cap=round,fill=fillColor,fill opacity=0.80] (191.81, 38.89) --
	(194.27, 41.35) --
	(191.81, 43.81) --
	(189.35, 41.35) --
	cycle;

\path[draw=drawColor,draw opacity=0.80,line width= 0.4pt,line join=round,line cap=round,fill=fillColor,fill opacity=0.80] ( 42.23, 69.33) --
	( 44.68, 71.79) --
	( 42.23, 74.25) --
	( 39.77, 71.79) --
	cycle;
\definecolor{drawColor}{RGB}{218,165,32}

\path[draw=drawColor,draw opacity=0.80,line width= 0.5pt,line join=round] ( 42.23, 71.73) --
	(111.02, 71.79) --
	(151.42, 37.43) --
	(179.65, 40.03) --
	(191.81, 41.68);
\definecolor{drawColor}{RGB}{0,0,139}

\path[draw=drawColor,draw opacity=0.80,line width= 0.5pt,line join=round] ( 42.23, 71.78) --
	(111.02, 71.83) --
	(151.42, 41.87) --
	(179.65, 38.01) --
	(191.81, 44.48);
\definecolor{drawColor}{RGB}{255,0,255}

\path[draw=drawColor,draw opacity=0.80,line width= 0.5pt,line join=round] ( 42.23, 71.79) --
	(111.02, 71.80) --
	(151.42, 37.54) --
	(179.65, 39.56) --
	(191.81, 42.86);
\definecolor{drawColor}{RGB}{178,34,34}

\path[draw=drawColor,draw opacity=0.80,line width= 0.5pt,line join=round] ( 42.23, 71.74) --
	(111.02, 57.42) --
	(151.42, 52.30) --
	(179.65, 49.51) --
	(191.81, 42.67);
\definecolor{drawColor}{RGB}{67,205,128}

\path[draw=drawColor,draw opacity=0.80,line width= 0.5pt,line join=round] ( 42.23, 71.79) --
	(111.02, 59.95) --
	(151.42, 38.15) --
	(179.65, 39.67) --
	(191.81, 41.35);
\definecolor{drawColor}{RGB}{0,0,0}

\path[draw=drawColor,line width= 0.5pt,line join=round] ( 42.23, 36.82) --
	(111.02, 36.68) --
	(151.42, 36.74) --
	(179.65, 36.73) --
	(191.81, 36.75);

\path[draw=drawColor,line width= 0.5pt,line join=round] ( 42.23, 36.73) --
	(111.02, 36.70) --
	(151.42, 36.77) --
	(179.65, 36.75) --
	(191.81, 36.72);

\path[draw=drawColor,line width= 0.5pt,line join=round] ( 42.23, 36.69) --
	(111.02, 36.66) --
	(151.42, 36.66) --
	(179.65, 36.77) --
	(191.81, 36.77);

\path[draw=drawColor,line width= 0.5pt,line join=round] ( 42.23, 36.81) --
	(111.02, 36.72) --
	(151.42, 36.71) --
	(179.65, 36.78) --
	(191.81, 36.74);
\definecolor{drawColor}{gray}{0.60}

\path[draw=drawColor,line width= 0.5pt,line join=round,line cap=round] ( 42.40, 30.28) -- ( 42.40, 27.43);

\path[draw=drawColor,line width= 0.5pt,line join=round,line cap=round] ( 49.51, 30.28) -- ( 49.51, 27.43);

\path[draw=drawColor,line width= 0.5pt,line join=round,line cap=round] ( 54.56, 30.28) -- ( 54.56, 27.43);

\path[draw=drawColor,line width= 0.5pt,line join=round,line cap=round] ( 58.47, 30.28) -- ( 58.47, 27.43);

\path[draw=drawColor,line width= 0.5pt,line join=round,line cap=round] ( 61.67, 30.28) -- ( 61.67, 27.43);

\path[draw=drawColor,line width= 0.5pt,line join=round,line cap=round] ( 64.38, 30.28) -- ( 64.38, 27.43);

\path[draw=drawColor,line width= 0.5pt,line join=round,line cap=round] ( 66.72, 30.28) -- ( 66.72, 27.43);

\path[draw=drawColor,line width= 0.5pt,line join=round,line cap=round] ( 68.78, 30.28) -- ( 68.78, 27.43);

\path[draw=drawColor,line width= 0.5pt,line join=round,line cap=round] ( 70.63, 30.28) -- ( 70.63, 24.59);

\path[draw=drawColor,line width= 0.5pt,line join=round,line cap=round] ( 82.79, 30.28) -- ( 82.79, 27.43);

\path[draw=drawColor,line width= 0.5pt,line join=round,line cap=round] ( 89.90, 30.28) -- ( 89.90, 27.43);

\path[draw=drawColor,line width= 0.5pt,line join=round,line cap=round] ( 94.95, 30.28) -- ( 94.95, 27.43);

\path[draw=drawColor,line width= 0.5pt,line join=round,line cap=round] ( 98.87, 30.28) -- ( 98.87, 27.43);

\path[draw=drawColor,line width= 0.5pt,line join=round,line cap=round] (102.06, 30.28) -- (102.06, 27.43);

\path[draw=drawColor,line width= 0.5pt,line join=round,line cap=round] (104.77, 30.28) -- (104.77, 27.43);

\path[draw=drawColor,line width= 0.5pt,line join=round,line cap=round] (107.11, 30.28) -- (107.11, 27.43);

\path[draw=drawColor,line width= 0.5pt,line join=round,line cap=round] (109.18, 30.28) -- (109.18, 27.43);

\path[draw=drawColor,line width= 0.5pt,line join=round,line cap=round] (111.02, 30.28) -- (111.02, 24.59);

\path[draw=drawColor,line width= 0.5pt,line join=round,line cap=round] (123.18, 30.28) -- (123.18, 27.43);

\path[draw=drawColor,line width= 0.5pt,line join=round,line cap=round] (130.30, 30.28) -- (130.30, 27.43);

\path[draw=drawColor,line width= 0.5pt,line join=round,line cap=round] (135.34, 30.28) -- (135.34, 27.43);

\path[draw=drawColor,line width= 0.5pt,line join=round,line cap=round] (139.26, 30.28) -- (139.26, 27.43);

\path[draw=drawColor,line width= 0.5pt,line join=round,line cap=round] (142.46, 30.28) -- (142.46, 27.43);

\path[draw=drawColor,line width= 0.5pt,line join=round,line cap=round] (145.16, 30.28) -- (145.16, 27.43);

\path[draw=drawColor,line width= 0.5pt,line join=round,line cap=round] (147.50, 30.28) -- (147.50, 27.43);

\path[draw=drawColor,line width= 0.5pt,line join=round,line cap=round] (149.57, 30.28) -- (149.57, 27.43);

\path[draw=drawColor,line width= 0.5pt,line join=round,line cap=round] (151.42, 30.28) -- (151.42, 24.59);

\path[draw=drawColor,line width= 0.5pt,line join=round,line cap=round] (163.58, 30.28) -- (163.58, 27.43);

\path[draw=drawColor,line width= 0.5pt,line join=round,line cap=round] (170.69, 30.28) -- (170.69, 27.43);

\path[draw=drawColor,line width= 0.5pt,line join=round,line cap=round] (175.74, 30.28) -- (175.74, 27.43);

\path[draw=drawColor,line width= 0.5pt,line join=round,line cap=round] (179.65, 30.28) -- (179.65, 27.43);

\path[draw=drawColor,line width= 0.5pt,line join=round,line cap=round] (182.85, 30.28) -- (182.85, 27.43);

\path[draw=drawColor,line width= 0.5pt,line join=round,line cap=round] (185.55, 30.28) -- (185.55, 27.43);

\path[draw=drawColor,line width= 0.5pt,line join=round,line cap=round] (187.89, 30.28) -- (187.89, 27.43);

\path[draw=drawColor,line width= 0.5pt,line join=round,line cap=round] (189.96, 30.28) -- (189.96, 27.43);

\path[draw=drawColor,line width= 0.5pt,line join=round,line cap=round] (191.81, 30.28) -- (191.81, 24.59);
\end{scope}
\begin{scope}
\definecolor{drawColor}{gray}{0.30}

\node[text=drawColor,anchor=base west,inner sep=0pt, outer sep=0pt, scale=  0.80] at ( 16.91, 33.32) {10};

\node[text=drawColor,anchor=base west,inner sep=0pt, outer sep=0pt, scale=  0.56] at ( 24.91, 36.59) {-4};

\node[text=drawColor,anchor=base west,inner sep=0pt, outer sep=0pt, scale=  0.80] at ( 16.91, 44.19) {10};

\node[text=drawColor,anchor=base west,inner sep=0pt, outer sep=0pt, scale=  0.56] at ( 24.91, 47.46) {-3};

\node[text=drawColor,anchor=base west,inner sep=0pt, outer sep=0pt, scale=  0.80] at ( 16.91, 55.06) {10};

\node[text=drawColor,anchor=base west,inner sep=0pt, outer sep=0pt, scale=  0.56] at ( 24.91, 58.33) {-2};

\node[text=drawColor,anchor=base west,inner sep=0pt, outer sep=0pt, scale=  0.80] at ( 16.91, 65.93) {10};

\node[text=drawColor,anchor=base west,inner sep=0pt, outer sep=0pt, scale=  0.56] at ( 24.91, 69.20) {-1};
\end{scope}
\begin{scope}
\definecolor{drawColor}{gray}{0.60}

\path[draw=drawColor,line width= 0.6pt,line join=round] ( 31.77, 36.75) --
	( 34.52, 36.75);

\path[draw=drawColor,line width= 0.6pt,line join=round] ( 31.77, 47.62) --
	( 34.52, 47.62);

\path[draw=drawColor,line width= 0.6pt,line join=round] ( 31.77, 58.49) --
	( 34.52, 58.49);

\path[draw=drawColor,line width= 0.6pt,line join=round] ( 31.77, 69.37) --
	( 34.52, 69.37);
\end{scope}
\begin{scope}
\definecolor{drawColor}{gray}{0.60}

\path[draw=drawColor,line width= 0.6pt,line join=round] ( 70.63, 27.53) --
	( 70.63, 30.28);

\path[draw=drawColor,line width= 0.6pt,line join=round] (111.02, 27.53) --
	(111.02, 30.28);

\path[draw=drawColor,line width= 0.6pt,line join=round] (151.42, 27.53) --
	(151.42, 30.28);

\path[draw=drawColor,line width= 0.6pt,line join=round] (191.81, 27.53) --
	(191.81, 30.28);
\end{scope}
\begin{scope}
\definecolor{drawColor}{gray}{0.30}

\node[text=drawColor,anchor=base west,inner sep=0pt, outer sep=0pt, scale=  0.80] at ( 64.30, 16.13) {10};

\node[text=drawColor,anchor=base west,inner sep=0pt, outer sep=0pt, scale=  0.56] at ( 72.30, 19.40) {-3};

\node[text=drawColor,anchor=base west,inner sep=0pt, outer sep=0pt, scale=  0.80] at (104.69, 16.13) {10};

\node[text=drawColor,anchor=base west,inner sep=0pt, outer sep=0pt, scale=  0.56] at (112.69, 19.40) {-2};

\node[text=drawColor,anchor=base west,inner sep=0pt, outer sep=0pt, scale=  0.80] at (145.09, 16.13) {10};

\node[text=drawColor,anchor=base west,inner sep=0pt, outer sep=0pt, scale=  0.56] at (153.08, 19.40) {-1};

\node[text=drawColor,anchor=base west,inner sep=0pt, outer sep=0pt, scale=  0.80] at (186.41, 16.13) {10};

\node[text=drawColor,anchor=base west,inner sep=0pt, outer sep=0pt, scale=  0.56] at (194.41, 19.40) {0};
\end{scope}
\begin{scope}
\definecolor{drawColor}{RGB}{0,0,0}

\node[text=drawColor,anchor=base,inner sep=0pt, outer sep=0pt, scale=  0.80] at (117.02,  4.40) {Sparsity};
\end{scope}
\begin{scope}
\definecolor{drawColor}{RGB}{0,0,0}

\node[text=drawColor,rotate= 90.00,anchor=base,inner sep=0pt, outer sep=0pt, scale=  0.80] at (  8.36, 52.11) {MSE};
\end{scope}
\end{tikzpicture}

%% file: figs/results_singleshot_5_helix_posttrain.tikz
\begin{tikzpicture}[x=1pt,y=1pt,baseline={(0,0.4)}]
\definecolor{fillColor}{RGB}{255,255,255}
\begin{scope}
\definecolor{drawColor}{RGB}{218,165,32}

\path[draw=drawColor,line width= 0.6pt,line join=round] ( 57.45, 71.89) --
	( 57.58, 71.89);

\path[draw=drawColor,line width= 0.6pt,line join=round] ( 57.52, 71.89) --
	( 57.52, 71.79);

\path[draw=drawColor,line width= 0.6pt,line join=round] ( 57.45, 71.79) --
	( 57.58, 71.79);

\path[draw=drawColor,line width= 0.6pt,line join=round] (124.40, 61.36) --
	(124.53, 61.36);

\path[draw=drawColor,line width= 0.6pt,line join=round] (124.46, 61.36) --
	(124.46, 54.40);

\path[draw=drawColor,line width= 0.6pt,line join=round] (124.40, 54.40) --
	(124.53, 54.40);

\path[draw=drawColor,line width= 0.6pt,line join=round] (171.19, 38.02) --
	(171.32, 38.02);

\path[draw=drawColor,line width= 0.6pt,line join=round] (171.26, 38.02) --
	(171.26, 35.38);

\path[draw=drawColor,line width= 0.6pt,line join=round] (171.19, 35.38) --
	(171.32, 35.38);

\path[draw=drawColor,line width= 0.6pt,line join=round] (191.34, 38.79) --
	(191.48, 38.79);

\path[draw=drawColor,line width= 0.6pt,line join=round] (191.41, 38.79) --
	(191.41, 35.59);

\path[draw=drawColor,line width= 0.6pt,line join=round] (191.34, 35.59) --
	(191.48, 35.59);

\path[draw=drawColor,line width= 0.6pt,line join=round] ( 42.02, 71.94) --
	( 42.16, 71.94);

\path[draw=drawColor,line width= 0.6pt,line join=round] ( 42.09, 71.94) --
	( 42.09, 71.84);

\path[draw=drawColor,line width= 0.6pt,line join=round] ( 42.02, 71.84) --
	( 42.16, 71.84);
\definecolor{drawColor}{RGB}{0,0,139}

\path[draw=drawColor,line width= 0.6pt,line join=round] ( 57.58, 71.96) --
	( 57.72, 71.96);

\path[draw=drawColor,line width= 0.6pt,line join=round] ( 57.65, 71.96) --
	( 57.65, 71.86);

\path[draw=drawColor,line width= 0.6pt,line join=round] ( 57.58, 71.86) --
	( 57.72, 71.86);

\path[draw=drawColor,line width= 0.6pt,line join=round] (124.53, 66.69) --
	(124.66, 66.69);

\path[draw=drawColor,line width= 0.6pt,line join=round] (124.60, 66.69) --
	(124.60, 65.83);

\path[draw=drawColor,line width= 0.6pt,line join=round] (124.53, 65.83) --
	(124.66, 65.83);

\path[draw=drawColor,line width= 0.6pt,line join=round] (171.32, 40.71) --
	(171.46, 40.71);

\path[draw=drawColor,line width= 0.6pt,line join=round] (171.39, 40.71) --
	(171.39, 37.78);

\path[draw=drawColor,line width= 0.6pt,line join=round] (171.32, 37.78) --
	(171.46, 37.78);

\path[draw=drawColor,line width= 0.6pt,line join=round] (191.48, 37.39) --
	(191.61, 37.39);

\path[draw=drawColor,line width= 0.6pt,line join=round] (191.54, 37.39) --
	(191.54, 33.63);

\path[draw=drawColor,line width= 0.6pt,line join=round] (191.48, 33.63) --
	(191.61, 33.63);

\path[draw=drawColor,line width= 0.6pt,line join=round] ( 42.16, 71.91) --
	( 42.29, 71.91);

\path[draw=drawColor,line width= 0.6pt,line join=round] ( 42.22, 71.91) --
	( 42.22, 71.83);

\path[draw=drawColor,line width= 0.6pt,line join=round] ( 42.16, 71.83) --
	( 42.29, 71.83);
\definecolor{drawColor}{RGB}{255,0,255}

\path[draw=drawColor,line width= 0.6pt,line join=round] ( 57.72, 71.94) --
	( 57.85, 71.94);

\path[draw=drawColor,line width= 0.6pt,line join=round] ( 57.79, 71.94) --
	( 57.79, 71.82);

\path[draw=drawColor,line width= 0.6pt,line join=round] ( 57.72, 71.82) --
	( 57.85, 71.82);

\path[draw=drawColor,line width= 0.6pt,line join=round] (124.66, 49.06) --
	(124.80, 49.06);

\path[draw=drawColor,line width= 0.6pt,line join=round] (124.73, 49.06) --
	(124.73, 38.89);

\path[draw=drawColor,line width= 0.6pt,line join=round] (124.66, 38.89) --
	(124.80, 38.89);

\path[draw=drawColor,line width= 0.6pt,line join=round] (171.46, 38.91) --
	(171.59, 38.91);

\path[draw=drawColor,line width= 0.6pt,line join=round] (171.52, 38.91) --
	(171.52, 34.01);

\path[draw=drawColor,line width= 0.6pt,line join=round] (171.46, 34.01) --
	(171.59, 34.01);

\path[draw=drawColor,line width= 0.6pt,line join=round] (191.61, 37.59) --
	(191.74, 37.59);

\path[draw=drawColor,line width= 0.6pt,line join=round] (191.68, 37.59) --
	(191.68, 34.88);

\path[draw=drawColor,line width= 0.6pt,line join=round] (191.61, 34.88) --
	(191.74, 34.88);

\path[draw=drawColor,line width= 0.6pt,line join=round] ( 42.29, 71.90) --
	( 42.42, 71.90);

\path[draw=drawColor,line width= 0.6pt,line join=round] ( 42.36, 71.90) --
	( 42.36, 71.81);

\path[draw=drawColor,line width= 0.6pt,line join=round] ( 42.29, 71.81) --
	( 42.42, 71.81);
\definecolor{drawColor}{RGB}{178,34,34}

\path[draw=drawColor,line width= 0.6pt,line join=round] ( 57.85, 67.88) --
	( 57.99, 67.88);

\path[draw=drawColor,line width= 0.6pt,line join=round] ( 57.92, 67.88) --
	( 57.92, 67.71);

\path[draw=drawColor,line width= 0.6pt,line join=round] ( 57.85, 67.71) --
	( 57.99, 67.71);

\path[draw=drawColor,line width= 0.6pt,line join=round] (124.80, 67.58) --
	(124.93, 67.58);

\path[draw=drawColor,line width= 0.6pt,line join=round] (124.86, 67.58) --
	(124.86, 67.44);

\path[draw=drawColor,line width= 0.6pt,line join=round] (124.80, 67.44) --
	(124.93, 67.44);

\path[draw=drawColor,line width= 0.6pt,line join=round] (171.59, 67.51) --
	(171.72, 67.51);

\path[draw=drawColor,line width= 0.6pt,line join=round] (171.66, 67.51) --
	(171.66, 67.25);

\path[draw=drawColor,line width= 0.6pt,line join=round] (171.59, 67.25) --
	(171.72, 67.25);

\path[draw=drawColor,line width= 0.6pt,line join=round] (191.74, 39.14) --
	(191.88, 39.14);

\path[draw=drawColor,line width= 0.6pt,line join=round] (191.81, 39.14) --
	(191.81, 35.65);

\path[draw=drawColor,line width= 0.6pt,line join=round] (191.74, 35.65) --
	(191.88, 35.65);

\path[draw=drawColor,line width= 0.6pt,line join=round] ( 42.42, 68.24) --
	( 42.56, 68.24);

\path[draw=drawColor,line width= 0.6pt,line join=round] ( 42.49, 68.24) --
	( 42.49, 67.76);

\path[draw=drawColor,line width= 0.6pt,line join=round] ( 42.42, 67.76) --
	( 42.56, 67.76);
\definecolor{drawColor}{RGB}{67,205,128}

\path[draw=drawColor,line width= 0.6pt,line join=round] ( 57.99, 69.27) --
	( 58.12, 69.27);

\path[draw=drawColor,line width= 0.6pt,line join=round] ( 58.05, 69.27) --
	( 58.05, 67.85);

\path[draw=drawColor,line width= 0.6pt,line join=round] ( 57.99, 67.85) --
	( 58.12, 67.85);

\path[draw=drawColor,line width= 0.6pt,line join=round] (124.93, 62.54) --
	(125.07, 62.54);

\path[draw=drawColor,line width= 0.6pt,line join=round] (125.00, 62.54) --
	(125.00, 59.86);

\path[draw=drawColor,line width= 0.6pt,line join=round] (124.93, 59.86) --
	(125.07, 59.86);

\path[draw=drawColor,line width= 0.6pt,line join=round] (171.72, 40.06) --
	(171.86, 40.06);

\path[draw=drawColor,line width= 0.6pt,line join=round] (171.79, 40.06) --
	(171.79, 37.48);

\path[draw=drawColor,line width= 0.6pt,line join=round] (171.72, 37.48) --
	(171.86, 37.48);

\path[draw=drawColor,line width= 0.6pt,line join=round] (191.88, 39.13) --
	(192.01, 39.13);

\path[draw=drawColor,line width= 0.6pt,line join=round] (191.94, 39.13) --
	(191.94, 35.93);

\path[draw=drawColor,line width= 0.6pt,line join=round] (191.88, 35.93) --
	(192.01, 35.93);

\path[draw=drawColor,line width= 0.6pt,line join=round] ( 42.56, 71.14) --
	( 42.69, 71.14);

\path[draw=drawColor,line width= 0.6pt,line join=round] ( 42.63, 71.14) --
	( 42.63, 69.69);

\path[draw=drawColor,line width= 0.6pt,line join=round] ( 42.56, 69.69) --
	( 42.69, 69.69);
\definecolor{drawColor}{RGB}{218,165,32}
\definecolor{fillColor}{RGB}{218,165,32}

\path[draw=drawColor,draw opacity=0.80,line width= 0.4pt,line join=round,line cap=round,fill=fillColor,fill opacity=0.80] ( 57.79, 74.89) --
	( 60.43, 70.32) --
	( 55.14, 70.32) --
	cycle;

\path[draw=drawColor,draw opacity=0.80,line width= 0.4pt,line join=round,line cap=round,fill=fillColor,fill opacity=0.80] (124.73, 61.92) --
	(127.37, 57.34) --
	(122.09, 57.34) --
	cycle;

\path[draw=drawColor,draw opacity=0.80,line width= 0.4pt,line join=round,line cap=round,fill=fillColor,fill opacity=0.80] (171.52, 39.90) --
	(174.17, 35.32) --
	(168.88, 35.32) --
	cycle;

\path[draw=drawColor,draw opacity=0.80,line width= 0.4pt,line join=round,line cap=round,fill=fillColor,fill opacity=0.80] (191.68, 40.46) --
	(194.32, 35.89) --
	(189.03, 35.89) --
	cycle;

\path[draw=drawColor,draw opacity=0.80,line width= 0.4pt,line join=round,line cap=round,fill=fillColor,fill opacity=0.80] ( 42.36, 74.94) --
	( 45.00, 70.37) --
	( 39.72, 70.37) --
	cycle;
\definecolor{drawColor}{RGB}{0,0,139}
\definecolor{fillColor}{RGB}{0,0,139}

\path[draw=drawColor,draw opacity=0.80,line width= 0.4pt,line join=round,line cap=round,fill=fillColor,fill opacity=0.80] ( 57.79, 69.45) --
	( 60.24, 71.91) --
	( 57.79, 74.37) --
	( 55.33, 71.91) --
	cycle;

\path[draw=drawColor,draw opacity=0.80,line width= 0.4pt,line join=round,line cap=round,fill=fillColor,fill opacity=0.80] (124.73, 63.81) --
	(127.19, 66.27) --
	(124.73, 68.73) --
	(122.27, 66.27) --
	cycle;

\path[draw=drawColor,draw opacity=0.80,line width= 0.4pt,line join=round,line cap=round,fill=fillColor,fill opacity=0.80] (171.52, 36.97) --
	(173.98, 39.43) --
	(171.52, 41.89) --
	(169.06, 39.43) --
	cycle;

\path[draw=drawColor,draw opacity=0.80,line width= 0.4pt,line join=round,line cap=round,fill=fillColor,fill opacity=0.80] (191.68, 33.35) --
	(194.14, 35.81) --
	(191.68, 38.27) --
	(189.22, 35.81) --
	cycle;

\path[draw=drawColor,draw opacity=0.80,line width= 0.4pt,line join=round,line cap=round,fill=fillColor,fill opacity=0.80] ( 42.36, 69.41) --
	( 44.82, 71.87) --
	( 42.36, 74.33) --
	( 39.90, 71.87) --
	cycle;
\definecolor{fillColor}{RGB}{255,0,255}

\path[fill=fillColor,fill opacity=0.80] ( 55.82, 69.92) --
	( 59.75, 69.92) --
	( 59.75, 73.84) --
	( 55.82, 73.84) --
	cycle;

\path[fill=fillColor,fill opacity=0.80] (122.77, 44.01) --
	(126.69, 44.01) --
	(126.69, 47.93) --
	(122.77, 47.93) --
	cycle;

\path[fill=fillColor,fill opacity=0.80] (169.56, 35.00) --
	(173.49, 35.00) --
	(173.49, 38.92) --
	(169.56, 38.92) --
	cycle;

\path[fill=fillColor,fill opacity=0.80] (189.71, 34.43) --
	(193.64, 34.43) --
	(193.64, 38.35) --
	(189.71, 38.35) --
	cycle;

\path[fill=fillColor,fill opacity=0.80] ( 40.40, 69.90) --
	( 44.32, 69.90) --
	( 44.32, 73.82) --
	( 40.40, 73.82) --
	cycle;
\definecolor{drawColor}{RGB}{178,34,34}
\definecolor{fillColor}{RGB}{178,34,34}

\path[draw=drawColor,draw opacity=0.80,line width= 0.4pt,line join=round,line cap=round,fill=fillColor,fill opacity=0.80] ( 57.79, 64.75) --
	( 60.43, 69.32) --
	( 55.14, 69.32) --
	cycle;

\path[draw=drawColor,draw opacity=0.80,line width= 0.4pt,line join=round,line cap=round,fill=fillColor,fill opacity=0.80] (124.73, 64.46) --
	(127.37, 69.04) --
	(122.09, 69.04) --
	cycle;

\path[draw=drawColor,draw opacity=0.80,line width= 0.4pt,line join=round,line cap=round,fill=fillColor,fill opacity=0.80] (171.52, 64.33) --
	(174.17, 68.91) --
	(168.88, 68.91) --
	cycle;

\path[draw=drawColor,draw opacity=0.80,line width= 0.4pt,line join=round,line cap=round,fill=fillColor,fill opacity=0.80] (191.68, 34.60) --
	(194.32, 39.18) --
	(189.03, 39.18) --
	cycle;

\path[draw=drawColor,draw opacity=0.80,line width= 0.4pt,line join=round,line cap=round,fill=fillColor,fill opacity=0.80] ( 42.36, 64.95) --
	( 45.00, 69.53) --
	( 39.72, 69.53) --
	cycle;
\definecolor{drawColor}{RGB}{67,205,128}
\definecolor{fillColor}{RGB}{67,205,128}

\path[draw=drawColor,draw opacity=0.80,line width= 0.4pt,line join=round,line cap=round,fill=fillColor,fill opacity=0.80] ( 57.79, 66.15) --
	( 60.24, 68.60) --
	( 57.79, 71.06) --
	( 55.33, 68.60) --
	cycle;

\path[draw=drawColor,draw opacity=0.80,line width= 0.4pt,line join=round,line cap=round,fill=fillColor,fill opacity=0.80] (124.73, 58.89) --
	(127.19, 61.35) --
	(124.73, 63.81) --
	(122.27, 61.35) --
	cycle;

\path[draw=drawColor,draw opacity=0.80,line width= 0.4pt,line join=round,line cap=round,fill=fillColor,fill opacity=0.80] (171.52, 36.46) --
	(173.98, 38.92) --
	(171.52, 41.37) --
	(169.06, 38.92) --
	cycle;

\path[draw=drawColor,draw opacity=0.80,line width= 0.4pt,line join=round,line cap=round,fill=fillColor,fill opacity=0.80] (191.68, 35.29) --
	(194.14, 37.75) --
	(191.68, 40.21) --
	(189.22, 37.75) --
	cycle;

\path[draw=drawColor,draw opacity=0.80,line width= 0.4pt,line join=round,line cap=round,fill=fillColor,fill opacity=0.80] ( 42.36, 68.00) --
	( 44.82, 70.46) --
	( 42.36, 72.92) --
	( 39.90, 70.46) --
	cycle;
\definecolor{drawColor}{RGB}{218,165,32}

\path[draw=drawColor,draw opacity=0.80,line width= 0.5pt,line join=round] ( 42.36, 71.89) --
	( 57.79, 71.84) --
	(124.73, 58.86) --
	(171.52, 36.85) --
	(191.68, 37.41);
\definecolor{drawColor}{RGB}{0,0,139}

\path[draw=drawColor,draw opacity=0.80,line width= 0.5pt,line join=round] ( 42.36, 71.87) --
	( 57.79, 71.91) --
	(124.73, 66.27) --
	(171.52, 39.43) --
	(191.68, 35.81);
\definecolor{drawColor}{RGB}{255,0,255}

\path[draw=drawColor,draw opacity=0.80,line width= 0.5pt,line join=round] ( 42.36, 71.86) --
	( 57.79, 71.88) --
	(124.73, 45.97) --
	(171.52, 36.96) --
	(191.68, 36.39);
\definecolor{drawColor}{RGB}{178,34,34}

\path[draw=drawColor,draw opacity=0.80,line width= 0.5pt,line join=round] ( 42.36, 68.00) --
	( 57.79, 67.80) --
	(124.73, 67.51) --
	(171.52, 67.38) --
	(191.68, 37.65);
\definecolor{drawColor}{RGB}{67,205,128}

\path[draw=drawColor,draw opacity=0.80,line width= 0.5pt,line join=round] ( 42.36, 70.46) --
	( 57.79, 68.60) --
	(124.73, 61.35) --
	(171.52, 38.92) --
	(191.68, 37.75);
\definecolor{drawColor}{RGB}{0,0,0}

\path[draw=drawColor,line width= 0.5pt,line join=round] ( 42.36, 32.36) --
	( 57.79, 32.33) --
	(124.73, 32.41) --
	(171.52, 32.31) --
	(191.68, 32.30);

\path[draw=drawColor,line width= 0.5pt,line join=round] ( 42.36, 32.34) --
	( 57.79, 32.34) --
	(124.73, 32.30) --
	(171.52, 32.26) --
	(191.68, 32.29);

\path[draw=drawColor,line width= 0.5pt,line join=round] ( 42.36, 32.33) --
	( 57.79, 32.34) --
	(124.73, 32.30) --
	(171.52, 32.33) --
	(191.68, 32.31);

\path[draw=drawColor,line width= 0.5pt,line join=round] ( 42.36, 32.38) --
	( 57.79, 32.30) --
	(124.73, 32.33) --
	(171.52, 32.34) --
	(191.68, 32.37);
\definecolor{drawColor}{gray}{0.60}

\path[draw=drawColor,line width= 0.5pt,line join=round,line cap=round] ( 37.63, 30.28) -- ( 37.63, 27.43);

\path[draw=drawColor,line width= 0.5pt,line join=round,line cap=round] ( 42.93, 30.28) -- ( 42.93, 27.43);

\path[draw=drawColor,line width= 0.5pt,line join=round,line cap=round] ( 47.42, 30.28) -- ( 47.42, 27.43);

\path[draw=drawColor,line width= 0.5pt,line join=round,line cap=round] ( 51.30, 30.28) -- ( 51.30, 27.43);

\path[draw=drawColor,line width= 0.5pt,line join=round,line cap=round] ( 54.72, 30.28) -- ( 54.72, 27.43);

\path[draw=drawColor,line width= 0.5pt,line join=round,line cap=round] ( 57.79, 30.28) -- ( 57.79, 24.59);

\path[draw=drawColor,line width= 0.5pt,line join=round,line cap=round] ( 77.94, 30.28) -- ( 77.94, 27.43);

\path[draw=drawColor,line width= 0.5pt,line join=round,line cap=round] ( 89.73, 30.28) -- ( 89.73, 27.43);

\path[draw=drawColor,line width= 0.5pt,line join=round,line cap=round] ( 98.09, 30.28) -- ( 98.09, 27.43);

\path[draw=drawColor,line width= 0.5pt,line join=round,line cap=round] (104.58, 30.28) -- (104.58, 27.43);

\path[draw=drawColor,line width= 0.5pt,line join=round,line cap=round] (109.88, 30.28) -- (109.88, 27.43);

\path[draw=drawColor,line width= 0.5pt,line join=round,line cap=round] (114.36, 30.28) -- (114.36, 27.43);

\path[draw=drawColor,line width= 0.5pt,line join=round,line cap=round] (118.24, 30.28) -- (118.24, 27.43);

\path[draw=drawColor,line width= 0.5pt,line join=round,line cap=round] (121.67, 30.28) -- (121.67, 27.43);

\path[draw=drawColor,line width= 0.5pt,line join=round,line cap=round] (124.73, 30.28) -- (124.73, 24.59);

\path[draw=drawColor,line width= 0.5pt,line join=round,line cap=round] (144.88, 30.28) -- (144.88, 27.43);

\path[draw=drawColor,line width= 0.5pt,line join=round,line cap=round] (156.67, 30.28) -- (156.67, 27.43);

\path[draw=drawColor,line width= 0.5pt,line join=round,line cap=round] (165.04, 30.28) -- (165.04, 27.43);

\path[draw=drawColor,line width= 0.5pt,line join=round,line cap=round] (171.52, 30.28) -- (171.52, 27.43);

\path[draw=drawColor,line width= 0.5pt,line join=round,line cap=round] (176.82, 30.28) -- (176.82, 27.43);

\path[draw=drawColor,line width= 0.5pt,line join=round,line cap=round] (181.31, 30.28) -- (181.31, 27.43);

\path[draw=drawColor,line width= 0.5pt,line join=round,line cap=round] (185.19, 30.28) -- (185.19, 27.43);

\path[draw=drawColor,line width= 0.5pt,line join=round,line cap=round] (188.61, 30.28) -- (188.61, 27.43);

\path[draw=drawColor,line width= 0.5pt,line join=round,line cap=round] (191.68, 30.28) -- (191.68, 24.59);
\end{scope}
\begin{scope}
\definecolor{drawColor}{gray}{0.30}

\node[text=drawColor,anchor=base west,inner sep=0pt, outer sep=0pt, scale=  0.80] at ( 16.91, 35.58) {10};

\node[text=drawColor,anchor=base west,inner sep=0pt, outer sep=0pt, scale=  0.56] at ( 24.91, 38.85) {-3};

\node[text=drawColor,anchor=base west,inner sep=0pt, outer sep=0pt, scale=  0.80] at ( 16.91, 48.91) {10};

\node[text=drawColor,anchor=base west,inner sep=0pt, outer sep=0pt, scale=  0.56] at ( 24.91, 52.18) {-2};

\node[text=drawColor,anchor=base west,inner sep=0pt, outer sep=0pt, scale=  0.80] at ( 16.91, 62.25) {10};

\node[text=drawColor,anchor=base west,inner sep=0pt, outer sep=0pt, scale=  0.56] at ( 24.91, 65.52) {-1};
\end{scope}
\begin{scope}
\definecolor{drawColor}{gray}{0.60}

\path[draw=drawColor,line width= 0.6pt,line join=round] ( 31.77, 39.01) --
	( 34.52, 39.01);

\path[draw=drawColor,line width= 0.6pt,line join=round] ( 31.77, 52.34) --
	( 34.52, 52.34);

\path[draw=drawColor,line width= 0.6pt,line join=round] ( 31.77, 65.68) --
	( 34.52, 65.68);
\end{scope}
\begin{scope}
\definecolor{drawColor}{gray}{0.60}

\path[draw=drawColor,line width= 0.6pt,line join=round] ( 57.79, 27.53) --
	( 57.79, 30.28);

\path[draw=drawColor,line width= 0.6pt,line join=round] (124.73, 27.53) --
	(124.73, 30.28);

\path[draw=drawColor,line width= 0.6pt,line join=round] (191.68, 27.53) --
	(191.68, 30.28);
\end{scope}
\begin{scope}
\definecolor{drawColor}{gray}{0.30}

\node[text=drawColor,anchor=base west,inner sep=0pt, outer sep=0pt, scale=  0.80] at ( 51.45, 16.13) {10};

\node[text=drawColor,anchor=base west,inner sep=0pt, outer sep=0pt, scale=  0.56] at ( 59.45, 19.40) {-2};

\node[text=drawColor,anchor=base west,inner sep=0pt, outer sep=0pt, scale=  0.80] at (118.40, 16.13) {10};

\node[text=drawColor,anchor=base west,inner sep=0pt, outer sep=0pt, scale=  0.56] at (126.40, 19.40) {-1};

\node[text=drawColor,anchor=base west,inner sep=0pt, outer sep=0pt, scale=  0.80] at (186.28, 16.13) {10};

\node[text=drawColor,anchor=base west,inner sep=0pt, outer sep=0pt, scale=  0.56] at (194.28, 19.40) {0};
\end{scope}
\begin{scope}
\definecolor{drawColor}{RGB}{0,0,0}

\node[text=drawColor,anchor=base,inner sep=0pt, outer sep=0pt, scale=  0.80] at (117.02,  4.40) {Sparsity};
\end{scope}
\begin{scope}
\definecolor{drawColor}{RGB}{0,0,0}

\node[text=drawColor,rotate= 90.00,anchor=base,inner sep=0pt, outer sep=0pt, scale=  0.80] at (  8.36, 52.11) {MSE};
\end{scope}
\end{tikzpicture}

%% file: figs/multishot.tex
\begin{figure}
\centering
    \begin{subfigure}[t]{0.49\textwidth}
    \centering
		\ifpdf
		\begin{tikzpicture}
		\begin{axis}[
		jonas line,
		xmode=log,
		width = 7cm,
		height = 3.5cm,
		xlabel = {}, 
		ylabel= {Accuracy},
		xmin=0.004, xmax=1.1,
		ymin=0, ymax=1,
		legend columns=2,
		x label style 		= {at={(axis description cs:0.5,-0.1)}, anchor=north, font=\scriptsize},
		y label style 		= {at={(axis description cs:0.05,0.6)},  anchor=south, font=\scriptsize},
		legend style={nodes={scale=0.9, transform shape}, at={(0.98,0.95)}, anchor=north east, row sep=-1.4pt, font=\tiny}
		]
		
		\addplot+[forget plot, eda errorbarcolored, y dir=plus, y explicit]
		table[x=magSparsity, y=magMeanPostprune, y error=magErrorPlusPostprune] {expres/multishot/rescircleMultishot.tsv};
		\addplot+[eda errorbarcolored, y dir=minus, y explicit]
		table[x=magSparsity, y=magMeanPostprune, y error=magErrorMinusPostprune] {expres/multishot/rescircleMultishot.tsv};
		
		\addplot+[forget plot, eda errorbarcolored, y dir=plus, y explicit]
		table[x expr=\thisrow{magSparsity}*1.05, y=synflowMeanPostprune, y error=synflowErrorPlusPostprune] {expres/multishot/rescircleMultishot.tsv};
		\addplot+[eda errorbarcolored, y dir=minus, y explicit]
		table[x expr=\thisrow{magSparsity}*1.05, y=synflowMeanPostprune, y error=synflowErrorMinusPostprune] {expres/multishot/rescircleMultishot.tsv};
		
		\addplot+[forget plot, eda errorbarcolored, y dir=plus, y explicit]
		table[x expr=\thisrow{magSparsity}*1.1, y=randMeanPostprune, y error=randErrorPlusPostprune] {expres/multishot/rescircleMultishot.tsv};
		\addplot+[eda errorbarcolored, y dir=minus, y explicit]
		table[x expr=\thisrow{magSparsity}*1.1, y=randMeanPostprune, y error=randErrorMinusPostprune] {expres/multishot/rescircleMultishot.tsv};
		
		\addplot+[forget plot, eda errorbarcolored, y dir=plus, y explicit]
		table[x expr=\thisrow{magSparsity}*1.15, y=snipMeanPostprune, y error=snipErrorPlusPostprune] {expres/multishot/rescircleMultishot.tsv};
		\addplot+[eda errorbarcolored, y dir=minus, y explicit]
		table[x expr=\thisrow{magSparsity}*1.15, y=snipMeanPostprune, y error=snipErrorMinusPostprune] {expres/multishot/rescircleMultishot.tsv};
		
		\addplot+[forget plot, eda errorbarcolored, y dir=plus, y explicit]
		table[x expr=\thisrow{magSparsity}*1.2, y=graspMeanPostprune, y error=graspErrorPlusPostprune] {expres/multishot/rescircleMultishot.tsv};
		\addplot+[eda errorbarcolored, y dir=minus, y explicit]
		table[x expr=\thisrow{magSparsity}*1.2, y=graspMeanPostprune, y error=graspErrorMinusPostprune] {expres/multishot/rescircleMultishot.tsv};
		
		\addplot+[forget plot, eda errorbarcolored, y dir=plus, y explicit]
		table[x expr=\thisrow{magSparsity}*1.25, y=edgepopupMeanPostprune, y error=edgepopupErrorPlusPostprune] {expres/multishot/rescircleMultishot.tsv};
		\addplot+[eda errorbarcolored, y dir=minus, y explicit]
		table[x expr=\thisrow{magSparsity}*1.25, y=edgepopupMeanPostprune, y error=edgepopupErrorMinusPostprune] {expres/multishot/rescircleMultishot.tsv};
		
		\addplot+[forget plot, dotted, crimson, eda errorbarcolored, y dir=plus, y explicit]
		table[x expr=\thisrow{magSparsity}*1.3, y=edgepopup_plusMeanPostprune, y error=edgepopup_plusErrorPlusPostprune] {expres/multishot/rescircleMultishot.tsv};
		\addplot+[dotted, crimson, eda errorbarcolored, y dir=minus, y explicit]
		table[x expr=\thisrow{magSparsity}*1.3, y=edgepopup_plusMeanPostprune, y error=edgepopup_plusErrorMinusPostprune] {expres/multishot/rescircleMultishot.tsv};
		
        \addplot[mark=none, dashed, black, samples=2] coordinates {(0.004,0.99) (1.0,0.99)};
		
		\end{axis}
		\end{tikzpicture}
		\fi
    \end{subfigure}
    \begin{subfigure}[t]{0.49\textwidth}
    \centering
		\ifpdf
		\begin{tikzpicture}
		\begin{axis}[
		jonas line,
		xmode=log,
        y axis line style 	= {opacity=0},
        y tick style = {draw=none},
        yticklabels={,,},
		width = 7cm,
		height = 3.5cm,
		xlabel = {}, 
		xmin=0.004, xmax=1.1,
		ymin=0, ymax=1,
		legend columns=2,
		x label style 		= {at={(axis description cs:0.5,-0.1)}, anchor=north, font=\scriptsize},
		y label style 		= {at={(axis description cs:0.0,0.6)},  anchor=south, font=\scriptsize},
		legend style={nodes={scale=0.9, transform shape}, at={(0.98,0.95)}, anchor=north east, row sep=-1.4pt, font=\tiny}
		]
		
		\addplot+[forget plot,eda errorbarcolored, y dir=plus, y explicit]
		table[x=magSparsity, y=magMeanPosttrain, y error=magErrorPlusPosttrain] {expres/multishot/rescircleMultishot.tsv};
		\addplot+[eda errorbarcolored, y dir=minus, y explicit]
		table[x=magSparsity, y=magMeanPosttrain, y error=magErrorMinusPosttrain] {expres/multishot/rescircleMultishot.tsv};
		
		\addplot+[forget plot, eda errorbarcolored, y dir=plus, y explicit]
		table[x expr=\thisrow{magSparsity}*1.05, y=synflowMeanPosttrain, y error=synflowErrorPlusPosttrain] {expres/multishot/rescircleMultishot.tsv};
		\addplot+[eda errorbarcolored, y dir=minus, y explicit]
		table[x expr=\thisrow{magSparsity}*1.05, y=synflowMeanPosttrain, y error=synflowErrorMinusPosttrain] {expres/multishot/rescircleMultishot.tsv};
		
		\addplot+[forget plot, eda errorbarcolored, y dir=plus, y explicit]
		table[x expr=\thisrow{magSparsity}*1.1, y=randMeanPosttrain, y error=randErrorPlusPosttrain] {expres/multishot/rescircleMultishot.tsv};
		\addplot+[eda errorbarcolored, y dir=minus, y explicit]
		table[x expr=\thisrow{magSparsity}*1.1, y=randMeanPosttrain, y error=randErrorMinusPosttrain] {expres/multishot/rescircleMultishot.tsv};
		
		\addplot+[forget plot, eda errorbarcolored, y dir=plus, y explicit]
		table[x expr=\thisrow{magSparsity}*1.15, y=snipMeanPosttrain, y error=snipErrorPlusPosttrain] {expres/multishot/rescircleMultishot.tsv};
		\addplot+[eda errorbarcolored, y dir=minus, y explicit]
		table[x expr=\thisrow{magSparsity}*1.15, y=snipMeanPosttrain, y error=snipErrorMinusPosttrain] {expres/multishot/rescircleMultishot.tsv};
		
		\addplot+[forget plot, eda errorbarcolored, y dir=plus, y explicit]
		table[x expr=\thisrow{magSparsity}*1.2, y=graspMeanPosttrain, y error=graspErrorPlusPosttrain] {expres/multishot/rescircleMultishot.tsv};
		\addplot+[eda errorbarcolored, y dir=minus, y explicit]
		table[x expr=\thisrow{magSparsity}*1.2, y=graspMeanPosttrain, y error=graspErrorMinusPosttrain] {expres/multishot/rescircleMultishot.tsv};

        \addplot[mark=none, dashed, black, samples=2] coordinates {(0.004,0.99) (1.0,0.99)};
		
		\end{axis}
		\end{tikzpicture}
		\fi
    \end{subfigure}
    \begin{subfigure}[t]{0.49\textwidth}
    \centering
		\ifpdf
		\begin{tikzpicture}
		\begin{axis}[
		jonas line,
		ytick={0.0001,0.001,0.01,0.1,1},
		xmode=log,
		ymode=log,
		width = 7cm,
		height = 3.5cm,
		xlabel = {}, 
		ylabel= {MSE},
		xmin=0.0001, xmax=1.1,
		ymax=1,
		legend columns=2,
		x label style 		= {at={(axis description cs:0.5,-0.1)}, anchor=north, font=\scriptsize},
		y label style 		= {at={(axis description cs:0.0,0.6)},  anchor=south, font=\scriptsize},
		legend style={nodes={scale=0.9, transform shape}, at={(0.5,0.6)}, anchor=north, row sep=-1.4pt, font=\tiny}
		]
		
		\addplot+[forget plot, eda errorbarcolored, y dir=plus, y explicit]
		table[x=magSparsity, y=magMeanPostprune, y error=magErrorPlusPostprune] {expres/multishot/resreluMultishot.tsv};
		\addplot+[eda errorbarcolored, y dir=minus, y explicit]
		table[x=magSparsity, y=magMeanPostprune, y error=magErrorMinusPostprune] {expres/multishot/resreluMultishot.tsv};
		
		\addplot+[forget plot, eda errorbarcolored, y dir=plus, y explicit]
		table[x expr=\thisrow{magSparsity}*1.05, y=synflowMeanPostprune, y error=synflowErrorPlusPostprune] {expres/multishot/resreluMultishot.tsv};
		\addplot+[eda errorbarcolored, y dir=minus, y explicit]
		table[x expr=\thisrow{magSparsity}*1.05, y=synflowMeanPostprune, y error=synflowErrorMinusPostprune] {expres/multishot/resreluMultishot.tsv};
		
		\addplot+[forget plot, eda errorbarcolored, y dir=plus, y explicit]
		table[x expr=\thisrow{magSparsity}*1.1, y=randMeanPostprune, y error=randErrorPlusPostprune] {expres/multishot/resreluMultishot.tsv};
		\addplot+[eda errorbarcolored, y dir=minus, y explicit]
		table[x expr=\thisrow{magSparsity}*1.1, y=randMeanPostprune, y error=randErrorMinusPostprune] {expres/multishot/resreluMultishot.tsv};
		
		\addplot+[forget plot, eda errorbarcolored, y dir=plus, y explicit]
		table[x expr=\thisrow{magSparsity}*1.15, y=snipMeanPostprune, y error=snipErrorPlusPostprune] {expres/multishot/resreluMultishot.tsv};
		\addplot+[eda errorbarcolored, y dir=minus, y explicit]
		table[x expr=\thisrow{magSparsity}*1.15, y=snipMeanPostprune, y error=snipErrorMinusPostprune] {expres/multishot/resreluMultishot.tsv};
		
		\addplot+[forget plot, eda errorbarcolored, y dir=plus, y explicit]
		table[x expr=\thisrow{magSparsity}*1.2, y=graspMeanPostprune, y error=graspErrorPlusPostprune] {expres/multishot/resreluMultishot.tsv};
		\addplot+[eda errorbarcolored, y dir=minus, y explicit]
		table[x expr=\thisrow{magSparsity}*1.2, y=graspMeanPostprune, y error=graspErrorMinusPostprune] {expres/multishot/resreluMultishot.tsv};
		
		\addplot+[forget plot, eda errorbarcolored, y dir=plus, y explicit]
		table[x expr=\thisrow{magSparsity}*1.25, y=edgepopupMeanPostprune, y error=edgepopupErrorPlusPostprune] {expres/multishot/resreluMultishot.tsv};
		\addplot+[eda errorbarcolored, y dir=minus, y explicit]
		table[x expr=\thisrow{magSparsity}*1.25, y=edgepopupMeanPostprune, y error=edgepopupErrorMinusPostprune] {expres/multishot/resreluMultishot.tsv};
		
		\addplot+[forget plot, dotted, crimson, eda errorbarcolored, y dir=plus, y explicit]
		table[x expr=\thisrow{magSparsity}*1.3, y=edgepopup_plusMeanPostprune, y error=edgepopup_plusErrorPlusPostprune] {expres/multishot/resreluMultishot.tsv};
		\addplot+[dotted, crimson,eda errorbarcolored, y dir=minus, y explicit]
		table[x expr=\thisrow{magSparsity}*1.3, y=edgepopup_plusMeanPostprune, y error=edgepopup_plusErrorMinusPostprune] {expres/multishot/resreluMultishot.tsv};
		
        \addplot[mark=none, dashed, black, samples=2] coordinates {(0.0001,0.00010) (1.0,0.00010)};
		
		\end{axis}
		\end{tikzpicture}
		\fi
    \end{subfigure}
    \begin{subfigure}[t]{0.49\textwidth}
    \centering
		\ifpdf
		\begin{tikzpicture}
		\begin{axis}[
		jonas line,
		ytick={0.0001,0.001,0.01,0.1,1},
		xmode=log,
		ymode=log,
        y axis line style 	= {opacity=0},
        y tick style = {draw=none},
        yticklabels={,,},
		width = 7cm,
		height = 3.5cm,
		xlabel = {}, 
		xmin=0.0001, xmax=1.1,
		ymax=1,
		legend columns=2,
		x label style 		= {at={(axis description cs:0.5,-0.1)}, anchor=north, font=\scriptsize},
		y label style 		= {at={(axis description cs:0.0,0.6)},  anchor=south, font=\scriptsize},
		legend style={nodes={scale=0.9, transform shape}, at={(0.98,0.95)}, anchor=north east, row sep=-1.4pt, font=\tiny}
		]
		
		\addplot+[forget plot,eda errorbarcolored, y dir=plus, y explicit]
		table[x=magSparsity, y=magMeanPosttrain, y error=magErrorPlusPosttrain] {expres/multishot/resreluMultishot.tsv};
		\addplot+[eda errorbarcolored, y dir=minus, y explicit]
		table[x=magSparsity, y=magMeanPosttrain, y error=magErrorMinusPosttrain] {expres/multishot/resreluMultishot.tsv};
		
		\addplot+[forget plot, eda errorbarcolored, y dir=plus, y explicit]
		table[x expr=\thisrow{magSparsity}*1.05, y=synflowMeanPosttrain, y error=synflowErrorPlusPosttrain] {expres/multishot/resreluMultishot.tsv};
		\addplot+[eda errorbarcolored, y dir=minus, y explicit]
		table[x expr=\thisrow{magSparsity}*1.05, y=synflowMeanPosttrain, y error=synflowErrorMinusPosttrain] {expres/multishot/resreluMultishot.tsv};
		
		\addplot+[forget plot, eda errorbarcolored, y dir=plus, y explicit]
		table[x expr=\thisrow{magSparsity}*1.1, y=randMeanPosttrain, y error=randErrorPlusPosttrain] {expres/multishot/resreluMultishot.tsv};
		\addplot+[eda errorbarcolored, y dir=minus, y explicit]
		table[x expr=\thisrow{magSparsity}*1.1, y=randMeanPosttrain, y error=randErrorMinusPosttrain] {expres/multishot/resreluMultishot.tsv};
		
		\addplot+[forget plot, eda errorbarcolored, y dir=plus, y explicit]
		table[x expr=\thisrow{magSparsity}*1.15, y=snipMeanPosttrain, y error=snipErrorPlusPosttrain] {expres/multishot/resreluMultishot.tsv};
		\addplot+[eda errorbarcolored, y dir=minus, y explicit]
		table[x expr=\thisrow{magSparsity}*1.15, y=snipMeanPosttrain, y error=snipErrorMinusPosttrain] {expres/multishot/resreluMultishot.tsv};
		
		\addplot+[forget plot, eda errorbarcolored, y dir=plus, y explicit]
		table[x expr=\thisrow{magSparsity}*1.2, y=graspMeanPosttrain, y error=graspErrorPlusPosttrain] {expres/multishot/resreluMultishot.tsv};
		\addplot+[eda errorbarcolored, y dir=minus, y explicit]
		table[x expr=\thisrow{magSparsity}*1.2, y=graspMeanPosttrain, y error=graspErrorMinusPosttrain] {expres/multishot/resreluMultishot.tsv};
		
        \addplot[mark=none, dashed, black, samples=2] coordinates {(0.0001,0.00010) (1.0,0.00010)};
		
		\end{axis}
		\end{tikzpicture}
		\fi
    \end{subfigure}
    \begin{subfigure}[t]{0.49\textwidth}
    \centering
		\ifpdf
		\begin{tikzpicture}
		\begin{axis}[
		jonas line,
		xmode=log,
		ymode=log,
		minor x tick num = 5,
		width = 7cm,
		height = 3.5cm,
		xlabel = {Sparsity}, 
		ylabel= {MSE},
		xmin=0.004, xmax=1.1,
		ymax=1,
		legend columns=2,
		x label style 		= {at={(axis description cs:0.5,-0.1)}, anchor=north, font=\scriptsize},
		y label style 		= {at={(axis description cs:0.0,0.6)},  anchor=south, font=\scriptsize},
		legend style={nodes={scale=0.9, transform shape}, at={(0.01,0.01)}, anchor=south west, row sep=-1.4pt, font=\tiny}
		]
		
		\addplot+[forget plot, eda errorbarcolored, y dir=plus, y explicit]
		table[x=magSparsity, y=magMeanPostprune, y error=magErrorPlusPostprune] {expres/multishot/reshelixMultishot.tsv};
		\addplot+[eda errorbarcolored, y dir=minus, y explicit]
		table[x=magSparsity, y=magMeanPostprune, y error=magErrorMinusPostprune] {expres/multishot/reshelixMultishot.tsv};
		\addlegendentry{\magnitude}
		
		\addplot+[forget plot, eda errorbarcolored, y dir=plus, y explicit]
		table[x expr=\thisrow{magSparsity}*1.05, y=synflowMeanPostprune, y error=synflowErrorPlusPostprune] {expres/multishot/reshelixMultishot.tsv};
		\addplot+[eda errorbarcolored, y dir=minus, y explicit]
		table[x expr=\thisrow{magSparsity}*1.05, y=synflowMeanPostprune, y error=synflowErrorMinusPostprune] {expres/multishot/reshelixMultishot.tsv};
		\addlegendentry{\synflow}
		
		\addplot+[forget plot, eda errorbarcolored, y dir=plus, y explicit]
		table[x expr=\thisrow{magSparsity}*1.1, y=randMeanPostprune, y error=randErrorPlusPostprune] {expres/multishot/reshelixMultishot.tsv};
		\addplot+[eda errorbarcolored, y dir=minus, y explicit]
		table[x expr=\thisrow{magSparsity}*1.1, y=randMeanPostprune, y error=randErrorMinusPostprune] {expres/multishot/reshelixMultishot.tsv};
		\addlegendentry{\rand}
		
		\addplot+[forget plot, eda errorbarcolored, y dir=plus, y explicit]
		table[x expr=\thisrow{magSparsity}*1.15, y=snipMeanPostprune, y error=snipErrorPlusPostprune] {expres/multishot/reshelixMultishot.tsv};
		\addplot+[eda errorbarcolored, y dir=minus, y explicit]
		table[x expr=\thisrow{magSparsity}*1.15, y=snipMeanPostprune, y error=snipErrorMinusPostprune] {expres/multishot/reshelixMultishot.tsv};
		\addlegendentry{\snip}
		
		\addplot+[forget plot, eda errorbarcolored, y dir=plus, y explicit]
		table[x expr=\thisrow{magSparsity}*1.2, y=graspMeanPostprune, y error=graspErrorPlusPostprune] {expres/multishot/reshelixMultishot.tsv};
		\addplot+[eda errorbarcolored, y dir=minus, y explicit]
		table[x expr=\thisrow{magSparsity}*1.2, y=graspMeanPostprune, y error=graspErrorMinusPostprune] {expres/multishot/reshelixMultishot.tsv};
		\addlegendentry{\grasp}
		
		\addplot+[forget plot, eda errorbarcolored, y dir=plus, y explicit]
		table[x expr=\thisrow{magSparsity}*1.25, y=edgepopupMeanPostprune, y error=edgepopupErrorPlusPostprune] {expres/multishot/reshelixMultishot.tsv};
		\addplot+[eda errorbarcolored, y dir=minus, y explicit]
		table[x expr=\thisrow{magSparsity}*1.25, y=edgepopupMeanPostprune, y error=edgepopupErrorMinusPostprune] {expres/multishot/reshelixMultishot.tsv};
		\addlegendentry{\edgepopup}
		
		\addplot+[dotted, crimson, forget plot, eda errorbarcolored, y dir=plus, y explicit]
		table[x expr=\thisrow{magSparsity}*1.3, y=edgepopup_plusMeanPostprune, y error=edgepopup_plusErrorPlusPostprune] {expres/multishot/reshelixMultishot.tsv};
		\addplot+[dotted, crimson, eda errorbarcolored, y dir=minus, y explicit]
		table[x expr=\thisrow{magSparsity}*1.3, y=edgepopup_plusMeanPostprune, y error=edgepopup_plusErrorMinusPostprune] {expres/multishot/reshelixMultishot.tsv};
		\addlegendentry{\edgepopupscaled}
		
        \addplot[mark=none, dashed, black, samples=2] coordinates {(0.004,0.00031) (1.0,0.00031)};
        \addlegendentry{True Ticket}
		
		\end{axis}
		\end{tikzpicture}
		\fi
    \end{subfigure}
    \begin{subfigure}[t]{0.49\textwidth}
    \centering
		\ifpdf
		\begin{tikzpicture}
		\begin{axis}[
		jonas line,
		xmode=log,
		ymode=log,
		minor x tick num = 5,
        y axis line style 	= {opacity=0},
        y tick style = {draw=none},
        yticklabels={,,},
		width = 7cm,
		height = 3.5cm,
		xlabel = {Sparsity}, 
		xmin=0.004, xmax=1.1,
		ymax=1,
		legend columns=2,
		x label style 		= {at={(axis description cs:0.5,-0.1)}, anchor=north, font=\scriptsize},
		y label style 		= {at={(axis description cs:0.0,0.6)},  anchor=south, font=\scriptsize},
		legend style={nodes={scale=0.9, transform shape}, at={(0.98,0.95)}, anchor=north east, row sep=-1.4pt, font=\tiny}
		]
		
		\addplot+[forget plot,eda errorbarcolored, y dir=plus, y explicit]
		table[x=magSparsity, y=magMeanPosttrain, y error=magErrorPlusPosttrain] {expres/multishot/reshelixMultishot.tsv};
		\addplot+[eda errorbarcolored, y dir=minus, y explicit]
		table[x=magSparsity, y=magMeanPosttrain, y error=magErrorMinusPosttrain] {expres/multishot/reshelixMultishot.tsv};
		
		\addplot+[forget plot, eda errorbarcolored, y dir=plus, y explicit]
		table[x expr=\thisrow{magSparsity}*1.05, y=synflowMeanPosttrain, y error=synflowErrorPlusPosttrain] {expres/multishot/reshelixMultishot.tsv};
		\addplot+[eda errorbarcolored, y dir=minus, y explicit]
		table[x expr=\thisrow{magSparsity}*1.05, y=synflowMeanPosttrain, y error=synflowErrorMinusPosttrain] {expres/multishot/reshelixMultishot.tsv};
		
		\addplot+[forget plot, eda errorbarcolored, y dir=plus, y explicit]
		table[x expr=\thisrow{magSparsity}*1.1, y=randMeanPosttrain, y error=randErrorPlusPosttrain] {expres/multishot/reshelixMultishot.tsv};
		\addplot+[eda errorbarcolored, y dir=minus, y explicit]
		table[x expr=\thisrow{magSparsity}*1.1, y=randMeanPosttrain, y error=randErrorMinusPosttrain] {expres/multishot/reshelixMultishot.tsv};
		
		\addplot+[forget plot, eda errorbarcolored, y dir=plus, y explicit]
		table[x expr=\thisrow{magSparsity}*1.15, y=snipMeanPosttrain, y error=snipErrorPlusPosttrain] {expres/multishot/reshelixMultishot.tsv};
		\addplot+[eda errorbarcolored, y dir=minus, y explicit]
		table[x expr=\thisrow{magSparsity}*1.15, y=snipMeanPosttrain, y error=snipErrorMinusPosttrain] {expres/multishot/reshelixMultishot.tsv};
		
		\addplot+[forget plot, eda errorbarcolored, y dir=plus, y explicit]
		table[x expr=\thisrow{magSparsity}*1.2, y=graspMeanPosttrain, y error=graspErrorPlusPosttrain] {expres/multishot/reshelixMultishot.tsv};
		\addplot+[eda errorbarcolored, y dir=minus, y explicit]
		table[x expr=\thisrow{magSparsity}*1.2, y=graspMeanPosttrain, y error=graspErrorMinusPosttrain] {expres/multishot/reshelixMultishot.tsv};
		
        \addplot[mark=none, dashed, black, samples=2] coordinates {(0.004,0.00031) (1.0,0.00031)};
		
		\end{axis}
		\end{tikzpicture}
		\fi
    \end{subfigure}
    \caption{\textit{Multishot results.} Performance for \texttt{Circle} (top), \texttt{ReLU} (middle), and \texttt{Helix} (bottom) for $10$ rounds of alternating pruning and training. We provide mean and obtained intervals (minimum and maximum) of accuracies of the final pruned network across $10$ repetitions before (strong ticket, left) and after (weak ticket, right)  final training. Baseline ticket accuracy is indicated in black.}\label{fig:multishot}
\end{figure}
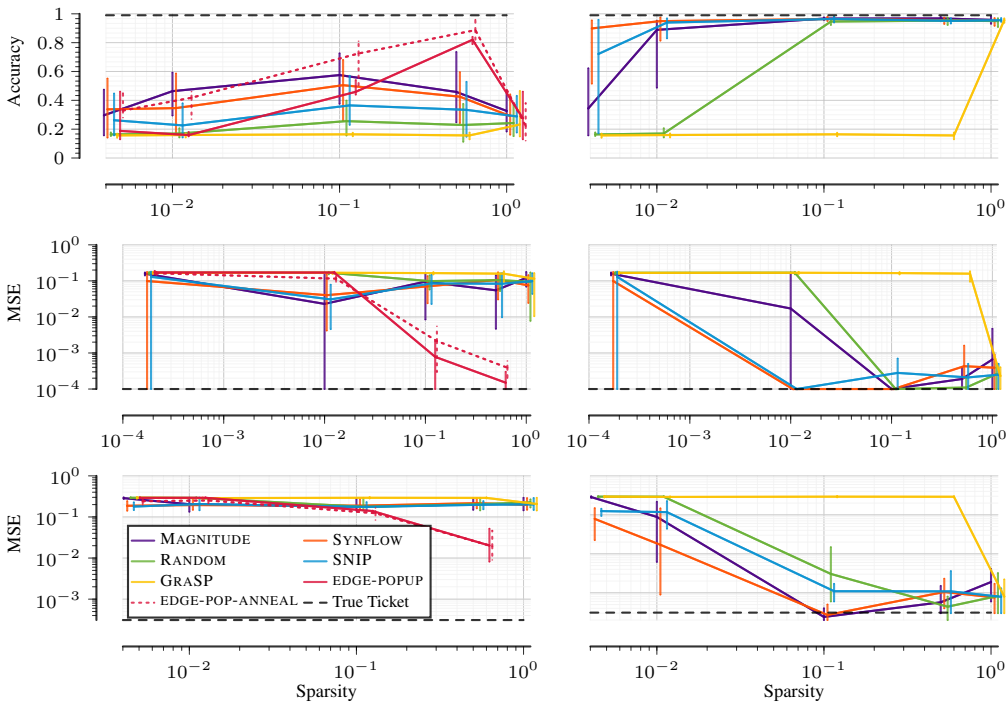

%% file: figs/results_6_circle_postprune.tikz
\begin{tikzpicture}[x=1pt,y=1pt]
\definecolor{fillColor}{RGB}{255,255,255}
\begin{scope}
\definecolor{drawColor}{RGB}{255,0,255}

\path[draw=drawColor,line width= 0.6pt,line join=round] ( 64.36, 38.85) --
	( 64.45, 38.85);

\path[draw=drawColor,line width= 0.6pt,line join=round] ( 64.41, 38.85) --
	( 64.41, 29.95);

\path[draw=drawColor,line width= 0.6pt,line join=round] ( 64.36, 29.95) --
	( 64.45, 29.95);

\path[draw=drawColor,line width= 0.6pt,line join=round] (117.61, 46.70) --
	(117.70, 46.70);

\path[draw=drawColor,line width= 0.6pt,line join=round] (117.65, 46.70) --
	(117.65, 36.70);

\path[draw=drawColor,line width= 0.6pt,line join=round] (117.61, 36.70) --
	(117.70, 36.70);

\path[draw=drawColor,line width= 0.6pt,line join=round] (154.83, 42.32) --
	(154.91, 42.32);

\path[draw=drawColor,line width= 0.6pt,line join=round] (154.87, 42.32) --
	(154.87, 33.75);

\path[draw=drawColor,line width= 0.6pt,line join=round] (154.83, 33.75) --
	(154.91, 33.75);

\path[draw=drawColor,line width= 0.6pt,line join=round] (170.85, 34.86) --
	(170.94, 34.86);

\path[draw=drawColor,line width= 0.6pt,line join=round] (170.90, 34.86) --
	(170.90, 28.90);

\path[draw=drawColor,line width= 0.6pt,line join=round] (170.85, 28.90) --
	(170.94, 28.90);

\path[draw=drawColor,line width= 0.6pt,line join=round] ( 42.52, 39.39) --
	( 42.61, 39.39);

\path[draw=drawColor,line width= 0.6pt,line join=round] ( 42.57, 39.39) --
	( 42.57, 28.62);

\path[draw=drawColor,line width= 0.6pt,line join=round] ( 42.52, 28.62) --
	( 42.61, 28.62);
\definecolor{drawColor}{RGB}{178,34,34}

\path[draw=drawColor,line width= 0.6pt,line join=round] ( 64.45, 26.12) --
	( 64.54, 26.12);

\path[draw=drawColor,line width= 0.6pt,line join=round] ( 64.50, 26.12) --
	( 64.50, 25.53);

\path[draw=drawColor,line width= 0.6pt,line join=round] ( 64.45, 25.53) --
	( 64.54, 25.53);

\path[draw=drawColor,line width= 0.6pt,line join=round] (117.70, 26.27) --
	(117.79, 26.27);

\path[draw=drawColor,line width= 0.6pt,line join=round] (117.74, 26.27) --
	(117.74, 25.83);

\path[draw=drawColor,line width= 0.6pt,line join=round] (117.70, 25.83) --
	(117.79, 25.83);

\path[draw=drawColor,line width= 0.6pt,line join=round] (154.91, 26.16) --
	(155.00, 26.16);

\path[draw=drawColor,line width= 0.6pt,line join=round] (154.96, 26.16) --
	(154.96, 25.23);

\path[draw=drawColor,line width= 0.6pt,line join=round] (154.91, 25.23) --
	(155.00, 25.23);

\path[draw=drawColor,line width= 0.6pt,line join=round] (170.94, 32.25) --
	(171.03, 32.25);

\path[draw=drawColor,line width= 0.6pt,line join=round] (170.99, 32.25) --
	(170.99, 26.10);

\path[draw=drawColor,line width= 0.6pt,line join=round] (170.94, 26.10) --
	(171.03, 26.10);

\path[draw=drawColor,line width= 0.6pt,line join=round] ( 42.61, 26.05) --
	( 42.70, 26.05);

\path[draw=drawColor,line width= 0.6pt,line join=round] ( 42.66, 26.05) --
	( 42.66, 25.37);

\path[draw=drawColor,line width= 0.6pt,line join=round] ( 42.61, 25.37) --
	( 42.70, 25.37);
\definecolor{drawColor}{RGB}{67,205,128}

\path[draw=drawColor,line width= 0.6pt,line join=round] ( 64.54, 31.86) --
	( 64.63, 31.86);

\path[draw=drawColor,line width= 0.6pt,line join=round] ( 64.58, 31.86) --
	( 64.58, 25.93);

\path[draw=drawColor,line width= 0.6pt,line join=round] ( 64.54, 25.93) --
	( 64.63, 25.93);

\path[draw=drawColor,line width= 0.6pt,line join=round] (117.79, 39.46) --
	(117.87, 39.46);

\path[draw=drawColor,line width= 0.6pt,line join=round] (117.83, 39.46) --
	(117.83, 31.06);

\path[draw=drawColor,line width= 0.6pt,line join=round] (117.79, 31.06) --
	(117.87, 31.06);

\path[draw=drawColor,line width= 0.6pt,line join=round] (155.00, 37.91) --
	(155.09, 37.91);

\path[draw=drawColor,line width= 0.6pt,line join=round] (155.05, 37.91) --
	(155.05, 29.79);

\path[draw=drawColor,line width= 0.6pt,line join=round] (155.00, 29.79) --
	(155.09, 29.79);

\path[draw=drawColor,line width= 0.6pt,line join=round] (171.03, 35.27) --
	(171.12, 35.27);

\path[draw=drawColor,line width= 0.6pt,line join=round] (171.08, 35.27) --
	(171.08, 28.23);

\path[draw=drawColor,line width= 0.6pt,line join=round] (171.03, 28.23) --
	(171.12, 28.23);

\path[draw=drawColor,line width= 0.6pt,line join=round] ( 42.70, 33.78) --
	( 42.79, 33.78);

\path[draw=drawColor,line width= 0.6pt,line join=round] ( 42.74, 33.78) --
	( 42.74, 27.29);

\path[draw=drawColor,line width= 0.6pt,line join=round] ( 42.70, 27.29) --
	( 42.79, 27.29);
\definecolor{drawColor}{RGB}{218,165,32}

\path[draw=drawColor,line width= 0.6pt,line join=round] ( 64.63, 43.15) --
	( 64.72, 43.15);

\path[draw=drawColor,line width= 0.6pt,line join=round] ( 64.67, 43.15) --
	( 64.67, 36.44);

\path[draw=drawColor,line width= 0.6pt,line join=round] ( 64.63, 36.44) --
	( 64.72, 36.44);

\path[draw=drawColor,line width= 0.6pt,line join=round] (117.87, 48.84) --
	(117.96, 48.84);

\path[draw=drawColor,line width= 0.6pt,line join=round] (117.92, 48.84) --
	(117.92, 41.00);

\path[draw=drawColor,line width= 0.6pt,line join=round] (117.87, 41.00) --
	(117.96, 41.00);

\path[draw=drawColor,line width= 0.6pt,line join=round] (155.09, 45.10) --
	(155.18, 45.10);

\path[draw=drawColor,line width= 0.6pt,line join=round] (155.14, 45.10) --
	(155.14, 34.00);

\path[draw=drawColor,line width= 0.6pt,line join=round] (155.09, 34.00) --
	(155.18, 34.00);

\path[draw=drawColor,line width= 0.6pt,line join=round] (171.12, 36.99) --
	(171.21, 36.99);

\path[draw=drawColor,line width= 0.6pt,line join=round] (171.16, 36.99) --
	(171.16, 30.02);

\path[draw=drawColor,line width= 0.6pt,line join=round] (171.12, 30.02) --
	(171.21, 30.02);

\path[draw=drawColor,line width= 0.6pt,line join=round] ( 42.79, 35.46) --
	( 42.88, 35.46);

\path[draw=drawColor,line width= 0.6pt,line join=round] ( 42.83, 35.46) --
	( 42.83, 28.75);

\path[draw=drawColor,line width= 0.6pt,line join=round] ( 42.79, 28.75) --
	( 42.88, 28.75);
\definecolor{drawColor}{RGB}{0,0,139}

\path[draw=drawColor,line width= 0.6pt,line join=round] ( 64.72, 26.88) --
	( 64.81, 26.88);

\path[draw=drawColor,line width= 0.6pt,line join=round] ( 64.76, 26.88) --
	( 64.76, 25.73);

\path[draw=drawColor,line width= 0.6pt,line join=round] ( 64.72, 25.73) --
	( 64.81, 25.73);

\path[draw=drawColor,line width= 0.6pt,line join=round] (117.96, 32.69) --
	(118.05, 32.69);

\path[draw=drawColor,line width= 0.6pt,line join=round] (118.01, 32.69) --
	(118.01, 27.81);

\path[draw=drawColor,line width= 0.6pt,line join=round] (117.96, 27.81) --
	(118.05, 27.81);

\path[draw=drawColor,line width= 0.6pt,line join=round] (155.18, 31.32) --
	(155.27, 31.32);

\path[draw=drawColor,line width= 0.6pt,line join=round] (155.22, 31.32) --
	(155.22, 26.71);

\path[draw=drawColor,line width= 0.6pt,line join=round] (155.18, 26.71) --
	(155.27, 26.71);

\path[draw=drawColor,line width= 0.6pt,line join=round] (171.21, 32.44) --
	(171.30, 32.44);

\path[draw=drawColor,line width= 0.6pt,line join=round] (171.25, 32.44) --
	(171.25, 26.85);

\path[draw=drawColor,line width= 0.6pt,line join=round] (171.21, 26.85) --
	(171.30, 26.85);

\path[draw=drawColor,line width= 0.6pt,line join=round] ( 42.88, 26.31) --
	( 42.97, 26.31);

\path[draw=drawColor,line width= 0.6pt,line join=round] ( 42.92, 26.31) --
	( 42.92, 25.72);

\path[draw=drawColor,line width= 0.6pt,line join=round] ( 42.88, 25.72) --
	( 42.97, 25.72);
\definecolor{fillColor}{RGB}{255,0,255}

\path[fill=fillColor,fill opacity=0.80] ( 62.67, 32.44) --
	( 66.59, 32.44) --
	( 66.59, 36.36) --
	( 62.67, 36.36) --
	cycle;

\path[fill=fillColor,fill opacity=0.80] (115.91, 39.74) --
	(119.84, 39.74) --
	(119.84, 43.66) --
	(115.91, 43.66) --
	cycle;

\path[fill=fillColor,fill opacity=0.80] (153.13, 36.07) --
	(157.05, 36.07) --
	(157.05, 40.00) --
	(153.13, 40.00) --
	cycle;

\path[fill=fillColor,fill opacity=0.80] (169.16, 29.91) --
	(173.08, 29.91) --
	(173.08, 33.84) --
	(169.16, 33.84) --
	cycle;

\path[fill=fillColor,fill opacity=0.80] ( 40.83, 32.04) --
	( 44.75, 32.04) --
	( 44.75, 35.97) --
	( 40.83, 35.97) --
	cycle;
\definecolor{drawColor}{RGB}{178,34,34}
\definecolor{fillColor}{RGB}{178,34,34}

\path[draw=drawColor,draw opacity=0.80,line width= 0.4pt,line join=round,line cap=round,fill=fillColor,fill opacity=0.80] ( 64.63, 22.78) --
	( 67.27, 27.35) --
	( 61.99, 27.35) --
	cycle;

\path[draw=drawColor,draw opacity=0.80,line width= 0.4pt,line join=round,line cap=round,fill=fillColor,fill opacity=0.80] (117.87, 23.00) --
	(120.52, 27.58) --
	(115.23, 27.58) --
	cycle;

\path[draw=drawColor,draw opacity=0.80,line width= 0.4pt,line join=round,line cap=round,fill=fillColor,fill opacity=0.80] (155.09, 22.64) --
	(157.73, 27.22) --
	(152.45, 27.22) --
	cycle;

\path[draw=drawColor,draw opacity=0.80,line width= 0.4pt,line join=round,line cap=round,fill=fillColor,fill opacity=0.80] (171.12, 26.12) --
	(173.76, 30.70) --
	(168.48, 30.70) --
	cycle;

\path[draw=drawColor,draw opacity=0.80,line width= 0.4pt,line join=round,line cap=round,fill=fillColor,fill opacity=0.80] ( 42.79, 22.66) --
	( 45.43, 27.24) --
	( 40.15, 27.24) --
	cycle;
\definecolor{drawColor}{RGB}{67,205,128}
\definecolor{fillColor}{RGB}{67,205,128}

\path[draw=drawColor,draw opacity=0.80,line width= 0.4pt,line join=round,line cap=round,fill=fillColor,fill opacity=0.80] ( 64.63, 26.44) --
	( 67.09, 28.90) --
	( 64.63, 31.36) --
	( 62.17, 28.90) --
	cycle;

\path[draw=drawColor,draw opacity=0.80,line width= 0.4pt,line join=round,line cap=round,fill=fillColor,fill opacity=0.80] (117.87, 32.80) --
	(120.33, 35.26) --
	(117.87, 37.72) --
	(115.42, 35.26) --
	cycle;

\path[draw=drawColor,draw opacity=0.80,line width= 0.4pt,line join=round,line cap=round,fill=fillColor,fill opacity=0.80] (155.09, 31.39) --
	(157.55, 33.85) --
	(155.09, 36.31) --
	(152.63, 33.85) --
	cycle;

\path[draw=drawColor,draw opacity=0.80,line width= 0.4pt,line join=round,line cap=round,fill=fillColor,fill opacity=0.80] (171.12, 29.29) --
	(173.58, 31.75) --
	(171.12, 34.21) --
	(168.66, 31.75) --
	cycle;

\path[draw=drawColor,draw opacity=0.80,line width= 0.4pt,line join=round,line cap=round,fill=fillColor,fill opacity=0.80] ( 42.79, 28.08) --
	( 45.25, 30.54) --
	( 42.79, 32.99) --
	( 40.33, 30.54) --
	cycle;
\definecolor{drawColor}{RGB}{218,165,32}
\definecolor{fillColor}{RGB}{218,165,32}

\path[draw=drawColor,draw opacity=0.80,line width= 0.4pt,line join=round,line cap=round,fill=fillColor,fill opacity=0.80] ( 64.63, 42.85) --
	( 67.27, 38.27) --
	( 61.99, 38.27) --
	cycle;

\path[draw=drawColor,draw opacity=0.80,line width= 0.4pt,line join=round,line cap=round,fill=fillColor,fill opacity=0.80] (117.87, 47.97) --
	(120.52, 43.39) --
	(115.23, 43.39) --
	cycle;

\path[draw=drawColor,draw opacity=0.80,line width= 0.4pt,line join=round,line cap=round,fill=fillColor,fill opacity=0.80] (155.09, 42.60) --
	(157.73, 38.02) --
	(152.45, 38.02) --
	cycle;

\path[draw=drawColor,draw opacity=0.80,line width= 0.4pt,line join=round,line cap=round,fill=fillColor,fill opacity=0.80] (171.12, 36.56) --
	(173.76, 31.98) --
	(168.48, 31.98) --
	cycle;

\path[draw=drawColor,draw opacity=0.80,line width= 0.4pt,line join=round,line cap=round,fill=fillColor,fill opacity=0.80] ( 42.79, 35.16) --
	( 45.43, 30.58) --
	( 40.15, 30.58) --
	cycle;
\definecolor{drawColor}{RGB}{0,0,139}
\definecolor{fillColor}{RGB}{0,0,139}

\path[draw=drawColor,draw opacity=0.80,line width= 0.4pt,line join=round,line cap=round,fill=fillColor,fill opacity=0.80] ( 64.63, 23.85) --
	( 67.09, 26.31) --
	( 64.63, 28.77) --
	( 62.17, 26.31) --
	cycle;

\path[draw=drawColor,draw opacity=0.80,line width= 0.4pt,line join=round,line cap=round,fill=fillColor,fill opacity=0.80] (117.87, 27.79) --
	(120.33, 30.25) --
	(117.87, 32.71) --
	(115.42, 30.25) --
	cycle;

\path[draw=drawColor,draw opacity=0.80,line width= 0.4pt,line join=round,line cap=round,fill=fillColor,fill opacity=0.80] (155.09, 26.56) --
	(157.55, 29.02) --
	(155.09, 31.47) --
	(152.63, 29.02) --
	cycle;

\path[draw=drawColor,draw opacity=0.80,line width= 0.4pt,line join=round,line cap=round,fill=fillColor,fill opacity=0.80] (171.12, 27.19) --
	(173.58, 29.65) --
	(171.12, 32.11) --
	(168.66, 29.65) --
	cycle;

\path[draw=drawColor,draw opacity=0.80,line width= 0.4pt,line join=round,line cap=round,fill=fillColor,fill opacity=0.80] ( 42.79, 23.56) --
	( 45.25, 26.02) --
	( 42.79, 28.48) --
	( 40.33, 26.02) --
	cycle;
\definecolor{drawColor}{RGB}{0,245,255}
\definecolor{fillColor}{RGB}{0,245,255}

\path[draw=drawColor,draw opacity=0.80,line width= 0.4pt,line join=round,line cap=round,fill=fillColor,fill opacity=0.80] ( 64.63, 25.81) circle (  1.96);

\path[draw=drawColor,draw opacity=0.80,line width= 0.4pt,line join=round,line cap=round,fill=fillColor,fill opacity=0.80] (117.87, 39.59) circle (  1.96);

\path[draw=drawColor,draw opacity=0.80,line width= 0.4pt,line join=round,line cap=round,fill=fillColor,fill opacity=0.80] (155.09, 56.16) circle (  1.96);

\path[draw=drawColor,draw opacity=0.80,line width= 0.4pt,line join=round,line cap=round,fill=fillColor,fill opacity=0.80] (171.12, 31.18) circle (  1.96);

\path[draw=drawColor,draw opacity=0.80,line width= 0.4pt,line join=round,line cap=round,fill=fillColor,fill opacity=0.80] ( 42.79, 27.14) circle (  1.96);
\definecolor{drawColor}{RGB}{255,0,255}

\path[draw=drawColor,draw opacity=0.80,line width= 0.5pt,line join=round] ( 42.79, 34.01) --
	( 64.63, 34.40) --
	(117.87, 41.70) --
	(155.09, 38.03) --
	(171.12, 31.88);
\definecolor{drawColor}{RGB}{178,34,34}

\path[draw=drawColor,draw opacity=0.80,line width= 0.5pt,line join=round] ( 42.79, 25.71) --
	( 64.63, 25.83) --
	(117.87, 26.05) --
	(155.09, 25.70) --
	(171.12, 29.18);
\definecolor{drawColor}{RGB}{67,205,128}

\path[draw=drawColor,draw opacity=0.80,line width= 0.5pt,line join=round] ( 42.79, 30.54) --
	( 64.63, 28.90) --
	(117.87, 35.26) --
	(155.09, 33.85) --
	(171.12, 31.75);
\definecolor{drawColor}{RGB}{218,165,32}

\path[draw=drawColor,draw opacity=0.80,line width= 0.5pt,line join=round] ( 42.79, 32.11) --
	( 64.63, 39.80) --
	(117.87, 44.92) --
	(155.09, 39.55) --
	(171.12, 33.51);
\definecolor{drawColor}{RGB}{0,0,139}

\path[draw=drawColor,draw opacity=0.80,line width= 0.5pt,line join=round] ( 42.79, 26.02) --
	( 64.63, 26.31) --
	(117.87, 30.25) --
	(155.09, 29.02) --
	(171.12, 29.65);
\definecolor{drawColor}{RGB}{0,245,255}

\path[draw=drawColor,draw opacity=0.80,line width= 0.5pt,line join=round] ( 42.79, 27.14) --
	( 64.63, 25.81) --
	(117.87, 39.59) --
	(155.09, 56.16) --
	(171.12, 31.18);

\path[draw=drawColor,draw opacity=0.80,line width= 0.4pt,line join=round,line cap=round,fill=fillColor,fill opacity=0.80] ( 64.63, 37.70) circle (  1.96);

\path[draw=drawColor,draw opacity=0.80,line width= 0.4pt,line join=round,line cap=round,fill=fillColor,fill opacity=0.80] (117.87, 51.80) circle (  1.96);

\path[draw=drawColor,draw opacity=0.80,line width= 0.4pt,line join=round,line cap=round,fill=fillColor,fill opacity=0.80] (155.09, 59.24) circle (  1.96);

\path[draw=drawColor,draw opacity=0.80,line width= 0.4pt,line join=round,line cap=round,fill=fillColor,fill opacity=0.80] (171.12, 28.70) circle (  1.96);

\path[draw=drawColor,draw opacity=0.80,line width= 0.4pt,line join=round,line cap=round,fill=fillColor,fill opacity=0.80] ( 42.79, 33.62) circle (  1.96);

\path[draw=drawColor,draw opacity=0.80,line width= 0.9pt,dash pattern=on 1pt off 3pt on 4pt off 3pt ,line join=round] ( 42.79, 33.62) --
	( 64.63, 37.70) --
	(117.87, 51.80) --
	(155.09, 59.24) --
	(171.12, 28.70);
\definecolor{drawColor}{RGB}{0,245,255}

\path[draw=drawColor,draw opacity=0.50,line width= 0.6pt,line join=round] ( 64.36, 39.60) --
	( 64.89, 39.60);

\path[draw=drawColor,draw opacity=0.50,line width= 0.6pt,line join=round] ( 64.63, 39.60) --
	( 64.63, 35.80);

\path[draw=drawColor,draw opacity=0.50,line width= 0.6pt,line join=round] ( 64.36, 35.80) --
	( 64.89, 35.80);

\path[draw=drawColor,draw opacity=0.50,line width= 0.6pt,line join=round] (117.61, 55.30) --
	(118.14, 55.30);

\path[draw=drawColor,draw opacity=0.50,line width= 0.6pt,line join=round] (117.87, 55.30) --
	(117.87, 48.30);

\path[draw=drawColor,draw opacity=0.50,line width= 0.6pt,line join=round] (117.61, 48.30) --
	(118.14, 48.30);

\path[draw=drawColor,draw opacity=0.50,line width= 0.6pt,line join=round] (154.83, 61.45) --
	(155.36, 61.45);

\path[draw=drawColor,draw opacity=0.50,line width= 0.6pt,line join=round] (155.09, 61.45) --
	(155.09, 57.03);

\path[draw=drawColor,draw opacity=0.50,line width= 0.6pt,line join=round] (154.83, 57.03) --
	(155.36, 57.03);

\path[draw=drawColor,draw opacity=0.50,line width= 0.6pt,line join=round] (170.85, 31.72) --
	(171.39, 31.72);

\path[draw=drawColor,draw opacity=0.50,line width= 0.6pt,line join=round] (171.12, 31.72) --
	(171.12, 25.68);

\path[draw=drawColor,draw opacity=0.50,line width= 0.6pt,line join=round] (170.85, 25.68) --
	(171.39, 25.68);

\path[draw=drawColor,draw opacity=0.50,line width= 0.6pt,line join=round] ( 42.52, 35.66) --
	( 43.05, 35.66);

\path[draw=drawColor,draw opacity=0.50,line width= 0.6pt,line join=round] ( 42.79, 35.66) --
	( 42.79, 31.58);

\path[draw=drawColor,draw opacity=0.50,line width= 0.6pt,line join=round] ( 42.52, 31.58) --
	( 43.05, 31.58);
\definecolor{drawColor}{RGB}{0,0,0}

\path[draw=drawColor,line width= 0.5pt,line join=round] ( 42.79, 63.43) --
	( 64.63, 63.37) --
	(117.87, 63.48) --
	(155.09, 63.43) --
	(171.12, 63.31);

\path[draw=drawColor,line width= 0.5pt,line join=round] ( 42.79, 63.53) --
	( 64.63, 63.50) --
	(117.87, 63.42) --
	(155.09, 63.21) --
	(171.12, 63.32);

\path[draw=drawColor,line width= 0.5pt,line join=round] ( 42.79, 63.44) --
	( 64.63, 63.37) --
	(117.87, 63.43) --
	(155.09, 63.29) --
	(171.12, 63.26);

\path[draw=drawColor,line width= 0.5pt,line join=round] ( 42.79, 63.46) --
	( 64.63, 63.52) --
	(117.87, 63.35) --
	(155.09, 63.27) --
	(171.12, 63.55);

\path[draw=drawColor,line width= 0.5pt,line join=round] ( 42.79, 63.36) --
	( 64.63, 63.43) --
	(117.87, 63.46) --
	(155.09, 63.40) --
	(171.12, 63.55);

\path[draw=drawColor,line width= 0.5pt,line join=round] ( 42.79, 63.36) --
	( 64.63, 63.43) --
	(117.87, 63.35) --
	(155.09, 63.33) --
	(171.12, 63.41);
\definecolor{drawColor}{gray}{0.60}

\path[draw=drawColor,line width= 0.5pt,line join=round,line cap=round] ( 36.79, 16.21) -- ( 36.79, 13.37);

\path[draw=drawColor,line width= 0.5pt,line join=round,line cap=round] ( 43.44, 16.21) -- ( 43.44, 13.37);

\path[draw=drawColor,line width= 0.5pt,line join=round,line cap=round] ( 48.60, 16.21) -- ( 48.60, 13.37);

\path[draw=drawColor,line width= 0.5pt,line join=round,line cap=round] ( 52.82, 16.21) -- ( 52.82, 13.37);

\path[draw=drawColor,line width= 0.5pt,line join=round,line cap=round] ( 56.38, 16.21) -- ( 56.38, 13.37);

\path[draw=drawColor,line width= 0.5pt,line join=round,line cap=round] ( 59.47, 16.21) -- ( 59.47, 13.37);

\path[draw=drawColor,line width= 0.5pt,line join=round,line cap=round] ( 62.19, 16.21) -- ( 62.19, 13.37);

\path[draw=drawColor,line width= 0.5pt,line join=round,line cap=round] ( 64.63, 16.21) -- ( 64.63, 10.52);

\path[draw=drawColor,line width= 0.5pt,line join=round,line cap=round] ( 80.66, 16.21) -- ( 80.66, 13.37);

\path[draw=drawColor,line width= 0.5pt,line join=round,line cap=round] ( 90.03, 16.21) -- ( 90.03, 13.37);

\path[draw=drawColor,line width= 0.5pt,line join=round,line cap=round] ( 96.69, 16.21) -- ( 96.69, 13.37);

\path[draw=drawColor,line width= 0.5pt,line join=round,line cap=round] (101.85, 16.21) -- (101.85, 13.37);

\path[draw=drawColor,line width= 0.5pt,line join=round,line cap=round] (106.06, 16.21) -- (106.06, 13.37);

\path[draw=drawColor,line width= 0.5pt,line join=round,line cap=round] (109.63, 16.21) -- (109.63, 13.37);

\path[draw=drawColor,line width= 0.5pt,line join=round,line cap=round] (112.71, 16.21) -- (112.71, 13.37);

\path[draw=drawColor,line width= 0.5pt,line join=round,line cap=round] (115.44, 16.21) -- (115.44, 13.37);

\path[draw=drawColor,line width= 0.5pt,line join=round,line cap=round] (117.87, 16.21) -- (117.87, 10.52);

\path[draw=drawColor,line width= 0.5pt,line join=round,line cap=round] (133.90, 16.21) -- (133.90, 13.37);

\path[draw=drawColor,line width= 0.5pt,line join=round,line cap=round] (143.28, 16.21) -- (143.28, 13.37);

\path[draw=drawColor,line width= 0.5pt,line join=round,line cap=round] (149.93, 16.21) -- (149.93, 13.37);

\path[draw=drawColor,line width= 0.5pt,line join=round,line cap=round] (155.09, 16.21) -- (155.09, 13.37);

\path[draw=drawColor,line width= 0.5pt,line join=round,line cap=round] (159.31, 16.21) -- (159.31, 13.37);

\path[draw=drawColor,line width= 0.5pt,line join=round,line cap=round] (162.87, 16.21) -- (162.87, 13.37);

\path[draw=drawColor,line width= 0.5pt,line join=round,line cap=round] (165.96, 16.21) -- (165.96, 13.37);

\path[draw=drawColor,line width= 0.5pt,line join=round,line cap=round] (168.68, 16.21) -- (168.68, 13.37);

\path[draw=drawColor,line width= 0.5pt,line join=round,line cap=round] (171.12, 16.21) -- (171.12, 10.52);
\end{scope}
\begin{scope}
\definecolor{drawColor}{gray}{0.30}

\node[text=drawColor,anchor=base east,inner sep=0pt, outer sep=0pt, scale=  0.80] at ( 31.13, 15.76) {0.00};

\node[text=drawColor,anchor=base east,inner sep=0pt, outer sep=0pt, scale=  0.80] at ( 31.13, 27.23) {0.25};

\node[text=drawColor,anchor=base east,inner sep=0pt, outer sep=0pt, scale=  0.80] at ( 31.13, 38.71) {0.50};

\node[text=drawColor,anchor=base east,inner sep=0pt, outer sep=0pt, scale=  0.80] at ( 31.13, 50.19) {0.75};

\node[text=drawColor,anchor=base east,inner sep=0pt, outer sep=0pt, scale=  0.80] at ( 31.13, 61.67) {1.00};
\end{scope}
\begin{scope}
\definecolor{drawColor}{gray}{0.60}

\path[draw=drawColor,line width= 0.6pt,line join=round] ( 33.33, 18.51) --
	( 36.08, 18.51);

\path[draw=drawColor,line width= 0.6pt,line join=round] ( 33.33, 29.99) --
	( 36.08, 29.99);

\path[draw=drawColor,line width= 0.6pt,line join=round] ( 33.33, 41.47) --
	( 36.08, 41.47);

\path[draw=drawColor,line width= 0.6pt,line join=round] ( 33.33, 52.95) --
	( 36.08, 52.95);

\path[draw=drawColor,line width= 0.6pt,line join=round] ( 33.33, 64.43) --
	( 36.08, 64.43);
\end{scope}
\begin{scope}
\definecolor{drawColor}{gray}{0.30}

\node[text=drawColor,anchor=base west,inner sep=0pt, outer sep=0pt, scale=  0.80] at ( 58.30,  2.07) {10};

\node[text=drawColor,anchor=base west,inner sep=0pt, outer sep=0pt, scale=  0.56] at ( 66.29,  5.34) {-2};

\node[text=drawColor,anchor=base west,inner sep=0pt, outer sep=0pt, scale=  0.80] at (111.54,  2.07) {10};

\node[text=drawColor,anchor=base west,inner sep=0pt, outer sep=0pt, scale=  0.56] at (119.54,  5.34) {-1};

\node[text=drawColor,anchor=base west,inner sep=0pt, outer sep=0pt, scale=  0.80] at (165.72,  2.07) {10};

\node[text=drawColor,anchor=base west,inner sep=0pt, outer sep=0pt, scale=  0.56] at (173.72,  5.34) {0};
\end{scope}
\begin{scope}
\definecolor{drawColor}{RGB}{0,0,0}

\node[text=drawColor,rotate= 90.00,anchor=base,inner sep=0pt, outer sep=0pt, scale=  0.80] at (  8.36, 41.47) {Accuracy};
\end{scope}
\end{tikzpicture}

%% file: figs/results_6_circle_posttrain.tikz
\begin{tikzpicture}[x=1pt,y=1pt]
\definecolor{fillColor}{RGB}{255,255,255}
\begin{scope}
\definecolor{drawColor}{RGB}{255,0,255}

\path[draw=drawColor,line width= 0.6pt,line join=round] ( 33.82, 63.69) --
	( 33.90, 63.69);

\path[draw=drawColor,line width= 0.6pt,line join=round] ( 33.86, 63.69) --
	( 33.86, 60.51);

\path[draw=drawColor,line width= 0.6pt,line join=round] ( 33.82, 60.51) --
	( 33.90, 60.51);

\path[draw=drawColor,line width= 0.6pt,line join=round] ( 86.94, 63.10) --
	( 87.03, 63.10);

\path[draw=drawColor,line width= 0.6pt,line join=round] ( 86.99, 63.10) --
	( 86.99, 62.41);

\path[draw=drawColor,line width= 0.6pt,line join=round] ( 86.94, 62.41) --
	( 87.03, 62.41);

\path[draw=drawColor,line width= 0.6pt,line join=round] (124.08, 62.92) --
	(124.16, 62.92);

\path[draw=drawColor,line width= 0.6pt,line join=round] (124.12, 62.92) --
	(124.12, 61.95);

\path[draw=drawColor,line width= 0.6pt,line join=round] (124.08, 61.95) --
	(124.16, 61.95);

\path[draw=drawColor,line width= 0.6pt,line join=round] (140.07, 62.66) --
	(140.16, 62.66);

\path[draw=drawColor,line width= 0.6pt,line join=round] (140.11, 62.66) --
	(140.11, 61.65);

\path[draw=drawColor,line width= 0.6pt,line join=round] (140.07, 61.65) --
	(140.16, 61.65);

\path[draw=drawColor,line width= 0.6pt,line join=round] ( 12.02, 64.21) --
	( 12.11, 64.21);

\path[draw=drawColor,line width= 0.6pt,line join=round] ( 12.07, 64.21) --
	( 12.07, 55.35);

\path[draw=drawColor,line width= 0.6pt,line join=round] ( 12.02, 55.35) --
	( 12.11, 55.35);
\definecolor{drawColor}{RGB}{178,34,34}

\path[draw=drawColor,line width= 0.6pt,line join=round] ( 33.90, 26.12) --
	( 33.99, 26.12);

\path[draw=drawColor,line width= 0.6pt,line join=round] ( 33.95, 26.12) --
	( 33.95, 25.53);

\path[draw=drawColor,line width= 0.6pt,line join=round] ( 33.90, 25.53) --
	( 33.99, 25.53);

\path[draw=drawColor,line width= 0.6pt,line join=round] ( 87.03, 26.27) --
	( 87.12, 26.27);

\path[draw=drawColor,line width= 0.6pt,line join=round] ( 87.07, 26.27) --
	( 87.07, 25.83);

\path[draw=drawColor,line width= 0.6pt,line join=round] ( 87.03, 25.83) --
	( 87.12, 25.83);

\path[draw=drawColor,line width= 0.6pt,line join=round] (124.16, 26.16) --
	(124.25, 26.16);

\path[draw=drawColor,line width= 0.6pt,line join=round] (124.21, 26.16) --
	(124.21, 25.23);

\path[draw=drawColor,line width= 0.6pt,line join=round] (124.16, 25.23) --
	(124.25, 25.23);

\path[draw=drawColor,line width= 0.6pt,line join=round] (140.16, 62.73) --
	(140.24, 62.73);

\path[draw=drawColor,line width= 0.6pt,line join=round] (140.20, 62.73) --
	(140.20, 62.09);

\path[draw=drawColor,line width= 0.6pt,line join=round] (140.16, 62.09) --
	(140.24, 62.09);

\path[draw=drawColor,line width= 0.6pt,line join=round] ( 12.11, 26.05) --
	( 12.20, 26.05);

\path[draw=drawColor,line width= 0.6pt,line join=round] ( 12.16, 26.05) --
	( 12.16, 25.37);

\path[draw=drawColor,line width= 0.6pt,line join=round] ( 12.11, 25.37) --
	( 12.20, 25.37);
\definecolor{drawColor}{RGB}{67,205,128}

\path[draw=drawColor,line width= 0.6pt,line join=round] ( 33.99, 62.95) --
	( 34.08, 62.95);

\path[draw=drawColor,line width= 0.6pt,line join=round] ( 34.04, 62.95) --
	( 34.04, 60.35);

\path[draw=drawColor,line width= 0.6pt,line join=round] ( 33.99, 60.35) --
	( 34.08, 60.35);

\path[draw=drawColor,line width= 0.6pt,line join=round] ( 87.12, 63.17) --
	( 87.21, 63.17);

\path[draw=drawColor,line width= 0.6pt,line join=round] ( 87.16, 63.17) --
	( 87.16, 62.42);

\path[draw=drawColor,line width= 0.6pt,line join=round] ( 87.12, 62.42) --
	( 87.21, 62.42);

\path[draw=drawColor,line width= 0.6pt,line join=round] (124.25, 62.56) --
	(124.34, 62.56);

\path[draw=drawColor,line width= 0.6pt,line join=round] (124.30, 62.56) --
	(124.30, 61.83);

\path[draw=drawColor,line width= 0.6pt,line join=round] (124.25, 61.83) --
	(124.34, 61.83);

\path[draw=drawColor,line width= 0.6pt,line join=round] (140.24, 62.69) --
	(140.33, 62.69);

\path[draw=drawColor,line width= 0.6pt,line join=round] (140.29, 62.69) --
	(140.29, 62.08);

\path[draw=drawColor,line width= 0.6pt,line join=round] (140.24, 62.08) --
	(140.33, 62.08);

\path[draw=drawColor,line width= 0.6pt,line join=round] ( 12.20, 59.72) --
	( 12.29, 59.72);

\path[draw=drawColor,line width= 0.6pt,line join=round] ( 12.25, 59.72) --
	( 12.25, 43.58);

\path[draw=drawColor,line width= 0.6pt,line join=round] ( 12.20, 43.58) --
	( 12.29, 43.58);
\definecolor{drawColor}{RGB}{218,165,32}

\path[draw=drawColor,line width= 0.6pt,line join=round] ( 34.08, 63.97) --
	( 34.17, 63.97);

\path[draw=drawColor,line width= 0.6pt,line join=round] ( 34.12, 63.97) --
	( 34.12, 54.65);

\path[draw=drawColor,line width= 0.6pt,line join=round] ( 34.08, 54.65) --
	( 34.17, 54.65);

\path[draw=drawColor,line width= 0.6pt,line join=round] ( 87.21, 63.14) --
	( 87.30, 63.14);

\path[draw=drawColor,line width= 0.6pt,line join=round] ( 87.25, 63.14) --
	( 87.25, 62.68);

\path[draw=drawColor,line width= 0.6pt,line join=round] ( 87.21, 62.68) --
	( 87.30, 62.68);

\path[draw=drawColor,line width= 0.6pt,line join=round] (124.34, 63.29) --
	(124.43, 63.29);

\path[draw=drawColor,line width= 0.6pt,line join=round] (124.39, 63.29) --
	(124.39, 62.62);

\path[draw=drawColor,line width= 0.6pt,line join=round] (124.34, 62.62) --
	(124.43, 62.62);

\path[draw=drawColor,line width= 0.6pt,line join=round] (140.33, 62.98) --
	(140.42, 62.98);

\path[draw=drawColor,line width= 0.6pt,line join=round] (140.38, 62.98) --
	(140.38, 62.02);

\path[draw=drawColor,line width= 0.6pt,line join=round] (140.33, 62.02) --
	(140.42, 62.02);

\path[draw=drawColor,line width= 0.6pt,line join=round] ( 12.29, 38.64) --
	( 12.38, 38.64);

\path[draw=drawColor,line width= 0.6pt,line join=round] ( 12.33, 38.64) --
	( 12.33, 29.97);

\path[draw=drawColor,line width= 0.6pt,line join=round] ( 12.29, 29.97) --
	( 12.38, 29.97);
\definecolor{drawColor}{RGB}{0,0,139}

\path[draw=drawColor,line width= 0.6pt,line join=round] ( 34.17, 26.88) --
	( 34.26, 26.88);

\path[draw=drawColor,line width= 0.6pt,line join=round] ( 34.21, 26.88) --
	( 34.21, 25.73);

\path[draw=drawColor,line width= 0.6pt,line join=round] ( 34.17, 25.73) --
	( 34.26, 25.73);

\path[draw=drawColor,line width= 0.6pt,line join=round] ( 87.30, 62.39) --
	( 87.38, 62.39);

\path[draw=drawColor,line width= 0.6pt,line join=round] ( 87.34, 62.39) --
	( 87.34, 61.53);

\path[draw=drawColor,line width= 0.6pt,line join=round] ( 87.30, 61.53) --
	( 87.38, 61.53);

\path[draw=drawColor,line width= 0.6pt,line join=round] (124.43, 62.81) --
	(124.52, 62.81);

\path[draw=drawColor,line width= 0.6pt,line join=round] (124.47, 62.81) --
	(124.47, 61.73);

\path[draw=drawColor,line width= 0.6pt,line join=round] (124.43, 61.73) --
	(124.52, 61.73);

\path[draw=drawColor,line width= 0.6pt,line join=round] (140.42, 62.74) --
	(140.51, 62.74);

\path[draw=drawColor,line width= 0.6pt,line join=round] (140.47, 62.74) --
	(140.47, 61.48);

\path[draw=drawColor,line width= 0.6pt,line join=round] (140.42, 61.48) --
	(140.51, 61.48);

\path[draw=drawColor,line width= 0.6pt,line join=round] ( 12.38, 26.31) --
	( 12.47, 26.31);

\path[draw=drawColor,line width= 0.6pt,line join=round] ( 12.42, 26.31) --
	( 12.42, 25.72);

\path[draw=drawColor,line width= 0.6pt,line join=round] ( 12.38, 25.72) --
	( 12.47, 25.72);
\definecolor{fillColor}{RGB}{255,0,255}

\path[fill=fillColor,fill opacity=0.80] ( 32.12, 60.14) --
	( 36.04, 60.14) --
	( 36.04, 64.06) --
	( 32.12, 64.06) --
	cycle;

\path[fill=fillColor,fill opacity=0.80] ( 85.24, 60.80) --
	( 89.17, 60.80) --
	( 89.17, 64.72) --
	( 85.24, 64.72) --
	cycle;

\path[fill=fillColor,fill opacity=0.80] (122.38, 60.48) --
	(126.30, 60.48) --
	(126.30, 64.40) --
	(122.38, 64.40) --
	cycle;

\path[fill=fillColor,fill opacity=0.80] (138.37, 60.19) --
	(142.30, 60.19) --
	(142.30, 64.12) --
	(138.37, 64.12) --
	cycle;

\path[fill=fillColor,fill opacity=0.80] ( 10.33, 57.82) --
	( 14.25, 57.82) --
	( 14.25, 61.74) --
	( 10.33, 61.74) --
	cycle;
\definecolor{drawColor}{RGB}{178,34,34}
\definecolor{fillColor}{RGB}{178,34,34}

\path[draw=drawColor,draw opacity=0.80,line width= 0.4pt,line join=round,line cap=round,fill=fillColor,fill opacity=0.80] ( 34.08, 22.78) --
	( 36.72, 27.35) --
	( 31.44, 27.35) --
	cycle;

\path[draw=drawColor,draw opacity=0.80,line width= 0.4pt,line join=round,line cap=round,fill=fillColor,fill opacity=0.80] ( 87.21, 23.00) --
	( 89.85, 27.58) --
	( 84.56, 27.58) --
	cycle;

\path[draw=drawColor,draw opacity=0.80,line width= 0.4pt,line join=round,line cap=round,fill=fillColor,fill opacity=0.80] (124.34, 22.64) --
	(126.98, 27.22) --
	(121.70, 27.22) --
	cycle;

\path[draw=drawColor,draw opacity=0.80,line width= 0.4pt,line join=round,line cap=round,fill=fillColor,fill opacity=0.80] (140.33, 59.36) --
	(142.98, 63.94) --
	(137.69, 63.94) --
	cycle;

\path[draw=drawColor,draw opacity=0.80,line width= 0.4pt,line join=round,line cap=round,fill=fillColor,fill opacity=0.80] ( 12.29, 22.66) --
	( 14.93, 27.24) --
	(  9.65, 27.24) --
	cycle;
\definecolor{drawColor}{RGB}{67,205,128}
\definecolor{fillColor}{RGB}{67,205,128}

\path[draw=drawColor,draw opacity=0.80,line width= 0.4pt,line join=round,line cap=round,fill=fillColor,fill opacity=0.80] ( 34.08, 59.19) --
	( 36.54, 61.65) --
	( 34.08, 64.11) --
	( 31.62, 61.65) --
	cycle;

\path[draw=drawColor,draw opacity=0.80,line width= 0.4pt,line join=round,line cap=round,fill=fillColor,fill opacity=0.80] ( 87.21, 60.34) --
	( 89.67, 62.80) --
	( 87.21, 65.25) --
	( 84.75, 62.80) --
	cycle;

\path[draw=drawColor,draw opacity=0.80,line width= 0.4pt,line join=round,line cap=round,fill=fillColor,fill opacity=0.80] (124.34, 59.74) --
	(126.80, 62.19) --
	(124.34, 64.65) --
	(121.88, 62.19) --
	cycle;

\path[draw=drawColor,draw opacity=0.80,line width= 0.4pt,line join=round,line cap=round,fill=fillColor,fill opacity=0.80] (140.33, 59.93) --
	(142.79, 62.39) --
	(140.33, 64.85) --
	(137.87, 62.39) --
	cycle;

\path[draw=drawColor,draw opacity=0.80,line width= 0.4pt,line join=round,line cap=round,fill=fillColor,fill opacity=0.80] ( 12.29, 49.19) --
	( 14.75, 51.65) --
	( 12.29, 54.11) --
	(  9.83, 51.65) --
	cycle;
\definecolor{drawColor}{RGB}{218,165,32}
\definecolor{fillColor}{RGB}{218,165,32}

\path[draw=drawColor,draw opacity=0.80,line width= 0.4pt,line join=round,line cap=round,fill=fillColor,fill opacity=0.80] ( 34.08, 62.36) --
	( 36.72, 57.79) --
	( 31.44, 57.79) --
	cycle;

\path[draw=drawColor,draw opacity=0.80,line width= 0.4pt,line join=round,line cap=round,fill=fillColor,fill opacity=0.80] ( 87.21, 65.96) --
	( 89.85, 61.38) --
	( 84.56, 61.38) --
	cycle;

\path[draw=drawColor,draw opacity=0.80,line width= 0.4pt,line join=round,line cap=round,fill=fillColor,fill opacity=0.80] (124.34, 66.00) --
	(126.98, 61.43) --
	(121.70, 61.43) --
	cycle;

\path[draw=drawColor,draw opacity=0.80,line width= 0.4pt,line join=round,line cap=round,fill=fillColor,fill opacity=0.80] (140.33, 65.55) --
	(142.98, 60.97) --
	(137.69, 60.97) --
	cycle;

\path[draw=drawColor,draw opacity=0.80,line width= 0.4pt,line join=round,line cap=round,fill=fillColor,fill opacity=0.80] ( 12.29, 37.36) --
	( 14.93, 32.78) --
	(  9.65, 32.78) --
	cycle;
\definecolor{drawColor}{RGB}{0,0,139}
\definecolor{fillColor}{RGB}{0,0,139}

\path[draw=drawColor,draw opacity=0.80,line width= 0.4pt,line join=round,line cap=round,fill=fillColor,fill opacity=0.80] ( 34.08, 23.85) --
	( 36.54, 26.31) --
	( 34.08, 28.77) --
	( 31.62, 26.31) --
	cycle;

\path[draw=drawColor,draw opacity=0.80,line width= 0.4pt,line join=round,line cap=round,fill=fillColor,fill opacity=0.80] ( 87.21, 59.50) --
	( 89.67, 61.96) --
	( 87.21, 64.42) --
	( 84.75, 61.96) --
	cycle;

\path[draw=drawColor,draw opacity=0.80,line width= 0.4pt,line join=round,line cap=round,fill=fillColor,fill opacity=0.80] (124.34, 59.81) --
	(126.80, 62.27) --
	(124.34, 64.73) --
	(121.88, 62.27) --
	cycle;

\path[draw=drawColor,draw opacity=0.80,line width= 0.4pt,line join=round,line cap=round,fill=fillColor,fill opacity=0.80] (140.33, 59.65) --
	(142.79, 62.11) --
	(140.33, 64.57) --
	(137.87, 62.11) --
	cycle;

\path[draw=drawColor,draw opacity=0.80,line width= 0.4pt,line join=round,line cap=round,fill=fillColor,fill opacity=0.80] ( 12.29, 23.56) --
	( 14.75, 26.02) --
	( 12.29, 28.48) --
	(  9.83, 26.02) --
	cycle;
\definecolor{drawColor}{RGB}{255,0,255}

\path[draw=drawColor,draw opacity=0.80,line width= 0.5pt,line join=round] ( 12.29, 59.78) --
	( 34.08, 62.10) --
	( 87.21, 62.76) --
	(124.34, 62.44) --
	(140.33, 62.15);
\definecolor{drawColor}{RGB}{178,34,34}

\path[draw=drawColor,draw opacity=0.80,line width= 0.5pt,line join=round] ( 12.29, 25.71) --
	( 34.08, 25.83) --
	( 87.21, 26.05) --
	(124.34, 25.70) --
	(140.33, 62.41);
\definecolor{drawColor}{RGB}{67,205,128}

\path[draw=drawColor,draw opacity=0.80,line width= 0.5pt,line join=round] ( 12.29, 51.65) --
	( 34.08, 61.65) --
	( 87.21, 62.80) --
	(124.34, 62.19) --
	(140.33, 62.39);
\definecolor{drawColor}{RGB}{218,165,32}

\path[draw=drawColor,draw opacity=0.80,line width= 0.5pt,line join=round] ( 12.29, 34.31) --
	( 34.08, 59.31) --
	( 87.21, 62.91) --
	(124.34, 62.95) --
	(140.33, 62.50);
\definecolor{drawColor}{RGB}{0,0,139}

\path[draw=drawColor,draw opacity=0.80,line width= 0.5pt,line join=round] ( 12.29, 26.02) --
	( 34.08, 26.31) --
	( 87.21, 61.96) --
	(124.34, 62.27) --
	(140.33, 62.11);
\definecolor{drawColor}{RGB}{0,0,0}

\path[draw=drawColor,line width= 0.5pt,line join=round] ( 12.29, 63.43) --
	( 34.08, 63.37) --
	( 87.21, 63.48) --
	(124.34, 63.43) --
	(140.33, 63.31);

\path[draw=drawColor,line width= 0.5pt,line join=round] ( 12.29, 63.53) --
	( 34.08, 63.50) --
	( 87.21, 63.42) --
	(124.34, 63.21) --
	(140.33, 63.32);

\path[draw=drawColor,line width= 0.5pt,line join=round] ( 12.29, 63.44) --
	( 34.08, 63.37) --
	( 87.21, 63.43) --
	(124.34, 63.29) --
	(140.33, 63.26);

\path[draw=drawColor,line width= 0.5pt,line join=round] ( 12.29, 63.46) --
	( 34.08, 63.52) --
	( 87.21, 63.35) --
	(124.34, 63.27) --
	(140.33, 63.55);

\path[draw=drawColor,line width= 0.5pt,line join=round] ( 12.29, 63.36) --
	( 34.08, 63.43) --
	( 87.21, 63.46) --
	(124.34, 63.40) --
	(140.33, 63.55);

\path[draw=drawColor,line width= 0.5pt,line join=round] ( 12.29, 63.36) --
	( 34.08, 63.43) --
	( 87.21, 63.35) --
	(124.34, 63.33) --
	(140.33, 63.41);
\definecolor{drawColor}{gray}{0.60}

\path[draw=drawColor,line width= 0.5pt,line join=round,line cap=round] (  6.30, 16.21) -- (  6.30, 13.37);

\path[draw=drawColor,line width= 0.5pt,line join=round,line cap=round] ( 12.94, 16.21) -- ( 12.94, 13.37);

\path[draw=drawColor,line width= 0.5pt,line join=round,line cap=round] ( 18.09, 16.21) -- ( 18.09, 13.37);

\path[draw=drawColor,line width= 0.5pt,line join=round,line cap=round] ( 22.29, 16.21) -- ( 22.29, 13.37);

\path[draw=drawColor,line width= 0.5pt,line join=round,line cap=round] ( 25.85, 16.21) -- ( 25.85, 13.37);

\path[draw=drawColor,line width= 0.5pt,line join=round,line cap=round] ( 28.93, 16.21) -- ( 28.93, 13.37);

\path[draw=drawColor,line width= 0.5pt,line join=round,line cap=round] ( 31.65, 16.21) -- ( 31.65, 13.37);

\path[draw=drawColor,line width= 0.5pt,line join=round,line cap=round] ( 34.08, 16.21) -- ( 34.08, 10.52);

\path[draw=drawColor,line width= 0.5pt,line join=round,line cap=round] ( 50.07, 16.21) -- ( 50.07, 13.37);

\path[draw=drawColor,line width= 0.5pt,line join=round,line cap=round] ( 59.43, 16.21) -- ( 59.43, 13.37);

\path[draw=drawColor,line width= 0.5pt,line join=round,line cap=round] ( 66.07, 16.21) -- ( 66.07, 13.37);

\path[draw=drawColor,line width= 0.5pt,line join=round,line cap=round] ( 71.21, 16.21) -- ( 71.21, 13.37);

\path[draw=drawColor,line width= 0.5pt,line join=round,line cap=round] ( 75.42, 16.21) -- ( 75.42, 13.37);

\path[draw=drawColor,line width= 0.5pt,line join=round,line cap=round] ( 78.98, 16.21) -- ( 78.98, 13.37);

\path[draw=drawColor,line width= 0.5pt,line join=round,line cap=round] ( 82.06, 16.21) -- ( 82.06, 13.37);

\path[draw=drawColor,line width= 0.5pt,line join=round,line cap=round] ( 84.78, 16.21) -- ( 84.78, 13.37);

\path[draw=drawColor,line width= 0.5pt,line join=round,line cap=round] ( 87.21, 16.21) -- ( 87.21, 10.52);

\path[draw=drawColor,line width= 0.5pt,line join=round,line cap=round] (103.20, 16.21) -- (103.20, 13.37);

\path[draw=drawColor,line width= 0.5pt,line join=round,line cap=round] (112.55, 16.21) -- (112.55, 13.37);

\path[draw=drawColor,line width= 0.5pt,line join=round,line cap=round] (119.19, 16.21) -- (119.19, 13.37);

\path[draw=drawColor,line width= 0.5pt,line join=round,line cap=round] (124.34, 16.21) -- (124.34, 13.37);

\path[draw=drawColor,line width= 0.5pt,line join=round,line cap=round] (128.55, 16.21) -- (128.55, 13.37);

\path[draw=drawColor,line width= 0.5pt,line join=round,line cap=round] (132.10, 16.21) -- (132.10, 13.37);

\path[draw=drawColor,line width= 0.5pt,line join=round,line cap=round] (135.18, 16.21) -- (135.18, 13.37);

\path[draw=drawColor,line width= 0.5pt,line join=round,line cap=round] (137.90, 16.21) -- (137.90, 13.37);

\path[draw=drawColor,line width= 0.5pt,line join=round,line cap=round] (140.33, 16.21) -- (140.33, 10.52);
\end{scope}
\begin{scope}
\definecolor{drawColor}{gray}{0.60}

\path[draw=drawColor,line width= 0.6pt,line join=round] (  2.85, 18.51) --
	(  5.60, 18.51);

\path[draw=drawColor,line width= 0.6pt,line join=round] (  2.85, 29.99) --
	(  5.60, 29.99);

\path[draw=drawColor,line width= 0.6pt,line join=round] (  2.85, 41.47) --
	(  5.60, 41.47);

\path[draw=drawColor,line width= 0.6pt,line join=round] (  2.85, 52.95) --
	(  5.60, 52.95);

\path[draw=drawColor,line width= 0.6pt,line join=round] (  2.85, 64.43) --
	(  5.60, 64.43);
\end{scope}
\begin{scope}
\definecolor{drawColor}{gray}{0.30}

\node[text=drawColor,anchor=base west,inner sep=0pt, outer sep=0pt, scale=  0.80] at ( 27.75,  2.07) {10};

\node[text=drawColor,anchor=base west,inner sep=0pt, outer sep=0pt, scale=  0.56] at ( 35.75,  5.34) {-2};

\node[text=drawColor,anchor=base west,inner sep=0pt, outer sep=0pt, scale=  0.80] at ( 80.88,  2.07) {10};

\node[text=drawColor,anchor=base west,inner sep=0pt, outer sep=0pt, scale=  0.56] at ( 88.87,  5.34) {-1};

\node[text=drawColor,anchor=base west,inner sep=0pt, outer sep=0pt, scale=  0.80] at (134.93,  2.07) {10};

\node[text=drawColor,anchor=base west,inner sep=0pt, outer sep=0pt, scale=  0.56] at (142.93,  5.34) {0};
\end{scope}
\begin{scope}
\definecolor{drawColor}{RGB}{0,0,0}

\node[text=drawColor,anchor=base west,inner sep=0pt, outer sep=0pt, scale=  0.80] at (163.53, 59.26) {Method};
\end{scope}
\begin{scope}
\definecolor{drawColor}{RGB}{255,0,255}

\path[draw=drawColor,line width= 0.6pt,line join=round] (164.13, 51.39) -- (168.95, 51.39);
\end{scope}
\begin{scope}
\definecolor{fillColor}{RGB}{255,0,255}

\path[fill=fillColor,fill opacity=0.80] (164.58, 49.43) --
	(168.50, 49.43) --
	(168.50, 53.35) --
	(164.58, 53.35) --
	cycle;
\end{scope}
\begin{scope}
\definecolor{drawColor}{RGB}{255,0,255}

\path[draw=drawColor,draw opacity=0.80,line width= 0.5pt,line join=round] (164.13, 51.39) -- (168.95, 51.39);
\end{scope}
\begin{scope}
\definecolor{drawColor}{RGB}{178,34,34}

\path[draw=drawColor,line width= 0.6pt,line join=round] (164.13, 45.21) -- (168.95, 45.21);
\end{scope}
\begin{scope}
\definecolor{drawColor}{RGB}{178,34,34}
\definecolor{fillColor}{RGB}{178,34,34}

\path[draw=drawColor,draw opacity=0.80,line width= 0.4pt,line join=round,line cap=round,fill=fillColor,fill opacity=0.80] (166.54, 42.16) --
	(169.18, 46.73) --
	(163.90, 46.73) --
	cycle;
\end{scope}
\begin{scope}
\definecolor{drawColor}{RGB}{178,34,34}

\path[draw=drawColor,draw opacity=0.80,line width= 0.5pt,line join=round] (164.13, 45.21) -- (168.95, 45.21);
\end{scope}
\begin{scope}
\definecolor{drawColor}{RGB}{67,205,128}

\path[draw=drawColor,line width= 0.6pt,line join=round] (164.13, 39.03) -- (168.95, 39.03);
\end{scope}
\begin{scope}
\definecolor{drawColor}{RGB}{67,205,128}
\definecolor{fillColor}{RGB}{67,205,128}

\path[draw=drawColor,draw opacity=0.80,line width= 0.4pt,line join=round,line cap=round,fill=fillColor,fill opacity=0.80] (166.54, 36.57) --
	(169.00, 39.03) --
	(166.54, 41.49) --
	(164.08, 39.03) --
	cycle;
\end{scope}
\begin{scope}
\definecolor{drawColor}{RGB}{67,205,128}

\path[draw=drawColor,draw opacity=0.80,line width= 0.5pt,line join=round] (164.13, 39.03) -- (168.95, 39.03);
\end{scope}
\begin{scope}
\definecolor{drawColor}{RGB}{218,165,32}

\path[draw=drawColor,line width= 0.6pt,line join=round] (164.13, 32.84) -- (168.95, 32.84);
\end{scope}
\begin{scope}
\definecolor{drawColor}{RGB}{218,165,32}
\definecolor{fillColor}{RGB}{218,165,32}

\path[draw=drawColor,draw opacity=0.80,line width= 0.4pt,line join=round,line cap=round,fill=fillColor,fill opacity=0.80] (166.54, 35.90) --
	(169.18, 31.32) --
	(163.90, 31.32) --
	cycle;
\end{scope}
\begin{scope}
\definecolor{drawColor}{RGB}{218,165,32}

\path[draw=drawColor,draw opacity=0.80,line width= 0.5pt,line join=round] (164.13, 32.84) -- (168.95, 32.84);
\end{scope}
\begin{scope}
\definecolor{drawColor}{RGB}{0,0,139}

\path[draw=drawColor,line width= 0.6pt,line join=round] (164.13, 26.66) -- (168.95, 26.66);
\end{scope}
\begin{scope}
\definecolor{drawColor}{RGB}{0,0,139}
\definecolor{fillColor}{RGB}{0,0,139}

\path[draw=drawColor,draw opacity=0.80,line width= 0.4pt,line join=round,line cap=round,fill=fillColor,fill opacity=0.80] (166.54, 24.20) --
	(169.00, 26.66) --
	(166.54, 29.12) --
	(164.08, 26.66) --
	cycle;
\end{scope}
\begin{scope}
\definecolor{drawColor}{RGB}{0,0,139}

\path[draw=drawColor,draw opacity=0.80,line width= 0.5pt,line join=round] (164.13, 26.66) -- (168.95, 26.66);
\end{scope}
\begin{scope}
\definecolor{drawColor}{RGB}{0,245,255}

\path[draw=drawColor,line width= 0.6pt,line join=round] (164.13, 20.48) -- (168.95, 20.48);
\end{scope}
\begin{scope}
\definecolor{drawColor}{RGB}{0,245,255}
\definecolor{fillColor}{RGB}{0,245,255}

\path[draw=drawColor,draw opacity=0.80,line width= 0.4pt,line join=round,line cap=round,fill=fillColor,fill opacity=0.80] (166.54, 20.48) circle (  1.96);
\end{scope}
\begin{scope}
\definecolor{drawColor}{RGB}{0,245,255}

\path[draw=drawColor,draw opacity=0.80,line width= 0.5pt,line join=round] (164.13, 20.48) -- (168.95, 20.48);
\end{scope}
\begin{scope}
\definecolor{drawColor}{RGB}{0,0,0}

\node[text=drawColor,anchor=base west,inner sep=0pt, outer sep=0pt, scale=  0.70] at (173.55, 48.98) {{ \footnotesize \textsc{Synflow}}};
\end{scope}
\begin{scope}
\definecolor{drawColor}{RGB}{0,0,0}

\node[text=drawColor,anchor=base west,inner sep=0pt, outer sep=0pt, scale=  0.70] at (173.55, 42.80) {{ \footnotesize \textsc{GraSP}}};
\end{scope}
\begin{scope}
\definecolor{drawColor}{RGB}{0,0,0}

\node[text=drawColor,anchor=base west,inner sep=0pt, outer sep=0pt, scale=  0.70] at (173.55, 36.62) {{ \footnotesize \textsc{SNIP}}};
\end{scope}
\begin{scope}
\definecolor{drawColor}{RGB}{0,0,0}

\node[text=drawColor,anchor=base west,inner sep=0pt, outer sep=0pt, scale=  0.70] at (173.55, 30.43) {{ \footnotesize \textsc{Magnitude}}};
\end{scope}
\begin{scope}
\definecolor{drawColor}{RGB}{0,0,0}

\node[text=drawColor,anchor=base west,inner sep=0pt, outer sep=0pt, scale=  0.70] at (173.55, 24.25) {{ \footnotesize \textsc{Random}}};
\end{scope}
\begin{scope}
\definecolor{drawColor}{RGB}{0,0,0}

\node[text=drawColor,anchor=base west,inner sep=0pt, outer sep=0pt, scale=  0.70] at (173.55, 18.07) {{ \footnotesize \textsc{edge-popup}}};
\end{scope}
\end{tikzpicture}

%% file: figs/results_6_relu_postprune.tikz
\begin{tikzpicture}[x=1pt,y=1pt]
\definecolor{fillColor}{RGB}{255,255,255}
\begin{scope}
\definecolor{drawColor}{RGB}{255,0,255}

\path[draw=drawColor,line width= 0.6pt,line join=round] (102.24, 41.88) --
	(102.30, 41.88);

\path[draw=drawColor,line width= 0.6pt,line join=round] (102.27, 41.88) --
	(102.27, 35.11);

\path[draw=drawColor,line width= 0.6pt,line join=round] (102.24, 35.11) --
	(102.30, 35.11);

\path[draw=drawColor,line width= 0.6pt,line join=round] (136.61, 43.03) --
	(136.66, 43.03);

\path[draw=drawColor,line width= 0.6pt,line join=round] (136.64, 43.03) --
	(136.64, 40.40);

\path[draw=drawColor,line width= 0.6pt,line join=round] (136.61, 40.40) --
	(136.66, 40.40);

\path[draw=drawColor,line width= 0.6pt,line join=round] (160.63, 44.39) --
	(160.68, 44.39);

\path[draw=drawColor,line width= 0.6pt,line join=round] (160.66, 44.39) --
	(160.66, 42.22);

\path[draw=drawColor,line width= 0.6pt,line join=round] (160.63, 42.22) --
	(160.68, 42.22);

\path[draw=drawColor,line width= 0.6pt,line join=round] (170.97, 43.35) --
	(171.03, 43.35);

\path[draw=drawColor,line width= 0.6pt,line join=round] (171.00, 43.35) --
	(171.00, 40.43);

\path[draw=drawColor,line width= 0.6pt,line join=round] (170.97, 40.43) --
	(171.03, 40.43);

\path[draw=drawColor,line width= 0.6pt,line join=round] ( 41.04, 44.53) --
	( 41.09, 44.53);

\path[draw=drawColor,line width= 0.6pt,line join=round] ( 41.07, 44.53) --
	( 41.07, 37.33);

\path[draw=drawColor,line width= 0.6pt,line join=round] ( 41.04, 37.33) --
	( 41.09, 37.33);
\definecolor{drawColor}{RGB}{178,34,34}

\path[draw=drawColor,line width= 0.6pt,line join=round] (102.30, 45.23) --
	(102.36, 45.23);

\path[draw=drawColor,line width= 0.6pt,line join=round] (102.33, 45.23) --
	(102.33, 45.03);

\path[draw=drawColor,line width= 0.6pt,line join=round] (102.30, 45.03) --
	(102.36, 45.03);

\path[draw=drawColor,line width= 0.6pt,line join=round] (136.66, 45.17) --
	(136.72, 45.17);

\path[draw=drawColor,line width= 0.6pt,line join=round] (136.69, 45.17) --
	(136.69, 44.94);

\path[draw=drawColor,line width= 0.6pt,line join=round] (136.66, 44.94) --
	(136.72, 44.94);

\path[draw=drawColor,line width= 0.6pt,line join=round] (160.68, 45.28) --
	(160.74, 45.28);

\path[draw=drawColor,line width= 0.6pt,line join=round] (160.71, 45.28) --
	(160.71, 44.51);

\path[draw=drawColor,line width= 0.6pt,line join=round] (160.68, 44.51) --
	(160.74, 44.51);

\path[draw=drawColor,line width= 0.6pt,line join=round] (171.03, 44.74) --
	(171.09, 44.74);

\path[draw=drawColor,line width= 0.6pt,line join=round] (171.06, 44.74) --
	(171.06, 42.33);

\path[draw=drawColor,line width= 0.6pt,line join=round] (171.03, 42.33) --
	(171.09, 42.33);

\path[draw=drawColor,line width= 0.6pt,line join=round] ( 41.09, 45.20) --
	( 41.15, 45.20);

\path[draw=drawColor,line width= 0.6pt,line join=round] ( 41.12, 45.20) --
	( 41.12, 44.99);

\path[draw=drawColor,line width= 0.6pt,line join=round] ( 41.09, 44.99) --
	( 41.15, 44.99);
\definecolor{drawColor}{RGB}{67,205,128}

\path[draw=drawColor,line width= 0.6pt,line join=round] (102.36, 40.74) --
	(102.41, 40.74);

\path[draw=drawColor,line width= 0.6pt,line join=round] (102.39, 40.74) --
	(102.39, 34.95);

\path[draw=drawColor,line width= 0.6pt,line join=round] (102.36, 34.95) --
	(102.41, 34.95);

\path[draw=drawColor,line width= 0.6pt,line join=round] (136.72, 43.80) --
	(136.78, 43.80);

\path[draw=drawColor,line width= 0.6pt,line join=round] (136.75, 43.80) --
	(136.75, 40.91);

\path[draw=drawColor,line width= 0.6pt,line join=round] (136.72, 40.91) --
	(136.78, 40.91);

\path[draw=drawColor,line width= 0.6pt,line join=round] (160.74, 43.93) --
	(160.80, 43.93);

\path[draw=drawColor,line width= 0.6pt,line join=round] (160.77, 43.93) --
	(160.77, 40.35);

\path[draw=drawColor,line width= 0.6pt,line join=round] (160.74, 40.35) --
	(160.80, 40.35);

\path[draw=drawColor,line width= 0.6pt,line join=round] (171.09, 44.17) --
	(171.14, 44.17);

\path[draw=drawColor,line width= 0.6pt,line join=round] (171.12, 44.17) --
	(171.12, 41.80);

\path[draw=drawColor,line width= 0.6pt,line join=round] (171.09, 41.80) --
	(171.14, 41.80);

\path[draw=drawColor,line width= 0.6pt,line join=round] ( 41.15, 45.41) --
	( 41.21, 45.41);

\path[draw=drawColor,line width= 0.6pt,line join=round] ( 41.18, 45.41) --
	( 41.18, 42.34);

\path[draw=drawColor,line width= 0.6pt,line join=round] ( 41.15, 42.34) --
	( 41.21, 42.34);
\definecolor{drawColor}{RGB}{218,165,32}

\path[draw=drawColor,line width= 0.6pt,line join=round] (102.41, 41.38) --
	(102.47, 41.38);

\path[draw=drawColor,line width= 0.6pt,line join=round] (136.78, 44.24) --
	(136.84, 44.24);

\path[draw=drawColor,line width= 0.6pt,line join=round] (136.81, 44.24) --
	(136.81, 41.30);

\path[draw=drawColor,line width= 0.6pt,line join=round] (136.78, 41.30) --
	(136.84, 41.30);

\path[draw=drawColor,line width= 0.6pt,line join=round] (160.80, 43.07) --
	(160.86, 43.07);

\path[draw=drawColor,line width= 0.6pt,line join=round] (160.83, 43.07) --
	(160.83, 35.94);

\path[draw=drawColor,line width= 0.6pt,line join=round] (160.80, 35.94) --
	(160.86, 35.94);

\path[draw=drawColor,line width= 0.6pt,line join=round] (171.14, 44.62) --
	(171.20, 44.62);

\path[draw=drawColor,line width= 0.6pt,line join=round] (171.17, 44.62) --
	(171.17, 43.09);

\path[draw=drawColor,line width= 0.6pt,line join=round] (171.14, 43.09) --
	(171.20, 43.09);

\path[draw=drawColor,line width= 0.6pt,line join=round] ( 41.21, 45.06) --
	( 41.27, 45.06);

\path[draw=drawColor,line width= 0.6pt,line join=round] ( 41.24, 45.06) --
	( 41.24, 44.79);

\path[draw=drawColor,line width= 0.6pt,line join=round] ( 41.21, 44.79) --
	( 41.27, 44.79);
\definecolor{drawColor}{RGB}{0,0,139}

\path[draw=drawColor,line width= 0.6pt,line join=round] (102.47, 45.26) --
	(102.53, 45.26);

\path[draw=drawColor,line width= 0.6pt,line join=round] (102.50, 45.26) --
	(102.50, 45.00);

\path[draw=drawColor,line width= 0.6pt,line join=round] (102.47, 45.00) --
	(102.53, 45.00);

\path[draw=drawColor,line width= 0.6pt,line join=round] (136.84, 43.78) --
	(136.89, 43.78);

\path[draw=drawColor,line width= 0.6pt,line join=round] (136.87, 43.78) --
	(136.87, 42.45);

\path[draw=drawColor,line width= 0.6pt,line join=round] (136.84, 42.45) --
	(136.89, 42.45);

\path[draw=drawColor,line width= 0.6pt,line join=round] (160.86, 44.37) --
	(160.91, 44.37);

\path[draw=drawColor,line width= 0.6pt,line join=round] (160.89, 44.37) --
	(160.89, 42.10);

\path[draw=drawColor,line width= 0.6pt,line join=round] (160.86, 42.10) --
	(160.91, 42.10);

\path[draw=drawColor,line width= 0.6pt,line join=round] (171.20, 44.29) --
	(171.26, 44.29);

\path[draw=drawColor,line width= 0.6pt,line join=round] (171.23, 44.29) --
	(171.23, 41.01);

\path[draw=drawColor,line width= 0.6pt,line join=round] (171.20, 41.01) --
	(171.26, 41.01);

\path[draw=drawColor,line width= 0.6pt,line join=round] ( 41.27, 45.23) --
	( 41.32, 45.23);

\path[draw=drawColor,line width= 0.6pt,line join=round] ( 41.30, 45.23) --
	( 41.30, 45.01);

\path[draw=drawColor,line width= 0.6pt,line join=round] ( 41.27, 45.01) --
	( 41.32, 45.01);
\definecolor{fillColor}{RGB}{255,0,255}

\path[fill=fillColor,fill opacity=0.80] (100.45, 37.91) --
	(104.38, 37.91) --
	(104.38, 41.83) --
	(100.45, 41.83) --
	cycle;

\path[fill=fillColor,fill opacity=0.80] (134.82, 39.98) --
	(138.74, 39.98) --
	(138.74, 43.91) --
	(134.82, 43.91) --
	cycle;

\path[fill=fillColor,fill opacity=0.80] (158.84, 41.50) --
	(162.76, 41.50) --
	(162.76, 45.43) --
	(158.84, 45.43) --
	cycle;

\path[fill=fillColor,fill opacity=0.80] (169.18, 40.21) --
	(173.11, 40.21) --
	(173.11, 44.13) --
	(169.18, 44.13) --
	cycle;

\path[fill=fillColor,fill opacity=0.80] ( 39.25, 40.51) --
	( 43.17, 40.51) --
	( 43.17, 44.43) --
	( 39.25, 44.43) --
	cycle;
\definecolor{drawColor}{RGB}{178,34,34}
\definecolor{fillColor}{RGB}{178,34,34}

\path[draw=drawColor,draw opacity=0.80,line width= 0.4pt,line join=round,line cap=round,fill=fillColor,fill opacity=0.80] (102.41, 42.08) --
	(105.06, 46.66) --
	( 99.77, 46.66) --
	cycle;

\path[draw=drawColor,draw opacity=0.80,line width= 0.4pt,line join=round,line cap=round,fill=fillColor,fill opacity=0.80] (136.78, 42.01) --
	(139.42, 46.58) --
	(134.14, 46.58) --
	cycle;

\path[draw=drawColor,draw opacity=0.80,line width= 0.4pt,line join=round,line cap=round,fill=fillColor,fill opacity=0.80] (160.80, 41.87) --
	(163.44, 46.44) --
	(158.16, 46.44) --
	cycle;

\path[draw=drawColor,draw opacity=0.80,line width= 0.4pt,line join=round,line cap=round,fill=fillColor,fill opacity=0.80] (171.14, 40.68) --
	(173.79, 45.26) --
	(168.50, 45.26) --
	cycle;

\path[draw=drawColor,draw opacity=0.80,line width= 0.4pt,line join=round,line cap=round,fill=fillColor,fill opacity=0.80] ( 41.21, 42.04) --
	( 43.85, 46.62) --
	( 38.57, 46.62) --
	cycle;
\definecolor{drawColor}{RGB}{67,205,128}
\definecolor{fillColor}{RGB}{67,205,128}

\path[draw=drawColor,draw opacity=0.80,line width= 0.4pt,line join=round,line cap=round,fill=fillColor,fill opacity=0.80] (102.41, 36.43) --
	(104.87, 38.89) --
	(102.41, 41.35) --
	( 99.96, 38.89) --
	cycle;

\path[draw=drawColor,draw opacity=0.80,line width= 0.4pt,line join=round,line cap=round,fill=fillColor,fill opacity=0.80] (136.78, 40.17) --
	(139.24, 42.63) --
	(136.78, 45.09) --
	(134.32, 42.63) --
	cycle;

\path[draw=drawColor,draw opacity=0.80,line width= 0.4pt,line join=round,line cap=round,fill=fillColor,fill opacity=0.80] (160.80, 40.10) --
	(163.26, 42.56) --
	(160.80, 45.02) --
	(158.34, 42.56) --
	cycle;

\path[draw=drawColor,draw opacity=0.80,line width= 0.4pt,line join=round,line cap=round,fill=fillColor,fill opacity=0.80] (171.14, 40.71) --
	(173.60, 43.17) --
	(171.14, 45.63) --
	(168.68, 43.17) --
	cycle;

\path[draw=drawColor,draw opacity=0.80,line width= 0.4pt,line join=round,line cap=round,fill=fillColor,fill opacity=0.80] ( 41.21, 41.73) --
	( 43.67, 44.18) --
	( 41.21, 46.64) --
	( 38.75, 44.18) --
	cycle;
\definecolor{drawColor}{RGB}{218,165,32}
\definecolor{fillColor}{RGB}{218,165,32}

\path[draw=drawColor,draw opacity=0.80,line width= 0.4pt,line join=round,line cap=round,fill=fillColor,fill opacity=0.80] (102.41, 40.88) --
	(105.06, 36.30) --
	( 99.77, 36.30) --
	cycle;

\path[draw=drawColor,draw opacity=0.80,line width= 0.4pt,line join=round,line cap=round,fill=fillColor,fill opacity=0.80] (136.78, 46.11) --
	(139.42, 41.53) --
	(134.14, 41.53) --
	cycle;

\path[draw=drawColor,draw opacity=0.80,line width= 0.4pt,line join=round,line cap=round,fill=fillColor,fill opacity=0.80] (160.80, 44.06) --
	(163.44, 39.49) --
	(158.16, 39.49) --
	cycle;

\path[draw=drawColor,draw opacity=0.80,line width= 0.4pt,line join=round,line cap=round,fill=fillColor,fill opacity=0.80] (171.14, 46.98) --
	(173.79, 42.41) --
	(168.50, 42.41) --
	cycle;

\path[draw=drawColor,draw opacity=0.80,line width= 0.4pt,line join=round,line cap=round,fill=fillColor,fill opacity=0.80] ( 41.21, 47.98) --
	( 43.85, 43.40) --
	( 38.57, 43.40) --
	cycle;
\definecolor{drawColor}{RGB}{0,0,139}
\definecolor{fillColor}{RGB}{0,0,139}

\path[draw=drawColor,draw opacity=0.80,line width= 0.4pt,line join=round,line cap=round,fill=fillColor,fill opacity=0.80] (102.41, 42.68) --
	(104.87, 45.14) --
	(102.41, 47.60) --
	( 99.96, 45.14) --
	cycle;

\path[draw=drawColor,draw opacity=0.80,line width= 0.4pt,line join=round,line cap=round,fill=fillColor,fill opacity=0.80] (136.78, 40.72) --
	(139.24, 43.18) --
	(136.78, 45.64) --
	(134.32, 43.18) --
	cycle;

\path[draw=drawColor,draw opacity=0.80,line width= 0.4pt,line join=round,line cap=round,fill=fillColor,fill opacity=0.80] (160.80, 40.95) --
	(163.26, 43.40) --
	(160.80, 45.86) --
	(158.34, 43.40) --
	cycle;

\path[draw=drawColor,draw opacity=0.80,line width= 0.4pt,line join=round,line cap=round,fill=fillColor,fill opacity=0.80] (171.14, 40.54) --
	(173.60, 43.00) --
	(171.14, 45.46) --
	(168.68, 43.00) --
	cycle;

\path[draw=drawColor,draw opacity=0.80,line width= 0.4pt,line join=round,line cap=round,fill=fillColor,fill opacity=0.80] ( 41.21, 42.66) --
	( 43.67, 45.12) --
	( 41.21, 47.58) --
	( 38.75, 45.12) --
	cycle;
\definecolor{drawColor}{RGB}{0,245,255}
\definecolor{fillColor}{RGB}{0,245,255}

\path[draw=drawColor,draw opacity=0.80,line width= 0.4pt,line join=round,line cap=round,fill=fillColor,fill opacity=0.80] (102.41, 50.18) circle (  1.96);

\path[draw=drawColor,draw opacity=0.80,line width= 0.4pt,line join=round,line cap=round,fill=fillColor,fill opacity=0.80] (136.78, 50.20) circle (  1.96);

\path[draw=drawColor,draw opacity=0.80,line width= 0.4pt,line join=round,line cap=round,fill=fillColor,fill opacity=0.80] (160.80, 50.18) circle (  1.96);

\path[draw=drawColor,draw opacity=0.80,line width= 0.4pt,line join=round,line cap=round,fill=fillColor,fill opacity=0.80] (171.14, 50.18) circle (  1.96);

\path[draw=drawColor,draw opacity=0.80,line width= 0.4pt,line join=round,line cap=round,fill=fillColor,fill opacity=0.80] ( 41.21, 50.20) circle (  1.96);
\definecolor{drawColor}{RGB}{255,0,255}

\path[draw=drawColor,draw opacity=0.80,line width= 0.5pt,line join=round] ( 41.21, 42.47) --
	(102.41, 39.87) --
	(136.78, 41.95) --
	(160.80, 43.47) --
	(171.14, 42.17);
\definecolor{drawColor}{RGB}{178,34,34}

\path[draw=drawColor,draw opacity=0.80,line width= 0.5pt,line join=round] ( 41.21, 45.09) --
	(102.41, 45.13) --
	(136.78, 45.06) --
	(160.80, 44.92) --
	(171.14, 43.73);
\definecolor{drawColor}{RGB}{67,205,128}

\path[draw=drawColor,draw opacity=0.80,line width= 0.5pt,line join=round] ( 41.21, 44.18) --
	(102.41, 38.89) --
	(136.78, 42.63) --
	(160.80, 42.56) --
	(171.14, 43.17);
\definecolor{drawColor}{RGB}{218,165,32}

\path[draw=drawColor,draw opacity=0.80,line width= 0.5pt,line join=round] ( 41.21, 44.93) --
	(102.41, 37.83) --
	(136.78, 43.06) --
	(160.80, 41.01) --
	(171.14, 43.93);
\definecolor{drawColor}{RGB}{0,0,139}

\path[draw=drawColor,draw opacity=0.80,line width= 0.5pt,line join=round] ( 41.21, 45.12) --
	(102.41, 45.14) --
	(136.78, 43.18) --
	(160.80, 43.40) --
	(171.14, 43.00);
\definecolor{drawColor}{RGB}{0,245,255}

\path[draw=drawColor,draw opacity=0.80,line width= 0.5pt,line join=round] ( 41.21, 50.20) --
	(102.41, 50.18) --
	(136.78, 50.20) --
	(160.80, 50.18) --
	(171.14, 50.18);

\path[draw=drawColor,draw opacity=0.80,line width= 0.4pt,line join=round,line cap=round,fill=fillColor,fill opacity=0.80] (102.41, 50.19) circle (  1.96);

\path[draw=drawColor,draw opacity=0.80,line width= 0.4pt,line join=round,line cap=round,fill=fillColor,fill opacity=0.80] (136.78, 50.16) circle (  1.96);

\path[draw=drawColor,draw opacity=0.80,line width= 0.4pt,line join=round,line cap=round,fill=fillColor,fill opacity=0.80] (160.80, 50.22) circle (  1.96);

\path[draw=drawColor,draw opacity=0.80,line width= 0.4pt,line join=round,line cap=round,fill=fillColor,fill opacity=0.80] (171.14, 50.19) circle (  1.96);

\path[draw=drawColor,draw opacity=0.80,line width= 0.4pt,line join=round,line cap=round,fill=fillColor,fill opacity=0.80] ( 41.21, 50.19) circle (  1.96);

\path[draw=drawColor,draw opacity=0.80,line width= 0.9pt,dash pattern=on 1pt off 3pt on 4pt off 3pt ,line join=round] ( 41.21, 50.19) --
	(102.41, 50.19) --
	(136.78, 50.16) --
	(160.80, 50.22) --
	(171.14, 50.19);
\definecolor{drawColor}{RGB}{0,245,255}

\path[draw=drawColor,draw opacity=0.50,line width= 0.6pt,line join=round] (102.24, 50.25) --
	(102.59, 50.25);

\path[draw=drawColor,draw opacity=0.50,line width= 0.6pt,line join=round] (102.41, 50.25) --
	(102.41, 50.14);

\path[draw=drawColor,draw opacity=0.50,line width= 0.6pt,line join=round] (102.24, 50.14) --
	(102.59, 50.14);

\path[draw=drawColor,draw opacity=0.50,line width= 0.6pt,line join=round] (136.61, 50.21) --
	(136.95, 50.21);

\path[draw=drawColor,draw opacity=0.50,line width= 0.6pt,line join=round] (136.78, 50.21) --
	(136.78, 50.11);

\path[draw=drawColor,draw opacity=0.50,line width= 0.6pt,line join=round] (136.61, 50.11) --
	(136.95, 50.11);

\path[draw=drawColor,draw opacity=0.50,line width= 0.6pt,line join=round] (160.63, 50.28) --
	(160.97, 50.28);

\path[draw=drawColor,draw opacity=0.50,line width= 0.6pt,line join=round] (160.80, 50.28) --
	(160.80, 50.16);

\path[draw=drawColor,draw opacity=0.50,line width= 0.6pt,line join=round] (160.63, 50.16) --
	(160.97, 50.16);

\path[draw=drawColor,draw opacity=0.50,line width= 0.6pt,line join=round] (170.97, 50.23) --
	(171.32, 50.23);

\path[draw=drawColor,draw opacity=0.50,line width= 0.6pt,line join=round] (171.14, 50.23) --
	(171.14, 50.15);

\path[draw=drawColor,draw opacity=0.50,line width= 0.6pt,line join=round] (170.97, 50.15) --
	(171.32, 50.15);

\path[draw=drawColor,draw opacity=0.50,line width= 0.6pt,line join=round] ( 41.04, 50.24) --
	( 41.38, 50.24);

\path[draw=drawColor,draw opacity=0.50,line width= 0.6pt,line join=round] ( 41.21, 50.24) --
	( 41.21, 50.14);

\path[draw=drawColor,draw opacity=0.50,line width= 0.6pt,line join=round] ( 41.04, 50.14) --
	( 41.38, 50.14);
\definecolor{drawColor}{RGB}{0,0,0}

\path[draw=drawColor,line width= 0.5pt,line join=round] ( 41.21, 17.92) --
	(102.41, 17.86);

\path[draw=drawColor,line width= 0.5pt,line join=round] (102.41, 17.87) --
	(136.78, 17.89) --
	(160.80, 17.88) --
	(171.14, 17.87);
\definecolor{drawColor}{gray}{0.60}

\path[draw=drawColor,line width= 0.5pt,line join=round,line cap=round] ( 44.03, 16.21) -- ( 44.03, 13.37);

\path[draw=drawColor,line width= 0.5pt,line join=round,line cap=round] ( 50.08, 16.21) -- ( 50.08, 13.37);

\path[draw=drawColor,line width= 0.5pt,line join=round,line cap=round] ( 54.37, 16.21) -- ( 54.37, 13.37);

\path[draw=drawColor,line width= 0.5pt,line join=round,line cap=round] ( 57.70, 16.21) -- ( 57.70, 13.37);

\path[draw=drawColor,line width= 0.5pt,line join=round,line cap=round] ( 60.43, 16.21) -- ( 60.43, 13.37);

\path[draw=drawColor,line width= 0.5pt,line join=round,line cap=round] ( 62.73, 16.21) -- ( 62.73, 13.37);

\path[draw=drawColor,line width= 0.5pt,line join=round,line cap=round] ( 64.72, 16.21) -- ( 64.72, 13.37);

\path[draw=drawColor,line width= 0.5pt,line join=round,line cap=round] ( 66.48, 16.21) -- ( 66.48, 13.37);

\path[draw=drawColor,line width= 0.5pt,line join=round,line cap=round] ( 68.05, 16.21) -- ( 68.05, 10.52);

\path[draw=drawColor,line width= 0.5pt,line join=round,line cap=round] ( 78.39, 16.21) -- ( 78.39, 13.37);

\path[draw=drawColor,line width= 0.5pt,line join=round,line cap=round] ( 84.45, 16.21) -- ( 84.45, 13.37);

\path[draw=drawColor,line width= 0.5pt,line join=round,line cap=round] ( 88.74, 16.21) -- ( 88.74, 13.37);

\path[draw=drawColor,line width= 0.5pt,line join=round,line cap=round] ( 92.07, 16.21) -- ( 92.07, 13.37);

\path[draw=drawColor,line width= 0.5pt,line join=round,line cap=round] ( 94.79, 16.21) -- ( 94.79, 13.37);

\path[draw=drawColor,line width= 0.5pt,line join=round,line cap=round] ( 97.09, 16.21) -- ( 97.09, 13.37);

\path[draw=drawColor,line width= 0.5pt,line join=round,line cap=round] ( 99.08, 16.21) -- ( 99.08, 13.37);

\path[draw=drawColor,line width= 0.5pt,line join=round,line cap=round] (100.84, 16.21) -- (100.84, 13.37);

\path[draw=drawColor,line width= 0.5pt,line join=round,line cap=round] (102.41, 16.21) -- (102.41, 10.52);

\path[draw=drawColor,line width= 0.5pt,line join=round,line cap=round] (112.76, 16.21) -- (112.76, 13.37);

\path[draw=drawColor,line width= 0.5pt,line join=round,line cap=round] (118.81, 16.21) -- (118.81, 13.37);

\path[draw=drawColor,line width= 0.5pt,line join=round,line cap=round] (123.10, 16.21) -- (123.10, 13.37);

\path[draw=drawColor,line width= 0.5pt,line join=round,line cap=round] (126.43, 16.21) -- (126.43, 13.37);

\path[draw=drawColor,line width= 0.5pt,line join=round,line cap=round] (129.16, 16.21) -- (129.16, 13.37);

\path[draw=drawColor,line width= 0.5pt,line join=round,line cap=round] (131.46, 16.21) -- (131.46, 13.37);

\path[draw=drawColor,line width= 0.5pt,line join=round,line cap=round] (133.45, 16.21) -- (133.45, 13.37);

\path[draw=drawColor,line width= 0.5pt,line join=round,line cap=round] (135.21, 16.21) -- (135.21, 13.37);

\path[draw=drawColor,line width= 0.5pt,line join=round,line cap=round] (136.78, 16.21) -- (136.78, 10.52);

\path[draw=drawColor,line width= 0.5pt,line join=round,line cap=round] (147.12, 16.21) -- (147.12, 13.37);

\path[draw=drawColor,line width= 0.5pt,line join=round,line cap=round] (153.18, 16.21) -- (153.18, 13.37);

\path[draw=drawColor,line width= 0.5pt,line join=round,line cap=round] (157.47, 16.21) -- (157.47, 13.37);

\path[draw=drawColor,line width= 0.5pt,line join=round,line cap=round] (160.80, 16.21) -- (160.80, 13.37);

\path[draw=drawColor,line width= 0.5pt,line join=round,line cap=round] (163.52, 16.21) -- (163.52, 13.37);

\path[draw=drawColor,line width= 0.5pt,line join=round,line cap=round] (165.82, 16.21) -- (165.82, 13.37);

\path[draw=drawColor,line width= 0.5pt,line join=round,line cap=round] (167.81, 16.21) -- (167.81, 13.37);

\path[draw=drawColor,line width= 0.5pt,line join=round,line cap=round] (169.57, 16.21) -- (169.57, 13.37);

\path[draw=drawColor,line width= 0.5pt,line join=round,line cap=round] (171.14, 16.21) -- (171.14, 10.52);
\end{scope}
\begin{scope}
\definecolor{drawColor}{gray}{0.30}

\node[text=drawColor,anchor=base west,inner sep=0pt, outer sep=0pt, scale=  0.80] at ( 16.91, 14.42) {10};

\node[text=drawColor,anchor=base west,inner sep=0pt, outer sep=0pt, scale=  0.56] at ( 24.91, 17.69) {-4};

\node[text=drawColor,anchor=base west,inner sep=0pt, outer sep=0pt, scale=  0.80] at ( 16.91, 22.88) {10};

\node[text=drawColor,anchor=base west,inner sep=0pt, outer sep=0pt, scale=  0.56] at ( 24.91, 26.15) {-3};

\node[text=drawColor,anchor=base west,inner sep=0pt, outer sep=0pt, scale=  0.80] at ( 16.91, 31.34) {10};

\node[text=drawColor,anchor=base west,inner sep=0pt, outer sep=0pt, scale=  0.56] at ( 24.91, 34.61) {-2};

\node[text=drawColor,anchor=base west,inner sep=0pt, outer sep=0pt, scale=  0.80] at ( 16.91, 39.80) {10};

\node[text=drawColor,anchor=base west,inner sep=0pt, outer sep=0pt, scale=  0.56] at ( 24.91, 43.07) {-1};

\node[text=drawColor,anchor=base west,inner sep=0pt, outer sep=0pt, scale=  0.80] at ( 18.78, 48.25) {10};

\node[text=drawColor,anchor=base west,inner sep=0pt, outer sep=0pt, scale=  0.56] at ( 26.77, 51.52) {0};
\end{scope}
\begin{scope}
\definecolor{drawColor}{gray}{0.60}

\path[draw=drawColor,line width= 0.6pt,line join=round] ( 31.77, 17.85) --
	( 34.52, 17.85);

\path[draw=drawColor,line width= 0.6pt,line join=round] ( 31.77, 26.31) --
	( 34.52, 26.31);

\path[draw=drawColor,line width= 0.6pt,line join=round] ( 31.77, 34.77) --
	( 34.52, 34.77);

\path[draw=drawColor,line width= 0.6pt,line join=round] ( 31.77, 43.23) --
	( 34.52, 43.23);

\path[draw=drawColor,line width= 0.6pt,line join=round] ( 31.77, 51.69) --
	( 34.52, 51.69);
\end{scope}
\begin{scope}
\definecolor{drawColor}{gray}{0.30}

\node[text=drawColor,anchor=base west,inner sep=0pt, outer sep=0pt, scale=  0.80] at ( 61.72,  2.07) {10};

\node[text=drawColor,anchor=base west,inner sep=0pt, outer sep=0pt, scale=  0.56] at ( 69.72,  5.34) {-3};

\node[text=drawColor,anchor=base west,inner sep=0pt, outer sep=0pt, scale=  0.80] at ( 96.08,  2.07) {10};

\node[text=drawColor,anchor=base west,inner sep=0pt, outer sep=0pt, scale=  0.56] at (104.08,  5.34) {-2};

\node[text=drawColor,anchor=base west,inner sep=0pt, outer sep=0pt, scale=  0.80] at (130.45,  2.07) {10};

\node[text=drawColor,anchor=base west,inner sep=0pt, outer sep=0pt, scale=  0.56] at (138.45,  5.34) {-1};

\node[text=drawColor,anchor=base west,inner sep=0pt, outer sep=0pt, scale=  0.80] at (165.75,  2.07) {10};

\node[text=drawColor,anchor=base west,inner sep=0pt, outer sep=0pt, scale=  0.56] at (173.74,  5.34) {0};
\end{scope}
\begin{scope}
\definecolor{drawColor}{RGB}{0,0,0}

\node[text=drawColor,rotate= 90.00,anchor=base,inner sep=0pt, outer sep=0pt, scale=  0.80] at (  8.36, 34.24) {MSE};
\end{scope}
\end{tikzpicture}

%% file: figs/results_6_relu_posttrain.tikz
\begin{tikzpicture}[x=1pt,y=1pt]
\definecolor{fillColor}{RGB}{255,255,255}
\begin{scope}
\definecolor{drawColor}{RGB}{255,0,255}

\path[draw=drawColor,line width= 0.6pt,line join=round] ( 86.98, 17.85) --
	( 87.05, 17.85);

\path[draw=drawColor,line width= 0.6pt,line join=round] ( 87.02, 17.85) --
	( 87.02, 17.85);

\path[draw=drawColor,line width= 0.6pt,line join=round] ( 86.98, 17.85) --
	( 87.05, 17.85);

\path[draw=drawColor,line width= 0.6pt,line join=round] (128.29, 17.85) --
	(128.35, 17.85);

\path[draw=drawColor,line width= 0.6pt,line join=round] (128.32, 17.85) --
	(128.32, 17.85);

\path[draw=drawColor,line width= 0.6pt,line join=round] (128.29, 17.85) --
	(128.35, 17.85);

\path[draw=drawColor,line width= 0.6pt,line join=round] (157.15, 25.19) --
	(157.22, 25.19);

\path[draw=drawColor,line width= 0.6pt,line join=round] (157.19, 25.19) --
	(157.19, 18.60);

\path[draw=drawColor,line width= 0.6pt,line join=round] (157.15, 18.60) --
	(157.22, 18.60);

\path[draw=drawColor,line width= 0.6pt,line join=round] (169.59, 24.38) --
	(169.66, 24.38);

\path[draw=drawColor,line width= 0.6pt,line join=round] (169.62, 24.38) --
	(169.62, 20.20);

\path[draw=drawColor,line width= 0.6pt,line join=round] (169.59, 20.20) --
	(169.66, 20.20);

\path[draw=drawColor,line width= 0.6pt,line join=round] ( 13.42, 44.53) --
	( 13.49, 44.53);

\path[draw=drawColor,line width= 0.6pt,line join=round] ( 13.46, 44.53) --
	( 13.46, 37.33);

\path[draw=drawColor,line width= 0.6pt,line join=round] ( 13.42, 37.33) --
	( 13.49, 37.33);
\definecolor{drawColor}{RGB}{178,34,34}

\path[draw=drawColor,line width= 0.6pt,line join=round] ( 87.05, 45.23) --
	( 87.12, 45.23);

\path[draw=drawColor,line width= 0.6pt,line join=round] ( 87.09, 45.23) --
	( 87.09, 45.03);

\path[draw=drawColor,line width= 0.6pt,line join=round] ( 87.05, 45.03) --
	( 87.12, 45.03);

\path[draw=drawColor,line width= 0.6pt,line join=round] (128.35, 45.17) --
	(128.42, 45.17);

\path[draw=drawColor,line width= 0.6pt,line join=round] (128.39, 45.17) --
	(128.39, 44.94);

\path[draw=drawColor,line width= 0.6pt,line join=round] (128.35, 44.94) --
	(128.42, 44.94);

\path[draw=drawColor,line width= 0.6pt,line join=round] (157.22, 45.28) --
	(157.29, 45.28);

\path[draw=drawColor,line width= 0.6pt,line join=round] (157.26, 45.28) --
	(157.26, 44.51);

\path[draw=drawColor,line width= 0.6pt,line join=round] (157.22, 44.51) --
	(157.29, 44.51);

\path[draw=drawColor,line width= 0.6pt,line join=round] (169.66, 22.04) --
	(169.73, 22.04);

\path[draw=drawColor,line width= 0.6pt,line join=round] (169.69, 22.04) --
	(169.69, 19.27);

\path[draw=drawColor,line width= 0.6pt,line join=round] (169.66, 19.27) --
	(169.73, 19.27);

\path[draw=drawColor,line width= 0.6pt,line join=round] ( 13.49, 45.20) --
	( 13.56, 45.20);

\path[draw=drawColor,line width= 0.6pt,line join=round] ( 13.53, 45.20) --
	( 13.53, 44.99);

\path[draw=drawColor,line width= 0.6pt,line join=round] ( 13.49, 44.99) --
	( 13.56, 44.99);
\definecolor{drawColor}{RGB}{67,205,128}

\path[draw=drawColor,line width= 0.6pt,line join=round] ( 87.12, 17.85) --
	( 87.19, 17.85);

\path[draw=drawColor,line width= 0.6pt,line join=round] ( 87.16, 17.85) --
	( 87.16, 17.85);

\path[draw=drawColor,line width= 0.6pt,line join=round] ( 87.12, 17.85) --
	( 87.19, 17.85);

\path[draw=drawColor,line width= 0.6pt,line join=round] (128.42, 23.36) --
	(128.49, 23.36);

\path[draw=drawColor,line width= 0.6pt,line join=round] (128.46, 23.36) --
	(128.46, 18.27);

\path[draw=drawColor,line width= 0.6pt,line join=round] (128.42, 18.27) --
	(128.49, 18.27);

\path[draw=drawColor,line width= 0.6pt,line join=round] (157.29, 22.05) --
	(157.36, 22.05);

\path[draw=drawColor,line width= 0.6pt,line join=round] (157.33, 22.05) --
	(157.33, 18.08);

\path[draw=drawColor,line width= 0.6pt,line join=round] (157.29, 18.08) --
	(157.36, 18.08);

\path[draw=drawColor,line width= 0.6pt,line join=round] (169.73, 22.36) --
	(169.79, 22.36);

\path[draw=drawColor,line width= 0.6pt,line join=round] (169.76, 22.36) --
	(169.76, 19.56);

\path[draw=drawColor,line width= 0.6pt,line join=round] (169.73, 19.56) --
	(169.79, 19.56);

\path[draw=drawColor,line width= 0.6pt,line join=round] ( 13.56, 45.41) --
	( 13.63, 45.41);

\path[draw=drawColor,line width= 0.6pt,line join=round] ( 13.60, 45.41) --
	( 13.60, 42.34);

\path[draw=drawColor,line width= 0.6pt,line join=round] ( 13.56, 42.34) --
	( 13.63, 42.34);
\definecolor{drawColor}{RGB}{218,165,32}

\path[draw=drawColor,line width= 0.6pt,line join=round] ( 87.19, 41.07) --
	( 87.26, 41.07);

\path[draw=drawColor,line width= 0.6pt,line join=round] (128.49, 17.85) --
	(128.56, 17.85);

\path[draw=drawColor,line width= 0.6pt,line join=round] (128.53, 17.85) --
	(128.53, 17.85);

\path[draw=drawColor,line width= 0.6pt,line join=round] (128.49, 17.85) --
	(128.56, 17.85);

\path[draw=drawColor,line width= 0.6pt,line join=round] (157.36, 21.58) --
	(157.43, 21.58);

\path[draw=drawColor,line width= 0.6pt,line join=round] (157.40, 21.58) --
	(157.40, 18.01);

\path[draw=drawColor,line width= 0.6pt,line join=round] (157.36, 18.01) --
	(157.43, 18.01);

\path[draw=drawColor,line width= 0.6pt,line join=round] (169.79, 28.21) --
	(169.86, 28.21);

\path[draw=drawColor,line width= 0.6pt,line join=round] ( 13.63, 45.06) --
	( 13.70, 45.06);

\path[draw=drawColor,line width= 0.6pt,line join=round] ( 13.67, 45.06) --
	( 13.67, 44.79);

\path[draw=drawColor,line width= 0.6pt,line join=round] ( 13.63, 44.79) --
	( 13.70, 44.79);
\definecolor{drawColor}{RGB}{0,0,139}

\path[draw=drawColor,line width= 0.6pt,line join=round] ( 87.26, 45.26) --
	( 87.33, 45.26);

\path[draw=drawColor,line width= 0.6pt,line join=round] ( 87.29, 45.26) --
	( 87.29, 45.00);

\path[draw=drawColor,line width= 0.6pt,line join=round] ( 87.26, 45.00) --
	( 87.33, 45.00);

\path[draw=drawColor,line width= 0.6pt,line join=round] (128.56, 17.85) --
	(128.63, 17.85);

\path[draw=drawColor,line width= 0.6pt,line join=round] (128.60, 17.85) --
	(128.60, 17.85);

\path[draw=drawColor,line width= 0.6pt,line join=round] (128.56, 17.85) --
	(128.63, 17.85);

\path[draw=drawColor,line width= 0.6pt,line join=round] (157.43, 18.89) --
	(157.50, 18.89);

\path[draw=drawColor,line width= 0.6pt,line join=round] (169.86, 22.64) --
	(169.93, 22.64);

\path[draw=drawColor,line width= 0.6pt,line join=round] (169.90, 22.64) --
	(169.90, 19.40);

\path[draw=drawColor,line width= 0.6pt,line join=round] (169.86, 19.40) --
	(169.93, 19.40);

\path[draw=drawColor,line width= 0.6pt,line join=round] ( 13.70, 45.23) --
	( 13.77, 45.23);

\path[draw=drawColor,line width= 0.6pt,line join=round] ( 13.73, 45.23) --
	( 13.73, 45.01);

\path[draw=drawColor,line width= 0.6pt,line join=round] ( 13.70, 45.01) --
	( 13.77, 45.01);
\definecolor{fillColor}{RGB}{255,0,255}

\path[fill=fillColor,fill opacity=0.80] ( 85.23, 15.89) --
	( 89.15, 15.89) --
	( 89.15, 19.82) --
	( 85.23, 19.82) --
	cycle;

\path[fill=fillColor,fill opacity=0.80] (126.53, 15.89) --
	(130.45, 15.89) --
	(130.45, 19.82) --
	(126.53, 19.82) --
	cycle;

\path[fill=fillColor,fill opacity=0.80] (155.40, 21.25) --
	(159.32, 21.25) --
	(159.32, 25.17) --
	(155.40, 25.17) --
	cycle;

\path[fill=fillColor,fill opacity=0.80] (167.83, 20.89) --
	(171.76, 20.89) --
	(171.76, 24.81) --
	(167.83, 24.81) --
	cycle;

\path[fill=fillColor,fill opacity=0.80] ( 11.67, 40.51) --
	( 15.59, 40.51) --
	( 15.59, 44.43) --
	( 11.67, 44.43) --
	cycle;
\definecolor{drawColor}{RGB}{178,34,34}
\definecolor{fillColor}{RGB}{178,34,34}

\path[draw=drawColor,draw opacity=0.80,line width= 0.4pt,line join=round,line cap=round,fill=fillColor,fill opacity=0.80] ( 87.19, 42.08) --
	( 89.83, 46.66) --
	( 84.55, 46.66) --
	cycle;

\path[draw=drawColor,draw opacity=0.80,line width= 0.4pt,line join=round,line cap=round,fill=fillColor,fill opacity=0.80] (128.49, 42.01) --
	(131.14, 46.58) --
	(125.85, 46.58) --
	cycle;

\path[draw=drawColor,draw opacity=0.80,line width= 0.4pt,line join=round,line cap=round,fill=fillColor,fill opacity=0.80] (157.36, 41.87) --
	(160.00, 46.44) --
	(154.72, 46.44) --
	cycle;

\path[draw=drawColor,draw opacity=0.80,line width= 0.4pt,line join=round,line cap=round,fill=fillColor,fill opacity=0.80] (169.79, 17.86) --
	(172.44, 22.44) --
	(167.15, 22.44) --
	cycle;

\path[draw=drawColor,draw opacity=0.80,line width= 0.4pt,line join=round,line cap=round,fill=fillColor,fill opacity=0.80] ( 13.63, 42.04) --
	( 16.27, 46.62) --
	( 10.99, 46.62) --
	cycle;
\definecolor{drawColor}{RGB}{67,205,128}
\definecolor{fillColor}{RGB}{67,205,128}

\path[draw=drawColor,draw opacity=0.80,line width= 0.4pt,line join=round,line cap=round,fill=fillColor,fill opacity=0.80] ( 87.19, 15.39) --
	( 89.65, 17.85) --
	( 87.19, 20.31) --
	( 84.73, 17.85) --
	cycle;

\path[draw=drawColor,draw opacity=0.80,line width= 0.4pt,line join=round,line cap=round,fill=fillColor,fill opacity=0.80] (128.49, 19.18) --
	(130.95, 21.64) --
	(128.49, 24.09) --
	(126.03, 21.64) --
	cycle;

\path[draw=drawColor,draw opacity=0.80,line width= 0.4pt,line join=round,line cap=round,fill=fillColor,fill opacity=0.80] (157.36, 18.12) --
	(159.82, 20.58) --
	(157.36, 23.04) --
	(154.90, 20.58) --
	cycle;

\path[draw=drawColor,draw opacity=0.80,line width= 0.4pt,line join=round,line cap=round,fill=fillColor,fill opacity=0.80] (169.79, 18.76) --
	(172.25, 21.22) --
	(169.79, 23.68) --
	(167.34, 21.22) --
	cycle;

\path[draw=drawColor,draw opacity=0.80,line width= 0.4pt,line join=round,line cap=round,fill=fillColor,fill opacity=0.80] ( 13.63, 41.73) --
	( 16.09, 44.18) --
	( 13.63, 46.64) --
	( 11.17, 44.18) --
	cycle;
\definecolor{drawColor}{RGB}{218,165,32}
\definecolor{fillColor}{RGB}{218,165,32}

\path[draw=drawColor,draw opacity=0.80,line width= 0.4pt,line join=round,line cap=round,fill=fillColor,fill opacity=0.80] ( 87.19, 39.80) --
	( 89.83, 35.22) --
	( 84.55, 35.22) --
	cycle;

\path[draw=drawColor,draw opacity=0.80,line width= 0.4pt,line join=round,line cap=round,fill=fillColor,fill opacity=0.80] (128.49, 20.90) --
	(131.14, 16.33) --
	(125.85, 16.33) --
	cycle;

\path[draw=drawColor,draw opacity=0.80,line width= 0.4pt,line join=round,line cap=round,fill=fillColor,fill opacity=0.80] (157.36, 23.26) --
	(160.00, 18.69) --
	(154.72, 18.69) --
	cycle;

\path[draw=drawColor,draw opacity=0.80,line width= 0.4pt,line join=round,line cap=round,fill=fillColor,fill opacity=0.80] (169.79, 27.84) --
	(172.44, 23.26) --
	(167.15, 23.26) --
	cycle;

\path[draw=drawColor,draw opacity=0.80,line width= 0.4pt,line join=round,line cap=round,fill=fillColor,fill opacity=0.80] ( 13.63, 47.98) --
	( 16.27, 43.40) --
	( 10.99, 43.40) --
	cycle;
\definecolor{drawColor}{RGB}{0,0,139}
\definecolor{fillColor}{RGB}{0,0,139}

\path[draw=drawColor,draw opacity=0.80,line width= 0.4pt,line join=round,line cap=round,fill=fillColor,fill opacity=0.80] ( 87.19, 42.68) --
	( 89.65, 45.14) --
	( 87.19, 47.60) --
	( 84.73, 45.14) --
	cycle;

\path[draw=drawColor,draw opacity=0.80,line width= 0.4pt,line join=round,line cap=round,fill=fillColor,fill opacity=0.80] (128.49, 15.39) --
	(130.95, 17.85) --
	(128.49, 20.31) --
	(126.03, 17.85) --
	cycle;

\path[draw=drawColor,draw opacity=0.80,line width= 0.4pt,line join=round,line cap=round,fill=fillColor,fill opacity=0.80] (157.36, 15.74) --
	(159.82, 18.20) --
	(157.36, 20.66) --
	(154.90, 18.20) --
	cycle;

\path[draw=drawColor,draw opacity=0.80,line width= 0.4pt,line join=round,line cap=round,fill=fillColor,fill opacity=0.80] (169.79, 18.90) --
	(172.25, 21.36) --
	(169.79, 23.82) --
	(167.34, 21.36) --
	cycle;

\path[draw=drawColor,draw opacity=0.80,line width= 0.4pt,line join=round,line cap=round,fill=fillColor,fill opacity=0.80] ( 13.63, 42.66) --
	( 16.09, 45.12) --
	( 13.63, 47.58) --
	( 11.17, 45.12) --
	cycle;
\definecolor{drawColor}{RGB}{255,0,255}

\path[draw=drawColor,draw opacity=0.80,line width= 0.5pt,line join=round] ( 13.63, 42.47) --
	( 87.19, 17.85) --
	(128.49, 17.85) --
	(157.36, 23.21) --
	(169.79, 22.85);
\definecolor{drawColor}{RGB}{178,34,34}

\path[draw=drawColor,draw opacity=0.80,line width= 0.5pt,line join=round] ( 13.63, 45.09) --
	( 87.19, 45.13) --
	(128.49, 45.06) --
	(157.36, 44.92) --
	(169.79, 20.91);
\definecolor{drawColor}{RGB}{67,205,128}

\path[draw=drawColor,draw opacity=0.80,line width= 0.5pt,line join=round] ( 13.63, 44.18) --
	( 87.19, 17.85) --
	(128.49, 21.64) --
	(157.36, 20.58) --
	(169.79, 21.22);
\definecolor{drawColor}{RGB}{218,165,32}

\path[draw=drawColor,draw opacity=0.80,line width= 0.5pt,line join=round] ( 13.63, 44.93) --
	( 87.19, 36.74) --
	(128.49, 17.85) --
	(157.36, 20.21) --
	(169.79, 24.78);
\definecolor{drawColor}{RGB}{0,0,139}

\path[draw=drawColor,draw opacity=0.80,line width= 0.5pt,line join=round] ( 13.63, 45.12) --
	( 87.19, 45.14) --
	(128.49, 17.85) --
	(157.36, 18.20) --
	(169.79, 21.36);
\definecolor{drawColor}{RGB}{0,0,0}

\path[draw=drawColor,line width= 0.5pt,line join=round] ( 13.63, 17.92) --
	( 87.19, 17.86);

\path[draw=drawColor,line width= 0.5pt,line join=round] ( 87.19, 17.87) --
	(128.49, 17.89) --
	(157.36, 17.88) --
	(169.79, 17.87);
\definecolor{drawColor}{gray}{0.60}

\path[draw=drawColor,line width= 0.5pt,line join=round,line cap=round] ( 17.02, 16.21) -- ( 17.02, 13.37);

\path[draw=drawColor,line width= 0.5pt,line join=round,line cap=round] ( 24.29, 16.21) -- ( 24.29, 13.37);

\path[draw=drawColor,line width= 0.5pt,line join=round,line cap=round] ( 29.45, 16.21) -- ( 29.45, 13.37);

\path[draw=drawColor,line width= 0.5pt,line join=round,line cap=round] ( 33.46, 16.21) -- ( 33.46, 13.37);

\path[draw=drawColor,line width= 0.5pt,line join=round,line cap=round] ( 36.73, 16.21) -- ( 36.73, 13.37);

\path[draw=drawColor,line width= 0.5pt,line join=round,line cap=round] ( 39.49, 16.21) -- ( 39.49, 13.37);

\path[draw=drawColor,line width= 0.5pt,line join=round,line cap=round] ( 41.89, 16.21) -- ( 41.89, 13.37);

\path[draw=drawColor,line width= 0.5pt,line join=round,line cap=round] ( 44.00, 16.21) -- ( 44.00, 13.37);

\path[draw=drawColor,line width= 0.5pt,line join=round,line cap=round] ( 45.89, 16.21) -- ( 45.89, 10.52);

\path[draw=drawColor,line width= 0.5pt,line join=round,line cap=round] ( 58.32, 16.21) -- ( 58.32, 13.37);

\path[draw=drawColor,line width= 0.5pt,line join=round,line cap=round] ( 65.59, 16.21) -- ( 65.59, 13.37);

\path[draw=drawColor,line width= 0.5pt,line join=round,line cap=round] ( 70.76, 16.21) -- ( 70.76, 13.37);

\path[draw=drawColor,line width= 0.5pt,line join=round,line cap=round] ( 74.76, 16.21) -- ( 74.76, 13.37);

\path[draw=drawColor,line width= 0.5pt,line join=round,line cap=round] ( 78.03, 16.21) -- ( 78.03, 13.37);

\path[draw=drawColor,line width= 0.5pt,line join=round,line cap=round] ( 80.79, 16.21) -- ( 80.79, 13.37);

\path[draw=drawColor,line width= 0.5pt,line join=round,line cap=round] ( 83.19, 16.21) -- ( 83.19, 13.37);

\path[draw=drawColor,line width= 0.5pt,line join=round,line cap=round] ( 85.30, 16.21) -- ( 85.30, 13.37);

\path[draw=drawColor,line width= 0.5pt,line join=round,line cap=round] ( 87.19, 16.21) -- ( 87.19, 10.52);

\path[draw=drawColor,line width= 0.5pt,line join=round,line cap=round] ( 99.62, 16.21) -- ( 99.62, 13.37);

\path[draw=drawColor,line width= 0.5pt,line join=round,line cap=round] (106.90, 16.21) -- (106.90, 13.37);

\path[draw=drawColor,line width= 0.5pt,line join=round,line cap=round] (112.06, 16.21) -- (112.06, 13.37);

\path[draw=drawColor,line width= 0.5pt,line join=round,line cap=round] (116.06, 16.21) -- (116.06, 13.37);

\path[draw=drawColor,line width= 0.5pt,line join=round,line cap=round] (119.33, 16.21) -- (119.33, 13.37);

\path[draw=drawColor,line width= 0.5pt,line join=round,line cap=round] (122.09, 16.21) -- (122.09, 13.37);

\path[draw=drawColor,line width= 0.5pt,line join=round,line cap=round] (124.49, 16.21) -- (124.49, 13.37);

\path[draw=drawColor,line width= 0.5pt,line join=round,line cap=round] (126.60, 16.21) -- (126.60, 13.37);

\path[draw=drawColor,line width= 0.5pt,line join=round,line cap=round] (128.49, 16.21) -- (128.49, 10.52);

\path[draw=drawColor,line width= 0.5pt,line join=round,line cap=round] (140.93, 16.21) -- (140.93, 13.37);

\path[draw=drawColor,line width= 0.5pt,line join=round,line cap=round] (148.20, 16.21) -- (148.20, 13.37);

\path[draw=drawColor,line width= 0.5pt,line join=round,line cap=round] (153.36, 16.21) -- (153.36, 13.37);

\path[draw=drawColor,line width= 0.5pt,line join=round,line cap=round] (157.36, 16.21) -- (157.36, 13.37);

\path[draw=drawColor,line width= 0.5pt,line join=round,line cap=round] (160.63, 16.21) -- (160.63, 13.37);

\path[draw=drawColor,line width= 0.5pt,line join=round,line cap=round] (163.40, 16.21) -- (163.40, 13.37);

\path[draw=drawColor,line width= 0.5pt,line join=round,line cap=round] (165.79, 16.21) -- (165.79, 13.37);

\path[draw=drawColor,line width= 0.5pt,line join=round,line cap=round] (167.90, 16.21) -- (167.90, 13.37);

\path[draw=drawColor,line width= 0.5pt,line join=round,line cap=round] (169.79, 16.21) -- (169.79, 10.52);
\end{scope}
\begin{scope}
\definecolor{drawColor}{gray}{0.60}

\path[draw=drawColor,line width= 0.6pt,line join=round] (  2.85, 17.85) --
	(  5.60, 17.85);

\path[draw=drawColor,line width= 0.6pt,line join=round] (  2.85, 26.31) --
	(  5.60, 26.31);

\path[draw=drawColor,line width= 0.6pt,line join=round] (  2.85, 34.77) --
	(  5.60, 34.77);

\path[draw=drawColor,line width= 0.6pt,line join=round] (  2.85, 43.23) --
	(  5.60, 43.23);

\path[draw=drawColor,line width= 0.6pt,line join=round] (  2.85, 51.69) --
	(  5.60, 51.69);
\end{scope}
\begin{scope}
\definecolor{drawColor}{gray}{0.30}

\node[text=drawColor,anchor=base west,inner sep=0pt, outer sep=0pt, scale=  0.80] at ( 39.56,  2.07) {10};

\node[text=drawColor,anchor=base west,inner sep=0pt, outer sep=0pt, scale=  0.56] at ( 47.56,  5.34) {-3};

\node[text=drawColor,anchor=base west,inner sep=0pt, outer sep=0pt, scale=  0.80] at ( 80.86,  2.07) {10};

\node[text=drawColor,anchor=base west,inner sep=0pt, outer sep=0pt, scale=  0.56] at ( 88.86,  5.34) {-2};

\node[text=drawColor,anchor=base west,inner sep=0pt, outer sep=0pt, scale=  0.80] at (122.16,  2.07) {10};

\node[text=drawColor,anchor=base west,inner sep=0pt, outer sep=0pt, scale=  0.56] at (130.16,  5.34) {-1};

\node[text=drawColor,anchor=base west,inner sep=0pt, outer sep=0pt, scale=  0.80] at (164.40,  2.07) {10};

\node[text=drawColor,anchor=base west,inner sep=0pt, outer sep=0pt, scale=  0.56] at (172.39,  5.34) {0};
\end{scope}
\end{tikzpicture}

%% file: figs/results_6_helix_postprune.tikz
\begin{tikzpicture}[x=1pt,y=1pt,baseline={(0,0.4)}]
\definecolor{fillColor}{RGB}{255,255,255}
\begin{scope}
\definecolor{drawColor}{RGB}{255,0,255}

\path[draw=drawColor,line width= 0.6pt,line join=round] ( 62.34, 61.46) --
	( 62.43, 61.46);

\path[draw=drawColor,line width= 0.6pt,line join=round] ( 62.39, 61.46) --
	( 62.39, 60.30);

\path[draw=drawColor,line width= 0.6pt,line join=round] ( 62.34, 60.30) --
	( 62.43, 60.30);

\path[draw=drawColor,line width= 0.6pt,line join=round] (116.56, 61.82) --
	(116.65, 61.82);

\path[draw=drawColor,line width= 0.6pt,line join=round] (116.60, 61.82) --
	(116.60, 60.60);

\path[draw=drawColor,line width= 0.6pt,line join=round] (116.56, 60.60) --
	(116.65, 60.60);

\path[draw=drawColor,line width= 0.6pt,line join=round] (154.45, 62.31) --
	(154.54, 62.31);

\path[draw=drawColor,line width= 0.6pt,line join=round] (154.50, 62.31) --
	(154.50, 61.24);

\path[draw=drawColor,line width= 0.6pt,line join=round] (154.45, 61.24) --
	(154.54, 61.24);

\path[draw=drawColor,line width= 0.6pt,line join=round] (170.77, 61.96) --
	(170.86, 61.96);

\path[draw=drawColor,line width= 0.6pt,line join=round] (170.82, 61.96) --
	(170.82, 60.84);

\path[draw=drawColor,line width= 0.6pt,line join=round] (170.77, 60.84) --
	(170.86, 60.84);

\path[draw=drawColor,line width= 0.6pt,line join=round] ( 41.04, 62.36) --
	( 41.13, 62.36);

\path[draw=drawColor,line width= 0.6pt,line join=round] ( 41.08, 62.36) --
	( 41.08, 60.86);

\path[draw=drawColor,line width= 0.6pt,line join=round] ( 41.04, 60.86) --
	( 41.13, 60.86);
\definecolor{drawColor}{RGB}{178,34,34}

\path[draw=drawColor,line width= 0.6pt,line join=round] ( 62.43, 62.99) --
	( 62.52, 62.99);

\path[draw=drawColor,line width= 0.6pt,line join=round] ( 62.48, 62.99) --
	( 62.48, 62.84);

\path[draw=drawColor,line width= 0.6pt,line join=round] ( 62.43, 62.84) --
	( 62.52, 62.84);

\path[draw=drawColor,line width= 0.6pt,line join=round] (116.65, 63.04) --
	(116.74, 63.04);

\path[draw=drawColor,line width= 0.6pt,line join=round] (116.69, 63.04) --
	(116.69, 62.92);

\path[draw=drawColor,line width= 0.6pt,line join=round] (116.65, 62.92) --
	(116.74, 62.92);

\path[draw=drawColor,line width= 0.6pt,line join=round] (154.54, 63.00) --
	(154.63, 63.00);

\path[draw=drawColor,line width= 0.6pt,line join=round] (154.59, 63.00) --
	(154.59, 62.91);

\path[draw=drawColor,line width= 0.6pt,line join=round] (154.54, 62.91) --
	(154.63, 62.91);

\path[draw=drawColor,line width= 0.6pt,line join=round] (170.86, 62.35) --
	(170.95, 62.35);

\path[draw=drawColor,line width= 0.6pt,line join=round] (170.91, 62.35) --
	(170.91, 60.74);

\path[draw=drawColor,line width= 0.6pt,line join=round] (170.86, 60.74) --
	(170.95, 60.74);

\path[draw=drawColor,line width= 0.6pt,line join=round] ( 41.13, 63.06) --
	( 41.22, 63.06);

\path[draw=drawColor,line width= 0.6pt,line join=round] ( 41.17, 63.06) --
	( 41.17, 62.90);

\path[draw=drawColor,line width= 0.6pt,line join=round] ( 41.13, 62.90) --
	( 41.22, 62.90);
\definecolor{drawColor}{RGB}{67,205,128}

\path[draw=drawColor,line width= 0.6pt,line join=round] ( 62.52, 62.36) --
	( 62.61, 62.36);

\path[draw=drawColor,line width= 0.6pt,line join=round] ( 62.57, 62.36) --
	( 62.57, 60.69);

\path[draw=drawColor,line width= 0.6pt,line join=round] ( 62.52, 60.69) --
	( 62.61, 60.69);

\path[draw=drawColor,line width= 0.6pt,line join=round] (116.74, 61.42) --
	(116.83, 61.42);

\path[draw=drawColor,line width= 0.6pt,line join=round] (116.78, 61.42) --
	(116.78, 60.51);

\path[draw=drawColor,line width= 0.6pt,line join=round] (116.74, 60.51) --
	(116.83, 60.51);

\path[draw=drawColor,line width= 0.6pt,line join=round] (154.63, 61.96) --
	(154.72, 61.96);

\path[draw=drawColor,line width= 0.6pt,line join=round] (154.68, 61.96) --
	(154.68, 60.92);

\path[draw=drawColor,line width= 0.6pt,line join=round] (154.63, 60.92) --
	(154.72, 60.92);

\path[draw=drawColor,line width= 0.6pt,line join=round] (170.95, 62.18) --
	(171.04, 62.18);

\path[draw=drawColor,line width= 0.6pt,line join=round] (171.00, 62.18) --
	(171.00, 60.77);

\path[draw=drawColor,line width= 0.6pt,line join=round] (170.95, 60.77) --
	(171.04, 60.77);

\path[draw=drawColor,line width= 0.6pt,line join=round] ( 41.22, 61.51) --
	( 41.31, 61.51);

\path[draw=drawColor,line width= 0.6pt,line join=round] ( 41.26, 61.51) --
	( 41.26, 60.58);

\path[draw=drawColor,line width= 0.6pt,line join=round] ( 41.22, 60.58) --
	( 41.31, 60.58);
\definecolor{drawColor}{RGB}{218,165,32}

\path[draw=drawColor,line width= 0.6pt,line join=round] ( 62.61, 62.47) --
	( 62.70, 62.47);

\path[draw=drawColor,line width= 0.6pt,line join=round] ( 62.66, 62.47) --
	( 62.66, 60.41);

\path[draw=drawColor,line width= 0.6pt,line join=round] ( 62.61, 60.41) --
	( 62.70, 60.41);

\path[draw=drawColor,line width= 0.6pt,line join=round] (116.83, 61.69) --
	(116.92, 61.69);

\path[draw=drawColor,line width= 0.6pt,line join=round] (116.87, 61.69) --
	(116.87, 60.31);

\path[draw=drawColor,line width= 0.6pt,line join=round] (116.83, 60.31) --
	(116.92, 60.31);

\path[draw=drawColor,line width= 0.6pt,line join=round] (154.72, 62.06) --
	(154.81, 62.06);

\path[draw=drawColor,line width= 0.6pt,line join=round] (154.77, 62.06) --
	(154.77, 60.97);

\path[draw=drawColor,line width= 0.6pt,line join=round] (154.72, 60.97) --
	(154.81, 60.97);

\path[draw=drawColor,line width= 0.6pt,line join=round] (171.04, 62.28) --
	(171.14, 62.28);

\path[draw=drawColor,line width= 0.6pt,line join=round] (171.09, 62.28) --
	(171.09, 61.00);

\path[draw=drawColor,line width= 0.6pt,line join=round] (171.04, 61.00) --
	(171.14, 61.00);

\path[draw=drawColor,line width= 0.6pt,line join=round] ( 41.31, 63.00) --
	( 41.40, 63.00);

\path[draw=drawColor,line width= 0.6pt,line join=round] ( 41.35, 63.00) --
	( 41.35, 62.81);

\path[draw=drawColor,line width= 0.6pt,line join=round] ( 41.31, 62.81) --
	( 41.40, 62.81);
\definecolor{drawColor}{RGB}{0,0,139}

\path[draw=drawColor,line width= 0.6pt,line join=round] ( 62.70, 63.01) --
	( 62.79, 63.01);

\path[draw=drawColor,line width= 0.6pt,line join=round] ( 62.75, 63.01) --
	( 62.75, 62.91);

\path[draw=drawColor,line width= 0.6pt,line join=round] ( 62.70, 62.91) --
	( 62.79, 62.91);

\path[draw=drawColor,line width= 0.6pt,line join=round] (116.92, 61.67) --
	(117.01, 61.67);

\path[draw=drawColor,line width= 0.6pt,line join=round] (116.96, 61.67) --
	(116.96, 60.22);

\path[draw=drawColor,line width= 0.6pt,line join=round] (116.92, 60.22) --
	(117.01, 60.22);

\path[draw=drawColor,line width= 0.6pt,line join=round] (154.81, 62.30) --
	(154.90, 62.30);

\path[draw=drawColor,line width= 0.6pt,line join=round] (154.86, 62.30) --
	(154.86, 60.71);

\path[draw=drawColor,line width= 0.6pt,line join=round] (154.81, 60.71) --
	(154.90, 60.71);

\path[draw=drawColor,line width= 0.6pt,line join=round] (171.14, 62.33) --
	(171.23, 62.33);

\path[draw=drawColor,line width= 0.6pt,line join=round] (171.18, 62.33) --
	(171.18, 61.58);

\path[draw=drawColor,line width= 0.6pt,line join=round] (171.14, 61.58) --
	(171.23, 61.58);

\path[draw=drawColor,line width= 0.6pt,line join=round] ( 41.40, 63.03) --
	( 41.49, 63.03);

\path[draw=drawColor,line width= 0.6pt,line join=round] ( 41.44, 63.03) --
	( 41.44, 62.94);

\path[draw=drawColor,line width= 0.6pt,line join=round] ( 41.40, 62.94) --
	( 41.49, 62.94);
\definecolor{fillColor}{RGB}{255,0,255}

\path[fill=fillColor,fill opacity=0.80] ( 60.65, 58.96) --
	( 64.58, 58.96) --
	( 64.58, 62.89) --
	( 60.65, 62.89) --
	cycle;

\path[fill=fillColor,fill opacity=0.80] (114.87, 59.30) --
	(118.79, 59.30) --
	(118.79, 63.22) --
	(114.87, 63.22) --
	cycle;

\path[fill=fillColor,fill opacity=0.80] (152.76, 59.85) --
	(156.69, 59.85) --
	(156.69, 63.77) --
	(152.76, 63.77) --
	cycle;

\path[fill=fillColor,fill opacity=0.80] (169.08, 59.47) --
	(173.01, 59.47) --
	(173.01, 63.40) --
	(169.08, 63.40) --
	cycle;

\path[fill=fillColor,fill opacity=0.80] ( 39.35, 59.72) --
	( 43.27, 59.72) --
	( 43.27, 63.64) --
	( 39.35, 63.64) --
	cycle;
\definecolor{drawColor}{RGB}{178,34,34}
\definecolor{fillColor}{RGB}{178,34,34}

\path[draw=drawColor,draw opacity=0.80,line width= 0.4pt,line join=round,line cap=round,fill=fillColor,fill opacity=0.80] ( 62.61, 59.86) --
	( 65.26, 64.44) --
	( 59.97, 64.44) --
	cycle;

\path[draw=drawColor,draw opacity=0.80,line width= 0.4pt,line join=round,line cap=round,fill=fillColor,fill opacity=0.80] (116.83, 59.93) --
	(119.47, 64.50) --
	(114.19, 64.50) --
	cycle;

\path[draw=drawColor,draw opacity=0.80,line width= 0.4pt,line join=round,line cap=round,fill=fillColor,fill opacity=0.80] (154.72, 59.90) --
	(157.37, 64.48) --
	(152.08, 64.48) --
	cycle;

\path[draw=drawColor,draw opacity=0.80,line width= 0.4pt,line join=round,line cap=round,fill=fillColor,fill opacity=0.80] (171.04, 58.58) --
	(173.69, 63.16) --
	(168.40, 63.16) --
	cycle;

\path[draw=drawColor,draw opacity=0.80,line width= 0.4pt,line join=round,line cap=round,fill=fillColor,fill opacity=0.80] ( 41.31, 59.93) --
	( 43.95, 64.51) --
	( 38.67, 64.51) --
	cycle;
\definecolor{drawColor}{RGB}{67,205,128}
\definecolor{fillColor}{RGB}{67,205,128}

\path[draw=drawColor,draw opacity=0.80,line width= 0.4pt,line join=round,line cap=round,fill=fillColor,fill opacity=0.80] ( 62.61, 59.16) --
	( 65.07, 61.62) --
	( 62.61, 64.07) --
	( 60.15, 61.62) --
	cycle;

\path[draw=drawColor,draw opacity=0.80,line width= 0.4pt,line join=round,line cap=round,fill=fillColor,fill opacity=0.80] (116.83, 58.54) --
	(119.29, 60.99) --
	(116.83, 63.45) --
	(114.37, 60.99) --
	cycle;

\path[draw=drawColor,draw opacity=0.80,line width= 0.4pt,line join=round,line cap=round,fill=fillColor,fill opacity=0.80] (154.72, 59.01) --
	(157.18, 61.47) --
	(154.72, 63.93) --
	(152.27, 61.47) --
	cycle;

\path[draw=drawColor,draw opacity=0.80,line width= 0.4pt,line join=round,line cap=round,fill=fillColor,fill opacity=0.80] (171.04, 59.08) --
	(173.50, 61.54) --
	(171.04, 64.00) --
	(168.59, 61.54) --
	cycle;

\path[draw=drawColor,draw opacity=0.80,line width= 0.4pt,line join=round,line cap=round,fill=fillColor,fill opacity=0.80] ( 41.31, 58.61) --
	( 43.77, 61.07) --
	( 41.31, 63.53) --
	( 38.85, 61.07) --
	cycle;
\definecolor{drawColor}{RGB}{218,165,32}
\definecolor{fillColor}{RGB}{218,165,32}

\path[draw=drawColor,draw opacity=0.80,line width= 0.4pt,line join=round,line cap=round,fill=fillColor,fill opacity=0.80] ( 62.61, 64.63) --
	( 65.26, 60.05) --
	( 59.97, 60.05) --
	cycle;

\path[draw=drawColor,draw opacity=0.80,line width= 0.4pt,line join=round,line cap=round,fill=fillColor,fill opacity=0.80] (116.83, 64.11) --
	(119.47, 59.53) --
	(114.19, 59.53) --
	cycle;

\path[draw=drawColor,draw opacity=0.80,line width= 0.4pt,line join=round,line cap=round,fill=fillColor,fill opacity=0.80] (154.72, 64.61) --
	(157.37, 60.03) --
	(152.08, 60.03) --
	cycle;

\path[draw=drawColor,draw opacity=0.80,line width= 0.4pt,line join=round,line cap=round,fill=fillColor,fill opacity=0.80] (171.04, 64.75) --
	(173.69, 60.17) --
	(168.40, 60.17) --
	cycle;

\path[draw=drawColor,draw opacity=0.80,line width= 0.4pt,line join=round,line cap=round,fill=fillColor,fill opacity=0.80] ( 41.31, 65.96) --
	( 43.95, 61.38) --
	( 38.67, 61.38) --
	cycle;
\definecolor{drawColor}{RGB}{0,0,139}
\definecolor{fillColor}{RGB}{0,0,139}

\path[draw=drawColor,draw opacity=0.80,line width= 0.4pt,line join=round,line cap=round,fill=fillColor,fill opacity=0.80] ( 62.61, 60.50) --
	( 65.07, 62.96) --
	( 62.61, 65.42) --
	( 60.15, 62.96) --
	cycle;

\path[draw=drawColor,draw opacity=0.80,line width= 0.4pt,line join=round,line cap=round,fill=fillColor,fill opacity=0.80] (116.83, 58.55) --
	(119.29, 61.01) --
	(116.83, 63.47) --
	(114.37, 61.01) --
	cycle;

\path[draw=drawColor,draw opacity=0.80,line width= 0.4pt,line join=round,line cap=round,fill=fillColor,fill opacity=0.80] (154.72, 59.12) --
	(157.18, 61.58) --
	(154.72, 64.04) --
	(152.27, 61.58) --
	cycle;

\path[draw=drawColor,draw opacity=0.80,line width= 0.4pt,line join=round,line cap=round,fill=fillColor,fill opacity=0.80] (171.04, 59.51) --
	(173.50, 61.97) --
	(171.04, 64.43) --
	(168.59, 61.97) --
	cycle;

\path[draw=drawColor,draw opacity=0.80,line width= 0.4pt,line join=round,line cap=round,fill=fillColor,fill opacity=0.80] ( 41.31, 60.53) --
	( 43.77, 62.98) --
	( 41.31, 65.44) --
	( 38.85, 62.98) --
	cycle;
\definecolor{drawColor}{RGB}{0,245,255}
\definecolor{fillColor}{RGB}{0,245,255}

\path[draw=drawColor,draw opacity=0.80,line width= 0.4pt,line join=round,line cap=round,fill=fillColor,fill opacity=0.80] ( 62.61, 62.70) circle (  1.96);

\path[draw=drawColor,draw opacity=0.80,line width= 0.4pt,line join=round,line cap=round,fill=fillColor,fill opacity=0.80] (116.83, 60.75) circle (  1.96);

\path[draw=drawColor,draw opacity=0.80,line width= 0.4pt,line join=round,line cap=round,fill=fillColor,fill opacity=0.80] (154.72, 60.53) circle (  1.96);

\path[draw=drawColor,draw opacity=0.80,line width= 0.4pt,line join=round,line cap=round,fill=fillColor,fill opacity=0.80] (171.04, 63.99) circle (  1.96);

\path[draw=drawColor,draw opacity=0.80,line width= 0.4pt,line join=round,line cap=round,fill=fillColor,fill opacity=0.80] ( 41.31, 62.77) circle (  1.96);
\definecolor{drawColor}{RGB}{255,0,255}

\path[draw=drawColor,draw opacity=0.80,line width= 0.5pt,line join=round] ( 41.31, 61.68) --
	( 62.61, 60.93) --
	(116.83, 61.26) --
	(154.72, 61.81) --
	(171.04, 61.44);
\definecolor{drawColor}{RGB}{178,34,34}

\path[draw=drawColor,draw opacity=0.80,line width= 0.5pt,line join=round] ( 41.31, 62.98) --
	( 62.61, 62.91) --
	(116.83, 62.98) --
	(154.72, 62.95) --
	(171.04, 61.63);
\definecolor{drawColor}{RGB}{67,205,128}

\path[draw=drawColor,draw opacity=0.80,line width= 0.5pt,line join=round] ( 41.31, 61.07) --
	( 62.61, 61.62) --
	(116.83, 60.99) --
	(154.72, 61.47) --
	(171.04, 61.54);
\definecolor{drawColor}{RGB}{218,165,32}

\path[draw=drawColor,draw opacity=0.80,line width= 0.5pt,line join=round] ( 41.31, 62.91) --
	( 62.61, 61.58) --
	(116.83, 61.06) --
	(154.72, 61.56) --
	(171.04, 61.69);
\definecolor{drawColor}{RGB}{0,0,139}

\path[draw=drawColor,draw opacity=0.80,line width= 0.5pt,line join=round] ( 41.31, 62.98) --
	( 62.61, 62.96) --
	(116.83, 61.01) --
	(154.72, 61.58) --
	(171.04, 61.97);
\definecolor{drawColor}{RGB}{0,245,255}

\path[draw=drawColor,draw opacity=0.80,line width= 0.5pt,line join=round] ( 41.31, 62.77) --
	( 62.61, 62.70) --
	(116.83, 60.75) --
	(154.72, 60.53) --
	(171.04, 63.99);

\path[draw=drawColor,draw opacity=0.80,line width= 0.4pt,line join=round,line cap=round,fill=fillColor,fill opacity=0.80] ( 62.61, 62.37) circle (  1.96);

\path[draw=drawColor,draw opacity=0.80,line width= 0.4pt,line join=round,line cap=round,fill=fillColor,fill opacity=0.80] (116.83, 60.85) circle (  1.96);

\path[draw=drawColor,draw opacity=0.80,line width= 0.4pt,line join=round,line cap=round,fill=fillColor,fill opacity=0.80] (154.72, 60.08) circle (  1.96);

\path[draw=drawColor,draw opacity=0.80,line width= 0.4pt,line join=round,line cap=round,fill=fillColor,fill opacity=0.80] (171.04, 63.79) circle (  1.96);

\path[draw=drawColor,draw opacity=0.80,line width= 0.4pt,line join=round,line cap=round,fill=fillColor,fill opacity=0.80] ( 41.31, 62.54) circle (  1.96);

\path[draw=drawColor,draw opacity=0.80,line width= 0.9pt,dash pattern=on 1pt off 3pt on 4pt off 3pt ,line join=round] ( 41.31, 62.54) --
	( 62.61, 62.37) --
	(116.83, 60.85) --
	(154.72, 60.08) --
	(171.04, 63.79);
\definecolor{drawColor}{RGB}{0,245,255}

\path[draw=drawColor,draw opacity=0.50,line width= 0.6pt,line join=round] ( 62.34, 62.54) --
	( 62.88, 62.54);

\path[draw=drawColor,draw opacity=0.50,line width= 0.6pt,line join=round] ( 62.61, 62.54) --
	( 62.61, 62.20);

\path[draw=drawColor,draw opacity=0.50,line width= 0.6pt,line join=round] ( 62.34, 62.20) --
	( 62.88, 62.20);

\path[draw=drawColor,draw opacity=0.50,line width= 0.6pt,line join=round] (116.56, 60.98) --
	(117.10, 60.98);

\path[draw=drawColor,draw opacity=0.50,line width= 0.6pt,line join=round] (116.83, 60.98) --
	(116.83, 60.71);

\path[draw=drawColor,draw opacity=0.50,line width= 0.6pt,line join=round] (116.56, 60.71) --
	(117.10, 60.71);

\path[draw=drawColor,draw opacity=0.50,line width= 0.6pt,line join=round] (154.45, 60.24) --
	(155.00, 60.24);

\path[draw=drawColor,draw opacity=0.50,line width= 0.6pt,line join=round] (154.72, 60.24) --
	(154.72, 59.92);

\path[draw=drawColor,draw opacity=0.50,line width= 0.6pt,line join=round] (154.45, 59.92) --
	(155.00, 59.92);

\path[draw=drawColor,draw opacity=0.50,line width= 0.6pt,line join=round] (170.77, 64.64) --
	(171.32, 64.64);

\path[draw=drawColor,draw opacity=0.50,line width= 0.6pt,line join=round] (171.04, 64.64) --
	(171.04, 62.71);

\path[draw=drawColor,draw opacity=0.50,line width= 0.6pt,line join=round] (170.77, 62.71) --
	(171.32, 62.71);

\path[draw=drawColor,draw opacity=0.50,line width= 0.6pt,line join=round] ( 41.04, 62.62) --
	( 41.58, 62.62);

\path[draw=drawColor,draw opacity=0.50,line width= 0.6pt,line join=round] ( 41.31, 62.62) --
	( 41.31, 62.46);

\path[draw=drawColor,draw opacity=0.50,line width= 0.6pt,line join=round] ( 41.04, 62.46) --
	( 41.58, 62.46);
\definecolor{drawColor}{RGB}{0,0,0}

\path[draw=drawColor,line width= 0.5pt,line join=round] ( 41.31, 36.39) --
	( 62.61, 36.38) --
	(116.83, 36.42) --
	(154.72, 36.47) --
	(171.04, 36.41);

\path[draw=drawColor,line width= 0.5pt,line join=round] ( 41.31, 36.44) --
	( 62.61, 36.34) --
	(116.83, 36.34) --
	(154.72, 36.38) --
	(171.04, 36.47);

\path[draw=drawColor,line width= 0.5pt,line join=round] ( 41.31, 36.42) --
	( 62.61, 36.37) --
	(116.83, 36.39) --
	(154.72, 36.40) --
	(171.04, 36.39);

\path[draw=drawColor,line width= 0.5pt,line join=round] ( 41.31, 36.42) --
	( 62.61, 36.41) --
	(116.83, 36.39) --
	(154.72, 36.41) --
	(171.04, 36.37);

\path[draw=drawColor,line width= 0.5pt,line join=round] ( 41.31, 36.45) --
	( 62.61, 36.43) --
	(116.83, 36.40) --
	(154.72, 36.44) --
	(171.04, 36.46);

\path[draw=drawColor,line width= 0.5pt,line join=round] ( 41.31, 36.45) --
	( 62.61, 36.41) --
	(116.83, 36.38) --
	(154.72, 36.38) --
	(171.04, 36.44);
\definecolor{drawColor}{gray}{0.60}

\path[draw=drawColor,line width= 0.5pt,line join=round,line cap=round] ( 41.04, 30.28) -- ( 41.04, 27.43);

\path[draw=drawColor,line width= 0.5pt,line join=round,line cap=round] ( 46.29, 30.28) -- ( 46.29, 27.43);

\path[draw=drawColor,line width= 0.5pt,line join=round,line cap=round] ( 50.59, 30.28) -- ( 50.59, 27.43);

\path[draw=drawColor,line width= 0.5pt,line join=round,line cap=round] ( 54.22, 30.28) -- ( 54.22, 27.43);

\path[draw=drawColor,line width= 0.5pt,line join=round,line cap=round] ( 57.36, 30.28) -- ( 57.36, 27.43);

\path[draw=drawColor,line width= 0.5pt,line join=round,line cap=round] ( 60.13, 30.28) -- ( 60.13, 27.43);

\path[draw=drawColor,line width= 0.5pt,line join=round,line cap=round] ( 62.61, 30.28) -- ( 62.61, 24.59);

\path[draw=drawColor,line width= 0.5pt,line join=round,line cap=round] ( 78.93, 30.28) -- ( 78.93, 27.43);

\path[draw=drawColor,line width= 0.5pt,line join=round,line cap=round] ( 88.48, 30.28) -- ( 88.48, 27.43);

\path[draw=drawColor,line width= 0.5pt,line join=round,line cap=round] ( 95.25, 30.28) -- ( 95.25, 27.43);

\path[draw=drawColor,line width= 0.5pt,line join=round,line cap=round] (100.51, 30.28) -- (100.51, 27.43);

\path[draw=drawColor,line width= 0.5pt,line join=round,line cap=round] (104.80, 30.28) -- (104.80, 27.43);

\path[draw=drawColor,line width= 0.5pt,line join=round,line cap=round] (108.43, 30.28) -- (108.43, 27.43);

\path[draw=drawColor,line width= 0.5pt,line join=round,line cap=round] (111.58, 30.28) -- (111.58, 27.43);

\path[draw=drawColor,line width= 0.5pt,line join=round,line cap=round] (114.35, 30.28) -- (114.35, 27.43);

\path[draw=drawColor,line width= 0.5pt,line join=round,line cap=round] (116.83, 30.28) -- (116.83, 24.59);

\path[draw=drawColor,line width= 0.5pt,line join=round,line cap=round] (133.15, 30.28) -- (133.15, 27.43);

\path[draw=drawColor,line width= 0.5pt,line join=round,line cap=round] (142.70, 30.28) -- (142.70, 27.43);

\path[draw=drawColor,line width= 0.5pt,line join=round,line cap=round] (149.47, 30.28) -- (149.47, 27.43);

\path[draw=drawColor,line width= 0.5pt,line join=round,line cap=round] (154.72, 30.28) -- (154.72, 27.43);

\path[draw=drawColor,line width= 0.5pt,line join=round,line cap=round] (159.02, 30.28) -- (159.02, 27.43);

\path[draw=drawColor,line width= 0.5pt,line join=round,line cap=round] (162.65, 30.28) -- (162.65, 27.43);

\path[draw=drawColor,line width= 0.5pt,line join=round,line cap=round] (165.79, 30.28) -- (165.79, 27.43);

\path[draw=drawColor,line width= 0.5pt,line join=round,line cap=round] (168.56, 30.28) -- (168.56, 27.43);

\path[draw=drawColor,line width= 0.5pt,line join=round,line cap=round] (171.04, 30.28) -- (171.04, 24.59);
\end{scope}
\begin{scope}
\definecolor{drawColor}{gray}{0.30}

\node[text=drawColor,anchor=base west,inner sep=0pt, outer sep=0pt, scale=  0.80] at ( 16.91, 28.50) {10};

\node[text=drawColor,anchor=base west,inner sep=0pt, outer sep=0pt, scale=  0.56] at ( 24.91, 31.78) {-4};

\node[text=drawColor,anchor=base west,inner sep=0pt, outer sep=0pt, scale=  0.80] at ( 16.91, 37.46) {10};

\node[text=drawColor,anchor=base west,inner sep=0pt, outer sep=0pt, scale=  0.56] at ( 24.91, 40.73) {-3};

\node[text=drawColor,anchor=base west,inner sep=0pt, outer sep=0pt, scale=  0.80] at ( 16.91, 46.42) {10};

\node[text=drawColor,anchor=base west,inner sep=0pt, outer sep=0pt, scale=  0.56] at ( 24.91, 49.69) {-2};

\node[text=drawColor,anchor=base west,inner sep=0pt, outer sep=0pt, scale=  0.80] at ( 16.91, 55.37) {10};

\node[text=drawColor,anchor=base west,inner sep=0pt, outer sep=0pt, scale=  0.56] at ( 24.91, 58.64) {-1};
\end{scope}
\begin{scope}
\definecolor{drawColor}{gray}{0.60}

\path[draw=drawColor,line width= 0.6pt,line join=round] ( 31.77, 31.94) --
	( 34.52, 31.94);

\path[draw=drawColor,line width= 0.6pt,line join=round] ( 31.77, 40.89) --
	( 34.52, 40.89);

\path[draw=drawColor,line width= 0.6pt,line join=round] ( 31.77, 49.85) --
	( 34.52, 49.85);

\path[draw=drawColor,line width= 0.6pt,line join=round] ( 31.77, 58.80) --
	( 34.52, 58.80);
\end{scope}
\begin{scope}
\definecolor{drawColor}{gray}{0.30}

\node[text=drawColor,anchor=base west,inner sep=0pt, outer sep=0pt, scale=  0.80] at ( 56.28, 16.13) {10};

\node[text=drawColor,anchor=base west,inner sep=0pt, outer sep=0pt, scale=  0.56] at ( 64.28, 19.40) {-2};

\node[text=drawColor,anchor=base west,inner sep=0pt, outer sep=0pt, scale=  0.80] at (110.50, 16.13) {10};

\node[text=drawColor,anchor=base west,inner sep=0pt, outer sep=0pt, scale=  0.56] at (118.50, 19.40) {-1};

\node[text=drawColor,anchor=base west,inner sep=0pt, outer sep=0pt, scale=  0.80] at (165.65, 16.13) {10};

\node[text=drawColor,anchor=base west,inner sep=0pt, outer sep=0pt, scale=  0.56] at (173.64, 19.40) {0};
\end{scope}
\begin{scope}
\definecolor{drawColor}{RGB}{0,0,0}

\node[text=drawColor,anchor=base,inner sep=0pt, outer sep=0pt, scale=  0.80] at (106.18,  4.40) {Target sparsity};
\end{scope}
\begin{scope}
\definecolor{drawColor}{RGB}{0,0,0}

\node[text=drawColor,rotate= 90.00,anchor=base,inner sep=0pt, outer sep=0pt, scale=  0.80] at (  8.36, 48.50) {MSE};
\end{scope}
\end{tikzpicture}

%% file: figs/results_6_helix_posttrain.tikz
\begin{tikzpicture}[x=1pt,y=1pt,baseline={(0,0.4)}]
\definecolor{fillColor}{RGB}{255,255,255}
\begin{scope}
\definecolor{drawColor}{RGB}{255,0,255}

\path[draw=drawColor,line width= 0.6pt,line join=round] ( 39.03, 56.39) --
	( 39.14, 56.39);

\path[draw=drawColor,line width= 0.6pt,line join=round] (104.19, 36.76) --
	(104.30, 36.76);

\path[draw=drawColor,line width= 0.6pt,line join=round] (104.24, 36.76) --
	(104.24, 34.52);

\path[draw=drawColor,line width= 0.6pt,line join=round] (104.19, 34.52) --
	(104.30, 34.52);

\path[draw=drawColor,line width= 0.6pt,line join=round] (149.73, 42.57) --
	(149.84, 42.57);

\path[draw=drawColor,line width= 0.6pt,line join=round] (149.79, 42.57) --
	(149.79, 38.65);

\path[draw=drawColor,line width= 0.6pt,line join=round] (149.73, 38.65) --
	(149.84, 38.65);

\path[draw=drawColor,line width= 0.6pt,line join=round] (169.35, 41.11) --
	(169.46, 41.11);

\path[draw=drawColor,line width= 0.6pt,line join=round] (169.40, 41.11) --
	(169.40, 37.89);

\path[draw=drawColor,line width= 0.6pt,line join=round] (169.35, 37.89) --
	(169.46, 37.89);

\path[draw=drawColor,line width= 0.6pt,line join=round] ( 13.42, 60.31) --
	( 13.53, 60.31);

\path[draw=drawColor,line width= 0.6pt,line join=round] ( 13.48, 60.31) --
	( 13.48, 56.01);

\path[draw=drawColor,line width= 0.6pt,line join=round] ( 13.42, 56.01) --
	( 13.53, 56.01);
\definecolor{drawColor}{RGB}{178,34,34}

\path[draw=drawColor,line width= 0.6pt,line join=round] ( 39.14, 62.99) --
	( 39.25, 62.99);

\path[draw=drawColor,line width= 0.6pt,line join=round] ( 39.19, 62.99) --
	( 39.19, 62.84);

\path[draw=drawColor,line width= 0.6pt,line join=round] ( 39.14, 62.84) --
	( 39.25, 62.84);

\path[draw=drawColor,line width= 0.6pt,line join=round] (104.30, 63.04) --
	(104.41, 63.04);

\path[draw=drawColor,line width= 0.6pt,line join=round] (104.35, 63.04) --
	(104.35, 62.92);

\path[draw=drawColor,line width= 0.6pt,line join=round] (104.30, 62.92) --
	(104.41, 62.92);

\path[draw=drawColor,line width= 0.6pt,line join=round] (149.84, 63.00) --
	(149.95, 63.00);

\path[draw=drawColor,line width= 0.6pt,line join=round] (149.90, 63.00) --
	(149.90, 62.91);

\path[draw=drawColor,line width= 0.6pt,line join=round] (149.84, 62.91) --
	(149.95, 62.91);

\path[draw=drawColor,line width= 0.6pt,line join=round] (169.46, 41.85) --
	(169.57, 41.85);

\path[draw=drawColor,line width= 0.6pt,line join=round] (169.51, 41.85) --
	(169.51, 36.48);

\path[draw=drawColor,line width= 0.6pt,line join=round] (169.46, 36.48) --
	(169.57, 36.48);

\path[draw=drawColor,line width= 0.6pt,line join=round] ( 13.53, 63.06) --
	( 13.64, 63.06);

\path[draw=drawColor,line width= 0.6pt,line join=round] ( 13.59, 63.06) --
	( 13.59, 62.90);

\path[draw=drawColor,line width= 0.6pt,line join=round] ( 13.53, 62.90) --
	( 13.64, 62.90);
\definecolor{drawColor}{RGB}{67,205,128}

\path[draw=drawColor,line width= 0.6pt,line join=round] ( 39.25, 60.68) --
	( 39.36, 60.68);

\path[draw=drawColor,line width= 0.6pt,line join=round] ( 39.30, 60.68) --
	( 39.30, 57.54);

\path[draw=drawColor,line width= 0.6pt,line join=round] ( 39.25, 57.54) --
	( 39.36, 57.54);

\path[draw=drawColor,line width= 0.6pt,line join=round] (104.41, 42.16) --
	(104.52, 42.16);

\path[draw=drawColor,line width= 0.6pt,line join=round] (104.46, 42.16) --
	(104.46, 40.10);

\path[draw=drawColor,line width= 0.6pt,line join=round] (104.41, 40.10) --
	(104.52, 40.10);

\path[draw=drawColor,line width= 0.6pt,line join=round] (149.95, 43.26) --
	(150.06, 43.26);

\path[draw=drawColor,line width= 0.6pt,line join=round] (150.01, 43.26) --
	(150.01, 36.72);

\path[draw=drawColor,line width= 0.6pt,line join=round] (149.95, 36.72) --
	(150.06, 36.72);

\path[draw=drawColor,line width= 0.6pt,line join=round] (169.57, 41.01) --
	(169.68, 41.01);

\path[draw=drawColor,line width= 0.6pt,line join=round] (169.62, 41.01) --
	(169.62, 38.55);

\path[draw=drawColor,line width= 0.6pt,line join=round] (169.57, 38.55) --
	(169.68, 38.55);

\path[draw=drawColor,line width= 0.6pt,line join=round] ( 13.64, 60.02) --
	( 13.75, 60.02);

\path[draw=drawColor,line width= 0.6pt,line join=round] ( 13.70, 60.02) --
	( 13.70, 59.28);

\path[draw=drawColor,line width= 0.6pt,line join=round] ( 13.64, 59.28) --
	( 13.75, 59.28);
\definecolor{drawColor}{RGB}{218,165,32}

\path[draw=drawColor,line width= 0.6pt,line join=round] ( 39.36, 60.37) --
	( 39.46, 60.37);

\path[draw=drawColor,line width= 0.6pt,line join=round] ( 39.41, 60.37) --
	( 39.41, 53.73);

\path[draw=drawColor,line width= 0.6pt,line join=round] ( 39.36, 53.73) --
	( 39.46, 53.73);

\path[draw=drawColor,line width= 0.6pt,line join=round] (104.52, 36.08) --
	(104.62, 36.08);

\path[draw=drawColor,line width= 0.6pt,line join=round] (104.57, 36.08) --
	(104.57, 34.43);

\path[draw=drawColor,line width= 0.6pt,line join=round] (104.52, 34.43) --
	(104.62, 34.43);

\path[draw=drawColor,line width= 0.6pt,line join=round] (150.06, 40.18) --
	(150.17, 40.18);

\path[draw=drawColor,line width= 0.6pt,line join=round] (150.11, 40.18) --
	(150.11, 36.29);

\path[draw=drawColor,line width= 0.6pt,line join=round] (150.06, 36.29) --
	(150.17, 36.29);

\path[draw=drawColor,line width= 0.6pt,line join=round] (169.68, 44.87) --
	(169.78, 44.87);

\path[draw=drawColor,line width= 0.6pt,line join=round] (169.73, 44.87) --
	(169.73, 40.89);

\path[draw=drawColor,line width= 0.6pt,line join=round] (169.68, 40.89) --
	(169.78, 40.89);

\path[draw=drawColor,line width= 0.6pt,line join=round] ( 13.75, 62.98) --
	( 13.86, 62.98);

\path[draw=drawColor,line width= 0.6pt,line join=round] ( 13.80, 62.98) --
	( 13.80, 62.79);

\path[draw=drawColor,line width= 0.6pt,line join=round] ( 13.75, 62.79) --
	( 13.86, 62.79);
\definecolor{drawColor}{RGB}{0,0,139}

\path[draw=drawColor,line width= 0.6pt,line join=round] ( 39.46, 63.01) --
	( 39.57, 63.01);

\path[draw=drawColor,line width= 0.6pt,line join=round] ( 39.52, 63.01) --
	( 39.52, 62.91);

\path[draw=drawColor,line width= 0.6pt,line join=round] ( 39.46, 62.91) --
	( 39.57, 62.91);

\path[draw=drawColor,line width= 0.6pt,line join=round] (104.62, 47.92) --
	(104.73, 47.92);

\path[draw=drawColor,line width= 0.6pt,line join=round] (150.17, 38.85) --
	(150.28, 38.85);

\path[draw=drawColor,line width= 0.6pt,line join=round] (150.22, 38.85) --
	(150.22, 36.06);

\path[draw=drawColor,line width= 0.6pt,line join=round] (150.17, 36.06) --
	(150.28, 36.06);

\path[draw=drawColor,line width= 0.6pt,line join=round] (169.78, 42.33) --
	(169.89, 42.33);

\path[draw=drawColor,line width= 0.6pt,line join=round] (169.84, 42.33) --
	(169.84, 34.84);

\path[draw=drawColor,line width= 0.6pt,line join=round] (169.78, 34.84) --
	(169.89, 34.84);

\path[draw=drawColor,line width= 0.6pt,line join=round] ( 13.86, 63.03) --
	( 13.97, 63.03);

\path[draw=drawColor,line width= 0.6pt,line join=round] ( 13.91, 63.03) --
	( 13.91, 62.94);

\path[draw=drawColor,line width= 0.6pt,line join=round] ( 13.86, 62.94) --
	( 13.97, 62.94);
\definecolor{fillColor}{RGB}{255,0,255}

\path[fill=fillColor,fill opacity=0.80] ( 37.39, 50.81) --
	( 41.32, 50.81) --
	( 41.32, 54.73) --
	( 37.39, 54.73) --
	cycle;

\path[fill=fillColor,fill opacity=0.80] (102.55, 33.84) --
	(106.48, 33.84) --
	(106.48, 37.76) --
	(102.55, 37.76) --
	cycle;

\path[fill=fillColor,fill opacity=0.80] (148.10, 39.12) --
	(152.02, 39.12) --
	(152.02, 43.04) --
	(148.10, 43.04) --
	cycle;

\path[fill=fillColor,fill opacity=0.80] (167.71, 37.86) --
	(171.64, 37.86) --
	(171.64, 41.79) --
	(167.71, 41.79) --
	cycle;

\path[fill=fillColor,fill opacity=0.80] ( 11.79, 56.76) --
	( 15.71, 56.76) --
	( 15.71, 60.69) --
	( 11.79, 60.69) --
	cycle;
\definecolor{drawColor}{RGB}{178,34,34}
\definecolor{fillColor}{RGB}{178,34,34}

\path[draw=drawColor,draw opacity=0.80,line width= 0.4pt,line join=round,line cap=round,fill=fillColor,fill opacity=0.80] ( 39.36, 59.86) --
	( 42.00, 64.44) --
	( 36.71, 64.44) --
	cycle;

\path[draw=drawColor,draw opacity=0.80,line width= 0.4pt,line join=round,line cap=round,fill=fillColor,fill opacity=0.80] (104.52, 59.93) --
	(107.16, 64.50) --
	(101.87, 64.50) --
	cycle;

\path[draw=drawColor,draw opacity=0.80,line width= 0.4pt,line join=round,line cap=round,fill=fillColor,fill opacity=0.80] (150.06, 59.90) --
	(152.70, 64.48) --
	(147.42, 64.48) --
	cycle;

\path[draw=drawColor,draw opacity=0.80,line width= 0.4pt,line join=round,line cap=round,fill=fillColor,fill opacity=0.80] (169.68, 36.97) --
	(172.32, 41.55) --
	(167.03, 41.55) --
	cycle;

\path[draw=drawColor,draw opacity=0.80,line width= 0.4pt,line join=round,line cap=round,fill=fillColor,fill opacity=0.80] ( 13.75, 59.93) --
	( 16.39, 64.51) --
	( 11.11, 64.51) --
	cycle;
\definecolor{drawColor}{RGB}{67,205,128}
\definecolor{fillColor}{RGB}{67,205,128}

\path[draw=drawColor,draw opacity=0.80,line width= 0.4pt,line join=round,line cap=round,fill=fillColor,fill opacity=0.80] ( 39.36, 56.96) --
	( 41.81, 59.42) --
	( 39.36, 61.88) --
	( 36.90, 59.42) --
	cycle;

\path[draw=drawColor,draw opacity=0.80,line width= 0.4pt,line join=round,line cap=round,fill=fillColor,fill opacity=0.80] (104.52, 38.80) --
	(106.97, 41.26) --
	(104.52, 43.72) --
	(102.06, 41.26) --
	cycle;

\path[draw=drawColor,draw opacity=0.80,line width= 0.4pt,line join=round,line cap=round,fill=fillColor,fill opacity=0.80] (150.06, 38.77) --
	(152.52, 41.23) --
	(150.06, 43.69) --
	(147.60, 41.23) --
	cycle;

\path[draw=drawColor,draw opacity=0.80,line width= 0.4pt,line join=round,line cap=round,fill=fillColor,fill opacity=0.80] (169.68, 37.52) --
	(172.13, 39.98) --
	(169.68, 42.43) --
	(167.22, 39.98) --
	cycle;

\path[draw=drawColor,draw opacity=0.80,line width= 0.4pt,line join=round,line cap=round,fill=fillColor,fill opacity=0.80] ( 13.75, 57.21) --
	( 16.21, 59.67) --
	( 13.75, 62.13) --
	( 11.29, 59.67) --
	cycle;
\definecolor{drawColor}{RGB}{218,165,32}
\definecolor{fillColor}{RGB}{218,165,32}

\path[draw=drawColor,draw opacity=0.80,line width= 0.4pt,line join=round,line cap=round,fill=fillColor,fill opacity=0.80] ( 39.36, 61.38) --
	( 42.00, 56.80) --
	( 36.71, 56.80) --
	cycle;

\path[draw=drawColor,draw opacity=0.80,line width= 0.4pt,line join=round,line cap=round,fill=fillColor,fill opacity=0.80] (104.52, 38.39) --
	(107.16, 33.82) --
	(101.87, 33.82) --
	cycle;

\path[draw=drawColor,draw opacity=0.80,line width= 0.4pt,line join=round,line cap=round,fill=fillColor,fill opacity=0.80] (150.06, 41.76) --
	(152.70, 37.18) --
	(147.42, 37.18) --
	cycle;

\path[draw=drawColor,draw opacity=0.80,line width= 0.4pt,line join=round,line cap=round,fill=fillColor,fill opacity=0.80] (169.68, 46.42) --
	(172.32, 41.84) --
	(167.03, 41.84) --
	cycle;

\path[draw=drawColor,draw opacity=0.80,line width= 0.4pt,line join=round,line cap=round,fill=fillColor,fill opacity=0.80] ( 13.75, 65.94) --
	( 16.39, 61.36) --
	( 11.11, 61.36) --
	cycle;
\definecolor{drawColor}{RGB}{0,0,139}
\definecolor{fillColor}{RGB}{0,0,139}

\path[draw=drawColor,draw opacity=0.80,line width= 0.4pt,line join=round,line cap=round,fill=fillColor,fill opacity=0.80] ( 39.36, 60.50) --
	( 41.81, 62.96) --
	( 39.36, 65.42) --
	( 36.90, 62.96) --
	cycle;

\path[draw=drawColor,draw opacity=0.80,line width= 0.4pt,line join=round,line cap=round,fill=fillColor,fill opacity=0.80] (104.52, 42.77) --
	(106.97, 45.23) --
	(104.52, 47.69) --
	(102.06, 45.23) --
	cycle;

\path[draw=drawColor,draw opacity=0.80,line width= 0.4pt,line join=round,line cap=round,fill=fillColor,fill opacity=0.80] (150.06, 35.24) --
	(152.52, 37.70) --
	(150.06, 40.16) --
	(147.60, 37.70) --
	cycle;

\path[draw=drawColor,draw opacity=0.80,line width= 0.4pt,line join=round,line cap=round,fill=fillColor,fill opacity=0.80] (169.68, 37.71) --
	(172.13, 40.17) --
	(169.68, 42.63) --
	(167.22, 40.17) --
	cycle;

\path[draw=drawColor,draw opacity=0.80,line width= 0.4pt,line join=round,line cap=round,fill=fillColor,fill opacity=0.80] ( 13.75, 60.53) --
	( 16.21, 62.98) --
	( 13.75, 65.44) --
	( 11.29, 62.98) --
	cycle;
\definecolor{drawColor}{RGB}{255,0,255}

\path[draw=drawColor,draw opacity=0.80,line width= 0.5pt,line join=round] ( 13.75, 58.73) --
	( 39.36, 52.77) --
	(104.52, 35.80) --
	(150.06, 41.08) --
	(169.68, 39.82);
\definecolor{drawColor}{RGB}{178,34,34}

\path[draw=drawColor,draw opacity=0.80,line width= 0.5pt,line join=round] ( 13.75, 62.98) --
	( 39.36, 62.91) --
	(104.52, 62.98) --
	(150.06, 62.95) --
	(169.68, 40.02);
\definecolor{drawColor}{RGB}{67,205,128}

\path[draw=drawColor,draw opacity=0.80,line width= 0.5pt,line join=round] ( 13.75, 59.67) --
	( 39.36, 59.42) --
	(104.52, 41.26) --
	(150.06, 41.23) --
	(169.68, 39.98);
\definecolor{drawColor}{RGB}{218,165,32}

\path[draw=drawColor,draw opacity=0.80,line width= 0.5pt,line join=round] ( 13.75, 62.88) --
	( 39.36, 58.32) --
	(104.52, 35.34) --
	(150.06, 38.71) --
	(169.68, 43.37);
\definecolor{drawColor}{RGB}{0,0,139}

\path[draw=drawColor,draw opacity=0.80,line width= 0.5pt,line join=round] ( 13.75, 62.98) --
	( 39.36, 62.96) --
	(104.52, 45.23) --
	(150.06, 37.70) --
	(169.68, 40.17);
\definecolor{drawColor}{RGB}{0,0,0}

\path[draw=drawColor,line width= 0.5pt,line join=round] ( 13.75, 36.39) --
	( 39.36, 36.38) --
	(104.52, 36.42) --
	(150.06, 36.47) --
	(169.68, 36.41);

\path[draw=drawColor,line width= 0.5pt,line join=round] ( 13.75, 36.44) --
	( 39.36, 36.34) --
	(104.52, 36.34) --
	(150.06, 36.38) --
	(169.68, 36.47);

\path[draw=drawColor,line width= 0.5pt,line join=round] ( 13.75, 36.42) --
	( 39.36, 36.37) --
	(104.52, 36.39) --
	(150.06, 36.40) --
	(169.68, 36.39);

\path[draw=drawColor,line width= 0.5pt,line join=round] ( 13.75, 36.42) --
	( 39.36, 36.41) --
	(104.52, 36.39) --
	(150.06, 36.41) --
	(169.68, 36.37);

\path[draw=drawColor,line width= 0.5pt,line join=round] ( 13.75, 36.45) --
	( 39.36, 36.43) --
	(104.52, 36.40) --
	(150.06, 36.44) --
	(169.68, 36.46);

\path[draw=drawColor,line width= 0.5pt,line join=round] ( 13.75, 36.45) --
	( 39.36, 36.41) --
	(104.52, 36.38) --
	(150.06, 36.38) --
	(169.68, 36.44);
\definecolor{drawColor}{gray}{0.60}

\path[draw=drawColor,line width= 0.5pt,line join=round,line cap=round] ( 13.43, 30.28) -- ( 13.43, 27.43);

\path[draw=drawColor,line width= 0.5pt,line join=round,line cap=round] ( 19.74, 30.28) -- ( 19.74, 27.43);

\path[draw=drawColor,line width= 0.5pt,line join=round,line cap=round] ( 24.90, 30.28) -- ( 24.90, 27.43);

\path[draw=drawColor,line width= 0.5pt,line join=round,line cap=round] ( 29.26, 30.28) -- ( 29.26, 27.43);

\path[draw=drawColor,line width= 0.5pt,line join=round,line cap=round] ( 33.04, 30.28) -- ( 33.04, 27.43);

\path[draw=drawColor,line width= 0.5pt,line join=round,line cap=round] ( 36.37, 30.28) -- ( 36.37, 27.43);

\path[draw=drawColor,line width= 0.5pt,line join=round,line cap=round] ( 39.36, 30.28) -- ( 39.36, 24.59);

\path[draw=drawColor,line width= 0.5pt,line join=round,line cap=round] ( 58.97, 30.28) -- ( 58.97, 27.43);

\path[draw=drawColor,line width= 0.5pt,line join=round,line cap=round] ( 70.44, 30.28) -- ( 70.44, 27.43);

\path[draw=drawColor,line width= 0.5pt,line join=round,line cap=round] ( 78.59, 30.28) -- ( 78.59, 27.43);

\path[draw=drawColor,line width= 0.5pt,line join=round,line cap=round] ( 84.90, 30.28) -- ( 84.90, 27.43);

\path[draw=drawColor,line width= 0.5pt,line join=round,line cap=round] ( 90.06, 30.28) -- ( 90.06, 27.43);

\path[draw=drawColor,line width= 0.5pt,line join=round,line cap=round] ( 94.42, 30.28) -- ( 94.42, 27.43);

\path[draw=drawColor,line width= 0.5pt,line join=round,line cap=round] ( 98.20, 30.28) -- ( 98.20, 27.43);

\path[draw=drawColor,line width= 0.5pt,line join=round,line cap=round] (101.53, 30.28) -- (101.53, 27.43);

\path[draw=drawColor,line width= 0.5pt,line join=round,line cap=round] (104.52, 30.28) -- (104.52, 24.59);

\path[draw=drawColor,line width= 0.5pt,line join=round,line cap=round] (124.13, 30.28) -- (124.13, 27.43);

\path[draw=drawColor,line width= 0.5pt,line join=round,line cap=round] (135.60, 30.28) -- (135.60, 27.43);

\path[draw=drawColor,line width= 0.5pt,line join=round,line cap=round] (143.75, 30.28) -- (143.75, 27.43);

\path[draw=drawColor,line width= 0.5pt,line join=round,line cap=round] (150.06, 30.28) -- (150.06, 27.43);

\path[draw=drawColor,line width= 0.5pt,line join=round,line cap=round] (155.22, 30.28) -- (155.22, 27.43);

\path[draw=drawColor,line width= 0.5pt,line join=round,line cap=round] (159.58, 30.28) -- (159.58, 27.43);

\path[draw=drawColor,line width= 0.5pt,line join=round,line cap=round] (163.36, 30.28) -- (163.36, 27.43);

\path[draw=drawColor,line width= 0.5pt,line join=round,line cap=round] (166.69, 30.28) -- (166.69, 27.43);

\path[draw=drawColor,line width= 0.5pt,line join=round,line cap=round] (169.68, 30.28) -- (169.68, 24.59);
\end{scope}
\begin{scope}
\definecolor{drawColor}{gray}{0.60}

\path[draw=drawColor,line width= 0.6pt,line join=round] (  2.85, 31.94) --
	(  5.60, 31.94);

\path[draw=drawColor,line width= 0.6pt,line join=round] (  2.85, 40.89) --
	(  5.60, 40.89);

\path[draw=drawColor,line width= 0.6pt,line join=round] (  2.85, 49.85) --
	(  5.60, 49.85);

\path[draw=drawColor,line width= 0.6pt,line join=round] (  2.85, 58.80) --
	(  5.60, 58.80);
\end{scope}
\begin{scope}
\definecolor{drawColor}{gray}{0.30}

\node[text=drawColor,anchor=base west,inner sep=0pt, outer sep=0pt, scale=  0.80] at ( 33.02, 16.13) {10};

\node[text=drawColor,anchor=base west,inner sep=0pt, outer sep=0pt, scale=  0.56] at ( 41.02, 19.40) {-2};

\node[text=drawColor,anchor=base west,inner sep=0pt, outer sep=0pt, scale=  0.80] at ( 98.18, 16.13) {10};

\node[text=drawColor,anchor=base west,inner sep=0pt, outer sep=0pt, scale=  0.56] at (106.18, 19.40) {-1};

\node[text=drawColor,anchor=base west,inner sep=0pt, outer sep=0pt, scale=  0.80] at (164.28, 16.13) {10};

\node[text=drawColor,anchor=base west,inner sep=0pt, outer sep=0pt, scale=  0.56] at (172.27, 19.40) {0};
\end{scope}
\begin{scope}
\definecolor{drawColor}{RGB}{0,0,0}

\node[text=drawColor,anchor=base,inner sep=0pt, outer sep=0pt, scale=  0.80] at ( 91.71,  4.40) {Target sparsity};
\end{scope}
\end{tikzpicture}

%% file: figs/vgg_cifar_fig.tex
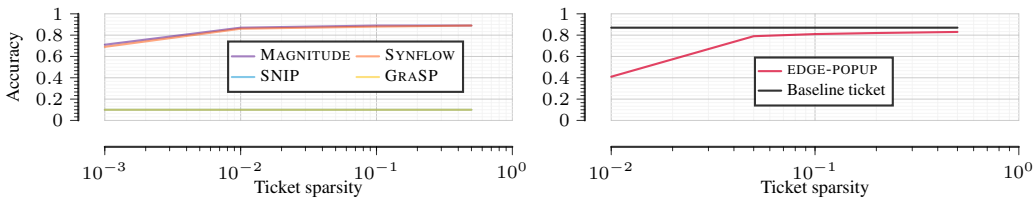
\begin{figure}
  \begin{subfigure}[t]{0.49\textwidth}
   \begin{tikzpicture}
    \begin{axis}[
     jonas line,
     xlabel = {Ticket sparsity}, 
     ylabel= {Accuracy},
     width = 7cm,
     height = 3cm,
     xmode=log,
     xmin=0.001, xmax=1,
     ymin=0, ymax=1,
     legend columns = 2,
     x label style = {at={(axis description cs:0.5,-0.1)}, anchor=north, font=\scriptsize},
     y label style = {at={(axis description cs:0.05,0.5)}, anchor=south, font=\scriptsize},
		legend style={nodes={scale=0.9, transform shape}, at={(0.3,0.75)}, anchor=north west, row sep=-1.4pt, font=\tiny}
    ]
	\addplot+[
     color=indigo(web),
     opacity=0.5,
     mark=triangle*,
    ]
    coordinates {
    (0.001,0.71)
    (0.01,0.87)
    (0.1,0.89)
    (0.5,0.89)
    }; \addlegendentry{\magnitude}
	\addplot+[
     color=internationalorange,
     mark=square*,
     opacity=0.5,
    ]
    coordinates {
    (0.001,0.69)
    (0.01,0.86)
    (0.1,0.88)
    (0.5,0.89)
    }; \addlegendentry{\synflow}
	\addplot+[
     color=richelectricblue,
     opacity=0.5,
     mark=diamond*,
    ]
    coordinates {
    (0.001,0.1)
    (0.01,0.1)
    (0.1,0.1)
    (0.5,0.1)
    }; \addlegendentry{\snip}
    \addplot+[
     color=goldenpoppy,
     mark=triangle*,
     opacity=0.5,
     mark options={rotate=180}
    ]
    coordinates {
    (0.001,0.1)
    (0.01,0.1)
    (0.1,0.1)
    (0.5,0.1)
    }; \addlegendentry{\grasp}
   \end{axis}
  \end{tikzpicture}%
  \end{subfigure}
  \hfill
  \begin{subfigure}[t]{0.49\textwidth}
   \begin{tikzpicture}
    \begin{axis}[
     jonas line,
     xlabel = {Ticket sparsity}, 
     ylabel= {},
     width = 7cm,
     height = 3cm,
     xmode=log,
     xmin=0.01, xmax=1,
     ymin=0, ymax=1,
     x label style = {at={(axis description cs:0.5,-0.1)}, anchor=north, font=\scriptsize},
     y label style = {at={(axis description cs:0.05,0.5)}, anchor=south, font=\scriptsize},
		legend style={nodes={scale=0.9, transform shape}, at={(0.35,0.6)}, anchor=north west, row sep=-1.4pt, font=\tiny}
    ]
	\addplot+[
    color=crimson,
     mark=*,
    ]
    coordinates {
    (0.01,0.41)
    (0.05,0.79)
    (0.1,0.81)
    (0.2,0.82)
    (0.5,0.83)
    };\addlegendentry{\edgepopup}
    \addplot[mark=asterisk, black] coordinates {(0.01,0.87) (0.5,0.87)};\addlegendentry{Baseline ticket}
   \end{axis}
  \end{tikzpicture}%
  \end{subfigure}
  \caption{\textit{VGG16 CIFAR10 results.} (Left) Performance for learned weak tickets. (Right) Performance of strong tickets discovered by \edgepopup for VGG with planted baseline ticket of sparsity $0.01$. Baseline ticket performance is indicated by black line. }\label{fig:vgg_edgepopup}
 \end{figure}

%% file: discussion.tex
\section{Discussion \& Conclusion}\label{sec:discussion}

We investigated the optimality of existing lottery ticket pruning methods and their potential for improvement, both regarding the discovery of strong tickets -- subnetworks that perform well at initialization -- as well as weak tickets -- subnetworks that require further training.
Recent works, in particular by \cite{frankle2021review}, evaluated pruning methods and showed that no single best method across considered settings and sparsities exists, also raising the issue of missing baselines.
To tackle this, we here proposed an algorithm that plants and hides target networks within a larger network, thus allowing to construct benchmarks for lottery tickets.
To construct parameter efficient tickets, we leveraged recent results that transferred the strong lottery ticket hypothesis to neural networks with arbitrary initial biases and proved the existence of lottery tickets under realistic conditions with respect to the network width and initialization scheme.
For three challenging tasks, we hand-crafted extremely sparse network topologies, planted them in large neural networks, and evaluated the state-of-the-art pruning methods in combination with different pruning strategies.

Intriguingly, we found that none of the approaches is able to discover strong tickets, in particular not the planted ticket,  in a single shot, and only at sparsity levels $\geq 0.1$ the methods are able to find weak tickets.
One of the problems that arises is layer collapse. While layer-wise pruning could prevent this collapse, it still results in an interruption of flow.
With a multishot approach, iteratively training, pruning, and resetting weights, certain methods were able to recover weak tickets for a classification task, even at sparsity similar to the planted ticket.
However, only \edgepopup was able to recover strong tickets in this setting, but only at sparsity orders of magnitude larger than the planted baseline.
Through annealing of the target sparsity, we were able to improve \edgepopup, discovering an order of magnitude sparser strong ticket than before, yet still not as sparse as the baseline.
Furthermore, we observed that all considered methods struggled more with regression than with classification tasks, where they recovered only weak tickets of at best $10\%$ sparsity.

These result on our benchmark data are coherent with reported as well as reproduced classification results on image data.
This shows that our benchmarks, while artificial in nature, reflect realistic conditions that result in similar trends as real world image data sets would.
Moreover, we have shown that our planting framework can also be used in a real data setting to answer open questions. 
For instance, by planting a trained weak ticket back into a CNN, we established that the failure of \edgepopup to discover extremely sparse strong lottery tickets is likely an algorithmic rather than a fundamental limitation.
%
This shows how our framework enables experiments beyond relative method comparisons that are typically conducted on standard image benchmark data.

As our results indicate, several major questions pertaining to neural network pruning are still open:
How can pruning approaches for weak tickets be improved to discover tickets of best possible sparsity?
How can we find weak tickets of high sparsity that match the performance of the large network without intermediate training rounds?
And how can we discover highly sparse strong lottery tickets?
We anticipate that our contribution can be used and extended to measure progress regarding these questions against independent sparse and well-performing baseline tickets. 

%% file: appendixtheory.tex
\section{Theory}
In the following section, we present the proofs of the theorems and lemmas of the main manuscript.

\subsection{Error propagation: Proof of Lemma~\ref{thm:approx}}
\begin{theorem*}
Assume $\epsilon > 0$ and let the target network $f$ and its approximation $f_{\epsilon}$ have the same architecture.
If every parameter $\theta$ of $f$ and corresponding $\theta_{\epsilon}$ of $f_{\epsilon}$ in layer $l$ fulfils $|\theta_{\epsilon} - \theta| \leq \epsilon_{l}$ for 
\begin{align}\label{eq:epsTheta2}
\epsilon_l := \epsilon \left(L \sqrt{m_{l}} \left(1+\sup_{x \in {[-1,1]}^{n_0}} \norm{\bm{x}^{(l-1)}}_{1}\right) \prod^L_{k=l+1} \left(\norm{\bm{W}^{(l)}}_{\infty} + \epsilon/L \right) \right)^{-1},
\end{align}
then it follows that $\norm{f-f_{\epsilon}}_{\infty} \leq \epsilon$. 
\end{theorem*}
\begin{proof}
Our objective is to bound $\norm{f-f_{\epsilon}}_{\infty} \leq \epsilon$.
We frequently use the triangle inequality and that $|\phi(x)-\phi(y)| \leq |x-y|$ is Lipschitz continuous with Lipschitz constant $1$ to derive
\begin{align*}
\norm{\bm{x^{(l)}} - \bm{x^{(l)}_{\epsilon}}}_2 & \leq \norm{\bm{h^{(l)}} - \bm{h^{(l)}_{\epsilon}}}_2 \\
& \leq  \norm{ \left(\bm{W^{(l)}} - \bm{W^{(l)}_{\epsilon}}\right) \bm{x^{(l-1)}} }_2   + \norm{ \bm{b^{(l)}}- \bm{b^{(l)}_{\epsilon}}}_2 + \norm{\bm{W^{(l)}_{\epsilon}} \left(\bm{x^{(l-1)}} - \bm{x^{(l-1)}_{\epsilon}}\right)}_2\\
    & \leq \epsilon_l \sqrt{m_{l}} \sup_{x \in {[-1,1]}^{n_0}} \norm{\bm{x}^{(l-1)}}_{1} + \epsilon_l \sqrt{m_{l}} + \left(\norm{\bm{W^{(l)}}}_{\infty} + \epsilon_l \right) \norm{\left(\bm{x^{(l-1)}} - \bm{x^{(l-1)}_{\epsilon}}\right)}_2
\end{align*}
with $\epsilon_l \leq \epsilon/L$. 
$m_{l}$ denotes the number of parameters in layer $l$ that are smaller than $\epsilon_l$ and $\norm{\bm{W}}_{\infty} =  \max_{i,j} |w_{i,j}|$.
Note that $m_{l} \leq n_l k_{l,\text{max}}$. 
The last inequality follows from the fact that all entries of the matrix $\left(\bm{W^{(l)}} - \bm{W^{(l)}_{\epsilon}}\right)$ and of the vector $(\bm{b^{(l)}}- \bm{b^{(l)}_{\epsilon}})$ are bounded by $\epsilon_l$ and maximally $m_l$ of these entries are non-zero.
Furthermore, $\norm{\bm{W^{(l)}_{\epsilon}}}_{\infty} \leq \left(\norm{\bm{W^{(l)}}}_{\infty} + \epsilon_l \right)$ follows again from the fact that each entry of $\left(\bm{W^{(l)}} - \bm{W^{(l)}_{\epsilon}}\right)$ is bounded by $\epsilon_l$.

Thus, at the last layer it holds for all $\bm{x} \in [-1,1]^{n_0}$ that
\begin{align*}
    \norm{f(x)-f_{\epsilon}(x)}_2 & = \norm{\bm{x^{(L)}} - \bm{x^{(L)}_{\epsilon}}}_2 \\
    & \leq \sum^L_{l=1} \epsilon_l \sqrt{m_{l}} \left(1+\sup_{x \in {[-1,1]}^{n_0}} \norm{\bm{x}^{(l-1)}}_{1}\right) \prod^L_{k=l+1} \left(\norm{\bm{W}^{(l)}}_{\infty} + \epsilon/L \right) \leq L \frac{\epsilon}{L} = \epsilon,
\end{align*}
using the definition of $\epsilon_l$ in the last step.
\end{proof}

\subsection{Existence of sparse lottery tickets: Proof of Theorem~\ref{thm:LTexistFullDepth}}\label{sec:existproof}
Next, we prove the following lower bound on the probability that a very sparse lottery ticket exists.
\begin{theorem*}
Assume that ${\epsilon \in {(0,1)}}$ and a target network $f$ with depth $L$ and architecture $\bar{n}$ are given. Each parameter of the larger deep neural network $f_{0}$ with depth $L$ and architecture $\bar{n}_0$ is initialized independently, uniformly at random with $w^{(l)}_{ij} \sim U{\left(\left[-\sigma^{(l)}_w, \sigma^{(l)}_w\right]\right)}$ and $b^{(l)}_{i} \sim U{\left(\left[- \prod^l_{k=1}\sigma^{(k)}_w, \prod^l_{k=1} \sigma^{(k)}_w\right]\right)}$. Then, $f_{0}$ contains a rescaled approximation $f_{\epsilon}$ of $f$ with probability at least
\begin{align}
  \mathbb{P}\left(\exists f_{\epsilon} \subset f_{0}: \ \norm{f- \lambda f_{\epsilon}}_{\infty} \leq \epsilon \right) \geq \prod^L_{l=1} \left(1 - \sum^{n_l}_{i=1}(1 - \epsilon^{k_i}_l )^{n_{l,0}} \right),
\end{align}
where $\epsilon_l$ is defined as in Eq.~(\ref{eq:epsTheta}) and the scaling factor is given by $\lambda = \prod^L_{l=1}1/\sigma^{(l)}_w$.
\end{theorem*}
\begin{proof}
As shown by \cite{nonzerobiases}, the scaling of the output by $\lambda$ simplifies the above parameter initialization to an equivalent setting, in which each parameter is distributed as $w_{ij}, b_i \sim {U[-1,1]}$, while the overall output is scaled by the stated scaling factor $\lambda$, again assuming that all parameters are bounded by $1-\epsilon$.
Each parameter in Layer $l$ needs to be approximated up to error $\epsilon_l$ according to Lemma~\ref{thm:approx}.
To match the same sparsity level of $f$, for each neuron $i$ in each Layer $l$ of $f$, we have to find exactly one neuron in the same layer (Layer $l$) of $f_0$ that represents $i$.
We start with matching neurons at Layer $1$ (given the input in Layer $0$) and proceed iteratively by matching neurons in Layer $l$ given the already matched neurons in Layer $l-1$.

Let us pick a random neuron in Layer $l$ of $f_0$.
How high is the probability that it is a match with a given target neuron $i$ in layer $l$ of $f$?
The neuron $i$ consists of $k_{i}$ parameters that have to be matched. 
Since the corresponding neurons in layer $l-1$ of $f$ and $f_0$ have already been matched according to our assumption, we only have one possible candidate $\theta_0$ for each of the $k_{i}$ parameters $\theta_i$.
For uniformly distributed parameters, we have $|\theta_i - \theta_0| \leq \epsilon_l$ with probability $\epsilon_l$.
For normally distributed $\theta_0 \sim \mathcal{N}(0,1)$, the probability is at least $\epsilon_l/2$ (as long as $|\theta_0 \pm \epsilon| \leq 1$.
This can be seen by Taylor approximation of the cdf of a standard normal $\Phi(z+\Delta z) - \Phi(z-\Delta z)$ in $z$. 
For the remainder of the proof, however, we assume uniformly distributed parameters.
Thus, all $k_i$ independent parameters are a match with probability $\epsilon^{k_i}_l$.
Accordingly, none of the available $n_{l,0}$ neurons in Layer $l$ is a match with probability $\left(1-\epsilon^{k_i}_l\right)^{n_{l,0}}$.

With the help of a union bound we can deduce that the probability that at least one of the neurons $i$ in Layer $l$ of $f$ has no match in $f_0$ is smaller or equal to $\sum^{n_l}_{i=1} \left(1-\epsilon^{k_i}_l\right)^{n_{l,0}}$.
Therefore, the converse probability that we find a match for every single neuron in Layer $l$ of $f$ is at least $1-\sum^{n_l}_{i=1} \left(1-\epsilon^{k_i}_l\right)^{n_{l,0}}$.

Since we have to guarantee a match for each single layer and the matching probability of a new layer is conditional on the previous layer, we obtain a lower bound on the existence probability of a lottery ticket by multiplying the layerwise bounds.
\end{proof}

This bound is only practical for very sparse target networks $f$ with neurons of small in-degrees $k_i$. 
It still shows that the existence of very sparse lottery tickets is possible under the right conditions. 
This extreme sparsity, however can pose significant challenges to pruning algorithms, as we also see in our experiments with planted solutions.
Inspired by this proof, we therefore explain next how to plant tickets (which could have variable sparsity levels).

\subsection{Planting algorithm}\label{sec:plantingalgo}
The idea of layerwise matching of neurons starting with the layer closest to the input can be transferred to planting.
The approach applies to general initial networks $f_0$ whose parameters have not necessarily been randomly initialized and captures the full variability of possible target solutions by allowing for different scaling factors per neuron.

We search in each layer of $f_0$ for suitable neurons that best match a target neuron in $f$, starting from the first hidden layer.
Given that we matched all neurons in layer $l-1$, we try to establish their connections to neurons in the current layer $l$.
A best match is decided by minimizing the l2-distance to its potential input parameters thereby adjusting for an optimal scaling factor.
For example, let neuron $i$ in Layer $l$ of the target $f$ have non-zero parameters $(b, \bm{w})$ that point to already matched neurons in Layer $l-1$.
With each of the matched neurons $j'$ in Layer $l-1$ has been previously associated a scaling factor $\lambda_{\text{old},j'}$ so that the corrected $\bm{\theta} = (b, \bm{w}\bm{\lambda_{\text{old}}})$ parameters would compute the correct neuron in $f_0$.
In Layer $l$ of $f_0$, each neuron $j$ that we have not matched yet could be a potential match for $i$.
Let the corresponding parameters of $j$ be $\bm{m}$.
The match quality between $i$ and $j$ is assessed by 
\begin{align}
q_{\theta}(m)=\norm{\bm{\theta}- \lambda(m)\bm{m}}_2, 
\end{align}
where $\lambda(m) =  \bm{\theta^T m}/\norm{\bm{m}}^2_2$ is the optimal scaling factor. 
The best matching parameters $m^* = \argmin_{m} q_{\theta}(m)$ are replaced by rescaled target parameters $\theta/\lambda(m^*)$ in $f_0$ and we remember the scaling factor $\lambda(m^*)$ to consider matches of neurons in Layer $l+1$.
Note that this rescaling is necessary to ensure that the neuron is properly hidden and attains similar values as other non-planted neurons in $f_0$. 
In addition to the provided pseudocode (Algorithm~\ref{alg:plant}), a Python implementation is submitted alongside the supplementary material.
%
%

\begin{algorithm}[t]
\DontPrintSemicolon
\SetAlgoLined
  \KwInput{target $f$, larger neural network $f_0$}
  \KwOutput{$f_0$ with planted $f$ ($f \subset f_0$), output scaling factors $\bm{\lambda}$}
  Initialize $\bm{\lambda_{\text{old}}} = [1]^{n_0}$; \tcp*{Scaling factors for input are 1.}
 \For{l = $1$ to $L-1$}{
    \For{all neurons $i$ of $f$ in Layer l}{
        $\theta:= (b, \bm{w}\bm{\lambda_{\text{old}}})$ \tcp*{non-zero parameters of neuron $i$ in $f$.}
        $m^* = \argmin_{m} q_{\theta}(m)$ \tcp*{Find best match for $i$ in $f_0$.}
        Replace parameters in $f_0$ by $\theta/\lambda(m^*)$; 
        Define $\lambda_n = \lambda(m^*)$. \tcp*{Remember scaling factor of $i$ in $f_0$.}
    }
    Replace $\bm{\lambda_{\text{old}}} = \bm{\lambda}$;
 } 
\caption{Planting}\label{alg:plant}
\end{algorithm}

\subsection{Construction of targets for planting}
We propose three examples for planting lottery tickets that pose different challenges for pruning algorithms.
Here, we explain the main ideas behind their construction in more detail. 
A Python implementation of related regression and classification problems is provided alongside the supplementary manuscript.

\subsubsection{\texttt{ReLU} unit}
Apart from a trivial function $f(x) = 0$, a univariate ReLU unit $f(x) = \phi(x) = \max(x,0)$ is the most sparse lottery ticket that is possible.
Assuming a mother network $f_0$ of depth $L$, a ReLU can be implemented with a single neuron per layer.
Any path through the network with positive weights $\prod^L_{l=1} \phi(w_{i_{l-1} i_l} x)$ defines a ReLU with scaling factor $\lambda = \prod^L_{l=1} w_{i_{l-1} i_l}$ for indices $i_l$ in Layer $l$ with $w_{i_{l-1} i_l} > 0$.

Note that each random path fulfills this criterion with probability $0.5^L$ so that even random pruning could have a considerable chance to find an optimal ticket.
A winning path exists with probability $\prod^{L}_{l=1} (1-0.5^{n_{l,0}})$, which is almost $1$ even in small mother networks.
Thus, planting is not really necessary in this case. 
Since not all pruning algorithms set biases to zero, however, we still set all randomly initialized biases along a winning path to zero to make the problem easier.

As we see in experiments, despite this simplification, pruning algorithms are severely challenged in finding an optimally sparse ticket.
Even though basic, a ReLU unit seems to be a suitable benchmark that is a common building block of other tickets.

\subsubsection{\texttt{Circle}}\label{sec:circleExplain}
For simplicity, we restrict ourselves to a 4-class classification problem with 2-dimensional input.  
The output is therefore 4-dimensional, where each output unit $f_c(x)$ corresponds to the probability $f_c(x)$ that an input $(x_1, x_2) \in [-1,1]^2$ belongs to the corresponding class $c$ with $c = 0, 1, 2, 3$. 
As common, this probability is computed assuming softmax activation functions in the last layer. 
The decision boundaries are defined in the last layer based on inputs of the form $g(x_1, x_2)$. 
The role of the first layers with ReLU activation functions of a \texttt{Circle} target $f$ is to compute the function $g(x_1, x_2) = x^2_1 + x^2_2$, which is fundamental to many problems, in particular to the computation of radial symmetric functions.

The high symmetry of $g(x_1, x_2)$ allows us to construct a particularly sparse representation by mirroring data points along axes as visualized in Figure~\ref{fig:circleExplainMain}~(a).
With the first two layers ($l=1,2$), we map each input vector $(x_1, x_2)$ to the first quadrant by defining
$x^{(1)}_1 = \phi\left(x_1\right) + \phi\left(-x_1\right) $ and $x^{(1)}_2 = \phi\left(x_2\right) + \phi\left(-x_2\right)$.
Thus, Layer $l=1$ consists of 4 neurons, i.e., $x^{(1)}_1 = \phi\left(x_1\right)$,  $ x^{(1)}_2 = \phi\left(-x_1\right)$, $x^{(1)}_3 = \phi\left(x_2\right)$,  $ x^{(1)}_4 = \phi\left(-x_2\right)$, while Layer $l=2$ consists of 2 neurons, i.e., $x^{(2)}_1 = \phi\left(x^{(1)}_1 + x^{(1)}_2 \right)$,  $ x^{(2)}_2 = \phi\left(x^{(1)}_3 + x^{(1)}_4 \right)$.

Each consecutive layer $l$ mirrors the previous layer $(x^{(l-1)}_1, x^{(l-1)}_2)$ along the axis $\bm{a^{(l)}} = (\cos(\pi/2^{l-1}), \sin(\pi/2^{l-1}))$.
It achieves this by mapping the neurons of the previous layer to three neurons, one representing the component of $\bm{x^{(l-1)}}$ that is parallel to $\bm{a^{(l)}}$, and two neurons that each represent the positive or negative signal component that is perpendicular to the axis $\bm{a^{(l)}} $. 
The last two neurons could be added to a single neuron in the next layer if we want to decrease the width of some layers to $2$ in between.
To take more advantage of the allowed depth $L$, we map three neurons immediately to the next three neurons that represent the mirroring by defining
\begin{align}
 & x^{(l)}_1 = \phi\left(a^{(l)}_1 x^{(l-1)}_1 +  a^{(l)}_2 x^{(l-1)}_2 - a^{(l)}_2 x^{(l-1)}_3\right), \ \ \  & x^{(l)}_2 = \phi\left(a^{(l)}_2 x^{(l-1)}_1 -  a^{(l)}_1 x^{(l-1)}_2 + a^{(l)}_1 x^{(l-1)}_3\right), \\
  & x^{(l)}_3 = \phi(-h^{(l)}_2).
\end{align}

If the depth of our network is high enough, we could use the parallel component $x^{(l)}_1$ as estimate of the radius of the input.
To enable higher precision for networks of smaller depth, however, we also apply to each remaining component a piecewise linear approximation of the univariate function $h(x) = x^2$ and add those two components.
Note that any univariate function can be easily approximated by a neural network of depth $L=2$.
The precise approach is explained in our next example.


\subsubsection{\texttt{Helix}}\label{sec:helixExplain}
To test the ability of pruning algorithms to detect lower dimensional submanifolds, we approximate a helix with three output coordinates $f_1(x)=(5\pi + 3\pi x) * \cos(5\pi + 3\pi x)/(8\pi)$, $f_2(x)=(5\pi + 3\pi x) * \sin(5\pi + 3\pi x)/(8\pi)$, and $f_3(x)=(5\pi + 3\pi x)/(8\pi)$ for 1-dimensional input $x \in [-1,1]$.
As we have observed that many pruning algorithms have the tendency to keep a higher number of neurons closer to the input (and sometimes also the output layer), we construct a ticket that has similar properties.
This should ease the task for pruning algorithms to find the planted winning ticket.

Each of the components $f_i(x)$ is an univariate function that we can approximate by an univariate deep neural network $n_i(x)$ that encodes a piece-wise linear function (see Figure~\ref{fig:circleExplainMain}~(c) for an explanation). 
As neural networks are generally overparameterized, we have multiple options to represent $n_i(x)$.
For simplicity, we write it as composition of the identity with a depth $L=2$ univariate network $g_i(x)$ of width $N$ in the hidden layer, which can be written as
\begin{align}\label{eq:univRep}
g_i(x) = \sum^{N}_{j = 1} a^{(i)}_{j} \phi(p_j(x - s_j)) + b^{(i)},
\end{align}
where the signs $p_j \in \{-1,1\}$ can be chosen arbitrarily (and we chose alternating signs to create diversity). 
The knots $\bm{s} = (s_j)_{j \in {[N]}}$ mark the boundaries of the linear regions and $\bm{a} = (a^{(i)}_j)_{j \in [N]}$ indicate changes in slopes $m^{(i)}_j = \left(f_i(s_{j+1})-f(s_{j})\right)/\left(s_{j+1}-s_{j}\right)$ (with $s_{N+1} := s_{N}+\epsilon$) from one linear region to the next. 
$a^{(i)}_j = m^{(i)}_j - m^{(i)}_{j-1}$ for $2 \leq i \leq N$, $a^{(i)}_1 = m^{(i)}_1$, and $b^{(i)} =  f_i(s_1) - \sum^{N}_{j = 1} a^{(i)}_{j} \phi(p_j(s_1 - s_j))$. 
Note that only the outer parameters $a^{(i)}_j$ are function specific, while the inner parameters $p_j$ and $s_j$ can be shared among the functions $f_i$.

We thus create a helix ticket by first mapping the input $x\in[-1,1]$ to $[0,2]$. 
This allows us to represent the identity in the later layers by $\phi(x) = x$, as $x\geq 0$. 
We can always compensate for the bias $+1$ by subtracting a bias $-1$ when needed. 
$f_3(x)=(5\pi + 3\pi x)/(8\pi)$ can therefore be represented by a path from the input to the output that only contains a single neuron per layer.
We concatenate this path with a neural network that consists of layers that approximate $f_1(x)$ and $f_2(x)$ and otherwise identity functions.
At Layer $l=2$, this network creates neurons of the form $\phi(p_j(x - s_j))$, where the knots $s_j$ mark an equidistant grid of $[0,2]$.
Layer $l=3$ creates two neurons, one corresponding to $x^{(2)}_1 = f_1(x)$ and one corresponding to $x^{(2)}_2 = f_2(x)$.
These can be computed by linear combination of the previous neurons using the parameters $a^{(i)}_j$ and $b^{(i)}$.
All the remaining layers basically encode the identity.



%% file: appendixexperiments.tex
\section{Experiments}

In this section, we discuss all relevant parameters and set-ups to reproduce the experimental results. Furthermore, we provide additional results obtained for singleshot learning spanning different architectures and pruning schemes.
All source code to run pruning algorithms and to generate the data is made publicly available.

\subsection{Hyperparameters and data}

For each experiment, we generate $n=10000$ samples, where input data is sampled from $[-1,1]$, for \texttt{ReLU} and \texttt{Helix} and from and $[-1,1]^3$ for \texttt{Circle}. 
The output for \texttt{ReLU} is computed by $f(x)=max(0,x)$. 
For \texttt{Helix}, we compute the three output coordinates as $f_1(x)=(5\pi + 3\pi x) * \cos(5\pi + 3\pi x)/(8\pi)$, $f_2(x)=(5\pi + 3\pi x) * \sin(5\pi + 3\pi x)/(8\pi)$, and $f_3(x)=(5\pi + 3\pi x)/(8\pi)$.
For \texttt{Circle}, we consider circles centered at the origin with radius $\sqrt{0.2}, \sqrt{0.5}$, and $\sqrt{0.7}$ as decision boundaries for the classes.
We additionally introduce a small amount of noise to simulate real world data more closely.
For \texttt{Circle} we flip approximately 1\% of samples to the next closest class, and for the two regression problems we introduce additive noise drawn from $\mathcal{N}(0,0.01)$ to each output dimension.
To assess the accuracy respectively mean squared error of the tickets and trained models, we split off 10\% of the data that acts as a hold out test set.

In general, all initial networks for each specific task are generated using the algorithm explained in Supp. A.
For \texttt{Circle}, we use $10$ knots for the piecewise linear approximation, and $30$ knots for the piecewise lienar approximations done in \texttt{Helix}.
To prune by \grasp,\snip,\synflow,\magnitude, and \rand and train the derived tickets, we use \texttt{Adam} \cite{kingma:15:adam} with a learning rate of $0.001$. We found that this learning rate performed well over all experiments, and leading to accurate models when there is no pruning.
It also corresponds to the default settings suggested by the authors of \synflow \cite{synflow}.
Training of the discovered tickets was done for $10$ epochs across all experiments, where we could always observe a convergence of the respective score on the validation sets (accuracy or MSE). We measured loss by MSE respectively cross entropy loss and used a batch size of $32$ for all experiments.
We report mean and confidence intervals across $10$ repetitions for multishot, and across $25$ repetions for the main singleshot experiments presented in Section \ref{sec:singleshot} measured on the hold out test set.

\paragraph{Singleshot pruning}

For singleshot pruning, we considered networks of depth $3,5,10$ each with layer width $100$ for all three data sets on target sparsities $\{0.01, 0.1, 0.5, 1\}$ and the sparsity of the ground truth ticket. Additionally, we tested for a network of depth $6$ and width $1000$ on \texttt{Circle} for the same sparsity levels.
As suggested by \cite{synflow}, we also test \synflow in combination with $100$ rounds of pruning for a network of depth $6$ and width $100$  on \texttt{Circle}.
For all additional singleshot experiments, we provide results in the next section.

\paragraph{Multishot pruning}

For multishot pruning, we alternated pruning and training for $10$ rounds, where each training step was carried out for $5$ epochs, which consistently lead to convergence of accuracy on the considered  \texttt{Circle} data set.
Similar to singleshot pruning, we considered target sparsities $\{0.01, 0.1, 0.5, 1\}$ and ground truth ticket sparsity.

\paragraph{\edgepopup pruning}

To prune with \edgepopup, we here used the parameters suggested in the original code of \cite{edgepopup}, which is SGD with momentum of $0.9$ and weight decay $0.0005$, combined with cosine annealing of the learning rate.
To establish a comparison to the multishot pruning results, we train the scores for $10$ epochs. Additionally, for the experiment extending \edgepopup by annealing the sparsity level, we slowly reduce the sparsity over time by $\rho^{i / 10}$, where $\rho$ is the desired network sparsity, and $i$ is the current epoch.

\subsection{Additional results on singleshot learning}

\paragraph{Model depth} In this section, we provide all results from the singleshot learning for depths $3,5,10$ and width $100$ in Figure \ref{fig:app_single3},\ref{fig:app_single5}, and \ref{fig:app_single10}, respectively. We observe that all methods have problems to deal with smaller networks, while the results for the larger networks are consistent.

\paragraph{Model width} To investigate the effect of layer size, we run an experiment with a network of depth $6$ and width $1000$ on \texttt{Circle} -- as the layer size is an important factor for the theoretical probability of the existence of tickets -- and report the results in Figure \ref{fig:app_1k}.
We observe that although the network is much larger, there is barely any change in comparison to the previous results.
Note that the results after training of tickets of individual sparsities cannot be directly compared directly to the other singleshot results, as the tickets are much larger (due to the much larger number of parameters) and hence easier to train.

\paragraph{Multiple pruning rounds} We report the results of running \synflow with $100$ rounds of pruning on a network of depth $6$ and width $100$ for \texttt{Circle} in Figure \ref{fig:app_synflow}. We find that there is again barely any change to the original singleshot results after pruning, but there is a slight increase in performance after training compared to the single-round singleshot results.

\paragraph{Noise experiments} To test the robustness of pruning algorithms to noise in the data, we considered the \texttt{Circle} problem with a network of depth $6$ and width $100$ and varied the amount of noise in the data to be $\{0, 0.001, 0.01, 0.1\}$.
We report the results before and after training in Fig. \ref{fig:noise}.

\subsection{Additional results on multishot learning}

To test whether we can reach an improvement of the performance of tickets discovered by \grasp using a multishot pruning approach and avoid layer collapse, we ran additional experiments using a local pruning rate. In particular, we used layer-wise pruning setting the target sparsity of the parameters of each individual layer to the global sparsity level.
We report results in Fig. \ref{fig:app_grasp_local}.

\input{figs/singleshot3_app}
\input{figs/singleshot5_app}
\input{figs/singleshot10_app}

\input{figs/width1k_app.tex}

\input{figs/synflow_100_app.tex}

\input{figs/noise_exp.tex}
\input{figs/grasp_local.tex}

%% file: figs/singleshot3_app.tex
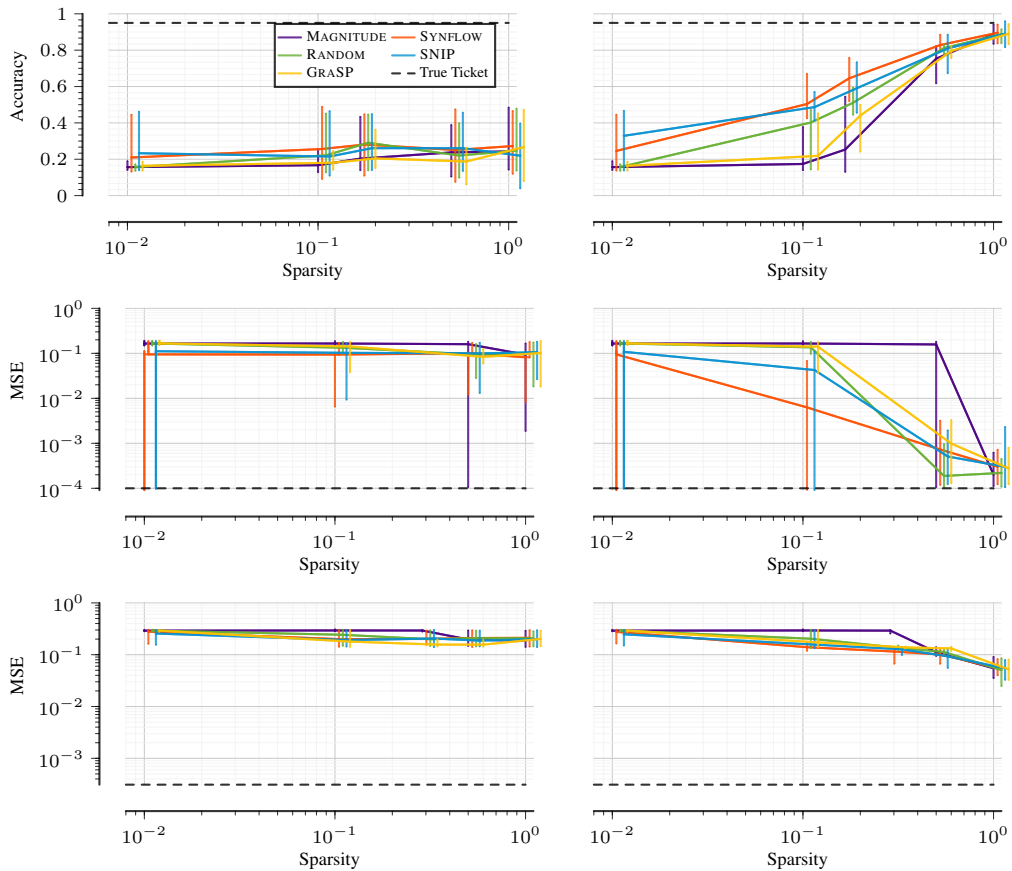
\begin{figure}
\centering
    \begin{subfigure}[t]{0.49\textwidth}
    \centering
		\ifpdf
		\begin{tikzpicture}
		\begin{axis}[
		jonas line,
		xmode=log,
		width = 7cm,
		height = 4cm,
		xlabel = {Sparsity}, 
		ylabel= {Accuracy},
		xmin=0.008, xmax=1.1,
		ymin=0, ymax=1,
		legend columns=2,
		x label style 		= {at={(axis description cs:0.5,-0.1)}, anchor=north, font=\scriptsize},
		y label style 		= {at={(axis description cs:0.05,0.6)},  anchor=south, font=\scriptsize},
		legend style={nodes={scale=0.8, transform shape}, at={(0.95,0.95)}, anchor=north east, row sep=-1.4pt, font=\tiny}
		]
		
		\addplot+[forget plot, eda errorbarcolored, y dir=plus, y explicit]
		table[x=magSparsity, y=magMeanPostprune, y error=magErrorPlusPostprune] {expres/grid/rescircleGridSingleshotDepth3.tsv};
		\addplot+[eda errorbarcolored, y dir=minus, y explicit]
		table[x=magSparsity, y=magMeanPostprune, y error=magErrorMinusPostprune] {expres/grid/rescircleGridSingleshotDepth3.tsv};
		\addlegendentry{\magnitude}
		
		\addplot+[forget plot, eda errorbarcolored, y dir=plus, y explicit]
		table[x expr=\thisrow{magSparsity}*1.05, y=synflowMeanPostprune, y error=synflowErrorPlusPostprune] {expres/grid/rescircleGridSingleshotDepth3.tsv};
		\addplot+[eda errorbarcolored, y dir=minus, y explicit]
		table[x expr=\thisrow{magSparsity}*1.05, y=synflowMeanPostprune, y error=synflowErrorMinusPostprune] {expres/grid/rescircleGridSingleshotDepth3.tsv};
		\addlegendentry{\synflow}
		
		\addplot+[forget plot, eda errorbarcolored, y dir=plus, y explicit]
		table[x expr=\thisrow{magSparsity}*1.1, y=randMeanPostprune, y error=randErrorPlusPostprune] {expres/grid/rescircleGridSingleshotDepth3.tsv};
		\addplot+[eda errorbarcolored, y dir=minus, y explicit]
		table[x expr=\thisrow{magSparsity}*1.1, y=randMeanPostprune, y error=randErrorMinusPostprune] {expres/grid/rescircleGridSingleshotDepth3.tsv};
		\addlegendentry{\rand}
		
		\addplot+[forget plot, eda errorbarcolored, y dir=plus, y explicit]
		table[x expr=\thisrow{magSparsity}*1.15, y=snipMeanPostprune, y error=snipErrorPlusPostprune] {expres/grid/rescircleGridSingleshotDepth3.tsv};
		\addplot+[eda errorbarcolored, y dir=minus, y explicit]
		table[x expr=\thisrow{magSparsity}*1.15, y=snipMeanPostprune, y error=snipErrorMinusPostprune] {expres/grid/rescircleGridSingleshotDepth3.tsv};
		\addlegendentry{\snip}
		
		\addplot+[forget plot, eda errorbarcolored, y dir=plus, y explicit]
		table[x expr=\thisrow{magSparsity}*1.2, y=graspMeanPostprune, y error=graspErrorPlusPostprune] {expres/grid/rescircleGridSingleshotDepth3.tsv};
		\addplot+[eda errorbarcolored, y dir=minus, y explicit]
		table[x expr=\thisrow{magSparsity}*1.2, y=graspMeanPostprune, y error=graspErrorMinusPostprune] {expres/grid/rescircleGridSingleshotDepth3.tsv};
		\addlegendentry{\grasp}
		
        \addplot[mark=none, dashed, black, samples=2] coordinates {(0.008,0.95) (1.0,0.95)};
        \addlegendentry{True Ticket}
		
		\end{axis}
		\end{tikzpicture}
		\fi
    \end{subfigure}
    \begin{subfigure}[t]{0.49\textwidth}
    \centering
		\ifpdf
		\begin{tikzpicture}
		\begin{axis}[
		jonas line,
		xmode=log,
        y axis line style 	= {opacity=0},
        y tick style = {draw=none},
        yticklabels={,,},
		width = 7cm,
		height = 4cm,
		xlabel = {Sparsity}, 
		xmin=0.008, xmax=1.1,
		ymin=0, ymax=1,
		legend columns=2,
		x label style 		= {at={(axis description cs:0.5,-0.1)}, anchor=north, font=\scriptsize},
		y label style 		= {at={(axis description cs:0.0,0.6)},  anchor=south, font=\scriptsize},
		legend style={nodes={scale=0.9, transform shape}, at={(0.98,0.95)}, anchor=north east, row sep=-1.4pt, font=\tiny}
		]
		
		\addplot+[forget plot, eda errorbarcolored, y dir=plus, y explicit]
		table[x=magSparsity, y=magMeanPosttrain, y error=magErrorPlusPosttrain] {expres/grid/rescircleGridSingleshotDepth3.tsv};
		\addplot+[eda errorbarcolored, y dir=minus, y explicit]
		table[x=magSparsity, y=magMeanPosttrain, y error=magErrorMinusPosttrain] {expres/grid/rescircleGridSingleshotDepth3.tsv};
		
		\addplot+[forget plot, eda errorbarcolored, y dir=plus, y explicit]
		table[x expr=\thisrow{magSparsity}*1.05, y=synflowMeanPosttrain, y error=synflowErrorPlusPosttrain] {expres/grid/rescircleGridSingleshotDepth3.tsv};
		\addplot+[eda errorbarcolored, y dir=minus, y explicit]
		table[x expr=\thisrow{magSparsity}*1.05, y=synflowMeanPosttrain, y error=synflowErrorMinusPosttrain] {expres/grid/rescircleGridSingleshotDepth3.tsv};
		
		\addplot+[forget plot, eda errorbarcolored, y dir=plus, y explicit]
		table[x expr=\thisrow{magSparsity}*1.1, y=randMeanPosttrain, y error=randErrorPlusPosttrain] {expres/grid/rescircleGridSingleshotDepth3.tsv};
		\addplot+[eda errorbarcolored, y dir=minus, y explicit]
		table[x expr=\thisrow{magSparsity}*1.1, y=randMeanPosttrain, y error=randErrorMinusPosttrain] {expres/grid/rescircleGridSingleshotDepth3.tsv};
		
		\addplot+[forget plot, eda errorbarcolored, y dir=plus, y explicit]
		table[x expr=\thisrow{magSparsity}*1.15, y=snipMeanPosttrain, y error=snipErrorPlusPosttrain] {expres/grid/rescircleGridSingleshotDepth3.tsv};
		\addplot+[eda errorbarcolored, y dir=minus, y explicit]
		table[x expr=\thisrow{magSparsity}*1.15, y=snipMeanPosttrain, y error=snipErrorMinusPosttrain] {expres/grid/rescircleGridSingleshotDepth3.tsv};
		
		\addplot+[forget plot, eda errorbarcolored, y dir=plus, y explicit]
		table[x expr=\thisrow{magSparsity}*1.2, y=graspMeanPosttrain, y error=graspErrorPlusPosttrain] {expres/grid/rescircleGridSingleshotDepth3.tsv};
		\addplot+[eda errorbarcolored, y dir=minus, y explicit]
		table[x expr=\thisrow{magSparsity}*1.2, y=graspMeanPosttrain, y error=graspErrorMinusPosttrain] {expres/grid/rescircleGridSingleshotDepth3.tsv};
		
        \addplot[mark=none, dashed, black, samples=2] coordinates {(0.008,0.95) (1.0,0.95)};
		
		\end{axis}
		\end{tikzpicture}
		\fi
    \end{subfigure}
    \begin{subfigure}[t]{0.49\textwidth}
    \centering
		\ifpdf
		\begin{tikzpicture}
		\begin{axis}[
		jonas line,
		xmode=log,
		ymode=log,
		width = 7cm,
		height = 4cm,
		xlabel = {Sparsity}, 
		ylabel= {MSE},
		xmin=0.008, xmax=1.1,
		ymax=1,
		legend columns=2,
		x label style 		= {at={(axis description cs:0.5,-0.1)}, anchor=north, font=\scriptsize},
		y label style 		= {at={(axis description cs:0.0,0.6)},  anchor=south, font=\scriptsize},
		legend style={nodes={scale=0.9, transform shape}, at={(0.98,0.95)}, anchor=north east, row sep=-1.4pt, font=\tiny}
		]
		
		\addplot+[forget plot, eda errorbarcolored, y dir=plus, y explicit]
		table[x=magSparsity, y=magMeanPostprune, y error=magErrorPlusPostprune] {expres/grid/resreluGridSingleshotDepth3.tsv};
		\addplot+[eda errorbarcolored, y dir=minus, y explicit]
		table[x=magSparsity, y=magMeanPostprune, y error=magErrorMinusPostprune] {expres/grid/resreluGridSingleshotDepth3.tsv};
		
		\addplot+[forget plot, eda errorbarcolored, y dir=plus, y explicit]
		table[x expr=\thisrow{magSparsity}*1.05, y=synflowMeanPostprune, y error=synflowErrorPlusPostprune] {expres/grid/resreluGridSingleshotDepth3.tsv};
		\addplot+[eda errorbarcolored, y dir=minus, y explicit]
		table[x expr=\thisrow{magSparsity}*1.0, y=synflowMeanPostprune, y error=synflowErrorMinusPostprune] {expres/grid/resreluGridSingleshotDepth3.tsv};
		
		\addplot+[forget plot, eda errorbarcolored, y dir=plus, y explicit]
		table[x expr=\thisrow{magSparsity}*1.1, y=randMeanPostprune, y error=randErrorPlusPostprune] {expres/grid/resreluGridSingleshotDepth3.tsv};
		\addplot+[eda errorbarcolored, y dir=minus, y explicit]
		table[x expr=\thisrow{magSparsity}*1.1, y=randMeanPostprune, y error=randErrorMinusPostprune] {expres/grid/resreluGridSingleshotDepth3.tsv};
		
		\addplot+[forget plot, eda errorbarcolored, y dir=plus, y explicit]
		table[x expr=\thisrow{magSparsity}*1.15, y=snipMeanPostprune, y error=snipErrorPlusPostprune] {expres/grid/resreluGridSingleshotDepth3.tsv};
		\addplot+[eda errorbarcolored, y dir=minus, y explicit]
		table[x expr=\thisrow{magSparsity}*1.15, y=snipMeanPostprune, y error=snipErrorMinusPostprune] {expres/grid/resreluGridSingleshotDepth3.tsv};
		
		\addplot+[forget plot, eda errorbarcolored, y dir=plus, y explicit]
		table[x expr=\thisrow{magSparsity}*1.2, y=graspMeanPostprune, y error=graspErrorPlusPostprune] {expres/grid/resreluGridSingleshotDepth3.tsv};
		\addplot+[eda errorbarcolored, y dir=minus, y explicit]
		table[x expr=\thisrow{magSparsity}*1.2, y=graspMeanPostprune, y error=graspErrorMinusPostprune] {expres/grid/resreluGridSingleshotDepth3.tsv};
		
        \addplot[mark=none, dashed, black, samples=2] coordinates {(0.008,0.0001) (1.0,0.0001)};
		
		\end{axis}
		\end{tikzpicture}
		\fi
    \end{subfigure}
    \begin{subfigure}[t]{0.49\textwidth}
    \centering
		\ifpdf
		\begin{tikzpicture}
		\begin{axis}[
		jonas line,
		xmode=log,
		ymode=log,
        y axis line style 	= {opacity=0},
        y tick style = {draw=none},
        yticklabels={,,},
		width = 7cm,
		height = 4cm,
		xlabel = {Sparsity}, 
		xmin=0.008, xmax=1.1,
		ymax=1,
		legend columns=2,
		x label style 		= {at={(axis description cs:0.5,-0.1)}, anchor=north, font=\scriptsize},
		y label style 		= {at={(axis description cs:0.0,0.6)},  anchor=south, font=\scriptsize},
		legend style={nodes={scale=0.9, transform shape}, at={(0.98,0.95)}, anchor=north east, row sep=-1.4pt, font=\tiny}
		]
		
		\addplot+[forget plot, eda errorbarcolored, y dir=plus, y explicit]
		table[x=magSparsity, y=magMeanPosttrain, y error=magErrorPlusPosttrain] {expres/grid/resreluGridSingleshotDepth3.tsv};
		\addplot+[eda errorbarcolored, y dir=minus, y explicit]
		table[x=magSparsity, y=magMeanPosttrain, y error=magErrorMinusPosttrain] {expres/grid/resreluGridSingleshotDepth3.tsv};
		
		\addplot+[forget plot, eda errorbarcolored, y dir=plus, y explicit]
		table[x expr=\thisrow{magSparsity}*1.05, y=synflowMeanPosttrain, y error=synflowErrorPlusPosttrain] {expres/grid/resreluGridSingleshotDepth3.tsv};
		\addplot+[eda errorbarcolored, y dir=minus, y explicit]
		table[x expr=\thisrow{magSparsity}*1.05, y=synflowMeanPosttrain, y error=synflowErrorMinusPosttrain] {expres/grid/resreluGridSingleshotDepth3.tsv};
		
		\addplot+[forget plot, eda errorbarcolored, y dir=plus, y explicit]
		table[x expr=\thisrow{magSparsity}*1.1, y=randMeanPosttrain, y error=randErrorPlusPosttrain] {expres/grid/resreluGridSingleshotDepth3.tsv};
		\addplot+[eda errorbarcolored, y dir=minus, y explicit]
		table[x expr=\thisrow{magSparsity}*1.1, y=randMeanPosttrain, y error=randErrorMinusPosttrain] {expres/grid/resreluGridSingleshotDepth3.tsv};
		
		\addplot+[forget plot, eda errorbarcolored, y dir=plus, y explicit]
		table[x expr=\thisrow{magSparsity}*1.15, y=snipMeanPosttrain, y error=snipErrorPlusPosttrain] {expres/grid/resreluGridSingleshotDepth3.tsv};
		\addplot+[eda errorbarcolored, y dir=minus, y explicit]
		table[x expr=\thisrow{magSparsity}*1.15, y=snipMeanPosttrain, y error=snipErrorMinusPosttrain] {expres/grid/resreluGridSingleshotDepth3.tsv};
		
		\addplot+[forget plot, eda errorbarcolored, y dir=plus, y explicit]
		table[x expr=\thisrow{magSparsity}*1.2, y=graspMeanPosttrain, y error=graspErrorPlusPosttrain] {expres/grid/resreluGridSingleshotDepth3.tsv};
		\addplot+[eda errorbarcolored, y dir=minus, y explicit]
		table[x expr=\thisrow{magSparsity}*1.2, y=graspMeanPosttrain, y error=graspErrorMinusPosttrain] {expres/grid/resreluGridSingleshotDepth3.tsv};
		
        \addplot[mark=none, dashed, black, samples=2] coordinates {(0.008,0.0001) (1.0,0.0001)};
		
		\end{axis}
		\end{tikzpicture}
		\fi
    \end{subfigure}
    \begin{subfigure}[t]{0.49\textwidth}
    \centering
		\ifpdf
		\begin{tikzpicture}
		\begin{axis}[
		jonas line,
		xmode=log,
		ymode=log,
		minor x tick num = 5,
		width = 7cm,
		height = 4cm,
		xlabel = {Sparsity}, 
		ylabel= {MSE},
		xmin=0.008, xmax=1.1,
		ymax=1,
		legend columns=2,
		x label style 		= {at={(axis description cs:0.5,-0.1)}, anchor=north, font=\scriptsize},
		y label style 		= {at={(axis description cs:0.0,0.6)},  anchor=south, font=\scriptsize},
		legend style={nodes={scale=0.9, transform shape}, at={(0.98,0.95)}, anchor=north east, row sep=-1.4pt, font=\tiny}
		]
		
		\addplot+[forget plot, eda errorbarcolored, y dir=plus, y explicit]
		table[x=magSparsity, y=magMeanPostprune, y error=magErrorPlusPostprune] {expres/grid/reshelixGridSingleshotDepth3.tsv};
		\addplot+[eda errorbarcolored, y dir=minus, y explicit]
		table[x=magSparsity, y=magMeanPostprune, y error=magErrorMinusPostprune] {expres/grid/reshelixGridSingleshotDepth3.tsv};
		
		\addplot+[forget plot, eda errorbarcolored, y dir=plus, y explicit]
		table[x expr=\thisrow{magSparsity}*1.05, y=synflowMeanPostprune, y error=synflowErrorPlusPostprune] {expres/grid/reshelixGridSingleshotDepth3.tsv};
		\addplot+[eda errorbarcolored, y dir=minus, y explicit]
		table[x expr=\thisrow{magSparsity}*1.05, y=synflowMeanPostprune, y error=synflowErrorMinusPostprune] {expres/grid/reshelixGridSingleshotDepth3.tsv};
		
		\addplot+[forget plot, eda errorbarcolored, y dir=plus, y explicit]
		table[x expr=\thisrow{magSparsity}*1.1, y=randMeanPostprune, y error=randErrorPlusPostprune] {expres/grid/reshelixGridSingleshotDepth3.tsv};
		\addplot+[eda errorbarcolored, y dir=minus, y explicit]
		table[x expr=\thisrow{magSparsity}*1.1, y=randMeanPostprune, y error=randErrorMinusPostprune] {expres/grid/reshelixGridSingleshotDepth3.tsv};
		
		\addplot+[forget plot, eda errorbarcolored, y dir=plus, y explicit]
		table[x expr=\thisrow{magSparsity}*1.15, y=snipMeanPostprune, y error=snipErrorPlusPostprune] {expres/grid/reshelixGridSingleshotDepth3.tsv};
		\addplot+[eda errorbarcolored, y dir=minus, y explicit]
		table[x expr=\thisrow{magSparsity}*1.15, y=snipMeanPostprune, y error=snipErrorMinusPostprune] {expres/grid/reshelixGridSingleshotDepth3.tsv};
		
		\addplot+[forget plot, eda errorbarcolored, y dir=plus, y explicit]
		table[x expr=\thisrow{magSparsity}*1.2, y=graspMeanPostprune, y error=graspErrorPlusPostprune] {expres/grid/reshelixGridSingleshotDepth3.tsv};
		\addplot+[eda errorbarcolored, y dir=minus, y explicit]
		table[x expr=\thisrow{magSparsity}*1.2, y=graspMeanPostprune, y error=graspErrorMinusPostprune] {expres/grid/reshelixGridSingleshotDepth3.tsv};
		
        \addplot[mark=none, dashed, black, samples=2] coordinates {(0.008,0.00031) (1.0,0.00031)};
		
		\end{axis}
		\end{tikzpicture}
		\fi
    \end{subfigure}
    \begin{subfigure}[t]{0.49\textwidth}
    \centering
		\ifpdf
		\begin{tikzpicture}
		\begin{axis}[
		jonas line,
		xmode=log,
		ymode=log,
		minor x tick num = 5,
        y axis line style 	= {opacity=0},
        y tick style = {draw=none},
        yticklabels={,,},
		width = 7cm,
		height = 4cm,
		xlabel = {Sparsity}, 
		xmin=0.008, xmax=1.1,
		ymax=1,
		legend columns=2,
		x label style 		= {at={(axis description cs:0.5,-0.1)}, anchor=north, font=\scriptsize},
		y label style 		= {at={(axis description cs:0.0,0.6)},  anchor=south, font=\scriptsize},
		legend style={nodes={scale=0.9, transform shape}, at={(0.98,0.95)}, anchor=north east, row sep=-1.4pt, font=\tiny}
		]
		
		\addplot+[forget plot, eda errorbarcolored, y dir=plus, y explicit]
		table[x=magSparsity, y=magMeanPosttrain, y error=magErrorPlusPosttrain] {expres/grid/reshelixGridSingleshotDepth3.tsv};
		\addplot+[eda errorbarcolored, y dir=minus, y explicit]
		table[x=magSparsity, y=magMeanPosttrain, y error=magErrorMinusPosttrain] {expres/grid/reshelixGridSingleshotDepth3.tsv};
		
		\addplot+[forget plot, eda errorbarcolored, y dir=plus, y explicit]
		table[x expr=\thisrow{magSparsity}*1.05, y=synflowMeanPosttrain, y error=synflowErrorPlusPosttrain] {expres/grid/reshelixGridSingleshotDepth3.tsv};
		\addplot+[eda errorbarcolored, y dir=minus, y explicit]
		table[x expr=\thisrow{magSparsity}*1.05, y=synflowMeanPosttrain, y error=synflowErrorMinusPosttrain] {expres/grid/reshelixGridSingleshotDepth3.tsv};
		
		\addplot+[forget plot, eda errorbarcolored, y dir=plus, y explicit]
		table[x expr=\thisrow{magSparsity}*1.1, y=randMeanPosttrain, y error=randErrorPlusPosttrain] {expres/grid/reshelixGridSingleshotDepth3.tsv};
		\addplot+[eda errorbarcolored, y dir=minus, y explicit]
		table[x expr=\thisrow{magSparsity}*1.1, y=randMeanPosttrain, y error=randErrorMinusPosttrain] {expres/grid/reshelixGridSingleshotDepth3.tsv};
		
		\addplot+[forget plot, eda errorbarcolored, y dir=plus, y explicit]
		table[x expr=\thisrow{magSparsity}*1.15, y=snipMeanPosttrain, y error=snipErrorPlusPosttrain] {expres/grid/reshelixGridSingleshotDepth3.tsv};
		\addplot+[eda errorbarcolored, y dir=minus, y explicit]
		table[x expr=\thisrow{magSparsity}*1.15, y=snipMeanPosttrain, y error=snipErrorMinusPosttrain] {expres/grid/reshelixGridSingleshotDepth3.tsv};
		
		\addplot+[forget plot, eda errorbarcolored, y dir=plus, y explicit]
		table[x expr=\thisrow{magSparsity}*1.2, y=graspMeanPosttrain, y error=graspErrorPlusPosttrain] {expres/grid/reshelixGridSingleshotDepth3.tsv};
		\addplot+[eda errorbarcolored, y dir=minus, y explicit]
		table[x expr=\thisrow{magSparsity}*1.2, y=graspMeanPosttrain, y error=graspErrorMinusPosttrain] {expres/grid/reshelixGridSingleshotDepth3.tsv};

        \addplot[mark=none, dashed, black, samples=2] coordinates {(0.008,0.00031) (1.0,0.00031)};
		
		\end{axis}
		\end{tikzpicture}
		\fi
    \end{subfigure}
    \caption{\textit{Singleshot results depth $3$.} Performance of discovered tickets for \texttt{Circle}, \texttt{ReLU}, and \texttt{Helix}  against target sparsities as mean and obtained intervals (minimum and maximum) across $25$ runs. In order of appearance from top to bottom: \texttt{Circle}, \texttt{ReLU}, and \texttt{Helix} post pruning (left) and post training performance (right). Baseline tickets have sparsities of $.16, .01,$ and $.29$, and their performance is given by black dashed line.}\label{fig:app_single3}
\end{figure}

%% file: figs/singleshot5_app.tex
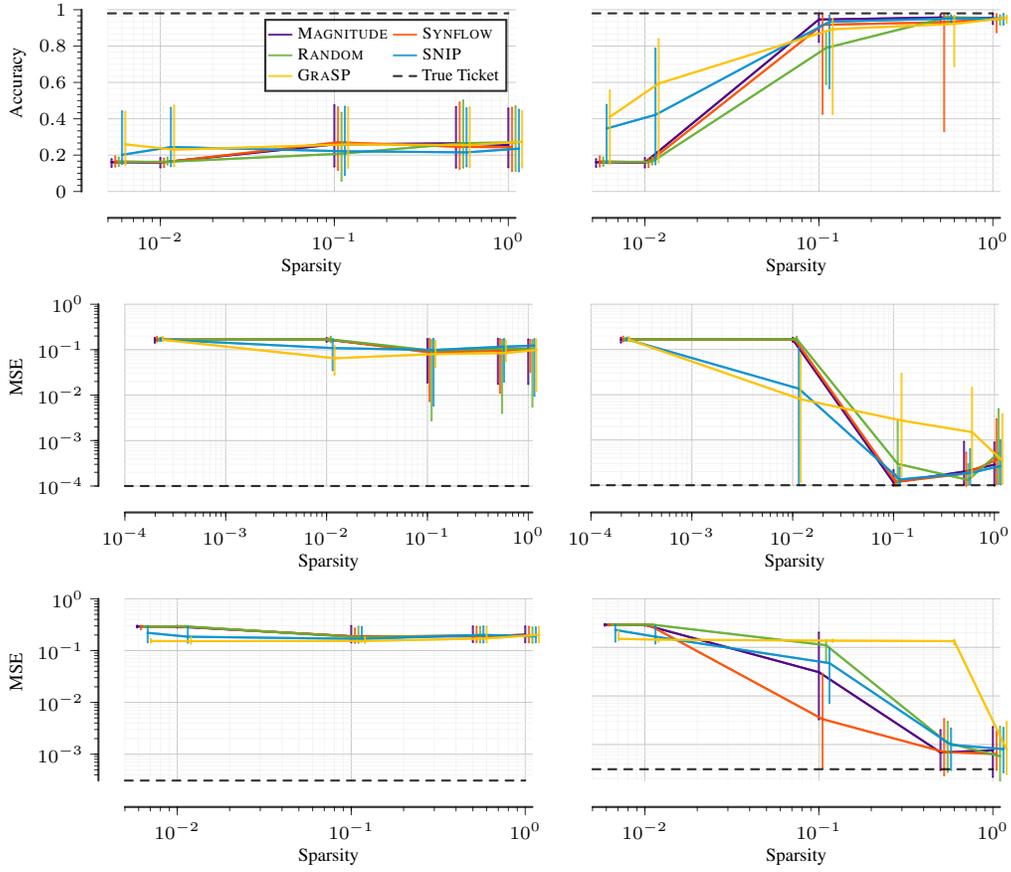
\begin{figure}
\centering
    \begin{subfigure}[t]{0.49\textwidth}
    \centering
		\ifpdf
		\begin{tikzpicture}
		\begin{axis}[
		jonas line,
		xmode=log,
		width = 7cm,
		height = 4cm,
		xlabel = {Sparsity}, 
		ylabel= {Accuracy},
		xmin=0.005, xmax=1.1,
		ymin=0, ymax=1,
		legend columns=2,
		x label style 		= {at={(axis description cs:0.5,-0.1)}, anchor=north, font=\scriptsize},
		y label style 		= {at={(axis description cs:0.05,0.6)},  anchor=south, font=\scriptsize},
		legend style={nodes={scale=0.9, transform shape}, at={(0.98,0.95)}, anchor=north east, row sep=-1.4pt, font=\tiny}
		]
		
		\addplot+[forget plot, eda errorbarcolored, y dir=plus, y explicit]
		table[x=magSparsity, y=magMeanPostprune, y error=magErrorPlusPostprune] {expres/grid/rescircleGridSingleshotDepth5.tsv};
		\addplot+[eda errorbarcolored, y dir=minus, y explicit]
		table[x=magSparsity, y=magMeanPostprune, y error=magErrorMinusPostprune] {expres/grid/rescircleGridSingleshotDepth5.tsv};
		\addlegendentry{\magnitude}
		
		\addplot+[forget plot, eda errorbarcolored, y dir=plus, y explicit]
		table[x expr=\thisrow{magSparsity}*1.05, y=synflowMeanPostprune, y error=synflowErrorPlusPostprune] {expres/grid/rescircleGridSingleshotDepth5.tsv};
		\addplot+[eda errorbarcolored, y dir=minus, y explicit]
		table[x expr=\thisrow{magSparsity}*1.05, y=synflowMeanPostprune, y error=synflowErrorMinusPostprune] {expres/grid/rescircleGridSingleshotDepth5.tsv};
		\addlegendentry{\synflow}
		
		\addplot+[forget plot, eda errorbarcolored, y dir=plus, y explicit]
		table[x expr=\thisrow{magSparsity}*1.1, y=randMeanPostprune, y error=randErrorPlusPostprune] {expres/grid/rescircleGridSingleshotDepth5.tsv};
		\addplot+[eda errorbarcolored, y dir=minus, y explicit]
		table[x expr=\thisrow{magSparsity}*1.1, y=randMeanPostprune, y error=randErrorMinusPostprune] {expres/grid/rescircleGridSingleshotDepth5.tsv};
		\addlegendentry{\rand}
		
		\addplot+[forget plot, eda errorbarcolored, y dir=plus, y explicit]
		table[x expr=\thisrow{magSparsity}*1.15, y=snipMeanPostprune, y error=snipErrorPlusPostprune] {expres/grid/rescircleGridSingleshotDepth5.tsv};
		\addplot+[eda errorbarcolored, y dir=minus, y explicit]
		table[x expr=\thisrow{magSparsity}*1.15, y=snipMeanPostprune, y error=snipErrorMinusPostprune] {expres/grid/rescircleGridSingleshotDepth5.tsv};
		\addlegendentry{\snip}
		
		\addplot+[forget plot, eda errorbarcolored, y dir=plus, y explicit]
		table[x expr=\thisrow{magSparsity}*1.2, y=graspMeanPostprune, y error=graspErrorPlusPostprune] {expres/grid/rescircleGridSingleshotDepth5.tsv};
		\addplot+[eda errorbarcolored, y dir=minus, y explicit]
		table[x expr=\thisrow{magSparsity}*1.2, y=graspMeanPostprune, y error=graspErrorMinusPostprune] {expres/grid/rescircleGridSingleshotDepth5.tsv};
		\addlegendentry{\grasp}
		
        \addplot[mark=none, dashed, black, samples=2] coordinates {(0.005,0.98) (1.0,0.98)};
        \addlegendentry{True Ticket}
		
		\end{axis}
		\end{tikzpicture}
		\fi
    \end{subfigure}
    \begin{subfigure}[t]{0.49\textwidth}
    \centering
		\ifpdf
		\begin{tikzpicture}
		\begin{axis}[
		jonas line,
		xmode=log,
        y axis line style 	= {opacity=0},
        y tick style = {draw=none},
        yticklabels={,,},
		width = 7cm,
		height = 4cm,
		xlabel = {Sparsity}, 
		xmin=0.005, xmax=1.1,
		ymin=0, ymax=1,
		legend columns=2,
		x label style 		= {at={(axis description cs:0.5,-0.1)}, anchor=north, font=\scriptsize},
		y label style 		= {at={(axis description cs:0.0,0.6)},  anchor=south, font=\scriptsize},
		legend style={nodes={scale=0.9, transform shape}, at={(0.98,0.95)}, anchor=north east, row sep=-1.4pt, font=\tiny}
		]
		
		\addplot+[forget plot, eda errorbarcolored, y dir=plus, y explicit]
		table[x=magSparsity, y=magMeanPosttrain, y error=magErrorPlusPosttrain] {expres/grid/rescircleGridSingleshotDepth5.tsv};
		\addplot+[eda errorbarcolored, y dir=minus, y explicit]
		table[x=magSparsity, y=magMeanPosttrain, y error=magErrorMinusPosttrain] {expres/grid/rescircleGridSingleshotDepth5.tsv};
		
		\addplot+[forget plot, eda errorbarcolored, y dir=plus, y explicit]
		table[x expr=\thisrow{magSparsity}*1.05, y=synflowMeanPosttrain, y error=synflowErrorPlusPosttrain] {expres/grid/rescircleGridSingleshotDepth5.tsv};
		\addplot+[eda errorbarcolored, y dir=minus, y explicit]
		table[x expr=\thisrow{magSparsity}*1.05, y=synflowMeanPosttrain, y error=synflowErrorMinusPosttrain] {expres/grid/rescircleGridSingleshotDepth5.tsv};
		
		\addplot+[forget plot, eda errorbarcolored, y dir=plus, y explicit]
		table[x expr=\thisrow{magSparsity}*1.1, y=randMeanPosttrain, y error=randErrorPlusPosttrain] {expres/grid/rescircleGridSingleshotDepth5.tsv};
		\addplot+[eda errorbarcolored, y dir=minus, y explicit]
		table[x expr=\thisrow{magSparsity}*1.1, y=randMeanPosttrain, y error=randErrorMinusPosttrain] {expres/grid/rescircleGridSingleshotDepth5.tsv};
		
		\addplot+[forget plot, eda errorbarcolored, y dir=plus, y explicit]
		table[x expr=\thisrow{magSparsity}*1.15, y=snipMeanPosttrain, y error=snipErrorPlusPosttrain] {expres/grid/rescircleGridSingleshotDepth5.tsv};
		\addplot+[eda errorbarcolored, y dir=minus, y explicit]
		table[x expr=\thisrow{magSparsity}*1.15, y=snipMeanPosttrain, y error=snipErrorMinusPosttrain] {expres/grid/rescircleGridSingleshotDepth5.tsv};
		
		\addplot+[forget plot, eda errorbarcolored, y dir=plus, y explicit]
		table[x expr=\thisrow{magSparsity}*1.2, y=graspMeanPosttrain, y error=graspErrorPlusPosttrain] {expres/grid/rescircleGridSingleshotDepth5.tsv};
		\addplot+[eda errorbarcolored, y dir=minus, y explicit]
		table[x expr=\thisrow{magSparsity}*1.2, y=graspMeanPosttrain, y error=graspErrorMinusPosttrain] {expres/grid/rescircleGridSingleshotDepth5.tsv};
		
        \addplot[mark=none, dashed, black, samples=2] coordinates {(0.005,0.98) (1.0,0.98)};
		
		\end{axis}
		\end{tikzpicture}
		\fi
    \end{subfigure}
    \begin{subfigure}[t]{0.49\textwidth}
    \centering
		\ifpdf
		\begin{tikzpicture}
		\begin{axis}[
		jonas line,
		xmode=log,
		ymode=log,
		width = 7cm,
		height = 4cm,
		xlabel = {Sparsity}, 
		ylabel= {MSE},
		xmin=0.0001, xmax=1.1,
		ymax=1,
		legend columns=2,
		x label style 		= {at={(axis description cs:0.5,-0.1)}, anchor=north, font=\scriptsize},
		y label style 		= {at={(axis description cs:0.0,0.6)},  anchor=south, font=\scriptsize},
		legend style={nodes={scale=0.9, transform shape}, at={(0.98,0.95)}, anchor=north east, row sep=-1.4pt, font=\tiny}
		]
		
		\addplot+[forget plot, eda errorbarcolored, y dir=plus, y explicit]
		table[x=magSparsity, y=magMeanPostprune, y error=magErrorPlusPostprune] {expres/grid/resreluGridSingleshotDepth5.tsv};
		\addplot+[eda errorbarcolored, y dir=minus, y explicit]
		table[x=magSparsity, y=magMeanPostprune, y error=magErrorMinusPostprune] {expres/grid/resreluGridSingleshotDepth5.tsv};
		
		\addplot+[forget plot, eda errorbarcolored, y dir=plus, y explicit]
		table[x expr=\thisrow{magSparsity}*1.05, y=synflowMeanPostprune, y error=synflowErrorPlusPostprune] {expres/grid/resreluGridSingleshotDepth5.tsv};
		\addplot+[eda errorbarcolored, y dir=minus, y explicit]
		table[x expr=\thisrow{magSparsity}*1.05, y=synflowMeanPostprune, y error=synflowErrorMinusPostprune] {expres/grid/resreluGridSingleshotDepth5.tsv};
		
		\addplot+[forget plot, eda errorbarcolored, y dir=plus, y explicit]
		table[x expr=\thisrow{magSparsity}*1.1, y=randMeanPostprune, y error=randErrorPlusPostprune] {expres/grid/resreluGridSingleshotDepth5.tsv};
		\addplot+[eda errorbarcolored, y dir=minus, y explicit]
		table[x expr=\thisrow{magSparsity}*1.1, y=randMeanPostprune, y error=randErrorMinusPostprune] {expres/grid/resreluGridSingleshotDepth5.tsv};
		
		\addplot+[forget plot, eda errorbarcolored, y dir=plus, y explicit]
		table[x expr=\thisrow{magSparsity}*1.15, y=snipMeanPostprune, y error=snipErrorPlusPostprune] {expres/grid/resreluGridSingleshotDepth5.tsv};
		\addplot+[eda errorbarcolored, y dir=minus, y explicit]
		table[x expr=\thisrow{magSparsity}*1.15, y=snipMeanPostprune, y error=snipErrorMinusPostprune] {expres/grid/resreluGridSingleshotDepth5.tsv};
		
		\addplot+[forget plot, eda errorbarcolored, y dir=plus, y explicit]
		table[x expr=\thisrow{magSparsity}*1.2, y=graspMeanPostprune, y error=graspErrorPlusPostprune] {expres/grid/resreluGridSingleshotDepth5.tsv};
		\addplot+[eda errorbarcolored, y dir=minus, y explicit]
		table[x expr=\thisrow{magSparsity}*1.2, y=graspMeanPostprune, y error=graspErrorMinusPostprune] {expres/grid/resreluGridSingleshotDepth5.tsv};
		
        \addplot[mark=none, dashed, black, samples=2] coordinates {(0.0001,0.00010) (1.0,0.00010)};
		
		\end{axis}
		\end{tikzpicture}
		\fi
    \end{subfigure}
    \begin{subfigure}[t]{0.49\textwidth}
    \centering
		\ifpdf
		\begin{tikzpicture}
		\begin{axis}[
		jonas line,
		xmode=log,
		ymode=log,
        y axis line style 	= {opacity=0},
        y tick style = {draw=none},
        yticklabels={,,},
		width = 7cm,
		height = 4cm,
		xlabel = {Sparsity}, 
		xmin=0.0001, xmax=1.1,
		ymax=1,
		legend columns=2,
		x label style 		= {at={(axis description cs:0.5,-0.1)}, anchor=north, font=\scriptsize},
		y label style 		= {at={(axis description cs:0.0,0.6)},  anchor=south, font=\scriptsize},
		legend style={nodes={scale=0.9, transform shape}, at={(0.98,0.95)}, anchor=north east, row sep=-1.4pt, font=\tiny}
		]
		
		\addplot+[forget plot, eda errorbarcolored, y dir=plus, y explicit]
		table[x=magSparsity, y=magMeanPosttrain, y error=magErrorPlusPosttrain] {expres/grid/resreluGridSingleshotDepth5.tsv};
		\addplot+[eda errorbarcolored, y dir=minus, y explicit]
		table[x=magSparsity, y=magMeanPosttrain, y error=magErrorMinusPosttrain] {expres/grid/resreluGridSingleshotDepth5.tsv};
		
		\addplot+[forget plot, eda errorbarcolored, y dir=plus, y explicit]
		table[x expr=\thisrow{magSparsity}*1.05, y=synflowMeanPosttrain, y error=synflowErrorPlusPosttrain] {expres/grid/resreluGridSingleshotDepth5.tsv};
		\addplot+[eda errorbarcolored, y dir=minus, y explicit]
		table[x expr=\thisrow{magSparsity}*1.05, y=synflowMeanPosttrain, y error=synflowErrorMinusPosttrain] {expres/grid/resreluGridSingleshotDepth5.tsv};
		
		\addplot+[forget plot, eda errorbarcolored, y dir=plus, y explicit]
		table[x expr=\thisrow{magSparsity}*1.1, y=randMeanPosttrain, y error=randErrorPlusPosttrain] {expres/grid/resreluGridSingleshotDepth5.tsv};
		\addplot+[eda errorbarcolored, y dir=minus, y explicit]
		table[x expr=\thisrow{magSparsity}*1.1, y=randMeanPosttrain, y error=randErrorMinusPosttrain] {expres/grid/resreluGridSingleshotDepth5.tsv};
		
		\addplot+[forget plot, eda errorbarcolored, y dir=plus, y explicit]
		table[x expr=\thisrow{magSparsity}*1.15, y=snipMeanPosttrain, y error=snipErrorPlusPosttrain] {expres/grid/resreluGridSingleshotDepth5.tsv};
		\addplot+[eda errorbarcolored, y dir=minus, y explicit]
		table[x expr=\thisrow{magSparsity}*1.15, y=snipMeanPosttrain, y error=snipErrorMinusPosttrain] {expres/grid/resreluGridSingleshotDepth5.tsv};
		
		\addplot+[forget plot, eda errorbarcolored, y dir=plus, y explicit]
		table[x expr=\thisrow{magSparsity}*1.2, y=graspMeanPosttrain, y error=graspErrorPlusPosttrain] {expres/grid/resreluGridSingleshotDepth5.tsv};
		\addplot+[eda errorbarcolored, y dir=minus, y explicit]
		table[x expr=\thisrow{magSparsity}*1.2, y=graspMeanPosttrain, y error=graspErrorMinusPosttrain] {expres/grid/resreluGridSingleshotDepth5.tsv};
		
        \addplot[mark=none, dashed, black, samples=2] coordinates {(0.0001,0.00010) (1.0,0.00010)};
		
		\end{axis}
		\end{tikzpicture}
		\fi
    \end{subfigure}
    \begin{subfigure}[t]{0.49\textwidth}
    \centering
		\ifpdf
		\begin{tikzpicture}
		\begin{axis}[
		jonas line,
		xmode=log,
		ymode=log,
		minor x tick num = 5,
		width = 7cm,
		height = 4cm,
		xlabel = {Sparsity}, 
		ylabel= {MSE},
		xmin=0.005, xmax=1.1,
		ymax=1,
		legend columns=2,
		x label style 		= {at={(axis description cs:0.5,-0.1)}, anchor=north, font=\scriptsize},
		y label style 		= {at={(axis description cs:0.0,0.6)},  anchor=south, font=\scriptsize},
		legend style={nodes={scale=0.9, transform shape}, at={(0.98,0.95)}, anchor=north east, row sep=-1.4pt, font=\tiny}
		]
		
		\addplot+[forget plot, eda errorbarcolored, y dir=plus, y explicit]
		table[x=magSparsity, y=magMeanPostprune, y error=magErrorPlusPostprune] {expres/grid/reshelixGridSingleshotDepth5.tsv};
		\addplot+[eda errorbarcolored, y dir=minus, y explicit]
		table[x=magSparsity, y=magMeanPostprune, y error=magErrorMinusPostprune] {expres/grid/reshelixGridSingleshotDepth5.tsv};
		
		\addplot+[forget plot, eda errorbarcolored, y dir=plus, y explicit]
		table[x expr=\thisrow{magSparsity}*1.05, y=synflowMeanPostprune, y error=synflowErrorPlusPostprune] {expres/grid/reshelixGridSingleshotDepth5.tsv};
		\addplot+[eda errorbarcolored, y dir=minus, y explicit]
		table[x expr=\thisrow{magSparsity}*1.05, y=synflowMeanPostprune, y error=synflowErrorMinusPostprune] {expres/grid/reshelixGridSingleshotDepth5.tsv};
		
		\addplot+[forget plot, eda errorbarcolored, y dir=plus, y explicit]
		table[x expr=\thisrow{magSparsity}*1.1, y=randMeanPostprune, y error=randErrorPlusPostprune] {expres/grid/reshelixGridSingleshotDepth5.tsv};
		\addplot+[eda errorbarcolored, y dir=minus, y explicit]
		table[x expr=\thisrow{magSparsity}*1.1, y=randMeanPostprune, y error=randErrorMinusPostprune] {expres/grid/reshelixGridSingleshotDepth5.tsv};
		
		\addplot+[forget plot, eda errorbarcolored, y dir=plus, y explicit]
		table[x expr=\thisrow{magSparsity}*1.15, y=snipMeanPostprune, y error=snipErrorPlusPostprune] {expres/grid/reshelixGridSingleshotDepth5.tsv};
		\addplot+[eda errorbarcolored, y dir=minus, y explicit]
		table[x expr=\thisrow{magSparsity}*1.15, y=snipMeanPostprune, y error=snipErrorMinusPostprune] {expres/grid/reshelixGridSingleshotDepth5.tsv};
		
		\addplot+[forget plot, eda errorbarcolored, y dir=plus, y explicit]
		table[x expr=\thisrow{magSparsity}*1.2, y=graspMeanPostprune, y error=graspErrorPlusPostprune] {expres/grid/reshelixGridSingleshotDepth5.tsv};
		\addplot+[eda errorbarcolored, y dir=minus, y explicit]
		table[x expr=\thisrow{magSparsity}*1.2, y=graspMeanPostprune, y error=graspErrorMinusPostprune] {expres/grid/reshelixGridSingleshotDepth5.tsv};
		
        \addplot[mark=none, dashed, black, samples=2] coordinates {(0.005,0.00031) (1.0,0.00031)};
		
		\end{axis}
		\end{tikzpicture}
		\fi
    \end{subfigure}
    \begin{subfigure}[t]{0.49\textwidth}
    \centering
		\ifpdf
		\begin{tikzpicture}
		\begin{axis}[
		jonas line,
		xmode=log,
		ymode=log,
		minor x tick num = 5,
        y axis line style 	= {opacity=0},
        y tick style = {draw=none},
        yticklabels={,,},
		width = 7cm,
		height = 4cm,
		xlabel = {Sparsity}, 
		xmin=0.005, xmax=1.1,
		ymax=1,
		legend columns=2,
		x label style 		= {at={(axis description cs:0.5,-0.1)}, anchor=north, font=\scriptsize},
		y label style 		= {at={(axis description cs:0.0,0.6)},  anchor=south, font=\scriptsize},
		legend style={nodes={scale=0.9, transform shape}, at={(0.98,0.95)}, anchor=north east, row sep=-1.4pt, font=\tiny}
		]
		
		\addplot+[forget plot, eda errorbarcolored, y dir=plus, y explicit]
		table[x=magSparsity, y=magMeanPosttrain, y error=magErrorPlusPosttrain] {expres/grid/reshelixGridSingleshotDepth5.tsv};
		\addplot+[eda errorbarcolored, y dir=minus, y explicit]
		table[x=magSparsity, y=magMeanPosttrain, y error=magErrorMinusPosttrain] {expres/grid/reshelixGridSingleshotDepth5.tsv};
		
		\addplot+[forget plot, eda errorbarcolored, y dir=plus, y explicit]
		table[x expr=\thisrow{magSparsity}*1.05, y=synflowMeanPosttrain, y error=synflowErrorPlusPosttrain] {expres/grid/reshelixGridSingleshotDepth5.tsv};
		\addplot+[eda errorbarcolored, y dir=minus, y explicit]
		table[x expr=\thisrow{magSparsity}*1.05, y=synflowMeanPosttrain, y error=synflowErrorMinusPosttrain] {expres/grid/reshelixGridSingleshotDepth5.tsv};
		
		\addplot+[forget plot, eda errorbarcolored, y dir=plus, y explicit]
		table[x expr=\thisrow{magSparsity}*1.1, y=randMeanPosttrain, y error=randErrorPlusPosttrain] {expres/grid/reshelixGridSingleshotDepth5.tsv};
		\addplot+[eda errorbarcolored, y dir=minus, y explicit]
		table[x expr=\thisrow{magSparsity}*1.1, y=randMeanPosttrain, y error=randErrorMinusPosttrain] {expres/grid/reshelixGridSingleshotDepth5.tsv};
		
		\addplot+[forget plot, eda errorbarcolored, y dir=plus, y explicit]
		table[x expr=\thisrow{magSparsity}*1.15, y=snipMeanPosttrain, y error=snipErrorPlusPosttrain] {expres/grid/reshelixGridSingleshotDepth5.tsv};
		\addplot+[eda errorbarcolored, y dir=minus, y explicit]
		table[x expr=\thisrow{magSparsity}*1.15, y=snipMeanPosttrain, y error=snipErrorMinusPosttrain] {expres/grid/reshelixGridSingleshotDepth5.tsv};
		
		\addplot+[forget plot, eda errorbarcolored, y dir=plus, y explicit]
		table[x expr=\thisrow{magSparsity}*1.2, y=graspMeanPosttrain, y error=graspErrorPlusPosttrain] {expres/grid/reshelixGridSingleshotDepth5.tsv};
		\addplot+[eda errorbarcolored, y dir=minus, y explicit]
		table[x expr=\thisrow{magSparsity}*1.2, y=graspMeanPosttrain, y error=graspErrorMinusPosttrain] {expres/grid/reshelixGridSingleshotDepth5.tsv};
		
        \addplot[mark=none, dashed, black, samples=2] coordinates {(0.005,0.00031) (1.0,0.00031)};
		
		\end{axis}
		\end{tikzpicture}
		\fi
    \end{subfigure}
    \caption{\textit{Singleshot results depth $5$.} Performance of discovered tickets for \texttt{Circle}, \texttt{ReLU}, and \texttt{Helix}  against target sparsities as mean and obtained intervals (minimum and maximum) across $25$ runs. In order of appearance from top to bottom: \texttt{Circle}, \texttt{ReLU}, and \texttt{Helix} post pruning (left) and post training performance (right). Baseline ticket with leftmost sparsity and performance given by black dashed line.}\label{fig:app_single5}
\end{figure}

%% file: figs/singleshot10_app.tex
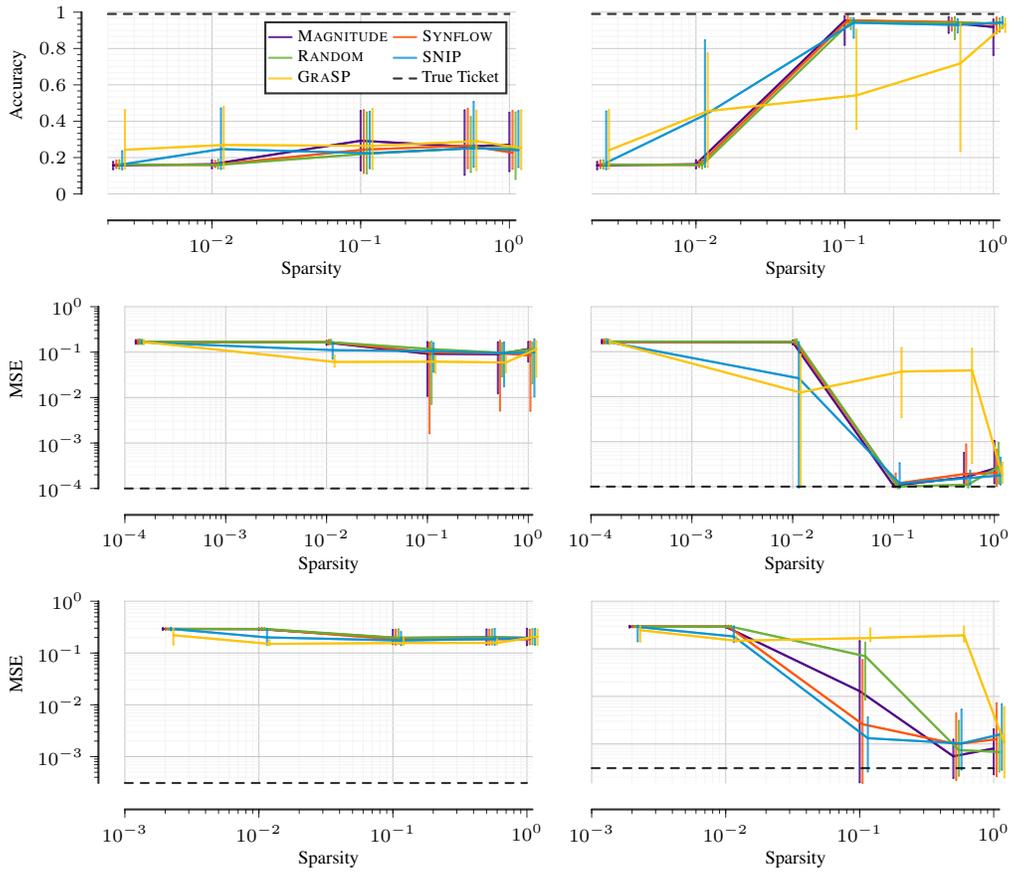
\begin{figure}
\centering
    \begin{subfigure}[t]{0.49\textwidth}
    \centering
		\ifpdf
		\begin{tikzpicture}
		\begin{axis}[
		jonas line,
		xmode=log,
		width = 7cm,
		height = 4cm,
		xlabel = {Sparsity}, 
		ylabel= {Accuracy},
		xmin=0.002, xmax=1.1,
		ymin=0, ymax=1,
		legend columns=2,
		x label style 		= {at={(axis description cs:0.5,-0.1)}, anchor=north, font=\scriptsize},
		y label style 		= {at={(axis description cs:0.05,0.6)},  anchor=south, font=\scriptsize},
		legend style={nodes={scale=0.9, transform shape}, at={(0.98,0.95)}, anchor=north east, row sep=-1.4pt, font=\tiny}
		]
		
		\addplot+[forget plot, eda errorbarcolored, y dir=plus, y explicit]
		table[x=magSparsity, y=magMeanPostprune, y error=magErrorPlusPostprune] {expres/grid/rescircleGridSingleshotDepth10.tsv};
		\addplot+[eda errorbarcolored, y dir=minus, y explicit]
		table[x=magSparsity, y=magMeanPostprune, y error=magErrorMinusPostprune] {expres/grid/rescircleGridSingleshotDepth10.tsv};
		\addlegendentry{\magnitude}
		
		\addplot+[forget plot, eda errorbarcolored, y dir=plus, y explicit]
		table[x expr=\thisrow{magSparsity}*1.05, y=synflowMeanPostprune, y error=synflowErrorPlusPostprune] {expres/grid/rescircleGridSingleshotDepth10.tsv};
		\addplot+[eda errorbarcolored, y dir=minus, y explicit]
		table[x expr=\thisrow{magSparsity}*1.05, y=synflowMeanPostprune, y error=synflowErrorMinusPostprune] {expres/grid/rescircleGridSingleshotDepth10.tsv};
		\addlegendentry{\synflow}
		
		\addplot+[forget plot, eda errorbarcolored, y dir=plus, y explicit]
		table[x expr=\thisrow{magSparsity}*1.1, y=randMeanPostprune, y error=randErrorPlusPostprune] {expres/grid/rescircleGridSingleshotDepth10.tsv};
		\addplot+[eda errorbarcolored, y dir=minus, y explicit]
		table[x expr=\thisrow{magSparsity}*1.1, y=randMeanPostprune, y error=randErrorMinusPostprune] {expres/grid/rescircleGridSingleshotDepth10.tsv};
		\addlegendentry{\rand}
		
		\addplot+[forget plot, eda errorbarcolored, y dir=plus, y explicit]
		table[x expr=\thisrow{magSparsity}*1.15, y=snipMeanPostprune, y error=snipErrorPlusPostprune] {expres/grid/rescircleGridSingleshotDepth10.tsv};
		\addplot+[eda errorbarcolored, y dir=minus, y explicit]
		table[x expr=\thisrow{magSparsity}*1.15, y=snipMeanPostprune, y error=snipErrorMinusPostprune] {expres/grid/rescircleGridSingleshotDepth10.tsv};
		\addlegendentry{\snip}
		
		\addplot+[forget plot, eda errorbarcolored, y dir=plus, y explicit]
		table[x expr=\thisrow{magSparsity}*1.2, y=graspMeanPostprune, y error=graspErrorPlusPostprune] {expres/grid/rescircleGridSingleshotDepth10.tsv};
		\addplot+[eda errorbarcolored, y dir=minus, y explicit]
		table[x expr=\thisrow{magSparsity}*1.2, y=graspMeanPostprune, y error=graspErrorMinusPostprune] {expres/grid/rescircleGridSingleshotDepth10.tsv};
		\addlegendentry{\grasp}
		
        \addplot[mark=none, dashed, black, samples=2] coordinates {(0.002,0.99) (1.0,0.99)};
        \addlegendentry{True Ticket}
		
		\end{axis}
		\end{tikzpicture}
		\fi
    \end{subfigure}
    \begin{subfigure}[t]{0.49\textwidth}
    \centering
		\ifpdf
		\begin{tikzpicture}
		\begin{axis}[
		jonas line,
		xmode=log,
        y axis line style 	= {opacity=0},
        y tick style = {draw=none},
        yticklabels={,,},
		width = 7cm,
		height = 4cm,
		xlabel = {Sparsity}, 
		xmin=0.002, xmax=1.1,
		ymin=0, ymax=1,
		legend columns=2,
		x label style 		= {at={(axis description cs:0.5,-0.1)}, anchor=north, font=\scriptsize},
		y label style 		= {at={(axis description cs:0.0,0.6)},  anchor=south, font=\scriptsize},
		legend style={nodes={scale=0.9, transform shape}, at={(0.98,0.95)}, anchor=north east, row sep=-1.4pt, font=\tiny}
		]
		
		\addplot+[forget plot, eda errorbarcolored, y dir=plus, y explicit]
		table[x=magSparsity, y=magMeanPosttrain, y error=magErrorPlusPosttrain] {expres/grid/rescircleGridSingleshotDepth10.tsv};
		\addplot+[eda errorbarcolored, y dir=minus, y explicit]
		table[x=magSparsity, y=magMeanPosttrain, y error=magErrorMinusPosttrain] {expres/grid/rescircleGridSingleshotDepth10.tsv};
		
		\addplot+[forget plot, eda errorbarcolored, y dir=plus, y explicit]
		table[x expr=\thisrow{magSparsity}*1.05, y=synflowMeanPosttrain, y error=synflowErrorPlusPosttrain] {expres/grid/rescircleGridSingleshotDepth10.tsv};
		\addplot+[eda errorbarcolored, y dir=minus, y explicit]
		table[x expr=\thisrow{magSparsity}*1.05, y=synflowMeanPosttrain, y error=synflowErrorMinusPosttrain] {expres/grid/rescircleGridSingleshotDepth10.tsv};
		
		\addplot+[forget plot, eda errorbarcolored, y dir=plus, y explicit]
		table[x expr=\thisrow{magSparsity}*1.1, y=randMeanPosttrain, y error=randErrorPlusPosttrain] {expres/grid/rescircleGridSingleshotDepth10.tsv};
		\addplot+[eda errorbarcolored, y dir=minus, y explicit]
		table[x expr=\thisrow{magSparsity}*1.1, y=randMeanPosttrain, y error=randErrorMinusPosttrain] {expres/grid/rescircleGridSingleshotDepth10.tsv};
		
		\addplot+[forget plot, eda errorbarcolored, y dir=plus, y explicit]
		table[x expr=\thisrow{magSparsity}*1.15, y=snipMeanPosttrain, y error=snipErrorPlusPosttrain] {expres/grid/rescircleGridSingleshotDepth10.tsv};
		\addplot+[eda errorbarcolored, y dir=minus, y explicit]
		table[x expr=\thisrow{magSparsity}*1.15, y=snipMeanPosttrain, y error=snipErrorMinusPosttrain] {expres/grid/rescircleGridSingleshotDepth10.tsv};
		
		\addplot+[forget plot, eda errorbarcolored, y dir=plus, y explicit]
		table[x expr=\thisrow{magSparsity}*1.2, y=graspMeanPosttrain, y error=graspErrorPlusPosttrain] {expres/grid/rescircleGridSingleshotDepth10.tsv};
		\addplot+[eda errorbarcolored, y dir=minus, y explicit]
		table[x expr=\thisrow{magSparsity}*1.2, y=graspMeanPosttrain, y error=graspErrorMinusPosttrain] {expres/grid/rescircleGridSingleshotDepth10.tsv};
		
        \addplot[mark=none, dashed, black, samples=2] coordinates {(0.002,0.99) (1.0,0.99)};
		
		\end{axis}
		\end{tikzpicture}
		\fi
    \end{subfigure}
    \begin{subfigure}[t]{0.49\textwidth}
    \centering
		\ifpdf
		\begin{tikzpicture}
		\begin{axis}[
		jonas line,
		xmode=log,
		ymode=log,
		width = 7cm,
		height = 4cm,
		xlabel = {Sparsity}, 
		ylabel= {MSE},
		xmin=0.0001, xmax=1.1,
		ymax=1,
		legend columns=2,
		x label style 		= {at={(axis description cs:0.5,-0.1)}, anchor=north, font=\scriptsize},
		y label style 		= {at={(axis description cs:0.0,0.6)},  anchor=south, font=\scriptsize},
		legend style={nodes={scale=0.9, transform shape}, at={(0.98,0.95)}, anchor=north east, row sep=-1.4pt, font=\tiny}
		]
		
		\addplot+[forget plot, eda errorbarcolored, y dir=plus, y explicit]
		table[x=magSparsity, y=magMeanPostprune, y error=magErrorPlusPostprune] {expres/grid/resreluGridSingleshotDepth10.tsv};
		\addplot+[eda errorbarcolored, y dir=minus, y explicit]
		table[x=magSparsity, y=magMeanPostprune, y error=magErrorMinusPostprune] {expres/grid/resreluGridSingleshotDepth10.tsv};
		
		\addplot+[forget plot, eda errorbarcolored, y dir=plus, y explicit]
		table[x expr=\thisrow{magSparsity}*1.05, y=synflowMeanPostprune, y error=synflowErrorPlusPostprune] {expres/grid/resreluGridSingleshotDepth10.tsv};
		\addplot+[eda errorbarcolored, y dir=minus, y explicit]
		table[x expr=\thisrow{magSparsity}*1.05, y=synflowMeanPostprune, y error=synflowErrorMinusPostprune] {expres/grid/resreluGridSingleshotDepth10.tsv};
		
		\addplot+[forget plot, eda errorbarcolored, y dir=plus, y explicit]
		table[x expr=\thisrow{magSparsity}*1.1, y=randMeanPostprune, y error=randErrorPlusPostprune] {expres/grid/resreluGridSingleshotDepth10.tsv};
		\addplot+[eda errorbarcolored, y dir=minus, y explicit]
		table[x expr=\thisrow{magSparsity}*1.1, y=randMeanPostprune, y error=randErrorMinusPostprune] {expres/grid/resreluGridSingleshotDepth10.tsv};
		
		\addplot+[forget plot, eda errorbarcolored, y dir=plus, y explicit]
		table[x expr=\thisrow{magSparsity}*1.15, y=snipMeanPostprune, y error=snipErrorPlusPostprune] {expres/grid/resreluGridSingleshotDepth10.tsv};
		\addplot+[eda errorbarcolored, y dir=minus, y explicit]
		table[x expr=\thisrow{magSparsity}*1.15, y=snipMeanPostprune, y error=snipErrorMinusPostprune] {expres/grid/resreluGridSingleshotDepth10.tsv};
		
		\addplot+[forget plot, eda errorbarcolored, y dir=plus, y explicit]
		table[x expr=\thisrow{magSparsity}*1.2, y=graspMeanPostprune, y error=graspErrorPlusPostprune] {expres/grid/resreluGridSingleshotDepth10.tsv};
		\addplot+[eda errorbarcolored, y dir=minus, y explicit]
		table[x expr=\thisrow{magSparsity}*1.2, y=graspMeanPostprune, y error=graspErrorMinusPostprune] {expres/grid/resreluGridSingleshotDepth10.tsv};
		
        \addplot[mark=none, dashed, black, samples=2] coordinates {(0.0001,0.00010) (1.0,0.00010)};
		
		\end{axis}
		\end{tikzpicture}
		\fi
    \end{subfigure}
    \begin{subfigure}[t]{0.49\textwidth}
    \centering
		\ifpdf
		\begin{tikzpicture}
		\begin{axis}[
		jonas line,
		xmode=log,
		ymode=log,
        y axis line style 	= {opacity=0},
        y tick style = {draw=none},
        yticklabels={,,},
		width = 7cm,
		height = 4cm,
		xlabel = {Sparsity}, 
		xmin=0.0001, xmax=1.1,
		ymax=1,
		legend columns=2,
		x label style 		= {at={(axis description cs:0.5,-0.1)}, anchor=north, font=\scriptsize},
		y label style 		= {at={(axis description cs:0.0,0.6)},  anchor=south, font=\scriptsize},
		legend style={nodes={scale=0.9, transform shape}, at={(0.98,0.95)}, anchor=north east, row sep=-1.4pt, font=\tiny}
		]
		
		\addplot+[forget plot, eda errorbarcolored, y dir=plus, y explicit]
		table[x=magSparsity, y=magMeanPosttrain, y error=magErrorPlusPosttrain] {expres/grid/resreluGridSingleshotDepth10.tsv};
		\addplot+[eda errorbarcolored, y dir=minus, y explicit]
		table[x=magSparsity, y=magMeanPosttrain, y error=magErrorMinusPosttrain] {expres/grid/resreluGridSingleshotDepth10.tsv};
		
		\addplot+[forget plot, eda errorbarcolored, y dir=plus, y explicit]
		table[x expr=\thisrow{magSparsity}*1.05, y=synflowMeanPosttrain, y error=synflowErrorPlusPosttrain] {expres/grid/resreluGridSingleshotDepth10.tsv};
		\addplot+[eda errorbarcolored, y dir=minus, y explicit]
		table[x expr=\thisrow{magSparsity}*1.05, y=synflowMeanPosttrain, y error=synflowErrorMinusPosttrain] {expres/grid/resreluGridSingleshotDepth10.tsv};
		
		\addplot+[forget plot, eda errorbarcolored, y dir=plus, y explicit]
		table[x expr=\thisrow{magSparsity}*1.1, y=randMeanPosttrain, y error=randErrorPlusPosttrain] {expres/grid/resreluGridSingleshotDepth10.tsv};
		\addplot+[eda errorbarcolored, y dir=minus, y explicit]
		table[x expr=\thisrow{magSparsity}*1.1, y=randMeanPosttrain, y error=randErrorMinusPosttrain] {expres/grid/resreluGridSingleshotDepth10.tsv};
		
		\addplot+[forget plot, eda errorbarcolored, y dir=plus, y explicit]
		table[x expr=\thisrow{magSparsity}*1.15, y=snipMeanPosttrain, y error=snipErrorPlusPosttrain] {expres/grid/resreluGridSingleshotDepth10.tsv};
		\addplot+[eda errorbarcolored, y dir=minus, y explicit]
		table[x expr=\thisrow{magSparsity}*1.15, y=snipMeanPosttrain, y error=snipErrorMinusPosttrain] {expres/grid/resreluGridSingleshotDepth10.tsv};
		
		\addplot+[forget plot, eda errorbarcolored, y dir=plus, y explicit]
		table[x expr=\thisrow{magSparsity}*1.2, y=graspMeanPosttrain, y error=graspErrorPlusPosttrain] {expres/grid/resreluGridSingleshotDepth10.tsv};
		\addplot+[eda errorbarcolored, y dir=minus, y explicit]
		table[x expr=\thisrow{magSparsity}*1.2, y=graspMeanPosttrain, y error=graspErrorMinusPosttrain] {expres/grid/resreluGridSingleshotDepth10.tsv};
		
        \addplot[mark=none, dashed, black, samples=2] coordinates {(0.0001,0.00010) (1.0,0.00010)};
		
		\end{axis}
		\end{tikzpicture}
		\fi
    \end{subfigure}
    \begin{subfigure}[t]{0.49\textwidth}
    \centering
		\ifpdf
		\begin{tikzpicture}
		\begin{axis}[
		jonas line,
		xmode=log,
		ymode=log,
		minor x tick num = 5,
		width = 7cm,
		height = 4cm,
		xlabel = {Sparsity}, 
		ylabel= {MSE},
		xmin=0.001, xmax=1.1,
		ymax=1,
		legend columns=2,
		x label style 		= {at={(axis description cs:0.5,-0.1)}, anchor=north, font=\scriptsize},
		y label style 		= {at={(axis description cs:0.0,0.6)},  anchor=south, font=\scriptsize},
		legend style={nodes={scale=0.9, transform shape}, at={(0.98,0.95)}, anchor=north east, row sep=-1.4pt, font=\tiny}
		]
		
		\addplot+[forget plot, eda errorbarcolored, y dir=plus, y explicit]
		table[x=magSparsity, y=magMeanPostprune, y error=magErrorPlusPostprune] {expres/grid/reshelixGridSingleshotDepth10.tsv};
		\addplot+[eda errorbarcolored, y dir=minus, y explicit]
		table[x=magSparsity, y=magMeanPostprune, y error=magErrorMinusPostprune] {expres/grid/reshelixGridSingleshotDepth10.tsv};
		
		\addplot+[forget plot, eda errorbarcolored, y dir=plus, y explicit]
		table[x expr=\thisrow{magSparsity}*1.05, y=synflowMeanPostprune, y error=synflowErrorPlusPostprune] {expres/grid/reshelixGridSingleshotDepth10.tsv};
		\addplot+[eda errorbarcolored, y dir=minus, y explicit]
		table[x expr=\thisrow{magSparsity}*1.05, y=synflowMeanPostprune, y error=synflowErrorMinusPostprune] {expres/grid/reshelixGridSingleshotDepth10.tsv};
		
		\addplot+[forget plot, eda errorbarcolored, y dir=plus, y explicit]
		table[x expr=\thisrow{magSparsity}*1.1, y=randMeanPostprune, y error=randErrorPlusPostprune] {expres/grid/reshelixGridSingleshotDepth10.tsv};
		\addplot+[eda errorbarcolored, y dir=minus, y explicit]
		table[x expr=\thisrow{magSparsity}*1.1, y=randMeanPostprune, y error=randErrorMinusPostprune] {expres/grid/reshelixGridSingleshotDepth10.tsv};
		
		\addplot+[forget plot, eda errorbarcolored, y dir=plus, y explicit]
		table[x expr=\thisrow{magSparsity}*1.15, y=snipMeanPostprune, y error=snipErrorPlusPostprune] {expres/grid/reshelixGridSingleshotDepth10.tsv};
		\addplot+[eda errorbarcolored, y dir=minus, y explicit]
		table[x expr=\thisrow{magSparsity}*1.15, y=snipMeanPostprune, y error=snipErrorMinusPostprune] {expres/grid/reshelixGridSingleshotDepth10.tsv};
		
		\addplot+[forget plot, eda errorbarcolored, y dir=plus, y explicit]
		table[x expr=\thisrow{magSparsity}*1.2, y=graspMeanPostprune, y error=graspErrorPlusPostprune] {expres/grid/reshelixGridSingleshotDepth10.tsv};
		\addplot+[eda errorbarcolored, y dir=minus, y explicit]
		table[x expr=\thisrow{magSparsity}*1.2, y=graspMeanPostprune, y error=graspErrorMinusPostprune] {expres/grid/reshelixGridSingleshotDepth10.tsv};
		
        \addplot[mark=none, dashed, black, samples=2] coordinates {(0.001,0.00031) (1.0,0.00031)};
		
		\end{axis}
		\end{tikzpicture}
		\fi
    \end{subfigure}
    \begin{subfigure}[t]{0.49\textwidth}
    \centering
		\ifpdf
		\begin{tikzpicture}
		\begin{axis}[
		jonas line,
		xmode=log,
		ymode=log,
		minor x tick num = 5,
        y axis line style 	= {opacity=0},
        y tick style = {draw=none},
        yticklabels={,,},
		width = 7cm,
		height = 4cm,
		xlabel = {Sparsity}, 
		xmin=0.001, xmax=1.1,
		ymax=1,
		legend columns=2,
		x label style 		= {at={(axis description cs:0.5,-0.1)}, anchor=north, font=\scriptsize},
		y label style 		= {at={(axis description cs:0.0,0.6)},  anchor=south, font=\scriptsize},
		legend style={nodes={scale=0.9, transform shape}, at={(0.98,0.95)}, anchor=north east, row sep=-1.4pt, font=\tiny}
		]
		
		\addplot+[forget plot, eda errorbarcolored, y dir=plus, y explicit]
		table[x=magSparsity, y=magMeanPosttrain, y error=magErrorPlusPosttrain] {expres/grid/reshelixGridSingleshotDepth10.tsv};
		\addplot+[eda errorbarcolored, y dir=minus, y explicit]
		table[x=magSparsity, y=magMeanPosttrain, y error=magErrorMinusPosttrain] {expres/grid/reshelixGridSingleshotDepth10.tsv};
		
		\addplot+[forget plot, eda errorbarcolored, y dir=plus, y explicit]
		table[x expr=\thisrow{magSparsity}*1.05, y=synflowMeanPosttrain, y error=synflowErrorPlusPosttrain] {expres/grid/reshelixGridSingleshotDepth10.tsv};
		\addplot+[eda errorbarcolored, y dir=minus, y explicit]
		table[x expr=\thisrow{magSparsity}*1.05, y=synflowMeanPosttrain, y error=synflowErrorMinusPosttrain] {expres/grid/reshelixGridSingleshotDepth10.tsv};
		
		\addplot+[forget plot, eda errorbarcolored, y dir=plus, y explicit]
		table[x expr=\thisrow{magSparsity}*1.1, y=randMeanPosttrain, y error=randErrorPlusPosttrain] {expres/grid/reshelixGridSingleshotDepth10.tsv};
		\addplot+[eda errorbarcolored, y dir=minus, y explicit]
		table[x expr=\thisrow{magSparsity}*1.1, y=randMeanPosttrain, y error=randErrorMinusPosttrain] {expres/grid/reshelixGridSingleshotDepth10.tsv};
		
		\addplot+[forget plot, eda errorbarcolored, y dir=plus, y explicit]
		table[x expr=\thisrow{magSparsity}*1.15, y=snipMeanPosttrain, y error=snipErrorPlusPosttrain] {expres/grid/reshelixGridSingleshotDepth10.tsv};
		\addplot+[eda errorbarcolored, y dir=minus, y explicit]
		table[x expr=\thisrow{magSparsity}*1.15, y=snipMeanPosttrain, y error=snipErrorMinusPosttrain] {expres/grid/reshelixGridSingleshotDepth10.tsv};
		
		\addplot+[forget plot, eda errorbarcolored, y dir=plus, y explicit]
		table[x expr=\thisrow{magSparsity}*1.2, y=graspMeanPosttrain, y error=graspErrorPlusPosttrain] {expres/grid/reshelixGridSingleshotDepth10.tsv};
		\addplot+[eda errorbarcolored, y dir=minus, y explicit]
		table[x expr=\thisrow{magSparsity}*1.2, y=graspMeanPosttrain, y error=graspErrorMinusPosttrain] {expres/grid/reshelixGridSingleshotDepth10.tsv};

        \addplot[mark=none, dashed, black, samples=2] coordinates {(0.001,0.00031) (1.0,0.00031)};
		
		\end{axis}
		\end{tikzpicture}
		\fi
    \end{subfigure}
    \caption{\textit{Singleshot results depth $10$.} Performance of discovered tickets for \texttt{Circle}, \texttt{ReLU}, and \texttt{Helix}  against target sparsities as mean and obtained intervals (minimum and maximum) across $25$ runs. In order of appearance from top to bottom: \texttt{Circle}, \texttt{ReLU}, and \texttt{Helix} post pruning (left) and post training performance (right). Baseline ticket has leftmost sparsity and its performance given by black dashed line.}\label{fig:app_single10}
\end{figure}

%% file: figs/width1k_app.tex
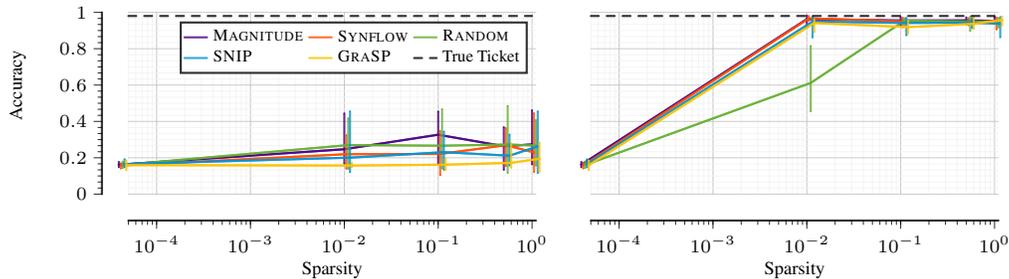
\begin{figure}
\centering
    \begin{subfigure}[t]{0.49\textwidth}
    \centering
		\ifpdf
		\begin{tikzpicture}
		\begin{axis}[
		jonas line,
		xmode=log,
		width = 7cm,
		height = 4cm,
		xlabel = {Sparsity}, 
		ylabel= {Accuracy},
		xmin=0.00005, xmax=1.1,
		ymin=0, ymax=1,
		legend columns=3,
		x label style 		= {at={(axis description cs:0.5,-0.1)}, anchor=north, font=\scriptsize},
		y label style 		= {at={(axis description cs:0.0,0.6)},  anchor=south, font=\scriptsize},
		legend style={nodes={scale=0.9, transform shape}, at={(0.98,0.95)}, anchor=north east, row sep=-1.4pt, font=\tiny}
		]
		
		\addplot+[forget plot, eda errorbarcolored, y dir=plus, y explicit]
		table[x=magSparsity, y=magMeanPostprune, y error=magErrorPlusPostprune] {expres/width1k/rescircleWidth1kSingleshot.tsv};
		\addplot+[eda errorbarcolored, y dir=minus, y explicit]
		table[x=magSparsity, y=magMeanPostprune, y error=magErrorMinusPostprune] {expres/width1k/rescircleWidth1kSingleshot.tsv};
		\addlegendentry{\magnitude}
		
		\addplot+[forget plot, eda errorbarcolored, y dir=plus, y explicit]
		table[x expr=\thisrow{magSparsity}*1.05, y=synflowMeanPostprune, y error=synflowErrorPlusPostprune] {expres/width1k/rescircleWidth1kSingleshot.tsv};
		\addplot+[eda errorbarcolored, y dir=minus, y explicit]
		table[x expr=\thisrow{magSparsity}*1.05, y=synflowMeanPostprune, y error=synflowErrorMinusPostprune] {expres/width1k/rescircleWidth1kSingleshot.tsv};
		\addlegendentry{\synflow}
		
		\addplot+[forget plot, eda errorbarcolored, y dir=plus, y explicit]
		table[x expr=\thisrow{magSparsity}*1.1, y=randMeanPostprune, y error=randErrorPlusPostprune] {expres/width1k/rescircleWidth1kSingleshot.tsv};
		\addplot+[eda errorbarcolored, y dir=minus, y explicit]
		table[x expr=\thisrow{magSparsity}*1.1, y=randMeanPostprune, y error=randErrorMinusPostprune] {expres/width1k/rescircleWidth1kSingleshot.tsv};
		\addlegendentry{\rand}
		
		\addplot+[forget plot, eda errorbarcolored, y dir=plus, y explicit]
		table[x expr=\thisrow{magSparsity}*1.15, y=snipMeanPostprune, y error=snipErrorPlusPostprune] {expres/width1k/rescircleWidth1kSingleshot.tsv};
		\addplot+[eda errorbarcolored, y dir=minus, y explicit]
		table[x expr=\thisrow{magSparsity}*1.15, y=snipMeanPostprune, y error=snipErrorMinusPostprune] {expres/width1k/rescircleWidth1kSingleshot.tsv};
		\addlegendentry{\snip}
		
		\addplot+[forget plot, eda errorbarcolored, y dir=plus, y explicit]
		table[x expr=\thisrow{magSparsity}*1.2, y=graspMeanPostprune, y error=graspErrorPlusPostprune] {expres/width1k/rescircleWidth1kSingleshot.tsv};
		\addplot+[eda errorbarcolored, y dir=minus, y explicit]
		table[x expr=\thisrow{magSparsity}*1.2, y=graspMeanPostprune, y error=graspErrorMinusPostprune] {expres/width1k/rescircleWidth1kSingleshot.tsv};
		\addlegendentry{\grasp}
		
        \addplot[mark=none, dashed, black, samples=2] coordinates {(0.00005,0.98) (1.0,0.98)};
        \addlegendentry{True Ticket}
		
		\end{axis}
		\end{tikzpicture}
		\fi
    \end{subfigure}
    \begin{subfigure}[t]{0.49\textwidth}
    \centering
		\ifpdf
		\begin{tikzpicture}
		\begin{axis}[
		jonas line,
		xmode=log,
        y axis line style 	= {opacity=0},
        y tick style = {draw=none},
        yticklabels={,,},
		width = 7cm,
		height = 4cm,
		xlabel = {Sparsity}, 
		ylabel= {},
		xmin=0.00005, xmax=1.1,
		ymin=0, ymax=1,
		legend columns=2,
		x label style 		= {at={(axis description cs:0.5,-0.1)}, anchor=north, font=\scriptsize},
		y label style 		= {at={(axis description cs:0.0,0.6)},  anchor=south, font=\scriptsize},
		legend style={nodes={scale=0.9, transform shape}, at={(0.01,0.95)}, anchor=north west, row sep=-1.4pt, font=\tiny}
		]
		
		\addplot+[forget plot, eda errorbarcolored, y dir=plus, y explicit]
		table[x=magSparsity, y=magMeanPosttrain, y error=magErrorPlusPosttrain] {expres/width1k/rescircleWidth1kSingleshot.tsv};
		\addplot+[eda errorbarcolored, y dir=minus, y explicit]
		table[x=magSparsity, y=magMeanPosttrain, y error=magErrorMinusPosttrain] {expres/width1k/rescircleWidth1kSingleshot.tsv};
		
		\addplot+[forget plot, eda errorbarcolored, y dir=plus, y explicit]
		table[x expr=\thisrow{magSparsity}*1.05, y=synflowMeanPosttrain, y error=synflowErrorPlusPosttrain] {expres/width1k/rescircleWidth1kSingleshot.tsv};
		\addplot+[eda errorbarcolored, y dir=minus, y explicit]
		table[x expr=\thisrow{magSparsity}*1.05, y=synflowMeanPosttrain, y error=synflowErrorMinusPosttrain] {expres/width1k/rescircleWidth1kSingleshot.tsv};
		
		\addplot+[forget plot, eda errorbarcolored, y dir=plus, y explicit]
		table[x expr=\thisrow{magSparsity}*1.1, y=randMeanPosttrain, y error=randErrorPlusPosttrain] {expres/width1k/rescircleWidth1kSingleshot.tsv};
		\addplot+[eda errorbarcolored, y dir=minus, y explicit]
		table[x expr=\thisrow{magSparsity}*1.1, y=randMeanPosttrain, y error=randErrorMinusPosttrain] {expres/width1k/rescircleWidth1kSingleshot.tsv};
		
		\addplot+[forget plot, eda errorbarcolored, y dir=plus, y explicit]
		table[x expr=\thisrow{magSparsity}*1.15, y=snipMeanPosttrain, y error=snipErrorPlusPosttrain] {expres/width1k/rescircleWidth1kSingleshot.tsv};
		\addplot+[eda errorbarcolored, y dir=minus, y explicit]
		table[x expr=\thisrow{magSparsity}*1.15, y=snipMeanPosttrain, y error=snipErrorMinusPosttrain] {expres/width1k/rescircleWidth1kSingleshot.tsv};
		
		\addplot+[forget plot, eda errorbarcolored, y dir=plus, y explicit]
		table[x expr=\thisrow{magSparsity}*1.2, y=graspMeanPosttrain, y error=graspErrorPlusPosttrain] {expres/width1k/rescircleWidth1kSingleshot.tsv};
		\addplot+[eda errorbarcolored, y dir=minus, y explicit]
		table[x expr=\thisrow{magSparsity}*1.2, y=graspMeanPosttrain, y error=graspErrorMinusPosttrain] {expres/width1k/rescircleWidth1kSingleshot.tsv};
		
        \addplot[mark=none, dashed, black, samples=2] coordinates {(0.00005,0.98) (1.0,0.98)};
		
		\end{axis}
		\end{tikzpicture}
		\fi
    \end{subfigure}
 \caption{\textit{Singleshot results, depth $6$ width $1000$.} Performance on test data are plotted for  \texttt{Circle} against target sparsities. We report mean and obtained intervals (minimum and maximum) across $10$ repetitions of ticket performance right after pruning (left) and after training (right). The baseline ticket performance is indicated by the black line, leftmost sparsity  correspond to planted ticket sparsity.}\label{fig:app_1k}
\end{figure}

%% file: figs/synflow_100_app.tex
\begin{figure}
\centering
    \begin{subfigure}[t]{0.49\textwidth}
    \centering
		\ifpdf
		\begin{tikzpicture}
		\begin{axis}[
		jonas line,
		xmode=log,
		width = 7cm,
		height = 4cm,
		xlabel = {Sparsity}, 
		ylabel= {Accuracy},
		xmin=0.003, xmax=1.1,
		ymin=0, ymax=1,
		legend columns=3,
		x label style 		= {at={(axis description cs:0.5,-0.1)}, anchor=north, font=\scriptsize},
		y label style 		= {at={(axis description cs:0.0,0.6)},  anchor=south, font=\scriptsize},
		legend style={nodes={scale=1, transform shape}, at={(0.98,0.95)}, anchor=north east, row sep=-1.4pt, font=\tiny}
		]
		
		\addplot+[internationalorange, forget plot, eda errorbarcolored, y dir=plus, y explicit]
		table[x expr=\thisrow{synflowSparsity}*1, y=synflowMeanPostprune, y error=synflowErrorPlusPostprune] {expres/SynflowExp/rescircleSynflow100roundsSingleshot.tsv};
		\addplot+[internationalorange, eda errorbarcolored, y dir=minus, y explicit]
		table[x expr=\thisrow{synflowSparsity}*1, y=synflowMeanPostprune, y error=synflowErrorMinusPostprune] {expres/SynflowExp/rescircleSynflow100roundsSingleshot.tsv};
		\addlegendentry{\synflow}
		
        \addplot[mark=none, dashed, black, samples=2] coordinates {(0.003,0.98) (1.0,0.98)};
        \addlegendentry{True Ticket}
		
		\end{axis}
		\end{tikzpicture}
		\fi
    \end{subfigure}
    \begin{subfigure}[t]{0.49\textwidth}
    \centering
		\ifpdf
		\begin{tikzpicture}
		\begin{axis}[
		jonas line,
		xmode=log,
		width = 7cm,
		height = 4cm,
		xlabel = {Sparsity}, 
		ylabel= {},
		xmin=0.003, xmax=1.1,
		ymin=0, ymax=1,
		legend columns=3,
		x label style 		= {at={(axis description cs:0.5,-0.1)}, anchor=north, font=\scriptsize},
		y label style 		= {at={(axis description cs:0.0,0.6)},  anchor=south, font=\scriptsize},
		legend style={nodes={scale=1, transform shape}, at={(0.01,0.95)}, anchor=north east, row sep=-1.4pt, font=\tiny}
		]

		\addplot+[internationalorange, forget plot, eda errorbarcolored, y dir=plus, y explicit]
		table[x expr=\thisrow{synflowSparsity}*1, y=synflowMeanPosttrain, y error=synflowErrorPlusPosttrain] {expres/SynflowExp/rescircleSynflow100roundsSingleshot.tsv};
		\addplot+[internationalorange, eda errorbarcolored, y dir=minus, y explicit]
		table[x expr=\thisrow{synflowSparsity}*1, y=synflowMeanPosttrain, y error=synflowErrorMinusPosttrain] {expres/SynflowExp/rescircleSynflow100roundsSingleshot.tsv};
		
        \addplot[mark=none, dashed, black, samples=2] coordinates {(0.003,0.98) (1.0,0.98)};
		
		\end{axis}
		\end{tikzpicture}
		\fi
    \end{subfigure}
 \caption{\textit{Singleshot results for \synflow with $100$ pruning rounds.} Performance on test data are plotted for  \texttt{Circle} against target sparsities. We report mean and obtained intervals (minimum and maximum) across $10$ repetitions of ticket performance right after pruning (left) and after training (right). The original network is of depth $6$ and width $100$. The baseline ticket performance is indicated by the black dashed line, leftmost sparsity corresponds to planted ticket sparsity.}\label{fig:app_synflow}
\end{figure}
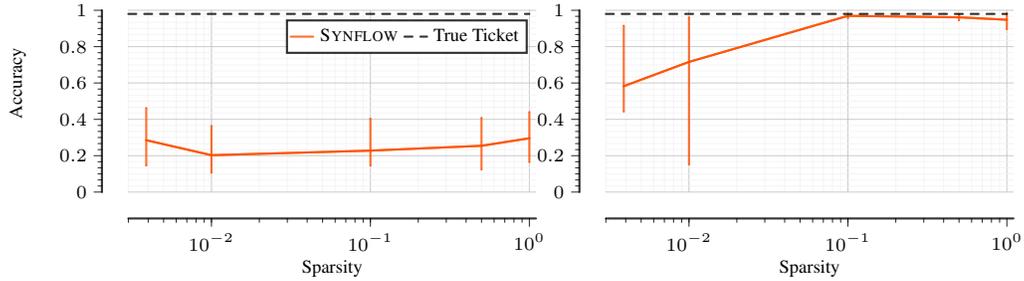

%% file: figs/noise_exp.tex
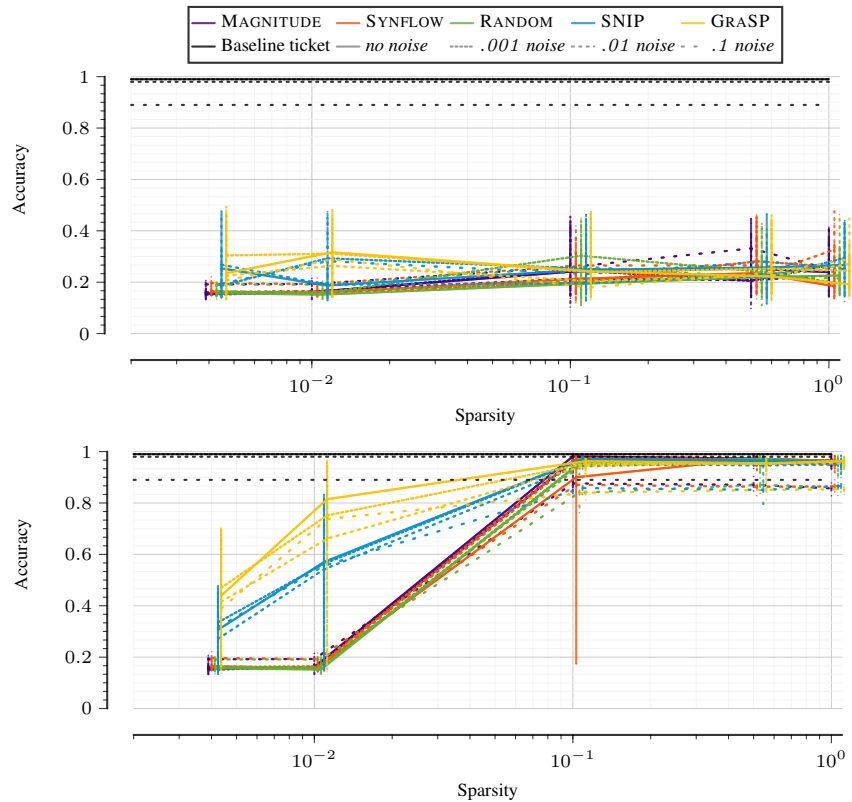
\begin{figure}
\centering
    \begin{subfigure}[t]{0.99\textwidth}
    \centering
		\ifpdf
		\begin{tikzpicture}
		\begin{axis}[
		jonas line,
		cycle multi list={
                mamba5
                    \nextlist
                solid, densely dotted, dotted, loosely dotted
            },
		xmode=log,
		width = 11cm,
		height = 5cm,
		xlabel = {Sparsity}, 
		ylabel= {Accuracy},
		xmin=0.002, xmax=1.1,
		ymin=0, ymax=1,
		legend columns=5,
		x label style 		= {at={(axis description cs:0.5,-0.1)}, anchor=north, font=\scriptsize},
		y label style 		= {at={(axis description cs:0.0,0.6)},  anchor=south, font=\scriptsize},
		legend style={nodes={scale=1, transform shape}, at={(0.5,1.05)}, anchor=south, row sep=-1.4pt}
		]

		\addlegendimage{no markers,indigo(web)}
		\addlegendentry{\magnitude}
		\addlegendimage{no markers,internationalorange}
		\addlegendentry{\synflow}
		\addlegendimage{no markers,green(ryb)}
		\addlegendentry{\rand}
		\addlegendimage{no markers,richelectricblue}
		\addlegendentry{\snip}
		\addlegendimage{no markers,goldenpoppy}
		\addlegendentry{\grasp}
		\addlegendimage{no markers,black}
		\addlegendentry{Baseline ticket}
		\addlegendimage{no markers,gray, solid}
		\addlegendentry{\textit{no noise}}
		\addlegendimage{no markers,gray, densely dotted}
		\addlegendentry{\textit{$\mathit{.001}$ noise}}
		\addlegendimage{no markers,gray, dotted}
		\addlegendentry{\textit{$\mathit{.01}$ noise}}
		\addlegendimage{no markers,gray, loosely dotted}
		\addlegendentry{\textit{$\mathit{.1}$ noise}}

		\addplot+[forget plot, eda errorbarcolored, y dir=plus, y explicit]
		table[x=magSparsity, y=magMeanPostprune, y error=magErrorPlusPostprune] {expres/noise/rescircleMultishotNoise0.tsv};
		\addplot+[eda errorbarcolored, y dir=minus, y explicit]
		table[x=magSparsity, y=magMeanPostprune, y error=magErrorMinusPostprune] {expres/noise/rescircleMultishotNoise0.tsv};
		
		\addplot+[forget plot, eda errorbarcolored, y dir=plus, y explicit]
		table[x=magSparsity, y=magMeanPostprune, y error=magErrorPlusPostprune] {expres/noise/rescircleMultishotNoise0.001.tsv};
		\addplot+[eda errorbarcolored, y dir=minus, y explicit]
		table[x=magSparsity, y=magMeanPostprune, y error=magErrorMinusPostprune] {expres/noise/rescircleMultishotNoise0.001.tsv};
		
		\addplot+[forget plot, eda errorbarcolored, y dir=plus, y explicit]
		table[x=magSparsity, y=magMeanPostprune, y error=magErrorPlusPostprune] {expres/noise/rescircleMultishotNoise0.01.tsv};
		\addplot+[eda errorbarcolored, y dir=minus, y explicit]
		table[x=magSparsity, y=magMeanPostprune, y error=magErrorMinusPostprune] {expres/noise/rescircleMultishotNoise0.01.tsv};
		
		\addplot+[forget plot, eda errorbarcolored, y dir=plus, y explicit]
		table[x=magSparsity, y=magMeanPostprune, y error=magErrorPlusPostprune] {expres/noise/rescircleMultishotNoise0.1.tsv};
		\addplot+[eda errorbarcolored, y dir=minus, y explicit]
		table[x=magSparsity, y=magMeanPostprune, y error=magErrorMinusPostprune] {expres/noise/rescircleMultishotNoise0.1.tsv};

		\addplot+[forget plot, eda errorbarcolored, y dir=plus, y explicit]
		table[x expr=\thisrow{magSparsity}*1.05, y=synflowMeanPostprune, y error=synflowErrorPlusPostprune] {expres/noise/rescircleMultishotNoise0.tsv};
		\addplot+[eda errorbarcolored, y dir=minus, y explicit]
		table[x expr=\thisrow{magSparsity}*1.05, y=synflowMeanPostprune, y error=synflowErrorMinusPostprune] {expres/noise/rescircleMultishotNoise0.tsv};
		
		\addplot+[forget plot, eda errorbarcolored, y dir=plus, y explicit]
		table[x expr=\thisrow{magSparsity}*1.05, y=synflowMeanPostprune, y error=synflowErrorPlusPostprune] {expres/noise/rescircleMultishotNoise0.001.tsv};
		\addplot+[eda errorbarcolored, y dir=minus, y explicit]
		table[x expr=\thisrow{magSparsity}*1.05, y=synflowMeanPostprune, y error=synflowErrorMinusPostprune] {expres/noise/rescircleMultishotNoise0.001.tsv};
		
		\addplot+[forget plot, eda errorbarcolored, y dir=plus, y explicit]
		table[x expr=\thisrow{magSparsity}*1.05, y=synflowMeanPostprune, y error=synflowErrorPlusPostprune] {expres/noise/rescircleMultishotNoise0.01.tsv};
		\addplot+[eda errorbarcolored, y dir=minus, y explicit]
		table[x expr=\thisrow{magSparsity}*1.05, y=synflowMeanPostprune, y error=synflowErrorMinusPostprune] {expres/noise/rescircleMultishotNoise0.01.tsv};
		
		\addplot+[forget plot, eda errorbarcolored, y dir=plus, y explicit]
		table[x expr=\thisrow{magSparsity}*1.05, y=synflowMeanPostprune, y error=synflowErrorPlusPostprune] {expres/noise/rescircleMultishotNoise0.1.tsv};
		\addplot+[eda errorbarcolored, y dir=minus, y explicit]
		table[x expr=\thisrow{magSparsity}*1.05, y=synflowMeanPostprune, y error=synflowErrorMinusPostprune] {expres/noise/rescircleMultishotNoise0.1.tsv};

		\addplot+[forget plot, eda errorbarcolored, y dir=plus, y explicit]
		table[x expr=\thisrow{magSparsity}*1.1, y=randMeanPostprune, y error=randErrorPlusPostprune] {expres/noise/rescircleMultishotNoise0.tsv};
		\addplot+[eda errorbarcolored, y dir=minus, y explicit]
		table[x expr=\thisrow{magSparsity}*1.1, y=randMeanPostprune, y error=randErrorMinusPostprune] {expres/noise/rescircleMultishotNoise0.tsv};
		
		\addplot+[forget plot, eda errorbarcolored, y dir=plus, y explicit]
		table[x expr=\thisrow{magSparsity}*1.1, y=randMeanPostprune, y error=randErrorPlusPostprune] {expres/noise/rescircleMultishotNoise0.001.tsv};
		\addplot+[eda errorbarcolored, y dir=minus, y explicit]
		table[x expr=\thisrow{magSparsity}*1.1, y=randMeanPostprune, y error=randErrorMinusPostprune] {expres/noise/rescircleMultishotNoise0.001.tsv};
		
		\addplot+[forget plot, eda errorbarcolored, y dir=plus, y explicit]
		table[x expr=\thisrow{magSparsity}*1.1, y=randMeanPostprune, y error=randErrorPlusPostprune] {expres/noise/rescircleMultishotNoise0.01.tsv};
		\addplot+[eda errorbarcolored, y dir=minus, y explicit]
		table[x expr=\thisrow{magSparsity}*1.1, y=randMeanPostprune, y error=randErrorMinusPostprune] {expres/noise/rescircleMultishotNoise0.01.tsv};
		
		\addplot+[forget plot, eda errorbarcolored, y dir=plus, y explicit]
		table[x expr=\thisrow{magSparsity}*1.1, y=randMeanPostprune, y error=randErrorPlusPostprune] {expres/noise/rescircleMultishotNoise0.1.tsv};
		\addplot+[eda errorbarcolored, y dir=minus, y explicit]
		table[x expr=\thisrow{magSparsity}*1.1, y=randMeanPostprune, y error=randErrorMinusPostprune] {expres/noise/rescircleMultishotNoise0.1.tsv};

		\addplot+[forget plot, eda errorbarcolored, y dir=plus, y explicit]
		table[x expr=\thisrow{magSparsity}*1.15, y=snipMeanPostprune, y error=snipErrorPlusPostprune] {expres/noise/rescircleMultishotNoise0.tsv};
		\addplot+[eda errorbarcolored, y dir=minus, y explicit]
		table[x expr=\thisrow{magSparsity}*1.15, y=snipMeanPostprune, y error=snipErrorMinusPostprune] {expres/noise/rescircleMultishotNoise0.tsv};
		
		\addplot+[forget plot, eda errorbarcolored, y dir=plus, y explicit]
		table[x expr=\thisrow{magSparsity}*1.15, y=snipMeanPostprune, y error=snipErrorPlusPostprune] {expres/noise/rescircleMultishotNoise0.001.tsv};
		\addplot+[eda errorbarcolored, y dir=minus, y explicit]
		table[x expr=\thisrow{magSparsity}*1.15, y=snipMeanPostprune, y error=snipErrorMinusPostprune] {expres/noise/rescircleMultishotNoise0.001.tsv};
		
		\addplot+[forget plot, eda errorbarcolored, y dir=plus, y explicit]
		table[x expr=\thisrow{magSparsity}*1.15, y=snipMeanPostprune, y error=snipErrorPlusPostprune] {expres/noise/rescircleMultishotNoise0.01.tsv};
		\addplot+[eda errorbarcolored, y dir=minus, y explicit]
		table[x expr=\thisrow{magSparsity}*1.15, y=snipMeanPostprune, y error=snipErrorMinusPostprune] {expres/noise/rescircleMultishotNoise0.01.tsv};
		
		\addplot+[forget plot, eda errorbarcolored, y dir=plus, y explicit]
		table[x expr=\thisrow{magSparsity}*1.15, y=snipMeanPostprune, y error=snipErrorPlusPostprune] {expres/noise/rescircleMultishotNoise0.1.tsv};
		\addplot+[eda errorbarcolored, y dir=minus, y explicit]
		table[x expr=\thisrow{magSparsity}*1.15, y=snipMeanPostprune, y error=snipErrorMinusPostprune] {expres/noise/rescircleMultishotNoise0.1.tsv};

		\addplot+[forget plot, eda errorbarcolored, y dir=plus, y explicit]
		table[x expr=\thisrow{magSparsity}*1.2, y=graspMeanPostprune, y error=graspErrorPlusPostprune] {expres/noise/rescircleMultishotNoise0.tsv};
		\addplot+[eda errorbarcolored, y dir=minus, y explicit]
		table[x expr=\thisrow{magSparsity}*1.2, y=graspMeanPostprune, y error=graspErrorMinusPostprune] {expres/noise/rescircleMultishotNoise0.tsv};
		
		\addplot+[forget plot, eda errorbarcolored, y dir=plus, y explicit]
		table[x expr=\thisrow{magSparsity}*1.2, y=graspMeanPostprune, y error=graspErrorPlusPostprune] {expres/noise/rescircleMultishotNoise0.001.tsv};
		\addplot+[eda errorbarcolored, y dir=minus, y explicit]
		table[x expr=\thisrow{magSparsity}*1.2, y=graspMeanPostprune, y error=graspErrorMinusPostprune] {expres/noise/rescircleMultishotNoise0.001.tsv};
		
		\addplot+[forget plot, eda errorbarcolored, y dir=plus, y explicit]
		table[x expr=\thisrow{magSparsity}*1.2, y=graspMeanPostprune, y error=graspErrorPlusPostprune] {expres/noise/rescircleMultishotNoise0.01.tsv};
		\addplot+[eda errorbarcolored, y dir=minus, y explicit]
		table[x expr=\thisrow{magSparsity}*1.2, y=graspMeanPostprune, y error=graspErrorMinusPostprune] {expres/noise/rescircleMultishotNoise0.01.tsv};
		
		\addplot+[forget plot, eda errorbarcolored, y dir=plus, y explicit]
		table[x expr=\thisrow{magSparsity}*1.2, y=graspMeanPostprune, y error=graspErrorPlusPostprune] {expres/noise/rescircleMultishotNoise0.1.tsv};
		\addplot+[eda errorbarcolored, y dir=minus, y explicit]
		table[x expr=\thisrow{magSparsity}*1.2, y=graspMeanPostprune, y error=graspErrorMinusPostprune] {expres/noise/rescircleMultishotNoise0.1.tsv};

        \addplot[mark=none, solid, black, samples=2] coordinates {(0.002,0.99) (1.0,0.99)};
        \addplot[mark=none, densely dotted, black, samples=2] coordinates {(0.002,0.99) (1.0,0.99)};
        \addplot[mark=none, dotted, black, samples=2] coordinates {(0.002,0.98) (1.0,0.98)};
        \addplot[mark=none, loosely dotted, black, samples=2] coordinates {(0.002,0.89) (1.0,0.89)};
		
		\end{axis}
		\end{tikzpicture}
		\fi
    \end{subfigure}
    \begin{subfigure}[t]{0.99\textwidth}
    \centering
		\ifpdf
		\begin{tikzpicture}
		\begin{axis}[
		jonas line,
		cycle multi list={
                mamba5
                    \nextlist
                solid, densely dotted, dotted, loosely dotted
            },
		xmode=log,
		width = 11cm,
		height = 5cm,
		xlabel = {Sparsity}, 
		ylabel= {Accuracy},
		xmin=0.002, xmax=1.1,
		ymin=0, ymax=1,
		legend columns=2,
		x label style 		= {at={(axis description cs:0.5,-0.1)}, anchor=north, font=\scriptsize},
		y label style 		= {at={(axis description cs:0.0,0.6)},  anchor=south, font=\scriptsize},
		legend style={nodes={scale=0.9, transform shape}, at={(0.98,0.95)}, anchor=north east, row sep=-1.4pt, font=\tiny}
		]

		\addplot+[forget plot, eda errorbarcolored, y dir=plus, y explicit]
		table[x=magSparsity, y=magMeanPosttrain, y error=magErrorPlusPosttrain] {expres/noise/rescircleMultishotNoise0.tsv};
		\addplot+[eda errorbarcolored, y dir=minus, y explicit]
		table[x=magSparsity, y=magMeanPosttrain, y error=magErrorMinusPosttrain] {expres/noise/rescircleMultishotNoise0.tsv};
		
		\addplot+[forget plot, eda errorbarcolored, y dir=plus, y explicit]
		table[x=magSparsity, y=magMeanPosttrain, y error=magErrorPlusPosttrain] {expres/noise/rescircleMultishotNoise0.001.tsv};
		\addplot+[eda errorbarcolored, y dir=minus, y explicit]
		table[x=magSparsity, y=magMeanPosttrain, y error=magErrorMinusPosttrain] {expres/noise/rescircleMultishotNoise0.001.tsv};
		
		\addplot+[forget plot, eda errorbarcolored, y dir=plus, y explicit]
		table[x=magSparsity, y=magMeanPosttrain, y error=magErrorPlusPosttrain] {expres/noise/rescircleMultishotNoise0.01.tsv};
		\addplot+[eda errorbarcolored, y dir=minus, y explicit]
		table[x=magSparsity, y=magMeanPosttrain, y error=magErrorMinusPosttrain] {expres/noise/rescircleMultishotNoise0.01.tsv};
		
		\addplot+[forget plot, eda errorbarcolored, y dir=plus, y explicit]
		table[x=magSparsity, y=magMeanPosttrain, y error=magErrorPlusPosttrain] {expres/noise/rescircleMultishotNoise0.1.tsv};
		\addplot+[eda errorbarcolored, y dir=minus, y explicit]
		table[x=magSparsity, y=magMeanPosttrain, y error=magErrorMinusPosttrain] {expres/noise/rescircleMultishotNoise0.1.tsv};

		\addplot+[forget plot, eda errorbarcolored, y dir=plus, y explicit]
		table[x expr=\thisrow{magSparsity}*1.03, y=synflowMeanPosttrain, y error=synflowErrorPlusPosttrain] {expres/noise/rescircleMultishotNoise0.tsv};
		\addplot+[eda errorbarcolored, y dir=minus, y explicit]
		table[x expr=\thisrow{magSparsity}*1.03, y=synflowMeanPosttrain, y error=synflowErrorMinusPosttrain] {expres/noise/rescircleMultishotNoise0.tsv};
		
		\addplot+[forget plot, eda errorbarcolored, y dir=plus, y explicit]
		table[x expr=\thisrow{magSparsity}*1.03, y=synflowMeanPosttrain, y error=synflowErrorPlusPosttrain] {expres/noise/rescircleMultishotNoise0.001.tsv};
		\addplot+[eda errorbarcolored, y dir=minus, y explicit]
		table[x expr=\thisrow{magSparsity}*1.03, y=synflowMeanPosttrain, y error=synflowErrorMinusPosttrain] {expres/noise/rescircleMultishotNoise0.001.tsv};
		
		\addplot+[forget plot, eda errorbarcolored, y dir=plus, y explicit]
		table[x expr=\thisrow{magSparsity}*1.03, y=synflowMeanPosttrain, y error=synflowErrorPlusPosttrain] {expres/noise/rescircleMultishotNoise0.01.tsv};
		\addplot+[eda errorbarcolored, y dir=minus, y explicit]
		table[x expr=\thisrow{magSparsity}*1.03, y=synflowMeanPosttrain, y error=synflowErrorMinusPosttrain] {expres/noise/rescircleMultishotNoise0.01.tsv};
		
		\addplot+[forget plot, eda errorbarcolored, y dir=plus, y explicit]
		table[x expr=\thisrow{magSparsity}*1.03, y=synflowMeanPosttrain, y error=synflowErrorPlusPosttrain] {expres/noise/rescircleMultishotNoise0.1.tsv};
		\addplot+[eda errorbarcolored, y dir=minus, y explicit]
		table[x expr=\thisrow{magSparsity}*1.03, y=synflowMeanPosttrain, y error=synflowErrorMinusPosttrain] {expres/noise/rescircleMultishotNoise0.1.tsv};

		\addplot+[forget plot, eda errorbarcolored, y dir=plus, y explicit]
		table[x expr=\thisrow{magSparsity}*1.06, y=randMeanPosttrain, y error=randErrorPlusPosttrain] {expres/noise/rescircleMultishotNoise0.tsv};
		\addplot+[eda errorbarcolored, y dir=minus, y explicit]
		table[x expr=\thisrow{magSparsity}*1.06, y=randMeanPosttrain, y error=randErrorMinusPosttrain] {expres/noise/rescircleMultishotNoise0.tsv};
		
		\addplot+[forget plot, eda errorbarcolored, y dir=plus, y explicit]
		table[x expr=\thisrow{magSparsity}*1.06, y=randMeanPosttrain, y error=randErrorPlusPosttrain] {expres/noise/rescircleMultishotNoise0.001.tsv};
		\addplot+[eda errorbarcolored, y dir=minus, y explicit]
		table[x expr=\thisrow{magSparsity}*1.06, y=randMeanPosttrain, y error=randErrorMinusPosttrain] {expres/noise/rescircleMultishotNoise0.001.tsv};
		
		\addplot+[forget plot, eda errorbarcolored, y dir=plus, y explicit]
		table[x expr=\thisrow{magSparsity}*1.06, y=randMeanPosttrain, y error=randErrorPlusPosttrain] {expres/noise/rescircleMultishotNoise0.01.tsv};
		\addplot+[eda errorbarcolored, y dir=minus, y explicit]
		table[x expr=\thisrow{magSparsity}*1.06, y=randMeanPosttrain, y error=randErrorMinusPosttrain] {expres/noise/rescircleMultishotNoise0.01.tsv};
		
		\addplot+[forget plot, eda errorbarcolored, y dir=plus, y explicit]
		table[x expr=\thisrow{magSparsity}*1.06, y=randMeanPosttrain, y error=randErrorPlusPosttrain] {expres/noise/rescircleMultishotNoise0.1.tsv};
		\addplot+[eda errorbarcolored, y dir=minus, y explicit]
		table[x expr=\thisrow{magSparsity}*1.06, y=randMeanPosttrain, y error=randErrorMinusPosttrain] {expres/noise/rescircleMultishotNoise0.1.tsv};

		\addplot+[forget plot, eda errorbarcolored, y dir=plus, y explicit]
		table[x expr=\thisrow{magSparsity}*1.09, y=snipMeanPosttrain, y error=snipErrorPlusPosttrain] {expres/noise/rescircleMultishotNoise0.tsv};
		\addplot+[eda errorbarcolored, y dir=minus, y explicit]
		table[x expr=\thisrow{magSparsity}*1.09, y=snipMeanPosttrain, y error=snipErrorMinusPosttrain] {expres/noise/rescircleMultishotNoise0.tsv};
		
		\addplot+[forget plot, eda errorbarcolored, y dir=plus, y explicit]
		table[x expr=\thisrow{magSparsity}*1.09, y=snipMeanPosttrain, y error=snipErrorPlusPosttrain] {expres/noise/rescircleMultishotNoise0.001.tsv};
		\addplot+[eda errorbarcolored, y dir=minus, y explicit]
		table[x expr=\thisrow{magSparsity}*1.09, y=snipMeanPosttrain, y error=snipErrorMinusPosttrain] {expres/noise/rescircleMultishotNoise0.001.tsv};
		
		\addplot+[forget plot, eda errorbarcolored, y dir=plus, y explicit]
		table[x expr=\thisrow{magSparsity}*1.09, y=snipMeanPosttrain, y error=snipErrorPlusPosttrain] {expres/noise/rescircleMultishotNoise0.01.tsv};
		\addplot+[eda errorbarcolored, y dir=minus, y explicit]
		table[x expr=\thisrow{magSparsity}*1.09, y=snipMeanPosttrain, y error=snipErrorMinusPosttrain] {expres/noise/rescircleMultishotNoise0.01.tsv};
		
		\addplot+[forget plot, eda errorbarcolored, y dir=plus, y explicit]
		table[x expr=\thisrow{magSparsity}*1.09, y=snipMeanPosttrain, y error=snipErrorPlusPosttrain] {expres/noise/rescircleMultishotNoise0.1.tsv};
		\addplot+[eda errorbarcolored, y dir=minus, y explicit]
		table[x expr=\thisrow{magSparsity}*1.09, y=snipMeanPosttrain, y error=snipErrorMinusPosttrain] {expres/noise/rescircleMultishotNoise0.1.tsv};

		\addplot+[forget plot, eda errorbarcolored, y dir=plus, y explicit]
		table[x expr=\thisrow{magSparsity}*1.12, y=graspMeanPosttrain, y error=graspErrorPlusPosttrain] {expres/noise/rescircleMultishotNoise0.tsv};
		\addplot+[eda errorbarcolored, y dir=minus, y explicit]
		table[x expr=\thisrow{magSparsity}*1.12, y=graspMeanPosttrain, y error=graspErrorMinusPosttrain] {expres/noise/rescircleMultishotNoise0.tsv};
		
		\addplot+[forget plot, eda errorbarcolored, y dir=plus, y explicit]
		table[x expr=\thisrow{magSparsity}*1.12, y=graspMeanPosttrain, y error=graspErrorPlusPosttrain] {expres/noise/rescircleMultishotNoise0.001.tsv};
		\addplot+[eda errorbarcolored, y dir=minus, y explicit]
		table[x expr=\thisrow{magSparsity}*1.12, y=graspMeanPosttrain, y error=graspErrorMinusPosttrain] {expres/noise/rescircleMultishotNoise0.001.tsv};
		
		\addplot+[forget plot, eda errorbarcolored, y dir=plus, y explicit]
		table[x expr=\thisrow{magSparsity}*1.12, y=graspMeanPosttrain, y error=graspErrorPlusPosttrain] {expres/noise/rescircleMultishotNoise0.01.tsv};
		\addplot+[eda errorbarcolored, y dir=minus, y explicit]
		table[x expr=\thisrow{magSparsity}*1.12, y=graspMeanPosttrain, y error=graspErrorMinusPosttrain] {expres/noise/rescircleMultishotNoise0.01.tsv};
		
		\addplot+[forget plot, eda errorbarcolored, y dir=plus, y explicit]
		table[x expr=\thisrow{magSparsity}*1.12, y=graspMeanPosttrain, y error=graspErrorPlusPosttrain] {expres/noise/rescircleMultishotNoise0.1.tsv};
		\addplot+[eda errorbarcolored, y dir=minus, y explicit]
		table[x expr=\thisrow{magSparsity}*1.12, y=graspMeanPosttrain, y error=graspErrorMinusPosttrain] {expres/noise/rescircleMultishotNoise0.1.tsv};

        \addplot[mark=none, solid, black, samples=2] coordinates {(0.002,0.99) (1.0,0.99)};
        \addplot[mark=none, densely dotted, black, samples=2] coordinates {(0.002,0.99) (1.0,0.99)};
        \addplot[mark=none, dotted, black, samples=2] coordinates {(0.002,0.98) (1.0,0.98)};
        \addplot[mark=none, loosely dotted, black, samples=2] coordinates {(0.002,0.89) (1.0,0.89)};
		
		
		\end{axis}
		\end{tikzpicture}
		\fi
    \end{subfigure}
    \caption{\textit{Varying noise.} Performance of methods for \texttt{Circle} with varying noise for $10$ rounds of alternating pruning and training. We report mean and obtained intervals (minimum and maximum) of accuracies of the final pruned network across $10$ repetitions before (left) and after (right) the final training. The noise level is indicated by line type. Baseline ticket accuracy is given in black.}\label{fig:noise}
\end{figure}

%% file: figs/grasp_local.tex
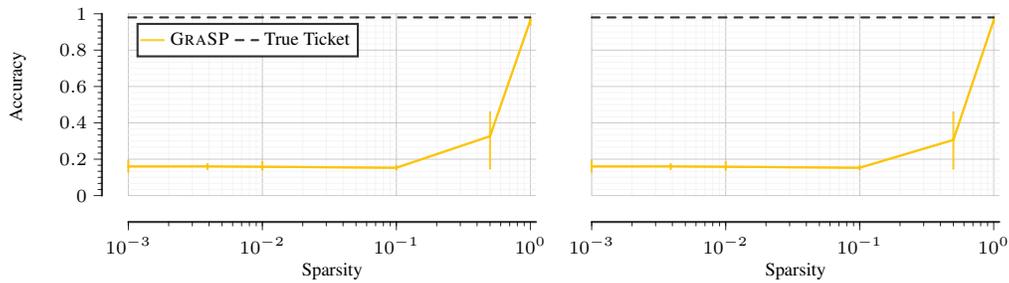
\begin{figure}
\centering
    \begin{subfigure}[t]{0.49\textwidth}
    \centering
		\ifpdf
		\begin{tikzpicture}
		\begin{axis}[
		jonas line,
		xmode=log,
		width = 7cm,
		height = 4cm,
		xlabel = {Sparsity}, 
		ylabel= {Accuracy},
		xmin=0.001, xmax=1.1,
		ymin=0, ymax=1,
		legend columns=2,
		x label style 		= {at={(axis description cs:0.5,-0.1)}, anchor=north, font=\scriptsize},
		y label style 		= {at={(axis description cs:0.0,0.6)},  anchor=south, font=\scriptsize},
		legend style={nodes={scale=1, transform shape}, at={(0.02,0.95)}, anchor=north west, row sep=-1.4pt, font=\tiny}
		]

		\addplot+[goldenpoppy, forget plot, eda errorbarcolored, y dir=plus, y explicit]
		table[x expr=\thisrow{graspSparsity}*1, y=graspMeanPostprune, y error=graspErrorPlusPostprune] {expres/multishot_local/rescircleMultishotLocal.tsv};
		\addplot+[goldenpoppy, eda errorbarcolored, y dir=minus, y explicit]
		table[x expr=\thisrow{graspSparsity}*1, y=graspMeanPostprune, y error=graspErrorMinusPostprune] {expres/multishot_local/rescircleMultishotLocal.tsv};
		\addlegendentry{\grasp}
		
        \addplot[mark=none, dashed, black, samples=2] coordinates {(0.001,0.98) (1.0,0.98)};
        \addlegendentry{True Ticket}
		
		\end{axis}
		\end{tikzpicture}
		\fi
    \end{subfigure}
    \begin{subfigure}[t]{0.49\textwidth}
    \centering
		\ifpdf
		\begin{tikzpicture}
		\begin{axis}[
		jonas line,
		xmode=log,
        y axis line style 	= {opacity=0},
        y tick style = {draw=none},
        yticklabels={,,},
		width = 7cm,
		height = 4cm,
		xlabel = {Sparsity}, 
		xmin=0.001, xmax=1.1,
		ymin=0, ymax=1,
		legend columns=2,
		x label style 		= {at={(axis description cs:0.5,-0.1)}, anchor=north, font=\scriptsize},
		y label style 		= {at={(axis description cs:0.0,0.6)},  anchor=south, font=\scriptsize},
		legend style={nodes={scale=0.9, transform shape}, at={(0.98,0.95)}, anchor=north east, row sep=-1.4pt, font=\tiny}
		]
		
		\addplot+[goldenpoppy, forget plot, eda errorbarcolored, y dir=plus, y explicit]
		table[x expr=\thisrow{graspSparsity}*1, y=graspMeanPosttrain, y error=graspErrorPlusPosttrain] {expres/multishot_local/rescircleMultishotLocal.tsv};
		\addplot+[goldenpoppy, eda errorbarcolored, y dir=minus, y explicit]
		table[x expr=\thisrow{graspSparsity}*1, y=graspMeanPosttrain, y error=graspErrorMinusPosttrain] {expres/multishot_local/rescircleMultishotLocal.tsv};
		
        \addplot[mark=none, dashed, black, samples=2] coordinates {(0.001,0.98) (1.0,0.98)};
		
		\end{axis}
		\end{tikzpicture}
		\fi
    \end{subfigure}
 \caption{\textit{Multishot with local pruning.} Performance on test data are plotted for  \texttt{Circle} against target sparsities. We report mean and obtained intervals (minimum and maximum) across $10$ repetitions of performance after pruning (left) and after training (right). Baseline ticket performance is indicated by the black line, second to left sparsity  correspond to planted ticket sparsity.}\label{fig:app_grasp_local}
\end{figure}